\documentclass{elsarticle}
\usepackage{amsmath}
\usepackage{amsfonts}
\usepackage{amssymb}
\usepackage{latexsym}
\usepackage{framed,multirow}
\usepackage{subcaption}
\usepackage{xspace}
\usepackage{float}
\usepackage[hidelinks]{hyperref}
\usepackage{subcaption}
\usepackage{color}
\usepackage{xcolor}
\usepackage{mdframed}
\usepackage{soul}
\usepackage{bbm}
\usepackage{tikz}
\usepackage{ifthen}
\usepackage{cases}
\usepackage{grffile}
\usepackage{xspace}

\journal{Journal of Computational Physics}
\usepackage[letterpaper,top=1in,bottom=1in,left=1in,right=1in]{geometry}

\usepackage{xcolor}
\definecolor{mygreen}{HTML}{006600}

\makeatletter
\DeclareRobustCommand\onedot{\futurelet\@let@token\@onedot}
\def\@onedot{\ifx\@let@token.\else.\null\fi\xspace}

\newcommand{\xreal}{x^R}
\newcommand{\ximag}{x^I}
\newcommand{\x}{\xreal + i \ximag}
\newcommand{\treal}{t^R}
\newcommand{\timag}{t^I}
\newcommand{\trealtil}{\tilde{t}^R}
\newcommand{\timagtil}{\tilde{t}^I}

\newcommand{\sign}[1]{\text{Sign}#1}
\newcommand{\rightcrit}{t_\text{crit,R}}
\newcommand{\leftcrit}{t_\text{crit,L}}

\newcounter{acntr}[section]
\renewcommand{\theacntr}{\thesection.\arabic{acntr}}
\makeatother
\newcommand{\RemarkCounter}[1]{%
  Remark
  \refstepcounter{acntr}%
  \theacntr%
  \label{#1}\xspace:\xspace
}

\usepackage{titlesec}
\titleformat{\paragraph}
{\itshape}{\theparagraph}{1em}{}
\titlespacing*{\paragraph}
{0pt}{3.25ex plus 1ex minus .2ex}{1.5ex plus .2ex}

\begin{document}
\begin{frontmatter}
\title{Addressing Discontinuous Root-Finding for Subsequent Differentiability in Machine Learning, Inverse Problems, and Control}
\author{Daniel Johnson\footnote{dansj@cs.stanford.edu, Corresponding author}}
\author{Ronald Fedkiw\footnote{fedkiw@cs.stanford.edu}}
\address{Stanford University, 353 Jane Stanford Way, Gates Computer Science, Stanford, CA, 94305, United States}
\begin{abstract}
There are many physical processes that have inherent discontinuities in their mathematical formulations.
This paper is motivated by the specific case of collisions between two rigid or deformable bodies and the intrinsic nature of that discontinuity. 
The impulse response to a collision is discontinuous with the lack of any response when no collision occurs, which causes difficulties for numerical approaches that require differentiability which are typical in machine learning, inverse problems, and control.
We theoretically and numerically demonstrate that the derivative of the collision time with respect to the parameters becomes infinite as one approaches the barrier separating colliding from not colliding, and use lifting to complexify the solution space so that solutions on the other side of the barrier are directly attainable as precise values. 
Subsequently, we mollify the barrier posed by the unbounded derivatives, so that one can tunnel back and forth in a smooth and reliable fashion facilitating the use of standard numerical approaches. 
Moreover, we illustrate that standard approaches fail in numerous ways mostly due to a lack of understanding of the mathematical nature of the problem (e.g.~typical backpropagation utilizes many rules of differentiation, but ignores L'Hopital's rule).
\end{abstract}
\end{frontmatter}

\section{Introduction}
\label{section:intro}

Recent excitement in deep learning has led to a plethora of interest in utilizing machine learning and data driven techniques for a wide variety of scientific disciplines including computational physics, see e.g. \cite{lee1990neural,sirignano2018dgm,raissi2018hidden, tripathy2018deep, winovich2019convpde,raissi2019physics,berg2019data,dal2020data,magiera2020constraint,huang2020learning,jagtap2020adaptive,geng2020coercing,ALUND2021109873}. In fact, JCP has dedicated an entire special issue to machine learning methods for physical systems \cite{jcpspecialissue}. In this paper, our focus is on the differentiability of one such physical phenomenon (i.e.~collisions).

In order to train a neural network, one typically formulates an objective function (i.e.~an energy or loss) that is subsequently minimized as a function of various neural network parameters. Many standard approaches to minimization iteratively use the Hessian or approximations to the Hessian or its inverse, often gaining efficiency by utilizing rank one updates of such approximations, see e.g.~\cite{fletcher1963rapidly,broyden1965class, broyden1967quasi, broyden1969new, fletcher1970new, goldfarb1970family, shanno1970conditioning,nocedal1980updating, liu1989limited, davidon1991variable, le2011optimization}.
In order to avoid dependence on the existence of second derivatives or robust approximations to them, the Hessian can be crudely approximated with the identity matrix in order to utilize gradient descent methods
\cite{robbins1951stochastic}.
Viewing gradient descent as forward Euler discretization of a gradient flow ordinary differential equation has led to a number of adaptive time-step numerical integration approaches including AdaGrad \cite{duchi2011adaptive}, RMSprop \cite{tieleman2012rmsprop}, and AdaDelta \cite{zeiler2012adadelta}. Moreover, the idea of using previous search directions to escape local minima led to the idea of momentum methods \cite{qian1999momentum} such as Nesterov \cite{nesterov1983method} and Adam \cite{kingma2014adam}.

Broad interest in machine and deep learning has led to the development of facilitating software such as PyTorch \cite{paszke2019pytorch}, TensorFlow \cite{abadi2016tensorflow}, Torch \cite{torch}, Caffe \cite{jia2014caffe}, Theano \cite{al2016theano}, Jax \cite{jax2018github}, etc. These software packages utilize automatic-differentiation \cite{baydin2017automatic}, specifically backpropagation \cite{schmidhuber2015deep}, to compute derivatives (as opposed to computing them analytically, numerically, or symbolically). This is accomplished by combining the derivatives of basic functions (e.g.~simple arithmetic, exponential, trigonometric, etc.) based on the various rules of differentiation, e.g.~product rule, quotient rule, chain rule, etc. There are (at least) three obvious flaws with this approach. Firstly, it is well known that it is quite difficult to write robust code for a number of common physical/mathematical problems at the heart of computational physics, e.g.~consider singular, ill-conditioned, and indefinite linear systems, robustness issues for SVD, positivity preservation and cavitation, limiters and TVD, interfaces, etc. Even the simple quadratic formula should be de-rationalized for robustness, see e.g.~\cite{heath2018scientific, harari2023computation, di1750produzioni}, and Cardano's formula has been found wanting for cubic equations, which are best solved via iterative methods and require double (as opposed to single) precision arithmetic, see e.g.~\cite{bridson2002robust}. In fact, some have argued that quadruple precision \cite{bailey2009high} or exact precision \cite{shewchuk1997adaptive} is required for various applications. Considering how much effort has been invested into devising robust numerical methods for various problems, it seems rather unlikely that the additional constraint of making such algorithms differentiable is readily obtainable. Secondly, the aforementioned software does not even do what it claims to do. That is, the basic functions are not actually being differentiated because the software ignores floating point and function approximation errors, e.g.~they differentiate $x+y$ instead of the actual $round(x+y)$, and use analytic derivatives of trigonometric, square root, and other functions that computer hardware estimates with various approximation errors.
Ignoring potentially adverse effects due to various floating point and function approximation errors seems unwise given the many issues uncovered by numerical analysts over the years.
Thirdly, these software packages have simplistic and overly idealistic modularity, which does not properly address issues that arise when disparate chunks of code are combined in various ways.
For example, \cite{johnson2023softwarebased} shows how both TensorFlow and PyTorch fail to properly treat the simple functions $x^2-4$ and $x-2$ when the former is divided by the later (aiming for $x+2$).
As is well known, the common remedy of adding a small number $\epsilon$ to the denominator (to avoid overflow when dividing by very small numbers) perturbs the result to incorrectly take on values near $0$ instead of $4$ when $x$ is near $2$;
unfortunately, both TensorFlow and PyTorch auto-differentiate this code to obtain derivatives with unbounded $O\left(\frac{1}{\epsilon}\right)$ errors.

The computational physics community has long embraced discontinuities via both mathematical formulations and numerical algorithms. Consider, for example, the use of the weak (integral) form for conservation laws in order to correctly model shocks and detonations, which do not possess the differentiability required for existence of the strong form, see e.g.~\cite{leveque1992numerical, toro2013riemann}. This has led to a variety of numerical approaches with discontinuous decision making, such as ENO \cite{shu1989efficient}. Also consider, for example, sharp interface methods for contact discontinuities and material interfaces when simulating compressible flows, incompressible flows, solids undergoing fracture, etc., see e.g.~level set methods \cite{osher2004level}, ghost fluids methods \cite{fedkiw1999non, liu2000boundary, kang2000boundary}, immersed interface methods \cite{li2006immersed}, XFEM \cite{moes1999finite}, etc. On the other hand, while respecting nondifferentiability and the need to correctly treat discontinuities, computational physics researchers have aimed for smoothness when desirable and/or appropriate. For example, implicit time integration and steady-state/quasistatic approaches to nonlinear hyperbolic partial differential equations typically lead to nonlinear system solves that require smoothness as well as direct or indirect access to various derivatives; thus, smooth or smoothed approximations to the governing equations are highly beneficial and often sought, see e.g.~\cite{jameson1991time,belov1995new,jiang1996efficient,belytschko2001computability,kadioglu2005second,teran2005robust,kwatra2009method,hughes2012finite,de2012nonlinear}. This motivated, for example, formulating WENO \cite{liu1994weighted,belytschko2013nonlinear} as a convex combination of the three ENO choices; however, practitioners have struggled to get the smoother WENO scheme to work well enough near shocks and detonations, and a popular approach is to use ENO near discontinuities and WENO in smoother regions of the flow, see e.g.~\cite{shu2020} and the references therein. Unfortunately, this careful consideration of the potentially adverse effects caused by smoothing is mostly absent from the software and proposed approaches of the computer science community, which typically naively assume that one can indiscriminantly smooth mathematical formulations of physical processes and/or differentiate code instead of equations, see e.g.~\cite{hu2019difftaichi}. Notable exceptions include: \cite{zhuang2020adaptive} where the authors explain how roundoff errors can cause backpropagation to fail when considering neural ordinary differential equations, \cite{suh2022differentiable} which discusses how stiffness and discontinuities can compromise the efficacy of differentiable simulators, and \cite{metz2021gradients} which discusses when differentiation based optimization algorithms struggle due to problems with the Jacobian.

In spite of the aforementioned issues, the machine and deep learning community still obtains rather impressive results on a variety of problems. This is likely because the direct penalization of deviations from the training data in their objective function helps to overcome other flaws in their approach. Particularly questionable is their justification for dropout \cite{srivastava2014dropout}, which claims that randomly setting derivatives to zero while training a neural network is somehow equivalent to averaging various good models together. Although model averaging is certainly justified when a number of good models exist, there is no justification that randomly dropping the subsets of a model that vanish when terminating dependencies by arbitrarily setting derivatives to zero leads to viable models. A better justification would be that that dropout coaxes the neural network to match the training data even when the derivatives are so poorly approximated that they are randomly set to zero; this calls into question the entire paradigm of using differentiability, backpropagation, and optimization for training neural networks. If one were to believe the theoretical justifications regarding differentiability, then at the very least one would ascertain error bounds on the derivatives and randomly perturb derivative estimates within those bounds instead of randomly setting derivatives to zero; additionally, derivatives that are nonexistent or that blow up towards infinite values should be more carefully addressed as well, rather than randomly setting them to zero and hoping for the best.

This paper was motivated by examining collisions between rigid and deformable bodies where the mathematical formulations and numerical methods are both known to be problematic due to differential inclusions \cite{stewart2011dynamics,stewart1996implicit} and issues with accurately solving cubic equations \cite{bridson2002robust, ferguson2021intersection, harari2023computation}. The concept of whether or not a collision occurs (i.e.~collision detection) is intrinsically discontinuous, and we show that this manifests itself as a derivative blowing up towards infinity as one approaches the decision boundary between colliding or not colliding. Any robust root-finding approach will require discontinuous (and thus non-differentiable) decision making, e.g.~consider the hybridization of Newton's method with bisection. Since both the equations and the numerical methods are not readily differentiable, backpropagating through the iterative solver seems unwise; instead, we follow an approach similar to \cite{geng2020coercing,agrawal2019differentiable} of differentiating the equations (this is typically referred to as an implicit layer, see e.g.~\cite{kolter2020nips} and the references therein). This is akin to using the pseudoinverse $A^+$ to represent the Jacobian $\frac{\partial x}{\partial b}$ from $x = A^+ b$ instead of backpropagating through whatever algorithm was used to solve $Ax=b$. For collision detection, this amounts to implicit differentiation of a cubic equation (note, \cite{liang2019differentiable, qiao2020scalable} took a similar, albeit incorrect, approach\footnote{\cite{liang2019differentiable} published an incorrect derivative; meanwhile, that derivative is set to be identically zero in their code (see https://github.com/williamljb/DifferentiableCloth). The code for \cite{qiao2020scalable} (see https://github.com/YilingQiao/diffsim) does not set the derivative to zero, but still uses the incorrect formula from \cite{liang2019differentiable}.}). Although implicit differentiation allows the derivatives to be obtained while still utilizing a state-of-the-art iterative solver with non-differentiable decision making intact, it does not address derivatives blowing up towards infinity. Unfortunately, as was pointed out in \cite{bolte2021nips}, the machine learning community has mostly ignored the conditions required to validate the use of the implicit function theorem. To properly address this for collision detection, we lift the solution space to $\mathbb{R}^2$ in order to allow the iterative solver to more readily work its way back and forth between real-valued roots representing collisions and complex-valued roots representing the absence of collisions; then, we mollify the barrier posed by the unbounded derivative in a fashion that does not pollute the accuracy or attainability of solutions when they exist. Notably, this is enabled by devising a new canonical form for cubic equations (perhaps not previously appearing in the literature).
\section{Preliminaries}
\label{section:motivation}
\newcommand{\blue}[1]{{\color{blue} #1}}
\definecolor{green}{rgb}{0,.4,0}
\newcommand{\green}[1]{{\color{green} #1}}
\newcommand{\red}[1]{{\color{red} #1}}

Collision detection and response is important to a wide variety of material modeling problems, where the materials may be approximated as multibody systems with each body either stiff enough to be treated as a rigid body or instead simulated with a deformable finite element approximation. The surface of each body can be discretized into a set of triangles; then, a collision occurs when either a vertex from one surface impacts a triangle of the other or when two edges collide. In both cases, a collision occurs when four points become coplanar. When the objects are close enough together, one can linearize the motion of these points via $\vec{x}_i (t) = \vec{x}_i^o + \vec{v}_i t$ where $\vec{x}_i (t)$ is the position of point $i$ at time $t$, $\vec{x}_i^o$ is the position before linearization, and $\vec{v}_i$ is the linearized velocity. Choosing $\vec{x}_1(t)$ as a frame of reference, the three edge vectors $\vec{x}_{21}(t) = \vec{x}_2(t) - \vec{x}_1(t)$, $\vec{x}_{31}(t) = \vec{x}_3(t) - \vec{x}_1(t)$, and $\vec{x}_{41}(t) = \vec{x}_4(t) - \vec{x}_1(t)$ describe a would-be tetrahedron (similar to Green strain \cite{teran2003finite}) where coplanarity is equivalent to the tetrahedron having zero volume, e.g. $\vec{x}_{21}(t) \times \vec{x}_{31}(t) \cdot \vec{x}_{41}(t) = 0$. This results in a cubic equation for $t$, which (as discussed in \cite{bridson2002robust}) requires a carefully designed/implemented iterative solver using double precision in order to guarantee that potential collisions are not missed. Given a time $t$ of coplanarity, the positions of the points are examined to determine if either the point is inside the triangle or the two edges overlap (depending on which case is being considered). If a collision occurs, the positions and velocities of bodies are used to determine a collision response.

When the object of interest is deformable, the particles are true degrees of freedom. When the body is rigid, the degrees of freedom are its center-of-mass translational and rotational velocity. In both cases, we refer to the degrees of freedom as $\vec{\lambda}$ for the sake of exposition. A typical goal might be to obtain a specific post-collision velocity by somehow modifying $\vec{\lambda}$ via controllable degrees of freedom. Although this seems feasible when considering collision response, changes in $\vec{\lambda}$ have no effect on post-collision velocities when there is no collision. In a real-world scenario, the prospect of missing a collision would inevitably motivate a change in strategy; instead of focusing on the final trajectory of the particle, one might turn their attention towards aiming to create a collision. Mathematically, missing the collision is equivalent to the desired root of a cubic equation being complex-valued instead of real-valued, and aiming to create a collision is equivalent to aiming to change the complex-valued root into a real-valued root. Although the proposition of collision or no collision at first appears binary and unavoidably non-differentiable, the real-world scenario would seem to indicate that the collision response perhaps could be differentiably connected to collision detection by formally lifting the root-finding problem to consider both real and complex roots. 

Let $t_\text{root}(\vec{\lambda})$ represent a solution/root to the aforementioned cubic equation, and consider minimizing an objective function 
\begin{equation}
\label{eq:loss} 
L(\vec{\lambda}) =
\frac{1}{2}\left(\sum_\text{roots}||t_\text{root}(\vec{\lambda}) - t_\text{root,L}||_2^2 +
\left|\left|
\vec{p}(\vec{\lambda}) - \vec{p}_L
\right|\right|_2^2
\right)
\end{equation}
where each root of interest $t_\text{root}(\vec{\lambda})$ may have an aspirational target value $t_\text{root,L}$ and the (column vector) coefficients of the cubic $\vec{p}(\vec{\lambda})$ may be regularized towards some $\vec{p}_L$.
For optimization, one would utilize the gradient 
\begin{equation}
\label{eq:loss_gradient} 
\begin{aligned}
\nabla_{\vec{\lambda}} L
= 
\frac{\partial \vec{p}}{\partial \vec{\lambda}}^T
\left(
\sum_\text{roots}
\frac{\partial t_\text{root}}{\partial \vec{p}}^T
(t_\text{root}(\vec{\lambda}) - t_\text{root,L})
+
\vec{p}(\vec{\lambda}) - \vec{p}_L
\right)
\end{aligned}
\end{equation}
where the $\partial$ represents the typical Jacobian and thus the gradients (being the transpose) reverse their order.
Here, $t_\text{root}$ is a column vector of separate real and imaginary parts, and $\frac{\partial t_\text{root}}{\partial \vec{p}}$ has two rows (one for the real part and one for imaginary part).

\newcommand{\raw}{\text{orig}}
In order to aid both the analysis and the numerics for $\frac{\partial t_\text{root}}{\partial \vec{p}}$, we guarantee that the maximum magnitude of any entry in $\vec{p}$ is bounded above by 1 simply by dividing by the largest entry $p_{\max}$ if it has magnitude larger than 1. Generally speaking, this division can be problematic (perhaps requiring asymptotic analysis) when entries of $\vec{p}$ are blowing up; however, since $\vec{p}$ is a function of $\vec{\lambda}$, this is problem specific and we leave it to the reader. Going forward, we will treat $\vec{p}$ as if it were bounded when considering $\frac{\partial t_\text{root}}{\partial \vec{p}}$, and use $\vec{p}_\raw$ to represent the original not necessarily bounded polynomial coefficients. Since $t_\text{root}(\vec{p}_\raw) = t_\text{root}(\vec{p})$, one can simply treat $t_\text{root}$ as a function of the bounded parameters instead. For the derivatives, one needs to replace $\frac{\partial t_\text{root}}{\partial \vec{p}_\raw}$ with $\frac{\partial t_\text{root}}{\partial \vec{p}}\frac{\partial \vec{p}}{\partial \vec{p}_\raw}$ where
\begin{equation}
    \label{eq:dpboundeddp} 
    \frac{\partial \vec{p}}{\partial \vec{p}_\raw}
    =
    \frac{1}{p_{\max}^2}
    \left(
    p_{\max} I - \vec{p}_\raw \hat{e}_k^T
    \right)
\end{equation}
and $\hat{e}_k$ is a standard unit basis vector (where $k$ is the index of $p_{\max}$ in $p_\raw$).
This leads to replacing $\frac{\partial \vec{p}}{\partial \vec{\lambda}}^T$ with $\frac{\partial \vec{p}_\raw}{\partial \vec{\lambda}}^T\frac{\partial \vec{p}}{\partial \vec{p}_\raw}^T$ in equation \ref{eq:loss_gradient}.
We stress that we still allow for the coefficients of the cubic (i.e. $\vec{p}_\raw$) to grow rather large, but believe that it is more tidy to address this with equation \ref{eq:dpboundeddp} and a bounded $\vec{p}$ in $\frac{\partial t_\text{root}}{\partial \vec{p}}$ than worrying about the case where $\vec{p}_\raw$ is large in $\frac{\partial t_\text{root}}{\partial \vec{p}_\raw}$. 
\section{Quadratic Equations}
\label{section:quadratic}
\newcommand{\tmax}{t_{\max}}

We motivate our approach by restricting the cubic equation $q t^3 + a t^2 + b t + c = 0$ to the simpler $q=0$ quadratic equation case. The roots of the quadratic equation are
\begin{equation}
\label{eq:quadraticroots}
t_\text{root}^\pm = \frac{-b \pm \sqrt{b^2-4ac}}{2a}
\end{equation}
which are real-valued when $b^2-4ac \geq 0$ (repeated when $b^2-4ac = 0$) and complex conjugates when $b^2-4ac < 0$. As discussed in Section \ref{section:motivation}, we guarantee that the polynomial coefficients have magnitudes less than or equal to 1; thus, the only numerically problematic case occurs when $a$ is small. As $a\to0$, $\frac{-b}{2a}$ can be any real number depending on the behavior of $b$ which may also go to zero and do so at speeds faster than, slower than, or commensurate with $a$; thus, repeated roots and the real part of complex roots may take on any value (perhaps even being unbounded). The imaginary parts of complex roots have equal and opposite sign and may bounded or unbounded. In the case of two distinct real roots, they may appear anywhere, i.e.~both bounded, one bounded and the other unbounded, or both unbounded (with the same or opposite signs).

\textit{
\RemarkCounter{re:}
For the sake of a practical implementation, one needs to establish the largest magnitude number $\tmax$ that can be used to represent a root (on a computer of interest). We refer to roots outside $(-\tmax,\tmax)$ as either unbounded, approaching $\pm \infty$, blowing up, etc. as appropriate. Notably, our approach fully handles this case, including the ability to drive these unbounded roots to smaller obtainable values when desired.
}

Letting the independent variable be $t_\text{root} = [\treal_\text{root}, \timag_\text{root}]^T$, one can write 
\begin{subequations}
\label{eq:diffquadraticformulaall}
\begin{align}
\label{eq:diffquadraticformula}
\frac{\partial \treal_\text{root}}{\partial \vec{p}}   
&=
\begin{bmatrix}
\frac{b\mp \sqrt{b^2-4ac}}{2a^2}
\mp \frac{c}{a \sqrt{b^2-4ac}}
&
\frac{-1}{2a} \pm \frac{b}{2a\sqrt{b^2-4ac}}
&
\frac{\mp 1}{\sqrt{b^2-4ac}} \\
\end{bmatrix}\\
\label{eq:implicitwithparams}
&=
\frac{-1}{\pm\sqrt{b^2-4ac}}
\begin{bmatrix}
\left(\frac{-b \pm\sqrt{b^2-4ac}}{2a}\right)^2
&
\frac{-b \pm\sqrt{b^2-4ac}}{2a}
&
1 \\
\end{bmatrix}\\
\label{eq:rework_to_implicit}
&=
\frac{-1}{\pm2\sqrt{-\tilde{c}}}
\begin{bmatrix}
\left(\treal_\text{root}\right)^2
&
\left(\treal_\text{root}\right)^1
&
\left(\treal_\text{root}\right)^0\\
\end{bmatrix}
\end{align}
\end{subequations}
when the roots are real (i.e.~with $\timag=0$);
here, $\tilde{c} = -\frac{1}{4}(b^2-4ac)\leq0$.
Equation \ref{eq:rework_to_implicit} elucidates a trivial and valid strategy covering the case when a root blows up; as $\treal_\text{root} \to \pm \infty$, the direction of $\frac{\partial \treal_\text{root}}{\partial \vec{p}}$ approaches $[\mp 1~0~0]$. This indicates the need to fix $a$, which makes sense since this degeneracy is caused by $a\to0$. The magnitude of $\frac{\partial \treal_\text{root}}{\partial \vec{p}}$ will also approach infinity; however, it can be clamped without changing the direction to some maximum allowable value that makes sense (for optimization). The magnitude can be similarly clamped when $\treal_\text{root}$ is bounded and $\tilde{c}\to0$, indicating the merging of two real roots into a repeated root.

A rather interesting case occurs as both $\tilde{c}\to0$ and $\treal_\text{root} \to 0$, i.e.~the roots are merging towards $\treal_\text{root}=0$. Although the aforementioned strategy robustly treats this via $\frac{\partial \treal_\text{root}}{\partial \vec{p}}$ having direction $[0~0~1]$ with a clamped magnitude, some form of L'Hospital's rule is required when one is interested in obtaining values for $\frac{\partial \treal_\text{root}}{\partial a}$ and $\frac{\partial \treal_\text{root}}{\partial b}$. Writing $b^2-4ac \to \gamma b^2$ treats the case when $b^2$ dominates $4ac$ by $\gamma=1$, the case when $4ac$ dominates $b^2$ by $\gamma\to\infty$, and co-dominance by $\gamma \in [0, 1) \cup (1, \infty)$ with $[0, 1)$ when $4ac>0$ and $(1, \infty)$ when $4ac < 0$. Substituting $b^2-4ac \to \gamma b^2$ into equation \ref{eq:quadraticroots} gives
\begin{equation}
\label{eq:realrootsgamma}
\treal_\text{root} \to \frac{-b}{2a}(1 \mp \sign{(b)}\sqrt{\gamma})\\
\end{equation}
where $\sign{(b)}$ is the sign of $b$.
Substituting equation \ref{eq:realrootsgamma} into equation \ref{eq:rework_to_implicit} gives
\begin{equation}
\label{eq:dtrdp4}
\frac{\partial \treal_\text{root}}{\partial \vec{p}}   
\to
\begin{bmatrix}
\frac{1}{4a^2}
\left(
\frac{\mp|b|}{\sqrt{\gamma}} + 2 b \mp |b|\sqrt{\gamma}
\right) 
&
\frac{-1}{2a}
\left(
\frac{\mp1}{\sign{(b)}\sqrt{\gamma}}
+
1
\right)
&
\frac{\mp1}{|b|\sqrt{\gamma}}\\
\end{bmatrix}   
\end{equation}
where $\gamma b^2 \to 0$ (and thus $b\sqrt{\gamma}\to0$) implies that $\frac{\partial \treal_\text{root}}{\partial c}$ always blows up. $\frac{\partial \treal_\text{root}}{\partial b}$ is indeterminate, ranging from $\frac{-1}{2a}$ as $\gamma\to\infty$ to various finite values for finite $\gamma$ to blowing up as $\gamma\to0$. The second two terms in $\frac{\partial \treal_\text{root}}{\partial a}$ vanish. When $\gamma\neq0$, the first term in $\frac{\partial \treal_\text{root}}{\partial a}$ also goes to zero implying that $\frac{\partial \treal_\text{root}}{\partial a} \to 0$. When $\gamma=0$, $b^2$ and $4ac$ co-dominate to cancel the $O(b^2)$ terms in $b^2-4ac$; however, even though $\gamma$ does not contain any $O(1)$ terms, it can still contain powers of $b$. When $\sqrt{\gamma}\to0$ slower than $b\to0$, $\frac{\partial \treal_\text{root}}{\partial a}\to0$. When $\sqrt{\gamma}\to0$ at the same speed as $b\to0$, $\frac{\partial \treal_\text{root}}{\partial a}$ is finite. When $\sqrt{\gamma}\to0$ faster than $b\to0$, $\frac{\partial \treal_\text{root}}{\partial a}$ blows up. Thus, $\frac{\partial \treal_\text{root}}{\partial a}$ is indeterminate. Although one might attempt to remove the set of measure zero sequences where $\gamma\to0$ by setting $\frac{\partial \treal_\text{root}}{\partial a}\to0$ in all cases, no similar strategy works for $\frac{\partial \treal_\text{root}}{\partial b}$.

\textit{
\RemarkCounter{re:}
Numerically, these asymptotics will manifest themselves via infinitesimal values for the parameters generated pseudo-randomly because of limited numerical precision, e.g.~$-4ac>>b^2$ is $\gamma\to\infty$, $b^2>>-4ac$ is $\gamma\to1$, and $b^2-4ac<<b^2$ is $\gamma\to0$. The obvious difficulty is that roundoff errors and the representability of small numbers are difficult to predict and control. 
}

Extending equation \ref{eq:rework_to_implicit} to include the identically zero imaginary part of the real roots gives
\begin{subequations}
\label{eq:dtdp1}
\begin{align}
\frac{\partial t_\text{root}}{\partial \vec{p}}
&=
\frac{-1}{\pm2\sqrt{-\tilde{c}}}
\begin{bmatrix}
\left(\treal_\text{root}\right)^2
&
\left(\treal_\text{root}\right)^1
&
\left(\treal_\text{root}\right)^0\\
0&0&0\\
\end{bmatrix}\\
&=
\frac{-1}{\pm2\sqrt{-\tilde{c}}}
\begin{bmatrix}
Re(z_\text{root}^2) & Re(z_\text{root}^1) & Re(z_\text{root}^0)\\
Im(z_\text{root}^2) & Im(z_\text{root}^1) & Im(z_\text{root}^0)\\
\end{bmatrix}
\end{align}
\end{subequations}
where $z_\text{root} = \treal_\text{root} + i \timag_\text{root}$, and $Re$ and $Im$ are the real and imaginary parts respectively. Equations \ref{eq:diffquadraticformulaall}-\ref{eq:dtdp1} and the related discussion are only valid in the halfspace where both roots are real.
In the halfspace where the roots are complex,
\begin{subequations}
\label{eq:dtdp2} 
\begin{align}
\frac{\partial t_\text{root}}{\partial \vec{p}}
&=
\begin{bmatrix}
\frac{b}{2a^2} 
&
\frac{-1}{2a}&
0 \\
\frac{\mp\sqrt{-(b^2-4ac)}}{2a^2} \pm \frac{c}{a\sqrt{-(b^2-4ac)}} & 
\mp\frac{b}{2a\sqrt{-(b^2-4ac)}} &
\frac{\pm1}{\sqrt{-(b^2-4ac)}}
\\
\end{bmatrix}\\
&=
\label{eq:dtdp2b} 
\frac{-1}{\pm2\sqrt{\tilde{c}}}
\begin{bmatrix}
	2\treal_\text{root}\timag_\text{root} & \timag_\text{root} & 0\\
	-\left((\treal_\text{root})^2 - (\timag_\text{root})^2\right) & -\treal_\text{root} & -1\\
\end{bmatrix}\\
&=
\frac{-1}{\pm2\sqrt{\tilde{c}}}
\begin{bmatrix}
    0 & 1\\
    -1 & 0\\
\end{bmatrix}
\begin{bmatrix}
	Re(z_\text{root}^2) & Re(z_\text{root}^1) & Re(z_\text{root}^0)\\
	Im(z_\text{root}^2) & Im(z_\text{root}^1) & Im(z_\text{root}^0)\\
\end{bmatrix}
\end{align}
\end{subequations}
instead.
Noting that $\pm2\sqrt{\tilde{c}} = \pm\sqrt{-(b^2-4ac)} = 2a\timag_\text{root}$, the real part of equation \ref{eq:dtdp2b} has direction $[2\treal_\text{root}~1~0]$ indicating the need to fix $a$ when $\treal_\text{root}$ is big. If either $\treal_\text{root}$ or $\timag_\text{root}$ is large, but not both, the imaginary part of equation \ref{eq:dtdp2b} also indicates fixing $a$; however, when $\treal_\text{root}$ and $\timag_\text{root}$ are both large, they may cancel making the second entry (i.e.~modify $b$) dominate. 

\textit{
\RemarkCounter{re:}
Strategically, modifying $b$ may be unwise in this case because it leads to real-valued roots whereas fixing $a$ can leave the roots complex-valued. Moreover, one needs to fix $a$ anyways (because $\treal_\text{root}$ is large) as indicated by the first row in equation \ref{eq:dtdp2b}.
}

Next, we revisit $\tilde{c} \to 0$ and $\treal_\text{root} \to 0$, i.e.~the roots merging to $\treal_\text{root}=0$ but from the complex halfspace this time. Here, $\gamma \in (-\infty,0]$ with $4ac$ dominating $b^2$ given by $\gamma\to-\infty$; in addition, $\gamma\to0$ from the $4ac>b^2$ side. Substituting $b^2-4ac \to \gamma b^2$ into equation \ref{eq:quadraticroots} gives
\begin{equation}
\label{eq:complexrootsgamma}
\begin{bmatrix}
    \treal_\text{root}\\
    \timag_\text{root}\\
\end{bmatrix}
\to
\frac{-b}{2a}
\begin{bmatrix}
1\\
\mp\sign{(b)}\sqrt{-\gamma}\\
\end{bmatrix}
\end{equation}
which when substituted into equation \ref{eq:dtdp2b} gives
\begin{equation}
\label{eq:} 
\frac{\partial t_\text{root}}{\partial \vec{p}}
\to
\begin{bmatrix}
	\frac{b}{2a^2} & \frac{-1}{2a} & 0\\
	\frac{1}{4a^2}\left(\frac{\pm |b|}{\sqrt{-\gamma}} \mp |b|\sqrt{-\gamma}\right) & \frac{-1}{2a}\left(\frac{\pm 1}{\sign{(b)}\sqrt{-\gamma}}\right) & \frac{\pm1}{|b|\sqrt{-\gamma}}\\
\end{bmatrix}
\end{equation}
where $\frac{\partial \timag_\text{root}}{\partial c}$ blows up, and both $\frac{\partial \timag_\text{root}}{\partial b}$ and $\frac{\partial \timag_\text{root}}{\partial a}$ are indeterminate (as in equation \ref{eq:dtrdp4}).

\subsection{Implicit Differentiation}
\label{subsection:quadratic-implicit-differentiation}
For the sake of exposition, we write the quadratic equation as
\begin{equation}
\label{eq:quadratic2d}
f(t_\text{root}; \vec{p}) = 
\begin{bmatrix}
	a (\treal_\text{root})^2 - a (\timag_\text{root})^2 + b \treal_\text{root} + c\\
	(2 a \treal_\text{root} + b) \timag_\text{root}\\
\end{bmatrix}
=
\vec{0}
\end{equation}
letting $\theta_1$ refer to the first variable (i.e.~$t_\text{root}$) and $\theta_2$ refer to the second variable (i.e.~$\vec{p}$)
so that the derivatives
\begin{subequations}
\label{eq:quadraticderivatives} 
\begin{align}
f_{\theta_1}(t_\text{root}; \vec{p}) &= 
\begin{bmatrix}
	2 a \treal_\text{root} + b & - 2 a \timag_\text{root}\\
	2 a \timag_\text{root} & 2 a \treal_\text{root} + b\\
\end{bmatrix}\\
f_{\theta_1}^{-1}(t_\text{root}; \vec{p}) &= 
\frac{1}{s}
\begin{bmatrix}
	2 a \treal_\text{root} + b & 2 a \timag_\text{root}\\
	-2 a \timag_\text{root} & 2 a \treal_\text{root} + b\\
\end{bmatrix}
\text{ where }
s = (2 a \treal_\text{root} + b)^2 + (2 a \timag_\text{root})^2\\
f_{\theta_2}(t_\text{root}; \vec{p}) &= 
\begin{bmatrix}
	Re(z_\text{root}^2) & Re(z_\text{root}^1) & Re(z_\text{root}^0)\\
	Im(z_\text{root}^2) & Im(z_\text{root}^1) & Im(z_\text{root}^0)\\
\end{bmatrix}
\end{align}
\end{subequations}
have compact notation.
The total derivative of equation \ref{eq:quadratic2d} is $f_{\theta_1}(t_\text{root}, \vec{p}) d \theta_1 + f_{\theta_2}(t_\text{root}, \vec{p}) d \theta_2 = 0$, which can be written as $d \theta_1 = - f_{\theta_1}^{-1}(t_\text{root}, \vec{p}) f_{\theta_2}(t_\text{root}, \vec{p}) d \theta_2$ or
\begin{equation}
\label{eq:implicitderivative} 
dt_\text{root}
=
-
\frac{1}{s}
\begin{bmatrix}
	2 a \treal_\text{root} + b & 2 a \timag_\text{root}\\
	-2 a \timag_\text{root} & 2 a \treal_\text{root} + b\\
\end{bmatrix}
\begin{bmatrix}
	Re(z_\text{root}^2) & Re(z_\text{root}^1) & Re(z_\text{root}^0)\\
	Im(z_\text{root}^2) & Im(z_\text{root}^1) & Im(z_\text{root}^0)\\
\end{bmatrix}
d\vec{p}
\end{equation}
when $f_{\theta_1}$ is invertible.
In the halfspace where the roots are real, $\timag_\text{root} = 0$ and $2a\treal_\text{root} +b = \pm2\sqrt{-\tilde{c}}$ showing the equivalence between equations \ref{eq:implicitderivative} and \ref{eq:dtdp1}b. In the complex halfspace, $2a\treal_\text{root} +b = 0$ and $2a\timag_\text{root} = \pm2\sqrt{\tilde{c}}$ showing the equivalence between equations \ref{eq:implicitderivative} and \ref{eq:dtdp2}c. Interestingly, even though $f_{\theta_1}$ is not invertible when $s=0$ (the repeated roots case) and thus equation \ref{eq:implicitderivative} is not formally derivable via the total derivative and the implicit function theorem, equation \ref{eq:implicitderivative} does match equations \ref{eq:dtdp1}b and \ref{eq:dtdp2}c.

\textit{
\RemarkCounter{re:}
Our treatment of $t_\text{root}$ as a column vector with separate real and imaginary parts means that the multiplication of two complex numbers $z_1$ and $z_2$ is
\begin{equation}
\label{eq:complexmul1}
z_1 z_2
= 
\begin{bmatrix}
	Re(z_1) & -Im(z_1)\\
	Im(z_1) & Re(z_1)\\
\end{bmatrix}
\begin{bmatrix}
	Re(z_2) \\ Im(z_2)\\
\end{bmatrix}
\end{equation}
allowing equation \ref{eq:implicitderivative} to be rewritten as 
\begin{equation}
\label{eq:t_root_differential}   
dz_\text{root} = \frac{-1}{2az_\text{root} + b}
\begin{bmatrix}
    z_\text{root}^2& z_\text{root}^1 & z_\text{root}^0
\end{bmatrix}
d\vec{p}
\end{equation}
since
\begin{equation}
\frac{1}{2 a z_\text{root} + b}	
=
\frac{1}{2 a (\treal_\text{root} + i \timag_\text{root}) + b}
=
\frac{1}{s} \left(2 a \treal_\text{root} + b - i 2a \timag_\text{root}\right)
.
\end{equation}
}
\section{Newton's Method}
\label{section:newtonsmethod}
When the roots are real, equation \ref{eq:quadraticroots} is typically evaluated using de-rationalization
\begin{subequations}
	\label{eq:quadform}
	\begin{align}
        t_\text{root}^+
		&=
		\begin{cases}
            \frac{-b+\sqrt{b^2-4ac}}{2a}
			& \text{if } b\leq0\\
			\frac{2c}{-b-\sqrt{b^2-4ac}}
			& \text{if }b>0\\
		\end{cases}\\
        t_\text{root}^-
		&=
		\begin{cases}
            \frac{-b-\sqrt{b^2-4ac}}{2a}
			& \text{if }b\geq0\\
			\frac{2c}{-b+\sqrt{b^2-4ac}}
			& \text{if } b<0\\
		\end{cases}
	\end{align}
\end{subequations}
to avoid catastrophic cancellation (see e.g.~\cite{heath2018scientific, harari2023computation, di1750produzioni}).
Prior works on cubic equations (e.g.~\cite{bridson2002robust}) found that Cardano's explicit formula lacked the accuracy required for collision detection and instead used iterative methods; moreover, \cite{bridson2002robust} stressed that double precision (rather than single precision) was required in order to detect collisions accurately enough for their cloth simulations. Since our consideration of the quadratic equation is merely a building block for the cubic equation, we utilize Newton's method. Note that it is important to use $t_\text{root}(\vec{p})$ instead of $t_\text{root}(\vec{p}_\raw)$, as discussed in Section \ref{section:motivation}, in order to avoid numerical issues with convergence (and detecting convergence) when using Newton's method; otherwise, we have observed Newton's method struggling to converge to prescribed tolerances due to cancellation issues. This makes sense since equation \ref{eq:quadratic2d} is linear in $\vec{p}$.

Newton's method for computing $t_n$ recursively from $t_{n-1}$ is
\begin{subequations}
\label{eq:newtons method} 
\begin{align}
t_n 
&= t_{n-1} - f_{\theta_1}^{-1}(t_{n-1}; \vec{p}) f(t_{n-1}; \vec{p})\\
&= t_{n-1} -
\frac{1}{s}
\begin{bmatrix}
	2 a \treal_{n-1} + b & 2 a \timag_{n-1}\\
	-2 a \timag_{n-1} & 2 a \treal_{n-1} + b\\
\end{bmatrix}
\begin{bmatrix}
	a (\treal_{n-1})^2 - a (\timag_{n-1})^2 + b \treal_{n-1} + c\\
	(2 a \treal_{n-1} + b) \timag_{n-1}\\
\end{bmatrix}
\end{align}
\end{subequations}
where $s = (2 a \treal_{n-1} + b)^2 + (2 a \timag_{n-1})^2$ and the iteration proceeds until $f(t_n, \vec{p})$ is small enough. When $a=0$ and $b\neq0$, equation \ref{eq:newtons method}b becomes
\begin{equation}
	\label{eq:complex newton update 1.5}
	\begin{aligned}
		t_{n}
		&=
		t_{n-1}-
		\frac{1}{b}
		\begin{bmatrix}
			b\treal_{n-1} + c\\
			b\timag_{n-1}\\
		\end{bmatrix}
	\end{aligned}
\end{equation}
which converges to $\treal_\text{root} = \frac{-c}{b}$ in one iteration. When $a=b=0$, $f(t; \vec{p})=c$ is a constant function where either nothing ($c\neq0$) or everything ($c=0$) is a root. When $a\neq0$, equation \ref{eq:newtons method}b can be rewritten as
\begin{equation}
	\label{eq:complex newton update 2}
	\begin{aligned}
		t_{n}
		&=
		\begin{bmatrix}
			-\frac{b}{2a}\\
			0\\
		\end{bmatrix}
		+
		\frac{1}{4a}
		\begin{bmatrix}
			\left( 1 - \frac{4\tilde{c}}{s} \right) \left(2a\treal_{n-1} + b\right)\\
			\left(1+\frac{4\tilde{c}}{s}\right)\left(2a\timag_{n-1}\right)\\
		\end{bmatrix}
	\end{aligned}
\end{equation}
using only algebraic manipulation.
In the case of repeated roots, $\tilde{c}=0$ and the convergence is linear (with the distance between $(\treal_n, \timag_n)$ and $(\frac{-b}{2a}, 0)$ cut in half every iteration) instead of quadratic. When $\tilde{c} \approx 0$ (but $\tilde{c}\neq0$) with roots close to $(\frac{-b}{2a}, 0)$, converging iterates give values of $s$ close to (perhaps even identically equal to) zero making $\frac{\tilde{c}}{s}$ problematic. Although $2a\treal_{n-1}+b$ and $2a\timag_{n-1}$ shrink like $\sqrt{s}$ and thus help to eliminate a vanishing $s$, $\frac{\tilde{c}}{\sqrt{s}}$ is still problematic. One remedy would be to set $\tilde{c} = 0$ (and eliminate $\frac{\tilde{c}}{s}$ from the computation) forcing a repeated root even when the roots are distinct, arguing that this is only a small perturbation of the distinct roots. 

For the sake of exposition, let $\tilde{c}\leq0$ so that only real-valued roots are relevant;
then equation \ref{eq:newtons method}b can be written as
\begin{equation}
\label{eq:lm_newton_with_error}
\treal_n = \treal_{n-1} - \frac{a (\treal_{n-1})^2 + b \treal_{n-1} + c + \delta}{2a\treal_{n-1} + b + \epsilon}
\end{equation}
where $\epsilon$ is used to avoid division by zero, and $\delta$ collects (all) the roundoff errors from computing the numerator and denominator, dividing, and subtracting from $\treal_{n-1}$.
Without loss of generality, equation \ref{eq:lm_newton_with_error} can be rewritten as
\begin{equation}
    \label{eq:lm_newton_with_error2}
    \bar{t}^R_n
    = \frac{a (\bar{t}^R_{n-1})^2 + \epsilon \bar{t}^R_{n-1} - \frac{\tilde{c}}{a} - \delta}{2a\bar{t}^R_{n-1} + \epsilon}   
\end{equation}
via the change of variables $\bar{t}^R = \treal + \frac{b}{2a}$.
Assume $a>0$ and $\bar{t}^R_{n-1}>0$, so that one would expect to converge to the positive root $\bar{t}_\text{root}^{R} \geq 0$. Assume that $\epsilon>0$ is chosen (properly) to match the sign of $2a\bar{t}^R_{n-1}$. Ignoring $\delta$, both the numerator and denominator remain strictly positive and convergence to the correct $\bar{t}_\text{root}^{R} \geq 0$ is guaranteed. On the other hand, the roundoff error $\delta$ can change the sign of the numerator when $\bar{t}^R_{n-1}$ and $\tilde{c}$ are small, which occurs when the roots are close together. This emphasizes the need to hybridize Newton's method with bisection in order to guarantee convergence to a desired root.

\sloppy
\textit{
\RemarkCounter{re:}
It is worth briefly discussing the addition of $\epsilon>0$ to $s$ in equation \ref{eq:newtons method}b (and thus equations \ref{eq:lm_newton_with_error} and \ref{eq:lm_newton_with_error2}). Newton's method discretizes $df(t_{n-1}; \vec{p}) = f_{\theta_1}(t_{n-1}; \vec{p}) dt$ with $df \approx 0 - f(t_{n-1}; \vec{p})$ and $dt \approx \Delta t_n = t_n - t_{n-1}$ leading to $f_{\theta_1}(t_{n-1}; \vec{p}) \Delta t_n = - f(t_{n-1}; \vec{p})$. Then, the normal equations
$f_{\theta_1}^T(t_{n-1}; \vec{p}) f_{\theta_1}(t_{n-1}; \vec{p}) \Delta t_n = - f_{\theta_1}^T(t_{n-1}; \vec{p}) f(t_{n-1}; \vec{p})$
reduce to 
$s \Delta t_n = - s f_{\theta_1}^{-1}(t_{n-1}; \vec{p}) f(t_{n-1}; \vec{p})$
since
$f_{\theta_1}^T = s f_{\theta_1}^{-1}$.
Dividing both sides by $s$ gives equation \ref{eq:newtons method}a as expected.
Instead, a Levenberg-Marquardt approach would modify the coefficient matrix to $f_{\theta_1}^T f_{\theta_1} + \epsilon I = (s+\epsilon)I$ illustrating that modifying $s$ to $s + \epsilon$ in equation \ref{eq:newtons method}b to ad hoc remove division by zero is formally equivalent to Levenberg-Marquardt.
}

In equation \ref{eq:newtons method}b, $t^I_{n-1}=0$ implies $t^I_{n}=0$;
in other words, iterates on the real axis are stuck on the real axis.
In addition, $t^R_{n-1}=-\frac{b}{2a}$ implies $t^R_{n}=-\frac{b}{2a}$;
in other words, such iterates can only obtain complex or repeated roots.
Equation \ref{eq:complex newton update 2} illustrates that $\treal_{n}$ becomes $\frac{-b}{2a}$ (and stays there) when $s = 4\tilde{c}$, which implies $\tilde{c}\geq0$ and thus complex or repeated roots; 
similarly, $\timag_n$ becomes zero (and stays there) when $s = -4\tilde{c}$, which implies $\tilde{c}\leq0$ and thus real or repeated roots. 
In both cases, this only happens when it should; 
however, numerical errors, initial guesses, etc.~may lead to $s=4\tilde{c}$ or $s=-4\tilde{c}$ erroneously.

\textit{
\RemarkCounter{re:}
The need to hybridize Newton's method with non-differentiable bisection in order to guarantee convergence to the desired root in the face of roundoff errors highlights the folly of aiming to make iterative solvers differentiable for the sake of backpropagation; moreover, the perturbation required in order to avoid spuriously getting stuck on the $\timag = 0$ and $\treal = \frac{-b}{2a}$ lines makes devising a differentiable iterative solver even more unlikely.
}

\textit{
\RemarkCounter{re:non-diff_decision}
Embracing a bit of non-differentiable decision-making allows for a straightforward approach. When $\tilde{c}<0$, both $\treal \in (-\infty, \frac{-b}{2a})$ and $\treal \in (\frac{-b}{2a}, \infty)$ are safe intervals for hybridizing Newton's method with bisection. When $\tilde{c}>0$, $\timag \in (-\infty, 0)$ and $\timag \in (0, \infty)$ are the safe intervals (with $\treal=\frac{-b}{2a}$).
}
\section{Bisection} 
\label{section:bisection}
\newcommand{\leftbi}{E_0^{\text{left}}}
\newcommand{\rightbi}{E_0^{\text{right}}}
\newcommand{\leftbione}{E_1^{\text{left}}}
\newcommand{\rightbione}{E_1^{\text{right}}}
\newcommand{\z}{H(f(t_0))}
\newcommand{\cvxp}{\eta}

Considering only real roots, a typical implementation of bisection (including derivative information) would proceed as follows. Given endpoints $\leftbi$ and $\rightbi$, the midpoint 
\begin{subequations}
    \label{eq:midpoint}
    \begin{align}
    t_0&=\frac{\leftbi + \rightbi}{2}\\
    \label{eq:midpointb}
    \frac{\partial t_0}{\partial \vec{p}}&=\frac{1}{2}\left(\frac{\partial \leftbi}{\partial \vec{p}} + \frac{\partial \rightbi}{\partial \vec{p}}\right)
    \end{align}
\end{subequations}
is chosen as the initial guess for the root.
Subsequently, the sign of $f(t_0)$ is used to branch the code.
Software infrastructures such as PyTorch \cite{paszke2019pytorch} and TensorFlow \cite{abadi2016tensorflow} compute derivatives by building a computational graph of dependencies, allowing one to ascertain the change in an output parameter with respect to an input parameter. The computational graph only contains branches that are taken, essentially adding a non-differentiable Heaviside function at every code branch.
Assuming $f(\leftbione)<0$ and $f(\rightbione)>0$ leads to 
\begin{subequations}
    \begin{align}
    \leftbione&= \z \leftbi + (1-\z) t_0\\
    \frac{\partial \leftbione}{\partial \vec{p}}
    \label{eq:dEL2}
    &= \left(\frac{1}{2}+\frac{\z}{2}\right)\frac{\partial \leftbi}{\partial \vec{p}} + \left(\frac{1}{2}-\frac{\z}{2}\right) \frac{\partial \rightbi}{\partial \vec{p}}\\
    \rightbione&= \z t_0 + (1-\z)\rightbi\\
    \frac{\partial \rightbione}{\partial \vec{p}}
    \label{eq:dER2}
    &= \frac{\z}{2}\frac{\partial \leftbi}{\partial \vec{p}} + \left(1-\frac{\z}{2}\right) \frac{\partial \rightbi}{\partial \vec{p}}
    \end{align}
\end{subequations}
where equation \ref{eq:midpointb} was used to simplify equations equations \ref{eq:dEL2} and \ref{eq:dER2}, and $H(f(t_0))$ is a piecewise constant Heaviside function with an identically zero derivative (almost everywhere).
Although one might attempt to smooth the Heaviside function by considering both branches, the bisection algorithm cannot be applied to intervals without a sign change. 

Proceeding recursively eventually leads to 
\begin{subequations}
    \begin{align}
    \label{eq:recursiveeqnforward}
	t_n &= \cvxp \leftbi + (1-\cvxp)E_0^{\text{right}}\\
    \label{eq:recursiveeqnbackward}
    \frac{\partial t_n}{\partial \vec{p}} &= \cvxp\frac{\partial \leftbi}{\partial \vec{p}} + (1-\cvxp) \frac{\partial \rightbi}{\partial \vec{p}}
    \end{align}
\end{subequations}
for some $\cvxp \in (0, 1)$;
however, equation \ref{eq:recursiveeqnbackward} is obviously incorrect and should actually be 
\begin{equation}
\label{eq:fullbider}
\frac{\partial t_n}{\partial \vec{p}}
= \cvxp\frac{\partial \leftbi}{\partial \vec{p}} + (1-\cvxp) \frac{\partial \rightbi}{\partial \vec{p}} + (\leftbi - \rightbi) \frac{\partial \cvxp}{\partial \vec{p}}
\end{equation}
based on equation \ref{eq:recursiveeqnforward}.
Along the lines of Remark \ref{re:non-diff_decision}, consider finding $\treal \in (\frac{-b}{2a}, \infty)$.
Substituting $\leftbi = \frac{-b}{2a}$, $\rightbi=K$ for large fixed constant $K$, and exact solution $t_n = \frac{-b+\sqrt{b^2-4ac}}{2a}$ into equation \ref{eq:recursiveeqnforward} leads to
\begin{equation}
\label{eq:cvxp}
\cvxp = 1 - \frac{\sqrt{b^2-4ac}}{b+2aK}   
\end{equation}
which satisfies equation \ref{eq:fullbider} (as expected), but not equation \ref{eq:recursiveeqnbackward} (since $\frac{d \cvxp}{d\vec{p}} \neq \vec{0}$).

\textit{
\RemarkCounter{re:}
Although the inability to execute bisection for non-taken code branches causes the typical software infrastructures to incorrectly obtain the result in equation \ref{eq:recursiveeqnbackward}, it appears that bisection could be made to be formally differentiable via an implicit layer (i.e.~see equations \ref{eq:recursiveeqnforward}, \ref{eq:fullbider}, and \ref{eq:cvxp}).
}

Comparing equations \ref{eq:recursiveeqnbackward} and \ref{eq:fullbider} leads to a strategy that allows the typical software infrastructures to obtain a correct derivative even while using the incorrect equation \ref{eq:recursiveeqnbackward} (we have verified this numerically). Setting
\begin{equation}
\label{eq:bisection-r-endpointnew}
\rightbi = \frac{-b}{2a} + K\frac{\sqrt{b^2-4ac}}{2a}
\end{equation}
with $K>1$ results in $\cvxp = 1-\frac{1}{K}$ and thus $\frac{d \cvxp}{d\vec{p}}=\vec{0}$.

\section{Differentiating Newton's Method}
\label{section:computingderivatives}
Consider any iterative solver where $t_n$ is a function of $t_{n-1}$ and the parameters $\vec{p}$; then, the total derivative can be written as
\begin{subequations}
\label{eq:total_derivativeall}
\begin{align}
\label{eq:total_derivative}
dt_n &= \frac{\partial t_n}{\partial t_{n-1}} dt_{n-1} + \frac{\partial t_n}{\partial \vec{p}} d\vec{p}\\
\label{eq:2d_dt1} 
\begin{bmatrix}
	d\treal_n\\ 
	d\timag_n\\
\end{bmatrix}
&=
\begin{bmatrix}
	\frac{\partial \treal_n}{\partial \treal_{n-1}}&
	\frac{\partial \treal_n}{\partial \timag_{n-1}}\\
	\frac{\partial \timag_n}{\partial \treal_{n-1}}&
	\frac{\partial \timag_n}{\partial \timag_{n-1}}\\
\end{bmatrix}
\begin{bmatrix}
	d\treal_{n-1}\\ 
	d\timag_{n-1}\\
\end{bmatrix}
+
\begin{bmatrix}
	\frac{\partial \treal_n}{\partial \vec{p}}\\
	\frac{\partial \timag_n}{\partial \vec{p}}\\
\end{bmatrix}
d \vec{p}
\end{align}
\end{subequations}
which becomes
\begin{equation}
\label{eq:total_derivative_rewrite} 
dt_n
= \prod_{j=0}^{n-1} \frac{\partial t_{j+1}}{\partial t_{j}} dt_0 + \sum_{i=1}^{n} \left(\prod_{j=i}^{n-1} \frac{\partial t_{j+1}}{\partial t_{j}}\right) \frac{\partial t_i}{\partial \vec{p}} d \vec{p} 
\end{equation}
using recursion.
Focusing on Newton's method from equation \ref{eq:newtons method}, 
\begin{subequations}
\label{eq:2d_dtdtall1} 
\begin{align}
\frac{\partial t_n}{\partial t_{n-1}} &=
\label{eq:2d_dtdt1} 
I
- \left(f_{\theta_1}^{-1}(t_{n-1}; \vec{p})\right)_{\theta_1} f(t_{n-1}; \vec{p}) - f^{-1}_{\theta_1}(t_{n-1}; \vec{p}) f_{\theta_1}(t_{n-1}; \vec{p})\\
&=
\label{eq:2d_dtdt_simplified1} 
- \left(f_{\theta_1}^{-1}(t_{n-1}; \vec{p})\right)_{\theta_1} f(t_{n-1}; \vec{p})
\end{align}
\end{subequations}
\begin{equation}
\label{eq:2d_dtdc1} 
\begin{aligned}
\frac{\partial t_n}{\partial \vec{p}} &=
- \left(f_{\theta_1}^{-1}(t_{n-1}; \vec{p})\right)_{\theta_2} f(t_{n-1}; \vec{p}) - f^{-1}_{\theta_1}(t_{n-1}; \vec{p}) f_{\theta_2}(t_{n-1}; \vec{p})
\end{aligned}
\end{equation}
taking some notational liberties for the sake of brevity. In equations \ref{eq:2d_dtdtall1}b and \ref{eq:2d_dtdc1}, the $\theta_1$ and $\theta_2$ subscripts to the far right of $f_{\theta_1}^{-1}(t_{n-1}; \vec{p})$ indicate replacing every term in the matrix and vector (respectively) with its appropriate row vector Jacobian without changing the dimension of the matrix or vector; then, the matrix-vector multiplications lead to a column of row vectors, which is treated as a matrix (as one would expect via tensor operations).

All of the terms in $\left(f_{\theta_1}^{-1}(t_{n-1}; \vec{p})\right)_{\theta_1}$ and $\left(f_{\theta_1}^{-1}(t_{n-1}; \vec{p})\right)_{\theta_2}$ can be accounted for via
\begin{subequations}
\label{eq:quadratic_terms1} 
\begin{align}
\frac{\partial f_{\theta_1}^{-1}(t_{n-1}; \vec{p})}{\partial \treal_{n-1}} 
&= 
\frac{1}{s}
\begin{bmatrix}
	2 a & 0\\
	0 & 2 a \\
\end{bmatrix}
-
\frac{8 a^2 \treal_{n-1}+ 4ab }{s}
f_{\theta_1}^{-1}(t_{n-1}; \vec{p})\\
\frac{\partial f_{\theta_1}^{-1}(t_{n-1}; \vec{p})}{\partial \timag_{n-1}} 
&= 
\frac{1}{s}
\begin{bmatrix}
	0 & 2 a\\
	-2 a & 0\\
\end{bmatrix}
-
\frac{8 a^2 \timag_{n-1} }{s}
f_{\theta_1}^{-1}(t_{n-1}; \vec{p})\\
\frac{\partial f_{\theta_1}^{-1}(t_{n-1}; \vec{p})}{\partial a}
&= 
\frac{1}{s}
\begin{bmatrix}
	2 \treal_{n-1} & 2 \timag_{n-1}\\
	-2 \timag_{n-1} & 2 \treal_{n-1}\\
\end{bmatrix}
-
\frac{8 a (\treal_{n-1})^2+4b\treal_{n-1} + 8 a (\timag_{n-1})^2}{s}
f_{\theta_1}^{-1}(t_{n-1}; \vec{p})\\
\frac{\partial f_{\theta_1}^{-1}(t_{n-1}; \vec{p})}{\partial b}
&=
\frac{1}{s}
\begin{bmatrix}
	1 & 0\\
	0 & 1\\
\end{bmatrix}
-
\frac{4 a \treal_{n-1}+2b}{s}
f_{\theta_1}^{-1}(t_{n-1}; \vec{p})\\
\frac{\partial f_{\theta_1}^{-1}(t_{n-1}; \vec{p})}{\partial c}  &= 
\begin{bmatrix}
0 & 0 \\	
0 & 0 \\	
\end{bmatrix}
\end{align}
\end{subequations}
where the columns of $\left(f_{\theta_1}^{-1}(t_{n-1}; \vec{p})\right)_{\theta_1}f(t_{n-1}; \vec{p})$ are formed by multiplying equations 34a-b by $f(t_{n-1}; \vec{p})$ and 
the columns of $\left(f_{\theta_1}^{-1}(t_{n-1}; \vec{p})\right)_{\theta_2}f(t_{n-1}; \vec{p})$ are formed by multiplying equations 34c-e by $f(t_{n-1}; \vec{p})$.
As long as $s$ remains bounded away from zero, 
$\left(f_{\theta_1}^{-1}(t_{n-1}; \vec{p})\right)_{\theta_1} f(t_{n-1}; \vec{p}) \to 0$ and $\left(f_{\theta_1}^{-1}(t_{n-1}; \vec{p})\right)_{\theta_2} f(t_{n-1}; \vec{p}) \to 0$ as one converges to a solution where $f(t_{n-1}; \vec{p}) \to 0$; then, equations \ref{eq:2d_dtdtall1} and \ref{eq:2d_dtdc1} behave like
\begin{equation}
\label{eq:dtdtlimit1}	
\frac{\partial t_n}{\partial t_{n-1}} \to \begin{bmatrix} 0 & 0\\0 & 0\\ \end{bmatrix}
\end{equation}
\begin{equation}
\label{eq:dtdclimit1}
\frac{\partial t_n}{\partial \vec{p}} \to - f^{-1}_{\theta_1}(t_{n-1}; \vec{p}) f_{\theta_2}(t_{n-1}; \vec{p})
\end{equation}
leading to 
\begin{equation}
\label{eq:2d_differential_result1} 
\begin{gathered}
dt_n
\to
-f^{-1}_{\theta_1}(t_{n-1}; \vec{p}) f_{\theta_2}(t_{n-1}; \vec{p})
d \vec{p}
\end{gathered}
\end{equation}
when plugged into equation \ref{eq:total_derivativeall}. This is consistent with the implicit differentiation in Section \ref{subsection:quadratic-implicit-differentiation}, i.e.~$d \theta_1 = - f_{\theta_1}^{-1}(t_\text{root}, \vec{p}) f_{\theta_2}(t_\text{root}, \vec{p}) d \theta_2$.

Next, consider the case where $s\to0$ and thus $\alpha = 2 a \treal_{n-1} + b \to 0$ and $\beta = 2 a \timag_{n-1} \to 0$.
If $a=0$, then $b=0$ and thus either everything or nothing is a root;
in such cases, Newton's method takes zero iterations either because the initial guess is considered to be a root or because the derivative is identically equal to zero.
Since backpropagating through Newton's method assuredly fails in this $a=0$ case, assume $a\neq0$
(note that our proposed method in Section \ref{section:quadraticproposedapproach} adequately deals with all degeneracies).
Although the iterative scheme can result in $s = \alpha^2 + \beta^2 = 0$ any time any iterate has $\treal_{n-1} = \frac{-b}{2a}$ and $\timag_{n-1} = 0$, the most problematic case is when one is converging to such a result (i.e.~a repeated root with $\tilde{c}=0$).
As discussed in Section \ref{section:quadratic}, the derivatives have indeterminacies when the repeated root is identically equal to zero;
thus, assume $b\neq0$ in order to avoid such cases.
Using the definitions of $\alpha$, $\beta$, and $\tilde{c}$ leads to
\begin{subequations}
\label{eq:terms_in_alpha_beta_1} 
\begin{align}
f(t_{n-1}; \vec{p}) &= 
\frac{1}{4a}
\begin{bmatrix}
	\alpha^2 - \beta^2\\
	2 \alpha \beta\\
\end{bmatrix}
+
\frac{1}{a}
\begin{bmatrix}
	\tilde{c}\\
	0\\
\end{bmatrix}\\
f_{\theta_1}(t_{n-1}; \vec{p}) &= 
\begin{bmatrix}
	\alpha & -\beta\\
	\beta & \alpha\\
\end{bmatrix}\\
f_{\theta_1}^{-1}(t_{n-1}; \vec{p}) &= 
\frac{1}{\alpha^2 + \beta^2}
\begin{bmatrix}
	\alpha & \beta\\
	-\beta & \alpha\\
\end{bmatrix}
\end{align}
\end{subequations}
and leveraging the inequalities
\begin{subequations}
\label{eq:bounds_1}
\begin{align}
\left| \frac{\alpha^2}{\alpha^2 + \beta^2} \right|
&\leq \left| \frac{\alpha^2}{\alpha^2} \right| = 1\\
\left| \frac{\beta^2}{\alpha^2 + \beta^2} \right|
&\leq \left| \frac{\beta^2}{\beta^2} \right| = 1\\
\left| \frac{\alpha\beta}{\alpha^2 + \beta^2} \right|
&\leq \max\left(\left|\frac{\alpha^2}{\alpha^2 + \beta^2}\right|, \left|\frac{\beta^2}{\alpha^2 + \beta^2} \right|\right) \leq 1
\end{align}
\end{subequations}
allows one to show that $\frac{1}{s} f(t_{n-1}; \vec{p})$ and $\sqrt{s} f_{\theta_1}^{-1}(t_{n-1}; \vec{p})$ are bounded when $\tilde{c}=0$ in equation \ref{eq:terms_in_alpha_beta_1}a; 
thus, $\frac{1}{\sqrt{s}}f_{\theta_1}^{-1}(t_{n-1}; \vec{p}) f(t_{n-1}; \vec{p})$ is bounded and $f_{\theta_1}^{-1}(t_{n-1}; \vec{p}) f(t_{n-1}; \vec{p})\to 0$ even as $s\to0$ (i.e., Newton's method converges even as $s\to0$, see equation \ref{eq:newtons method}a).
Next, consider multiplying the right hand sides of equations \ref{eq:quadratic_terms1}a-e by $f(t_{n-1}; \vec{p})$ in order form the columns of $\left(f_{\theta_1}^{-1}(t_{n-1}; \vec{p})\right)_{\theta_1} f(t_{n-1}; \vec{p})$ and $\left(f_{\theta_1}^{-1}(t_{n-1}; \vec{p})\right)_{\theta_2} f(t_{n-1}; \vec{p})$. Since $\frac{1}{s} f(t_{n-1}; \vec{p})$ and $\frac{1}{\sqrt{s}}f_{\theta_1}^{-1}(t_{n-1}; \vec{p}) f(t_{n-1}; \vec{p})$ are bounded, 
one need only consider $\frac{1}{\sqrt{s}}(8 a^2 \treal_{n-1} + 4ab) = \frac{1}{\sqrt{s}} 4a\alpha$, $\frac{1}{\sqrt{s}}(8 a^2 \timag_{n-1}) = \frac{1}{\sqrt{s}}4a\beta$, $\frac{1}{\sqrt{s}}(8 a (\treal_{n-1})^2 + 4b\treal_{n-1} + 8 a (\timag_{n-1})^2) = \frac{1}{\sqrt{s}}\frac{2}{a}(\alpha^2+\beta^2-b\alpha)$, and $\frac{1}{\sqrt{s}}(4 a \treal_{n-1} + 2b) = \frac{1}{\sqrt{s}}2\alpha$ which are all bounded according to equation \ref{eq:bounds_1}.
Thus, both $\left(f_{\theta_1}^{-1}(t_{n-1}; \vec{p})\right)_{\theta_1} f(t_{n-1}; \vec{p})$ and $\left(f_{\theta_1}^{-1}(t_{n-1}; \vec{p})\right)_{\theta_2} f(t_{n-1}; \vec{p})$ are bounded. Finally, consider $f^{-1}_{\theta_1}(t_{n-1}; \vec{p}) f_{\theta_2}(t_{n-1}; \vec{p})$, which can be rewritten as
\begin{equation}
\label{eq:ftheta1inverseftheta2_rewrite_alpha_beta1}
\frac{1}{s}
\begin{bmatrix}
	\alpha & \beta\\
	-\beta & \alpha\\
\end{bmatrix}
\begin{bmatrix}
	\frac{1}{4a^2} \left(\alpha^2 - 2\alpha b + b^2 - \beta^2 \right) & \frac{1}{2a}(\alpha - b) & 1\\
	\frac{1}{2a^2} \left(\alpha \beta - b \beta\right) & \frac{1}{2a}\beta & 0\\
\end{bmatrix}
\to
\frac{\alpha}{s}
\begin{bmatrix}
	\frac{b^2}{4a^2} & \frac{-b}{2a} & 1\\
	\frac{-b}{2a^2} \beta & \frac{1}{2a}\beta & 0\\
\end{bmatrix}
+
\frac{\beta}{s}
\begin{bmatrix}
	\frac{-b}{2a^2} \beta & \frac{1}{2a}\beta & 0\\
	-\frac{b^2}{4a^2} & \frac{b}{2a} & -1\\
\end{bmatrix}
\end{equation}
where the expression to the right splits the diagonal and off-diagonal components of $f_{\theta_1}^{-1}(t_{n-1}; \vec{p})$ into separate terms;
in addition, all the terms that vanish (when combined with non-vanishing terms) have been eliminated.
The first matrix bears similarity to equation \ref{eq:dtdp1} (when $\beta=0$), and the second matrix bears similarity to equation \ref{eq:dtdp2} (when $\alpha=0$).
As long as $\alpha\to0$ slower then $\beta^2\to0$, the first row in the first matrix blows up;
when $\beta$ is identically equal to zero, Newton's method is operating on real numbers only and only this top row is used.
As long as $\beta\to0$ slower then $\alpha^2\to0$, the second row in the second matrix blows up.
These unbounded terms dominate the bounded $\left(f_{\theta_1}^{-1}(t_{n-1}; \vec{p})\right)_{\theta_1} f(t_{n-1}; \vec{p})$ and $\left(f_{\theta_1}^{-1}(t_{n-1}; \vec{p})\right)_{\theta_2} f(t_{n-1}; \vec{p})$ terms, implying that $\frac{\partial t_n}{\partial t_{n-1}}$ can still be ignored and equations \ref{eq:dtdclimit1} and equation \ref{eq:2d_differential_result1} are still valid.

\textit{
\RemarkCounter{re:shownewtonbackprop}
We have shown that backpropagating through Newton's method results in a vanishing (or ignorable) contribution from the recursive $\frac{\partial t_{n}}{\partial t_{n-1}}$ term when Newton iteration makes sense (i.e.~$a$ and $b$ are not both zero), the derivatives make sense (i.e.~avoiding the indeterminacies of an identically equal to zero repeated root), the roots are distinct (i.e.~$s\neq0$), and the roots are repeated except under very special circumstances (i.e. $\alpha\to0$ as fast or faster than $\beta^2\to0$, and $\beta\to0$ as fast or faster than $\alpha^2\to0$).
}

\textit{
\RemarkCounter{re:}
The convergence of equation \ref{eq:total_derivativeall} to equation \ref{eq:2d_differential_result1} (which we have shown in the vast majority of cases) indicates that the recursive terms should (typically) make no contribution to $dt_n$ as $t_n \to t_\text{root}$.
Strategically, it makes little sense to aim for a robust implementation of backpropagation through Newton's method (including all of the degeneracies) that will at best (typically) do nothing.
}

\subsection{One Parameter Examples}
\label{subsection:computingderivatives-1p-examples}
The three parameter family $\vec{p} = [a, b, c]^T$ can be reduced to a two parameter family $\hat{p} = [\hat{b}, \hat{c}]^T$ by dividing the quadratic equation by $a$, resulting in $\hat{b} = \frac{b}{a}$ and $\hat{c} = \frac{c}{a}$; alternatively, division by $a$ can be avoided using a change of variables $\hat{t} = at$ to obtain $\hat{t}^2 + \hat{b} \hat{t} + \hat{c} = 0$ where $\hat{b} = b$ and $\hat{c} = ac$. The latter approach is used for the examples in this section, although the approaches are equivalent until the mapping between $\hat{p}$ and $\vec{p}$ is considered. The two parameter family $\hat{p}$ can be further reduced to a one parameter family $\tilde{p}$ (represented by a single scalar $\tilde{c}$). This can be accomplished either via the standard approach where $\tilde{t}^R = \treal + \frac{b}{2a}$ and $\tilde{t}^I = \timag$ yield $\tilde{t}^2 + \tilde{c} = 0$ with $\tilde{c} = \frac{-b^2}{4a^2} + \frac{c}{a}$, or via $\tilde{t}^R = a\treal + \frac{b}{2}$ and $\tilde{t}^I = a\timag$ to also obtain $\tilde{t}^2 + \tilde{c} = 0$ but with $\tilde{c} = \frac{-b^2}{4} + ac$ (which is the $\tilde{c}$ used throughout the paper); once again, the examples are indifferent until one considers the mapping between $\tilde{p}$ and $\vec{p}$.

Fixing $\tilde{a}=1$ and $\tilde{b}=0$ results in $d\tilde{a}=d\tilde{b}=0$, and equation \ref{eq:implicitderivative} reduces to 
\begin{equation}
\label{eq:2d_dt_root/dc} 
\begin{bmatrix}
	d\tilde{t}^R_\text{root}\\ 
	d\tilde{t}^I_\text{root}\\
\end{bmatrix}
=
-
\frac{1}{s}
\begin{bmatrix}
	2 \tilde{t}^R_\text{root} & 2 \tilde{t}^I_\text{root}\\
	-2 \tilde{t}^I_\text{root} & 2 \tilde{t}^R_\text{root}\\
\end{bmatrix}
\begin{bmatrix}
	1\\
	0\\
\end{bmatrix}
d\tilde{c}
= 
-\frac{1}{2\left((\tilde{t}^R_\text{root})^2 + (\tilde{t}^I_\text{root})^2\right)}
\begin{bmatrix}
	\tilde{t}^R_\text{root}\\
	-\tilde{t}^I_\text{root}\\
\end{bmatrix}
d\tilde{c}
\end{equation}
for $\tilde{t}^2 + \tilde{c} = 0$. When $\tilde{c}\leq0$, $\tilde{t}^R_\text{root} = \pm \sqrt{-\tilde{c}}$ and $\tilde{t}^I_\text{root} = 0$ reducing equation \ref{eq:2d_dt_root/dc} to
\begin{equation}
\label{eq:dt_root/dc} 
\begin{gathered}
d \tilde{t}^R_\text{root}
= -\frac{1}{2 \tilde{t}^R_\text{root}}d \tilde{c}
= \mp\frac{1}{2 \sqrt{-\tilde{c}}}d \tilde{c}
\end{gathered}
\end{equation}
where $\frac{\partial\tilde{t}^R_\text{root}}{\partial \tilde{c}} \to \mp \infty$ as the root gains multiplicity with $\tilde{c}\to0$.
When $\tilde{c}\geq0$, $\tilde{t}^R_\text{root} = 0$ and $\tilde{t}^I_\text{root} = \pm \sqrt{\tilde{c}}$ leading to $\frac{\partial\tilde{t}^I_\text{root}}{\partial \tilde{c}} \to \pm \infty$ as $\tilde{c}\to0$.
Here, we present results for the $\tilde{t}^R_\text{root} = \sqrt{-\tilde{c}}$ case noting that the results for $\tilde{t}^R_\text{root} = -\sqrt{-\tilde{c}}$ and $\tilde{t}^I_\text{root} = \pm\sqrt{\tilde{c}}$ are similar.
Plugging the final result of Newton iteration into the middle of equation \ref{eq:dt_root/dc} for $\tilde{t}^R_\text{root}$ gives
\begin{equation}
\label{eq:estimate} 
\begin{gathered}
\frac{\partial \tilde{t}^R_\text{root}}{\partial \tilde{c}} \approx
-\frac{1}{2\tilde{t}^R_{n}}
\end{gathered}
\end{equation}
after $n$ iterations of Newton's method.

\begin{figure}[H]
	\centering
	\begin{subfigure}[b]{0.475\textwidth}
		\centering
		\includegraphics[width=\textwidth]{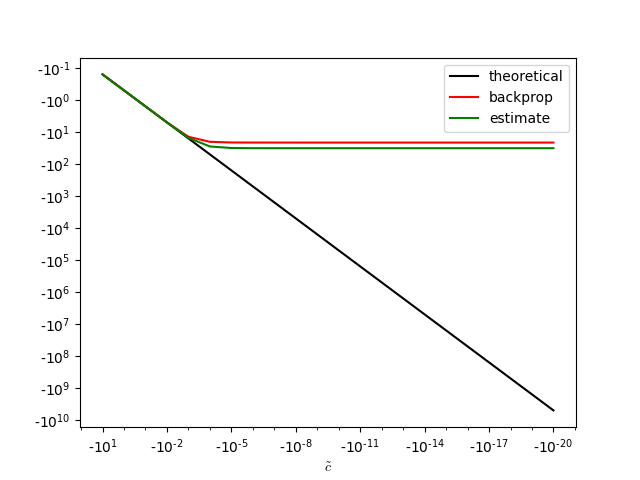}
		\caption{Derivatives (7 Newton iterations)} 
		\label{fig:}
	\end{subfigure}
	\hfill
	\begin{subfigure}[b]{0.475\textwidth}  
		\centering 
		\includegraphics[width=\textwidth]{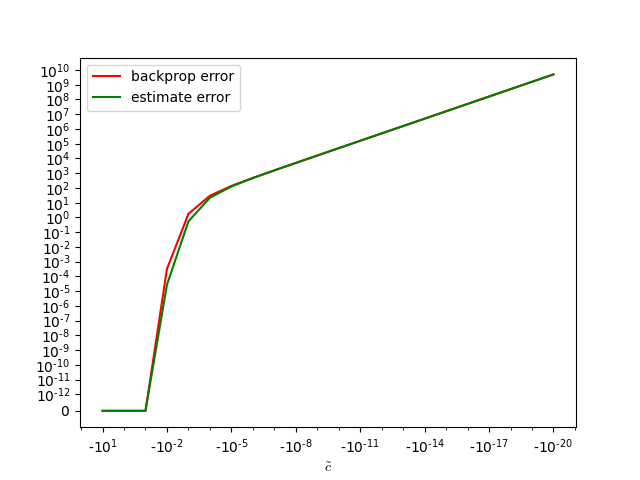}
		\caption{Errors (7 Newton iterations)} 
		\label{fig:}
	\end{subfigure}
	\vskip\baselineskip
	\begin{subfigure}[b]{0.475\textwidth}
		\centering
		\includegraphics[width=\textwidth]{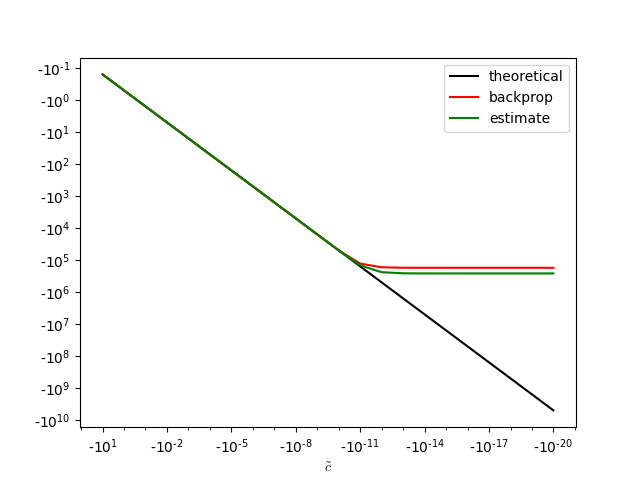}
		\caption{Derivatives (20 Newton iterations)} 
		\label{fig:}
	\end{subfigure}
	\hfill
	\begin{subfigure}[b]{0.475\textwidth}  
		\centering 
		\includegraphics[width=\textwidth]{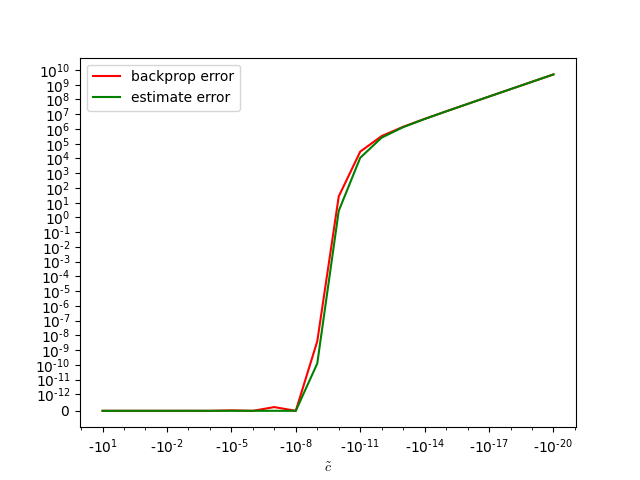}
		\caption{Errors (20 Newton iterations)} 
		\label{fig:}
	\end{subfigure}
	\vskip\baselineskip
	\begin{subfigure}[b]{0.475\textwidth}
		\centering
		\includegraphics[width=\textwidth]{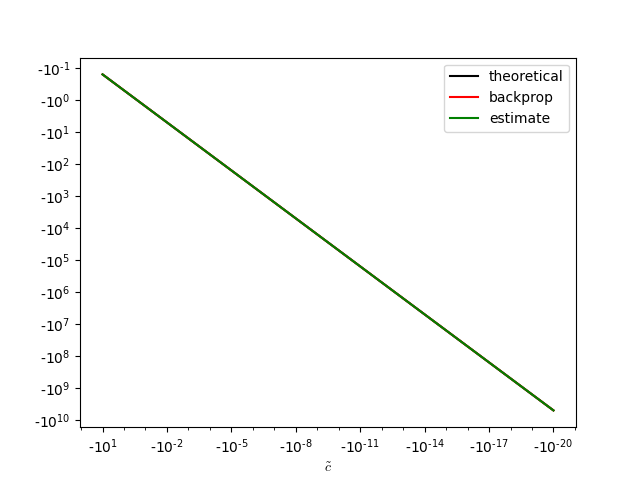}
		\caption{Derivatives (50 Newton iterations)} 
		\label{fig:}
	\end{subfigure}
	\hfill
	\begin{subfigure}[b]{0.475\textwidth}  
		\centering 
		\includegraphics[width=\textwidth]{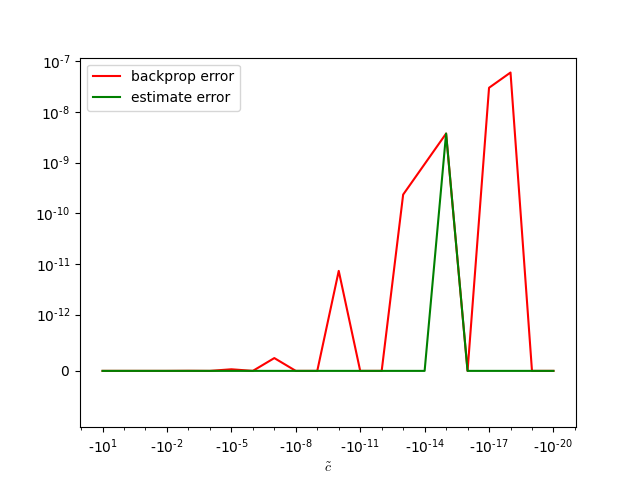}
		\caption{Errors (50 Newton iterations)} 
		\label{fig:}
	\end{subfigure}
	\caption{For $\tilde{t}^2 + \tilde{c}=0$, the $\frac{d \tilde{t}^R_\text{root}}{d \tilde{c}}$ derivatives (as a function of $\tilde{c}$ on a log-log scale) computed using backpropagation (red), the equation \ref{eq:estimate} estimate based on an implicit layer (green), and the theoretical value from equation \ref{eq:dt_root/dc} (black). The errors are computed by comparing to the theoretical value from equation \ref{eq:dt_root/dc}.} 
	\label{fig:newton_grad}
\end{figure}

\newcommand{\smallnum}{\epsilon}
For the sake of a baseline, we implemented backpropagation of Newton's method in Pytorch (and were careful to avoid the various degeneracies discussed earlier in this section).
Figure \ref{fig:newton_grad} shows the results obtained using 7, 20, and 50 Newton iterations. As compared to the theoretical value, one would be hard pressed to argue for the benefits of backpropagation over the estimate in equation \ref{eq:estimate} or vice versa, since both have commensurate errors. Although increasing the number of Newton iterations does eventually give the desired results on smaller and smaller values of $\tilde{c}$, these rather large numbers of Newton iterations would not typically be used by a practitioner. In order to demonstrate that obtaining a reasonable derivative approximation is significantly more difficult than obtaining an accurate root, we experimentally determine the number of Newton iterations required to reduce the relative error to 1\% for each. In order to avoid dividing by zero when computing the relative error for the root, we move the repeated root from $t=0$ to $t=1$ by using $t^2 - 2t + 1 - \smallnum = 0$ as $\smallnum\to0$. Focusing on the root to the right, i.e.~$\treal_\text{root} = 1 + \sqrt{\smallnum}$, equation \ref{eq:implicitderivative} gives $\frac{\partial \treal_\text{root}}{\partial c} = -\frac{1}{2\treal_\text{root} - 2} = -\frac{1}{2\sqrt{\smallnum}}$ which resembles equation \ref{eq:dt_root/dc}. Figure \ref{fig:newton_threshold} shows that the root itself is obtained to 1\% relative accuracy with 7 Newton iterations for varying values of $\smallnum$, even as the repeated root is approached; however, an excessive number of Newton iterations is required to approximate the derivative to an equivalent 1\% relative accuracy. We stress that 1\% relative error in the derivative is a rather large absolute error as compared to the same relative error in the root, since the derivative is approaching infinity while the root is bounded. Achieving commensurate absolute error for the derivative approximation would require even more Newton iterations.

\textit{
\RemarkCounter{re:}
Given all of the degeneracies discussed earlier in this section, as well as the similar accuracies (and inaccuracies) obtained using either the estimate in equation \ref{eq:estimate} or standard backpropagation, we prefer to avoid backpropagation and instead proceed by developing a theoretical approach (resembling equation \ref{eq:estimate}) that can be implemented via an implicit layer.  
}

\textit{
\RemarkCounter{re:}
If one utilizes the best numerical algorithms available (differentiable or not) and subsequently seeks equations to describe that code, then one is empowered to pursue regularizations of the code's governing equations without requiring changes to the algorithms themselves (one may think of this as constitutive/continuum modeling of code). For example, instead of attempting to differentiate a particularly sensitive implementation of an iterative solver such as MINRES \cite{paige1975solution} for a poorly conditioned and/or singular $Ax=b$, one can simply write $x = A^+ b$ and thus $\frac{\partial x}{\partial b}=A^+$ noting that the pseudoinverse contains all the desired derivatives regardless of the algorithm used to solve $Ax = b$; then, one can efficiently/robustly estimate $A^+$ to the desired accuracy using robust PCA/SVD approaches including the power method \cite{heath2018scientific}, Lanczos iteration \cite{lanczos1950iteration}, etc.
}

\begin{figure}[H]
	\centering
	\includegraphics[width=0.7\linewidth]{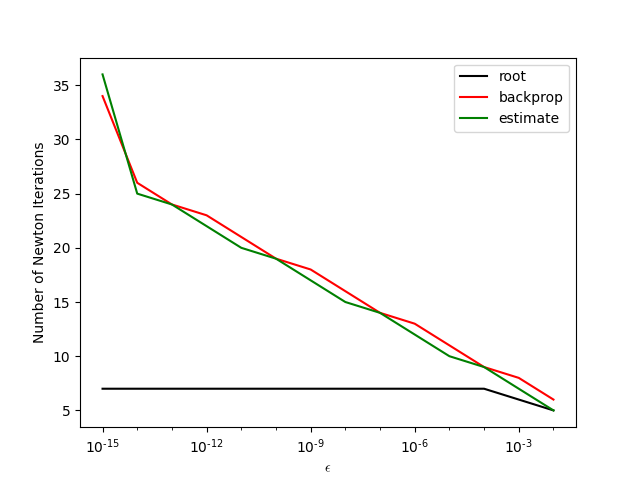}
	\caption{Only seven Newton iterations are required to obtain 1\% relative error for the root (black) for varying values of $\smallnum$ in $t^2 - 2t + 1 - \smallnum=0$. In contrast, both backpropagation (red) and the estimate based on the implicit layer (green) require an increasing number of Newton iterations to reach a 1\% relative error as $\smallnum\to0$.}
	\label{fig:newton_threshold}
\end{figure}

\newpage

\newcommand{\ttarget}{t_\text{root,L}}
\newcommand{\ttargethat}{\hat{t}_\text{root,L}}
\newcommand{\ttargettilde}{\tilde{t}_\text{root,L}}
\subsection{Two Parameter Examples}
\label{subsection:computingderivatives-2p-examples}
Returning to the two parameter quadratic equation from Section \ref{subsection:computingderivatives-1p-examples}, i.e.~$\hat{p}=[\hat{b}~\hat{c}]^T$, fixing $\hat{a}=1$ results in $d\hat{a}=0$ and equation \ref{eq:implicitderivative} reduces to 
\begin{equation}
\label{eq:implicitderivativehat} 
\begin{bmatrix}
d\hat{t}^R_\text{root}\\
d\hat{t}^I_\text{root}\\
\end{bmatrix}
=
-
\frac{1}{s}
\begin{bmatrix}
2 \hat{t}^R_\text{root} + \hat{b} & 2 \hat{t}^I_\text{root}\\
-2 \hat{t}^I_\text{root} & 2 \hat{t}^R_\text{root} + \hat{b}\\
\end{bmatrix}
\begin{bmatrix}
\hat{t}^R_\text{root} & 1\\
\hat{t}^I_\text{root} & 0\\
\end{bmatrix}
\begin{bmatrix}
d\hat{b}\\
d\hat{c}\\
\end{bmatrix}
\end{equation}
for $\hat{t}^2 + \hat{b}\hat{t} + \hat{c} = 0$.
Note that $\tilde{c} = \frac{-b^2}{4} + ac$ can still be used to classify the behavior.
When $\tilde{c}\leq0$, $\hat{t}^R_\text{root} =\frac{-\hat{b}}{2} \pm \sqrt{-\tilde{c}}$ and $\hat{t}^I_\text{root} = 0$ reducing equation \ref{eq:implicitderivativehat} to
\begin{equation}
\label{eq:dtrhatrootdbhatdchat} 
\begin{gathered}
d\hat{t}^R_\text{root}
=
-
\frac{1}{2 \hat{t}^R_\text{root}+\hat{b}}
\begin{bmatrix}
    \hat{t}^R_\text{root} & 1\\
\end{bmatrix}
\begin{bmatrix}
    d\hat{b}\\
    d\hat{c}\\
\end{bmatrix}
=
\mp
\frac{1}{2\sqrt{-\tilde{c}}}
\begin{bmatrix}
    \frac{-\hat{b}}{2} \pm \sqrt{-\tilde{c}} & 1\\
\end{bmatrix}
\begin{bmatrix}
    d\hat{b}\\
    d\hat{c}\\
\end{bmatrix}
\end{gathered}
\end{equation}
where $\frac{\partial \hat{t}^R_\text{root}}{\partial \hat{c}} \to \mp \infty$ as the root gains multiplicity with $\tilde{c}\to0$ (consistent with equation \ref{eq:dt_root/dc});
in addition, $\frac{\partial \hat{t}^R_\text{root}}{\partial \hat{b}} \to \pm \infty$ (depending on the sign of $\hat{b}$) as $\tilde{c}\to0$.
When $\tilde{c}\geq0$, $\hat{t}^R_\text{root} = \frac{-\hat{b}}{2}$ and $\hat{t}^I_\text{root} = \pm \sqrt{\tilde{c}}$ leading to $d\hat{t}^I_\text{root}$ behaving similarly to $d\hat{t}^R_\text{root}$ in equation \ref{eq:dtrhatrootdbhatdchat} (also blowing up as $\tilde{c}\to 0$).
Here, we present results for the $\hat{t}^R_\text{root} =\frac{-\hat{b}}{2} + \sqrt{-\tilde{c}}$ case noting that the results for $\hat{t}^R_\text{root} =\frac{-\hat{b}}{2} - \sqrt{-\tilde{c}}$ and the two complex roots are similar.
Plugging the final result of Newton iteration into the middle of equation \ref{eq:dtrhatrootdbhatdchat} gives
\begin{equation}
\label{eq:complexestimate} 
\begin{gathered}
d\hat{t}^R_\text{root}
\approx
-
\frac{1}{2 \hat{t}^R_n+\hat{b}}
\begin{bmatrix}
    \hat{t}^R_n & 1\\
\end{bmatrix}
\begin{bmatrix}
    d\hat{b}\\
    d\hat{c}\\
\end{bmatrix}
\end{gathered}
\end{equation}
after $n$ iterations of Newton's method.

Considering only real roots (with $\hat{t}^I_n=d\hat{t}^I_n=0$ for all $n$), equation \ref{eq:2d_dt1} reduces to 
\begin{equation}
\begin{gathered}
d\hat{t}^R_n
=
\frac{\partial \hat{t}^R_n}{\partial \hat{t}^R_{n-1}}
d\hat{t}^R_{n-1}
+
\frac{\partial \hat{t}^R_n}{\partial \hat{p}}
d \hat{p}
\end{gathered}
\end{equation}
while equations \ref{eq:quadratic_terms1}a and \ref{eq:quadratic_terms1}d reduce to 
\begin{equation}
\label{eq:quadtermsarewrite}
\frac{\partial f_{\theta_1}^{-1}(t_{n-1}; \hat{p})}{\partial \hat{t}^R_{n-1}}
= 
-
\frac{2}{(2 \hat{t}^R_{n-1}+ \hat{b})^2}
I
\end{equation}
and 
\begin{equation}
\label{eq:ftheta1inversedbhat}
\frac{\partial f_{\theta_1}^{-1}(t_{n-1}; \hat{p})}{\partial \hat{b}}
=
-
\frac{1}{(2\hat{t}^R_{n-1}+\hat{b})^2}I
\end{equation}
respectively.
Multiplying equations \ref{eq:quadtermsarewrite} and \ref{eq:ftheta1inversedbhat} by $f(t_{n-1}; \hat{p})$ leads to 
\begin{equation}
\label{eq:remark3a}
\frac{\partial \hat{t}^R_{n}}{\partial \hat{t}^R_{n-1}}   
=
\frac{2 ((\hat{t}^R_{n-1})^2 + \hat{b}\hat{t}^R_{n-1} + \hat{c})}{(2\hat{t}^R_{n-1} + \hat{b})^2}\\
= \frac{1}{2} + \frac{2\tilde{c}}{(2\hat{t}^R_{n-1} + \hat{b})^2}
\end{equation}
and
\begin{subequations}
\label{eq:remark3}
\begin{align}
\frac{\partial \hat{t}^R_{n}}{\partial \hat{p}}   
&=
\begin{bmatrix}
    \frac{(\hat{t}^R_{n-1})^2 + \hat{b}\hat{t}^R_{n-1} + \hat{c}}{(2\hat{t}^R_{n-1} + \hat{b})^2}
    & 0
\end{bmatrix}
-
\frac{1}{2\hat{t}^R_{n-1}+\hat{b}}
\begin{bmatrix}
    \hat{t}^R_{n-1} & 1
\end{bmatrix}\\
&=
\left(
\frac{1}{2} + \frac{2\tilde{c}}{(2\hat{t}^R_{n-1} + \hat{b})^2}
\right)
\begin{bmatrix}
    \frac{1}{2}
    & 0
\end{bmatrix}
-
\frac{1}{2\hat{t}^R_{n-1}+\hat{b}}
\begin{bmatrix}
    \hat{t}^R_{n-1} & 1
\end{bmatrix}
\end{align}
\end{subequations}
respectively (following the derivations earlier in this section).
Note that 
\begin{equation}
    \label{eq:lhopforbhat}
    \frac{1}{2} + \frac{2\tilde{c}}{(2\hat{t}^R_{n-1} + \hat{b})^2} \to 0
\end{equation}
as $\hat{t}^R_{n-1} \to \hat{t}^R_\text{root} =\frac{-\hat{b}}{2} + \sqrt{-\tilde{c}}$;
thus, 
$\frac{\partial \hat{t}^R_{n}}{\partial \hat{t}^R_{n-1}} \to 0$ and the first term in equation \ref{eq:remark3}b vanishes (leaving only the second term, which is identical to the estimate in equation \ref{eq:complexestimate}) as $\hat{t}^R_{n-1} \to \hat{t}^R_\text{root}$.  

As $\tilde{c}\to0$, indicating closeness to a repeated root, $\hat{t}^R_{n-1} \to \frac{-\hat{b}}{2}$ and L'Hopital's rule is required for equation \ref{eq:lhopforbhat};
in addition, $\frac{1}{2\hat{t}^R_{n-1}+\hat{b}}$ blows up.
To demonstrate this numerically, set $\hat{b}=-2$ and $\hat{c} = 1 + \epsilon$ (which also sets $\tilde{c} = \epsilon$) so that $\hat{t}^R_\text{root} = 1 \pm \epsilon$ is close to being a repeated root.
Table \ref{tab:bnonzero10} shows the results for 10 Newton iterations, and Table \ref{tab:bnonzero100} shows the results for 100 Newton iterations.
The results shown in Table \ref{tab:bnonzero100} substantiate our analysis;
furthermore, Table \ref{tab:bnonzero10} demonstrates the highly erroneous results obtained using only 10 Newton iterations, even though $\hat{t}^R_{10}$ has two significant digits of accuracy.
The results for $\frac{\partial \hat{t}^R_{n}}{\partial \hat{c}}$ were omitted for brevity, but behave as expected.

\begin{table}[H]
\centering
\begin{tabular}{|c|c|c|c|c|c|}
\hline
$\epsilon$&$(\hat{t}^R_\text{root})^+$&$\hat{t}^R_{10}$&$\frac{1}{2} + \frac{2\tilde{c}}{(2\hat{t}^R_{10} + \hat{b})^2}$&$-\frac{\hat{t}^R_{10}}{2\hat{t}^R_{10}+\hat{b}}$&Backprop $\frac{\partial \hat{t}^R_{10}}{\partial \hat{b}}$	 \\ \hline 
-7.203e-09	&	1.000e+00	&	1.009e+00	&	5.000e-01	&	-5.439e+01	&	-3.643e+01	 \\ \hline 
-3.023e-09	&	1.000e+00	&	1.009e+00	&	5.000e-01	&	-5.439e+01	&	-3.643e+01	 \\ \hline 
-4.170e-10	&	1.000e+00	&	1.009e+00	&	5.000e-01	&	-5.439e+01	&	-3.643e+01	 \\ \hline 
-1.468e-11	&	1.000e+00	&	1.009e+00	&	5.000e-01	&	-5.439e+01	&	-3.643e+01	 \\ \hline 
-1.144e-12	&	1.000e+00	&	1.009e+00	&	5.000e-01	&	-5.439e+01	&	-3.643e+01	 \\ \hline 
\end{tabular}
\caption{Even after 10 Newton iterations when $\hat{t}^R_{10}$ has two significant digits of accuracy, columns 4, 5, and 6 all give erroneous values (as compared to the more accurate values in Table \ref{tab:bnonzero100}).}
\label{tab:bnonzero10}
\end{table}

\begin{table}[H]
\centering
\begin{tabular}{|c|c|c|c|c|c|}
\hline
$\epsilon$&$(\hat{t}^R_\text{root})^+$&$\hat{t}^R_{100}$&$\frac{1}{2} + \frac{2\tilde{c}}{(2\hat{t}^R_{100} + \hat{b})^2}$&$-\frac{\hat{t}^R_{100}}{2\hat{t}^R_{100}+\hat{b}}$&Backprop $\frac{\partial \hat{t}^R_{100}}{\partial \hat{b}}$	 \\ \hline 
-7.203e-09	&	1.000e+00	&	1.000e+00	&	5.497e-09	&	-5.892e+03	&	-5.892e+03	 \\ \hline 
-3.023e-09	&	1.000e+00	&	1.000e+00	&	1.632e-08	&	-9.094e+03	&	-9.094e+03	 \\ \hline 
-4.170e-10	&	1.000e+00	&	1.000e+00	&	-1.121e-07	&	-2.449e+04	&	-2.449e+04	 \\ \hline 
-1.468e-11	&	1.000e+00	&	1.000e+00	&	3.689e-06	&	-1.305e+05	&	-1.305e+05	 \\ \hline 
-1.144e-12	&	1.000e+00	&	1.000e+00	&	-3.699e-05	&	-4.675e+05	&	-4.675e+05	 \\ \hline 
\end{tabular}
\caption{After 100 Newton iterations, the entries in column 4 are small and the estimates in column 5 match the results of backpropagation in column 6.}
\label{tab:bnonzero100}
\end{table}

Next, consider the even more problematic $\hat{b}\approx0$ case, and set $\hat{b}=10^{-4}$ and $\hat{c} = \epsilon$ (which also sets $\tilde{c} = \epsilon - 2.5 \times 10^{-9}$) so that $\hat{t}^R_\text{root} = -5\times 10^{-5} \pm \sqrt{2.5 \times 10^{-9}-\epsilon}\approx0$ is close to being a repeated root.
Table \ref{tab:bzero100_1} shows the results after 100 Newton iterations where the root is well-converged, equation \ref{eq:lhopforbhat} is valid, and the estimate from equation \ref{eq:complexestimate} well-matches the results from backpropagation for $\frac{\partial \hat{t}^R}{\partial \hat{c}}$. 
Table \ref{tab:bzero100_2} is a continuation of Table \ref{tab:bzero100_1} (with matching rows) and shows $\gamma = \frac{\hat{b}^2-4\hat{c}}{\hat{b}^2} = \frac{-4\tilde{c}}{\hat{b}^2}$ along with expressions from equation \ref{eq:dtrdp4}.
Note that the second column in Table \ref{tab:bzero100_2} matches the last two columns in Table \ref{tab:bzero100_1} (as expected).
The agreement of equation \ref{eq:dtrdp4}, the equation \ref{eq:complexestimate} estimate, and backpropagation in the last three rows of Table \ref{tab:bzero100_2} numerically validates our discussion of the non-removable singularity.

\textit{
\RemarkCounter{re:}
Simply switching from backpropagation to an implicit layer (or similar use of the implicit function theorem) is not enough to deal with the inherent non-removable singularity.
This makes differentiating the equations instead of the code even more important since understanding the fundamental structure of the equations may be necessary in order to remedy indeterminate derivatives. 
}

\begin{table}[H]
\centering
\begin{tabular}{|c|c|c|c|c|c|c|}
\hline
$\epsilon$&$\tilde{c}$&$(\hat{t}^R_\text{root})^+$&$\hat{t}^R_{100}$&$\frac{1}{2} + \frac{2\tilde{c}}{(2\hat{t}^R_{100} + \hat{b})^2}$	&$-\frac{1}{2\hat{t}^R_{100}+\hat{b}}$&Backprop $\frac{\partial \hat{t}^R_{100}}{\partial \hat{c}}$ \\ \hline 
-1.977e-06	&	-1.979e-06	&	1.357e-03	&	1.357e-03	&	-1.110e-16	 &	-3.554e+02	&	-3.554e+02\\ \hline 
-9.335e-08	&	-9.585e-08	&	2.596e-04	&	2.596e-04	&	5.551e-17	&	-1.615e+03	&	-1.615e+03 \\ \hline 
-2.159e-08	&	-2.409e-08	&	1.052e-04	&	1.052e-04	&	1.110e-16	&	-3.221e+03	&	-3.221e+03 \\ \hline 
3.743e-10	&	-2.126e-09	&	-3.894e-06	&	-3.894e-06	&	-1.110e-16	&	-1.084e+04	&	-1.084e+04 \\ \hline 
2.003e-09	&	-4.970e-10	&	-2.771e-05	&	-2.771e-05	&	3.331e-16 &	-2.243e+04	&	-2.243e+04\\ \hline 
2.316e-09	&	-1.836e-10	&	-3.645e-05	&	-3.645e-05	&	1.221e-15	&	-3.690e+04	&	-3.690e+04 \\ \hline 
2.486e-09	&	-1.367e-11	&	-4.630e-05	&	-4.630e-05	&	5.218e-15	&	-1.352e+05	&	-1.352e+05 \\ \hline 
\end{tabular}
\caption{For a problematically small $\hat{b}=10^{-4}$, each row shows a small $\hat{c}$ leading to a small $\tilde{c}$ indicating that $\hat{t}^R_\text{root}$ is close to being a repeated root. See Table \ref{tab:bzero100_2} for a continuation of the rows.}
\label{tab:bzero100_1}
\end{table}

\begin{table}[H]
\centering
\begin{tabular}{|c|c|c|c|c|}
\hline
$\gamma$&$\frac{-1}{|\hat{b}|\sqrt{\gamma}}$&$\frac{-1}{2}\left(\frac{-1}{\sign{(\hat{b})}\sqrt{\gamma}}+1\right)$&$-\frac{\hat{t}^R_{100}}{2\hat{t}^R_{100}+\hat{b}}$&Backprop $\frac{\partial \hat{t}^R_{100}}{\partial \hat{b}}$	 \\ \hline 
7.917e+02&	-3.554e+02	&	-4.822e-01	&	-4.822e-01	&	-4.822e-01			 \\ \hline 
3.834e+01&	-1.615e+03	&	-4.192e-01	&	-4.192e-01	&	-4.192e-01			 \\ \hline 
9.637e+00&	-3.221e+03	&	-3.389e-01	&	-3.389e-01	&	-3.389e-01			 \\ \hline 
8.503e-01&	-1.084e+04	&	4.223e-02	&	4.223e-02	&	4.223e-02			 \\ \hline 
1.988e-01&	-2.243e+04	&	6.214e-01	&	6.214e-01	&	6.214e-01			 \\ \hline 
7.346e-02&	-3.690e+04	&	1.345e+00	&	1.345e+00	&	1.345e+00			 \\ \hline 
5.469e-03&	-1.352e+05	&	6.261e+00	&	6.261e+00	&	6.261e+00			 \\ \hline 
\end{tabular}
\caption{Equation \ref{eq:dtrdp4} (columns 2 and 5), the equation \ref{eq:complexestimate} estimate (columns 3 and 6), and backpropagation (columns 4 and 7) all agree with each other as well as the non-removable nature of the derivatives with respect to $\hat{b}$ and the large magnitude of the derivatives with respect to $\hat{c}$.}
\label{tab:bzero100_2}
\end{table}

\section{Branch Selection}
\label{section:branchselection}
Correctly identifying the branches of $t_\text{root}^\pm$ is important, since at least one of them will appear in the objective function where it needs to be differentiated in order to obtain a search direction.  
Let $\ttarget$ designate a desired target value for $t_\text{root}$, and consider the one-parameter quadratic equation $\tilde{t}^2+\tilde{c} = 0$.
In this one-parameter quadratic equation, equation \ref{eq:quadratic2d} (top) leads to $\tilde{c} = -(\ttargettilde^R)^2 + (\ttargettilde^I)^2$.
When the target root is real-valued, $\tilde{c} = -(\ttargettilde^R)^2$ leads to two real roots $(\tilde{t}^R_\text{root})^\pm = \pm \sqrt{-\tilde{c}} = \pm |\ttargettilde^R|$; importantly, only one of $(\tilde{t}_\text{root}^R)^\pm$ matches $\ttargettilde^R$.
In other words, one needs to work with $(\tilde{t}^R_\text{root})^+$ when $\ttargettilde^R > 0$ and $(\tilde{t}^R_\text{root})^-$ when $\ttargettilde^R < 0$, while both work when $\ttargettilde^R = 0$.
When the target root is complex-valued, equation \ref{eq:quadratic2d} (bottom) leads to $\ttargettilde^R=0$ and thus $\tilde{c} = (\ttargettilde^I)^2$; then, $(\tilde{t}_\text{root}^I)^\pm = \pm \sqrt{\tilde{c}} = \pm |\ttargettilde^I|$. 
In other words, one needs to work with $(\tilde{t}^I_\text{root})^+$ when $\ttargettilde^I > 0$ and $(\tilde{t}^I_\text{root})^-$ when $\ttargettilde^I < 0$.

\begin{figure}[H]
	\centering
	\includegraphics[width=0.7\linewidth]{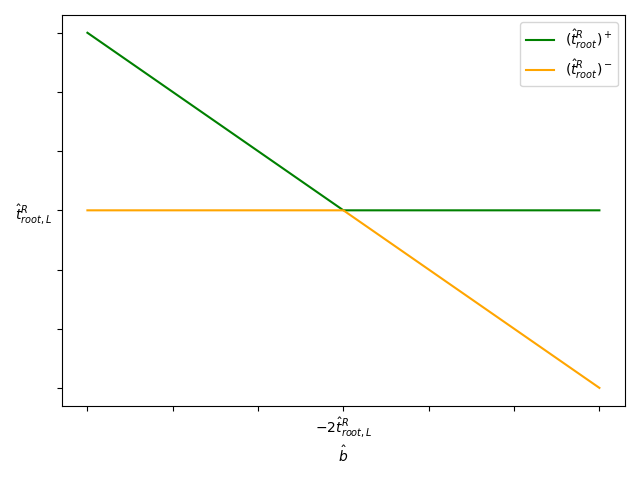}
	\caption{When $\hat{b}\geq -2\ttargethat^R$, $(\hat{t}^R_\text{root})^+ = \ttargethat^R$ and $(\hat{t}^R_\text{root})^- = -\hat{b} - \ttargethat^R$; otherwise, when $\hat{b}\leq -2\ttargethat^R$, $(\hat{t}^R_\text{root})^+ = -\hat{b} - \ttargethat^R$ and $(\hat{t}^R_\text{root})^- = \ttargethat^R$. See equation \ref{eq:troot_ttarget1}.}
	\label{fig:t-target}
\end{figure}

Next, consider the two-parameter quadratic equation $\hat{t}^2 +\hat{b} \hat{t}+\hat{c} = 0$ where equation \ref{eq:quadratic2d} (top) leads to $\hat{c} = -(\ttargethat^R)^2 +(\ttargethat^I)^2 - \ttargethat^R\hat{b}$.
Plugging this into equation \ref{eq:quadraticroots} leads to 
\begin{equation}
\label{eq:troot_ttarget1orig}
\hat{t}_\text{root}^\pm = \frac{-\hat{b} \pm \sqrt{\hat{b}^2+4(\ttargethat^R)^2 +4\ttargethat^R\hat{b}-4(\ttargethat^I)^2 }}{2}
\end{equation}
which becomes
\begin{equation}
\label{eq:troot_ttarget1}
(\hat{t}^R_\text{root})^\pm = \frac{-\hat{b} \pm |\hat{b} + 2 \ttargethat^R|}{2}
\end{equation}
when the target root is real-valued.
When $\hat{b}\geq -2\ttargethat^R$, $(\hat{t}^R_\text{root})^+ = \ttargethat^R$ and $(\hat{t}^R_\text{root})^- = -\hat{b} - \ttargethat^R$; otherwise, when $\hat{b}\leq -2\ttargethat^R$, $(\hat{t}^R_\text{root})^+ = -\hat{b} - \ttargethat^R$ and $(\hat{t}^R_\text{root})^- = \ttargethat^R$. See Figure \ref{fig:t-target}.
Here, the choice of which $(\hat{t}_\text{root}^R)^\pm$ to plug into the objective function depends on the value of $\hat{b}$, which itself has a one parameter set of potential values.
The three-parameter quadratic equation has an additional degree of freedom, since the one-parameter set of values for $\hat{b} = \frac{b}{a}$ comes from a two-parameter set of values for $a$ and $b$.
When the target root is complex-valued, equation \ref{eq:quadratic2d} (bottom) leads to $\ttargethat^R = -\frac{\hat{b}}{2}$, which uniquely determines $\hat{b}$; however, the three-parameter quadratic equation has an additional degree of freedom, since $\hat{b} = \frac{b}{a}$.
Substituting $\ttargethat^R = -\frac{\hat{b}}{2}$ into equation \ref{eq:troot_ttarget1orig} leads to
\begin{equation}
\begin{bmatrix}
	(\hat{t}^R_\text{root})^\pm\\
	(\hat{t}^I_\text{root})^\pm\\
\end{bmatrix}
= 
\begin{bmatrix}
	\frac{-\hat{b}}{2}\\
	\pm \left|\ttargethat^I\right|\\
\end{bmatrix}
\end{equation}
illustrating that branch selection is only needed for $(\hat{t}^I_\text{root})^\pm$.

\textit{
\RemarkCounter{re:}
A real-valued target root has a unique solution for the one-parameter quadratic equation, a one-parameter family of solutions for the two-parameter quadratic equation, and a two-parameter family of solutions for the three-parameter quadratic equation.
The real part of a complex-valued target root is identically zero (with no dependence on the target root) for the one-parameter quadratic equation, has a unique solution for the two-parameter quadratic equation, and has a one-parameter family of solutions for the three-parameter quadratic equation.
The imaginary part of a complex-valued target root always has a unique solution (for all three quadratic equations).
}

\begin{figure}[H]
	\centering
	\begin{subfigure}[b]{0.45\textwidth}
		\centering
		\includegraphics[width=\textwidth]{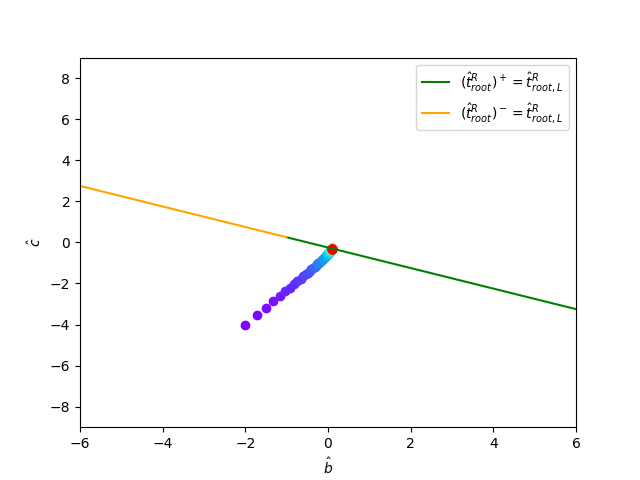}
		\label{fig:example-1a}
	\end{subfigure}
	\begin{subfigure}[b]{0.45\textwidth}
		\centering
		\includegraphics[width=\textwidth]{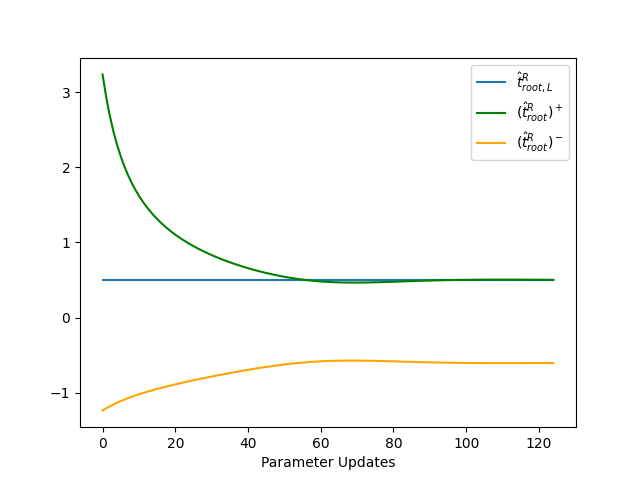}
		\label{fig:example-1b}
	\end{subfigure}
	\caption{(Left) The line $\hat{c} = - \ttargethat^R\hat{b} -(\ttargethat^R)^2$ of valid solutions with green denoting $(\hat{t}_\text{root}^R)^+ = \ttargethat^R$ and yellow denoting $(\hat{t}_\text{root}^R)^- = \ttargethat^R$. Each $(\hat{b}, \hat{c})$ parameter iterate is shown as a separate dot, color-coded from purple to red as iteration proceeds. (Right) The value of the $(\hat{t}_\text{root}^R)^+$ converges to the target value of $\ttargethat^R$ as the optimization proceeds.} 
	\label{fig:example-1}
\end{figure}

\begin{figure}[H]
	\centering
	\begin{subfigure}[b]{0.3\textwidth}
		\centering
		\includegraphics[width=\textwidth]{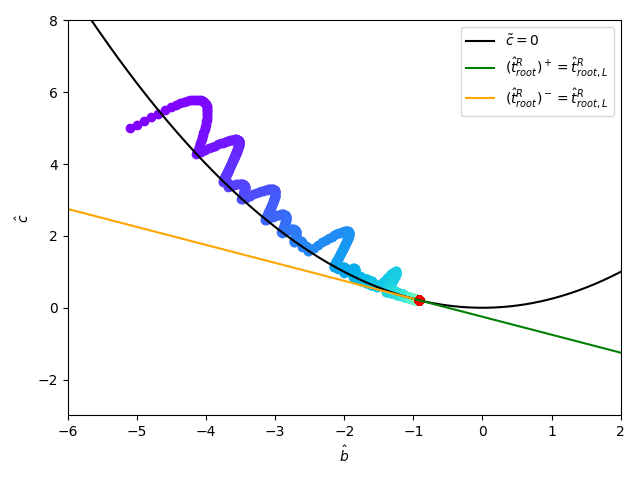}
		\label{fig:example-3a}
	\end{subfigure}
	\begin{subfigure}[b]{0.3\textwidth}
		\centering
		\includegraphics[width=\textwidth]{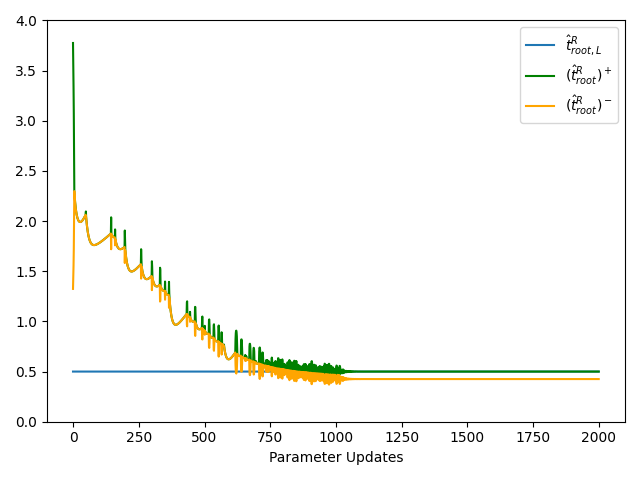}
		\label{fig:example-3b}
	\end{subfigure}
	\begin{subfigure}[b]{0.3\textwidth}
		\centering
		\includegraphics[width=\textwidth]{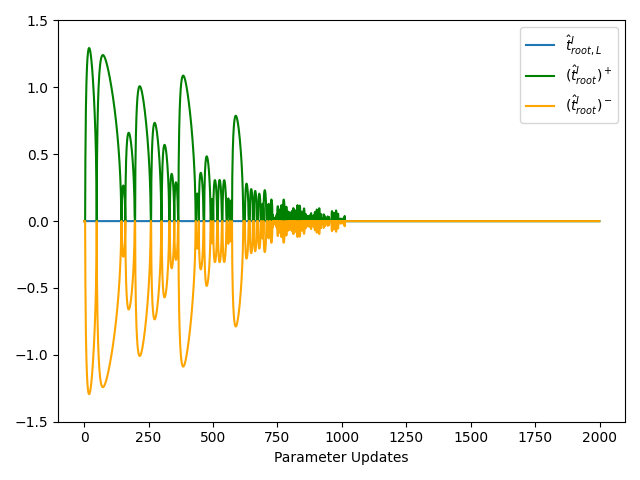}
		\label{fig:example-3c}
	\end{subfigure}
	\caption{(Left) The parameter iterates oscillate into the complex region (with boundary denoted by the parabola $\hat{c} = \frac{\hat{b}^2}{4}$ where $\tilde{c}=0$) as they proceed towards an valid solution on the $\hat{t}_\text{root}^+ = \ttargethat$ green ray. (Middle) The real parts of the roots. (Right) The imaginary part of the roots.
    } 
	\label{fig:example-3}
\end{figure}

\begin{figure}[H]
	\centering
	\begin{subfigure}[b]{0.3\textwidth}
		\centering
		\includegraphics[width=\textwidth]{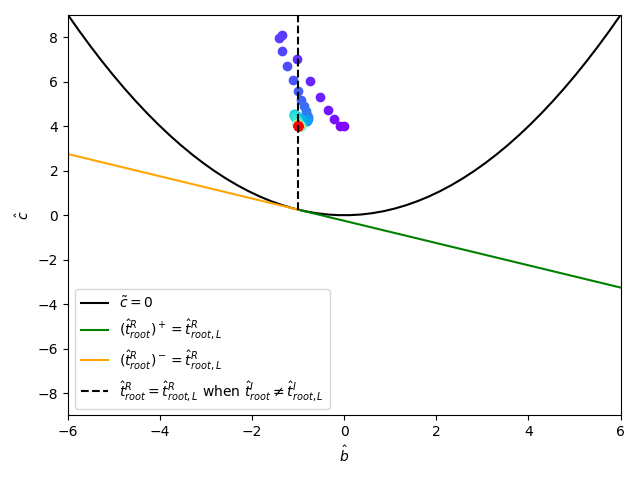}
		\label{fig:}
	\end{subfigure}
	\begin{subfigure}[b]{0.3\textwidth}
		\centering
		\includegraphics[width=\textwidth]{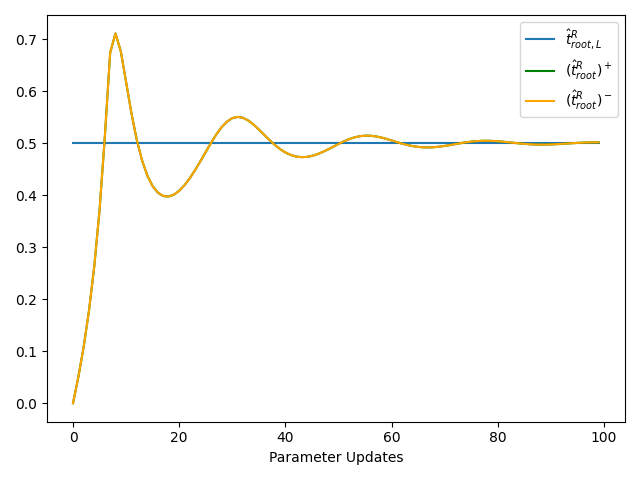}
		\label{fig:}
	\end{subfigure}
	\begin{subfigure}[b]{0.3\textwidth}
		\centering
		\includegraphics[width=\textwidth]{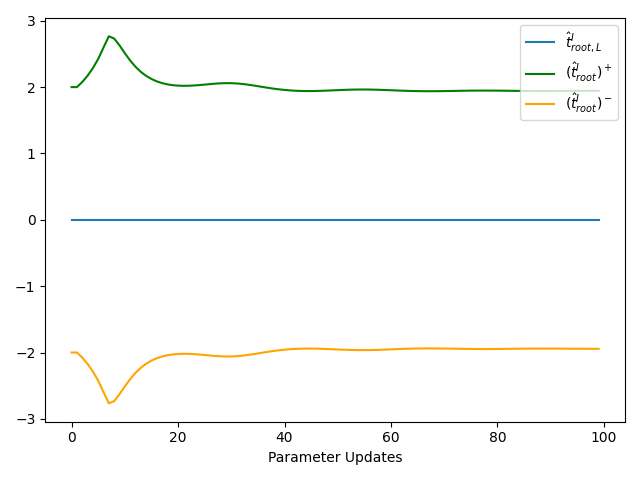}
		\label{fig:}
	\end{subfigure}
	\caption{(Left) One can erroneously converge to the $\hat{t}_\text{root}^R = \ttargethat^R$ ray in the complex region, if $\ttargethat^I$ is not explicitly set to zero in the objective function. (Middle) In the complex region, $(\hat{t}_\text{root}^R)^\pm$ coincide. (Right) In the complex region, $(\hat{t}_\text{root}^I)^\pm$ are equal and opposite.} 
	\label{fig:3.5a}
\end{figure}

Figure \ref{fig:example-1} plots the one-parameter family of solutions $(\hat{b}, \hat{c})$ corresponding to a target root $\ttargethat^R = \frac{1}{2}$, and shows the results obtained minimizing
\begin{equation}
\label{eq:lossagain} 
L(\hat{p}) =
\frac{1}{2}
||\hat{t}_\text{root}(\hat{p}) - \hat{t}_\text{root,L}||_2^2
\end{equation}
using Adam \cite{kingma2014adam} optimization with backpropagation (via PyTorch \cite{paszke2019pytorch}) starting from an initial guess of $(\hat{b}_0, \hat{c}_0) = (-2, -4)$. Since $(\hat{t}_\text{root}^R)^+$ is used in the objective function, the iteration converges to a point on the green ray (note that all points on the green ray are valid solutions). In Figure \ref{fig:example-3}, the initial guess is modified to $(\hat{b}_0, \hat{c}_0) = (-5.1, 5)$ so that the parameter iterates enter the complex region on their way to the green ray. This highlights the fact that one needs to consider complex roots in both the Newton iteration and the parameter optimization, even in the case where both the initial guess and the final solutions are real-valued. Moreover, one needs to explicitly set the imaginary part of the root to zero in the objective function, e.g.~
\begin{equation}
\label{eq:lossagain2d} 
L(\hat{p}) =
\frac{1}{2}
\left|\left|
\begin{bmatrix}
\hat{t}^R_\text{root}(\hat{p})\\  
\hat{t}^I_\text{root}(\hat{p})\\  
\end{bmatrix}
-
\begin{bmatrix}
\frac{1}{2} \\  
0 \\  
\end{bmatrix}
\right|\right|_2^2
,
\end{equation}
in order to enforce convergence to real-valued roots. Figure \ref{fig:3.5a} instead minimizes
\begin{equation}
\label{eq:badobjectivefunction} 
L(\hat{p}) =
\frac{1}{2}
\left|\left|
\hat{t}^R_\text{root}(\hat{p}) - \frac{1}{2} 
\right|\right|_2^2
\end{equation}
demonstrating that the optimization can otherwise converge to a complex-valued root with a real part matching the desired $\ttargethat^R$.
Finally, Figure \ref{fig:example19} demonstrates what happens when one switches from using $\hat{t}^+_\text{root}$ in the objective function to instead using $\hat{t}^-_\text{root}$ in the objective function during parameter optimization.


\foreach \subexample/\figcaption/\phasediagramnum in {normal/Figure \ref{fig:example-3}/1}{
\begin{figure}[H]
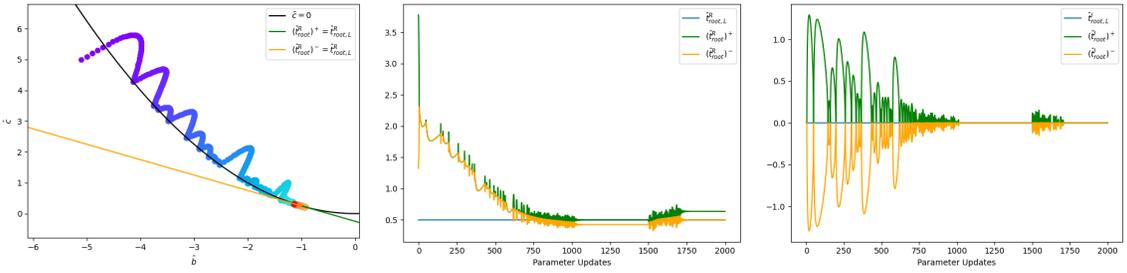

	\centering
    \foreach \deriv in {switch} {
	\begin{subfigure}[b]{0.3\textwidth}
		\centering
		\includegraphics[width=\textwidth]{figures/quadratic-section/smoothing-subsection/examples_3_parameter/example19/\deriv_\subexample_phase_diagram_\phasediagramnum.png}
		\label{fig:}
	\end{subfigure}
	\begin{subfigure}[b]{0.3\textwidth}
		\centering
		\includegraphics[width=\textwidth]{figures/quadratic-section/smoothing-subsection/examples_3_parameter/example19/\deriv_\subexample_roots_real.png}
		\label{fig:}
	\end{subfigure}
	\begin{subfigure}[b]{0.3\textwidth}
		\centering
		\includegraphics[width=\textwidth]{figures/quadratic-section/smoothing-subsection/examples_3_parameter/example19/\deriv_\subexample_roots_imag.png}
		\label{fig:}
	\end{subfigure}
    }
	\caption{
    For the first 1500 iterations, $\hat{t}^+_\text{root}$ is used in the objective function; afterwards, $\hat{t}^-_\text{root}$ is used instead. Note how the iterates move from the green to the yellow ray (in the left figure), and how $(\hat{t}^R_\text{root})^-$ converges to $\ttargethat$ after 1500 iterations (in the middle figure).
    }
	\label{fig:example19}
\end{figure}
}

\newpage
\section{Difficulties Near Repeated Roots}
\label{section:diffnearrepeatedroots}
\begin{figure}[H]
	\centering
	\begin{subfigure}[b]{0.45\textwidth}
		\centering
		\includegraphics[width=\textwidth]{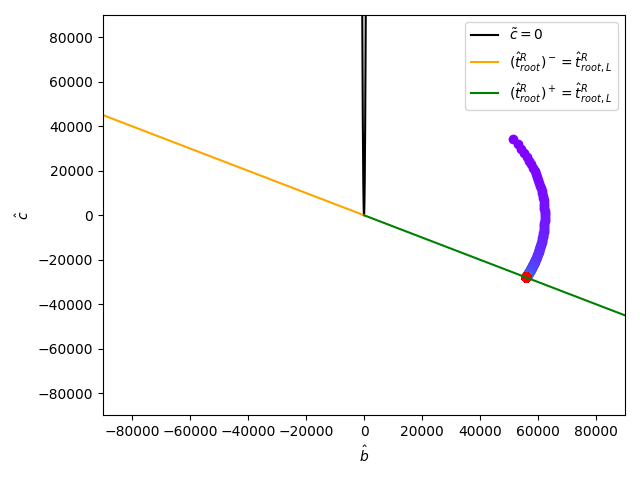}
		\label{fig:}
	\end{subfigure}
	\begin{subfigure}[b]{0.45\textwidth}
		\centering
		\includegraphics[width=\textwidth]{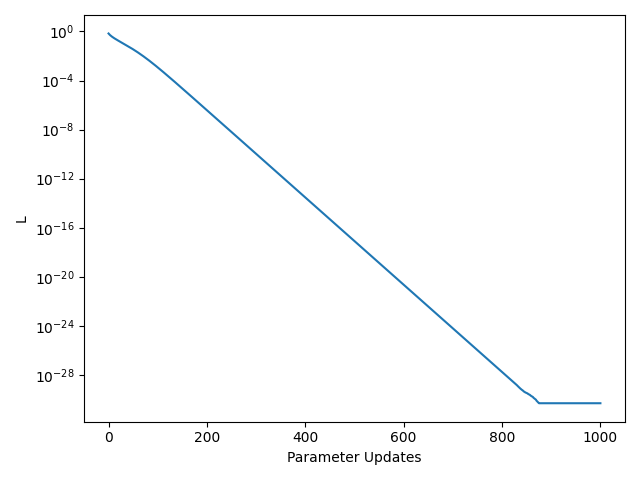}
		\label{fig:}
	\end{subfigure}
	\caption{Starting from iteration 1 in Table \ref{tab:exp4}, increasing the step size of SGD to $10^8$ allows it to converge to the green ray (in hundreds of iterations).} 
	\label{fig:convergences_sgd}
\end{figure}

\begin{figure}[H]
	\centering
	\begin{subfigure}[b]{0.45\textwidth}
		\centering
		\includegraphics[width=\textwidth]{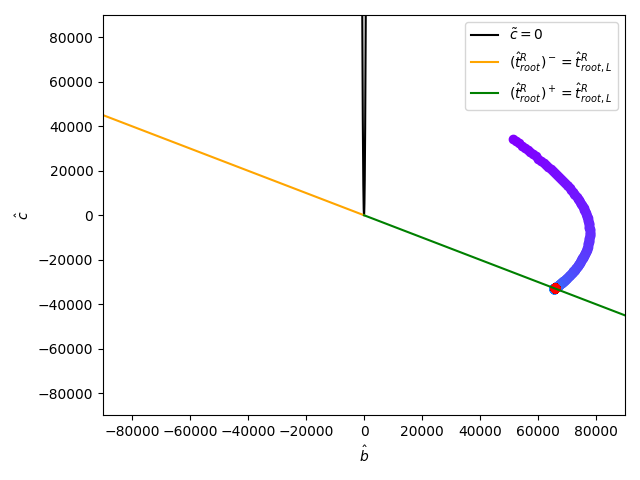}
		\label{fig:}
	\end{subfigure}
	\begin{subfigure}[b]{0.45\textwidth}
		\centering
		\includegraphics[width=\textwidth]{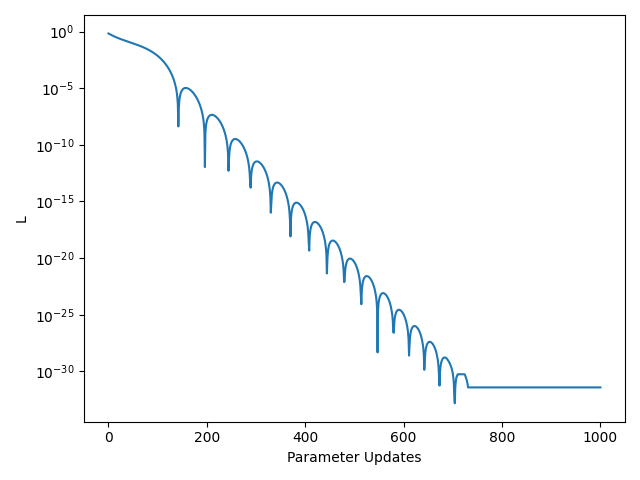}
		\label{fig:}
	\end{subfigure}
	\caption{Switching to Adam after the first iteration in Table \ref{tab:exp4} and increasing the step size to $10^3$ enables convergence to the green ray (in hundreds of iterations).} 
	\label{fig:convergences_adam}
\end{figure}

Many problems of interest will aim to either create or avoid a collision, and thus necessarily spend time iteratively wading back and forth between real and complex roots near $\tilde{c} = 0$, i.e.~near coalescence to a repeated root. From equation \ref{eq:lossagain},
\begin{equation}
\label{eq:dldphat}
\frac{\partial L}{\partial \hat{p}}
= \frac{\partial L}{\partial \hat{t}_\text{root}} \frac{\partial \hat{t}_\text{root}}{\partial \hat{p}}
= 
\begin{bmatrix}
	\hat{t}_\text{root}^R(\hat{p}) - \ttargethat^R &
	\hat{t}_\text{root}^I(\hat{p}) - \ttargethat^I\\
\end{bmatrix}
\frac{\partial \hat{t}_\text{root}}{\partial \hat{p}}
\end{equation}
where, from equation \ref{eq:implicitderivativehat}, one can write
\begin{subequations}
\label{eq:dthatdpcases}
\begin{align}
\frac{\partial \hat{t}_\text{root}}{\partial \hat{p}}
&=
\frac{-1}{\pm2\sqrt{-\tilde{c}}}
\begin{bmatrix}
-\frac{\hat{b}}{2}
&
1\\
0&0\\
\end{bmatrix}
-
\begin{bmatrix}
\frac{1}{2} & 0\\
0 & 0\\
\end{bmatrix}
&& \text{if }\tilde{c}<0\\
\frac{\partial \hat{t}_\text{root}}{\partial \hat{p}}
&=
\frac{1}{\pm2\sqrt{\tilde{c}}}
\begin{bmatrix}
0&0\\
-\frac{\hat{b}}{2}
&
1\\
\end{bmatrix}
-
\begin{bmatrix}
\frac{1}{2} & 0\\
0 & 0\\
\end{bmatrix}
&& \text{if }\tilde{c}>0
\end{align}
\end{subequations}
noting that $\tilde{c}<0$ has $\hat{t}^R_\text{root} = -\frac{\hat{b}}{2} \pm \sqrt{-\tilde{c}}$ and $\hat{t}^I_\text{root} = 0$ while $\tilde{c}>0$ has $\hat{t}^R_\text{root} = -\frac{\hat{b}}{2}$ and $\hat{t}^I_\text{root} = \pm \sqrt{\tilde{c}}$. Starting with an initial guess of $(\hat{b}_0, \hat{c}_0)=(-3,\frac{9}{4} - 10^{-12})$, so that $\tilde{c} = -\frac{\hat{b}^2}{4} + \hat{c} = -10^{-12}$ indicates closeness to a repeated root, Table \ref{tab:exp4} shows that gradient descent optimization (SGD) with backpropagation (via PyTorch) takes an erroneously large step in ($\hat{b}, \hat{c}$) because of the large values of $\frac{\partial L}{\partial \hat{b}}$ and $\frac{\partial L}{\partial \hat{c}}$. Afterwards, $\frac{\partial L}{\partial \hat{b}}$ and $\frac{\partial L}{\partial \hat{c}}$ are about $10^{9}$ times smaller than $\hat{b}$ and $\hat{c}$, and SGD is unable to make any significant progress towards the target root $\ttargethat^R = \frac{1}{2}$.
Both of these issues can be understood from equation \ref{eq:dthatdpcases}a. $\frac{\partial \hat{t}^R_\text{root}}{\partial \hat{c}} = \frac{-1}{\pm2\sqrt{-\tilde{c}}}$ blows up when $\tilde{c}$ is small and vanishes when $\tilde{c}$ is large.
$\frac{\partial \hat{t}^R_\text{root}}{\partial \hat{b}} = \frac{\hat{b}}{\pm4\sqrt{-\tilde{c}}} - \frac{1}{2}$ blows up when $\tilde{c}$ is small compared to $\hat{b}$; in addition, when $\hat{b}^2 >> \hat{c}$, $\tilde{c} \approx -\frac{\hat{b}^2}{4}$ and thus $\frac{\partial \hat{t}^R_\text{root}}{\partial \hat{b}}\approx\pm\frac{1}{2} \frac{\hat{b}}{|\hat{b}|} - \frac{1}{2}$ which is approximately zero for $(\hat{t}^R_\text{root})^+$ which is used in the objective function. Increasing the step size to a rather large value of $10^8$ allows SGD to recover and converge after hundreds of iterations, albeit to rather large values for ($\hat{b}, \hat{c}$) as shown in Figure \ref{fig:convergences_sgd}. For the sake of comparison, switching to Adam after the first iteration only required increasing the step size to $10^3$ (still far too large) in order to recover within a similar number of iterations (see Figure \ref{fig:convergences_adam}).

\begin{table}[H]
\centering
\begin{tabular}{|c|c|c|c|c|c|c|c|}
\hline
Iterations&$\hat{b}$&$\hat{c}$&$\tilde{c}$&$(\hat{t}^R_\text{root})^+$&$L$&$\frac{\partial L}{\partial \hat{b}}$&$\frac{\partial L}{\partial \hat{c}}$	 \\ \hline 
0	&	-3.000e+00	&	2.250e+00	&	-1.000e-12	&	1.500e+00	&	5.000e-01	&	-5.146e+05	&	-3.431e+05	 \\ \hline 
1	&	5.146e+04	&	3.431e+04	&	-6.619e+08	&	-6.668e-01	&	6.807e-01	&	-1.512e-05	&	2.268e-05	 \\ \hline 
2	&	5.146e+04	&	3.431e+04	&	-6.619e+08	&	-6.668e-01	&	6.807e-01	&	-1.512e-05	&	2.268e-05	 \\ \hline 
100000	&	5.146e+04	&	3.431e+04	&	-6.619e+08	&	-6.668e-01	&	6.807e-01	&	-1.512e-05	&	2.268e-05	 \\ \hline 
\end{tabular}
\caption{SGD takes an erroneously large step in ($\hat{b}, \hat{c}$) because of the large values of $\frac{\partial L}{\partial \hat{b}}$ and $\frac{\partial L}{\partial \hat{c}}$ close to the repeated root; subsequently, it struggles to recover.}
\label{tab:exp4}
\end{table}

\begin{figure}[H]
	\centering
	\begin{subfigure}[b]{0.45\textwidth}
		\centering
		\includegraphics[width=\textwidth]{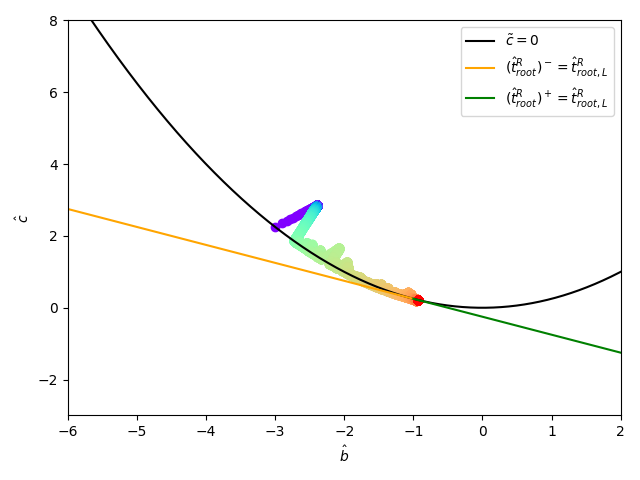}
		\label{fig:}
	\end{subfigure}
	\begin{subfigure}[b]{0.45\textwidth}
		\centering
		\includegraphics[width=\textwidth]{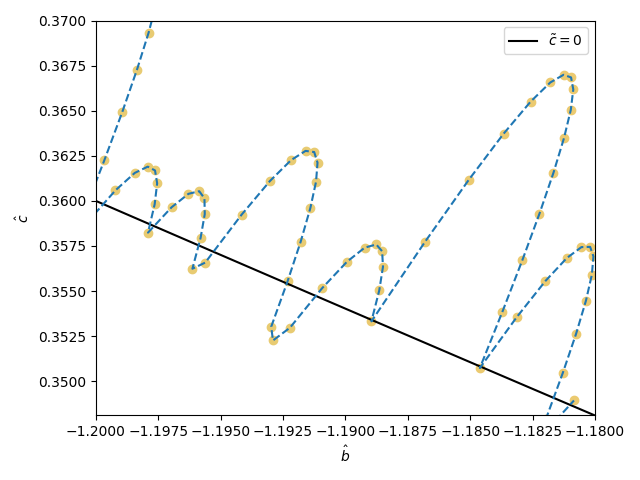}
		\label{fig:}
	\end{subfigure}
	\begin{subfigure}[b]{0.45\textwidth}
		\centering
		\includegraphics[width=\textwidth]{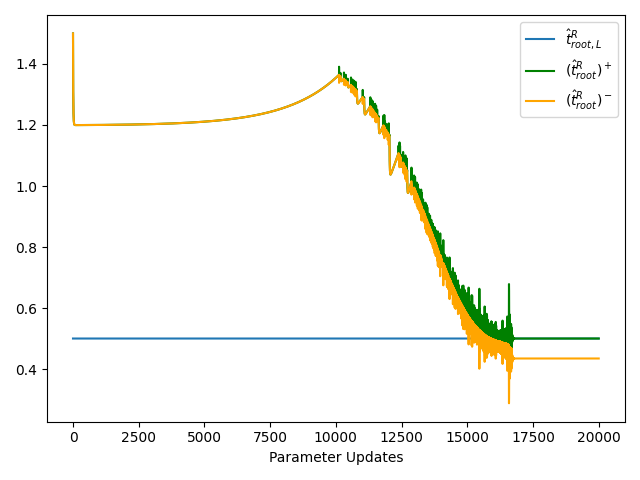}
		\label{fig:}
	\end{subfigure}
	\begin{subfigure}[b]{0.45\textwidth}
		\centering
		\includegraphics[width=\textwidth]{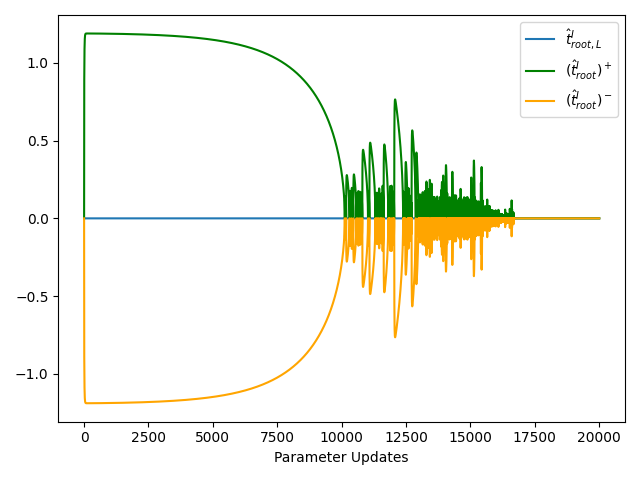}
		\label{fig:}
	\end{subfigure}
	\caption{Although Adam performs better than SGD, i.e.~not initially taking an erroneously large step in ($\hat{b}, \hat{c}$) because of the large values of $\frac{\partial L}{\partial \hat{b}}$ and $\frac{\partial L}{\partial \hat{c}}$ close to the repeated root case, it suffers from both oscillatory behavior and slow convergence. The top right subfigure shows a zoomed in view of the oscillatory behavior near the $\tilde{c}=0$ parabola.} 
	\label{fig:example-5a}
\end{figure}

\begin{figure}[H]
	\centering
	\begin{subfigure}[b]{0.45\textwidth}
		\centering
		\includegraphics[width=\textwidth]{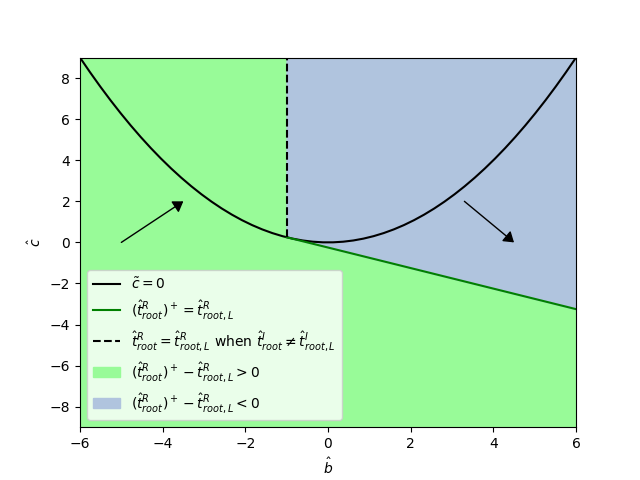}
		\label{fig:}
	\end{subfigure}
	\begin{subfigure}[b]{0.45\textwidth}
		\centering
		\includegraphics[width=\textwidth]{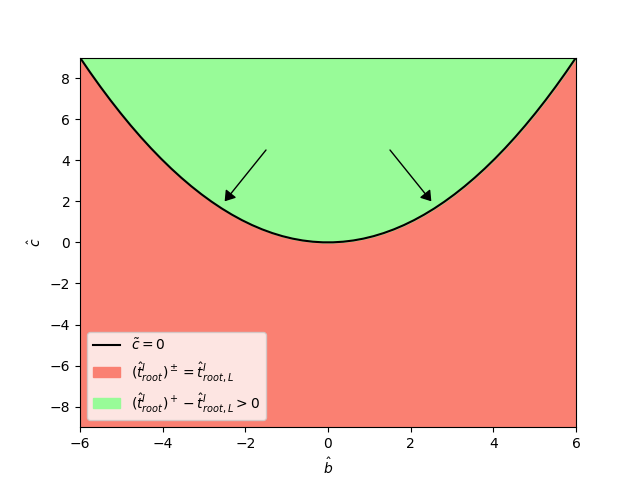}
		\label{fig:}
	\end{subfigure}
	\caption{
    These figures address using $\hat{t}_\text{root}^+$ in the objective function (the figures for using $\hat{t}_\text{root}^-$ illustrate similar behavior).
    When the roots are real (below the $\tilde{c}=0$ parabola), equation \ref{eq:grad_in_dldp}a indicates that a small $\tilde{c}$ makes the descent direction $-\frac{\partial L}{\partial \hat{p}}$ point in the same direction as $\frac{\partial \tilde{c}}{\partial \hat{p}}$ when $\frac{\partial L}{\partial \hat{t}^R_\text{root}}$ is positive (the green region) and opposite $\frac{\partial \tilde{c}}{\partial \hat{p}}$ when $\frac{\partial L}{\partial \hat{t}^R_\text{root}}$ is negative (the blue region). This is illustrated by the arrows in the left subfigure. When the roots are complex (above the $\tilde{c}=0$ parabola), equation \ref{eq:grad_in_dldp}b illustrates that a small $\tilde{c}$ makes $-\frac{\partial L}{\partial \hat{p}}$ point opposite $\frac{\partial \tilde{c}}{\partial \hat{p}}$ when $\frac{\partial L}{\partial \hat{t}^I_\text{root}}$ is positive (the green region). This is illustrated by the arrows in the right subfigure. See Remark \ref{re:specificform}.
    }
	\label{fig:phase_color_1}
\end{figure}

\begin{figure}[H]
	\centering
	\begin{subfigure}[b]{0.45\textwidth}
		\centering
		\includegraphics[width=\textwidth]{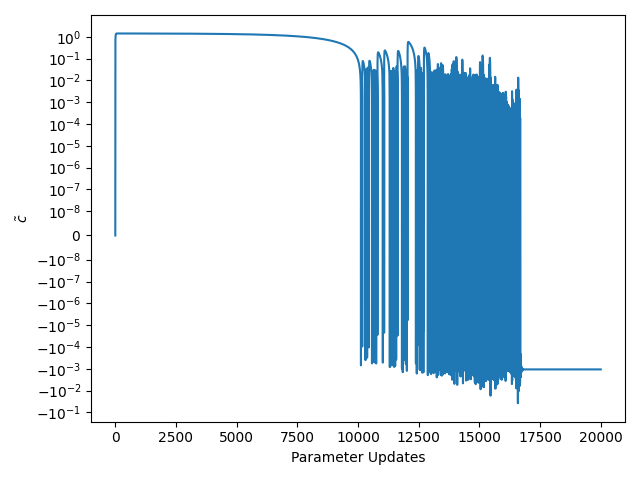}
		\label{fig:}
	\end{subfigure}
	\begin{subfigure}[b]{0.45\textwidth}
		\centering
		\includegraphics[width=\textwidth]{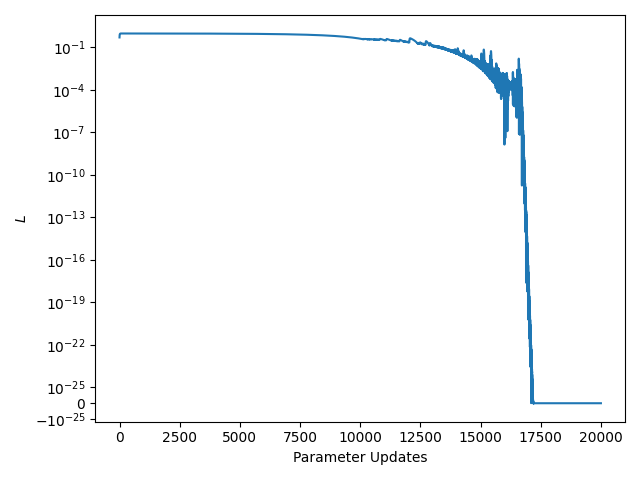}
		\label{fig:}
	\end{subfigure}
	\begin{subfigure}[b]{0.45\textwidth}
		\centering
		\includegraphics[width=\textwidth]{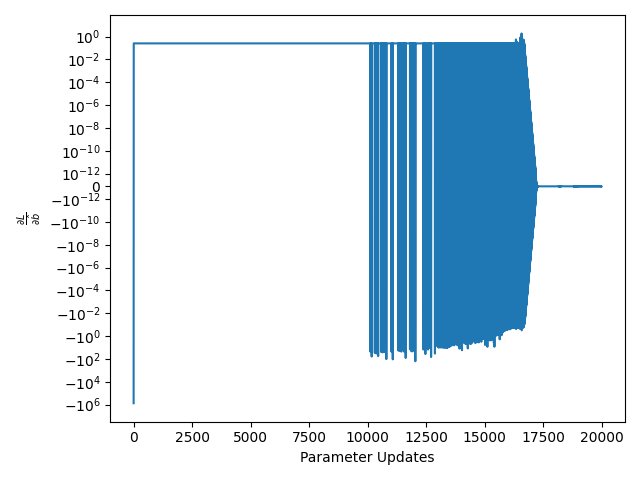}
		\label{fig:}
	\end{subfigure}
	\begin{subfigure}[b]{0.45\textwidth}
		\centering
		\includegraphics[width=\textwidth]{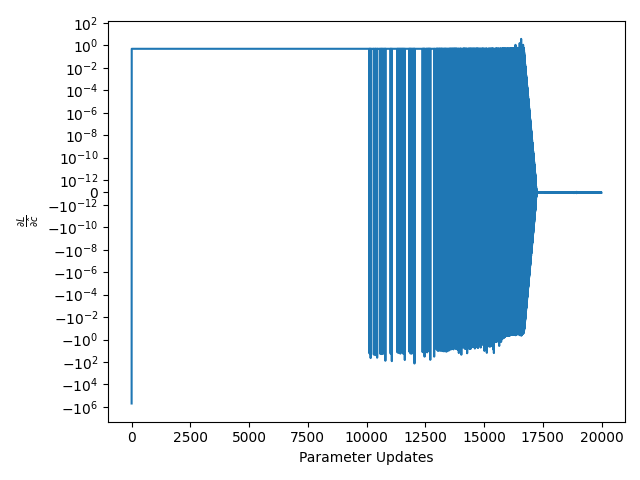}
		\label{fig:}
	\end{subfigure}
	\caption{
    After entering the blue region in the left subfigure of Figure \ref{fig:phase_color_1}, the iterates are no longer attracted to the $\tilde{c}=0$ parabola and can then converge to valid solutions (on the green ray).
    } 
	\label{fig:example-5b}
\end{figure}

Repeating the aforementioned example using Adam (instead of SGD) alleviates issues with erroneously jumping to large values of ($\hat{b},\hat{c}$); however, other issues lead to rather slow convergence (thousands of iterations). A representative example is shown in Figure \ref{fig:example-5a}.
Combining equations \ref{eq:dldphat} and \ref{eq:dthatdpcases} leads to
\begin{subequations}
\label{eq:grad_in_dldp}
\begin{align}
\frac{\partial L}{\partial \hat{p}}
&=
\frac{\partial L}{\partial \hat{t}^R_\text{root}}
\left(
\frac{-1}{\pm2\sqrt{-\tilde{c}}}
\frac{\partial \tilde{c}}{\partial \hat{p}}
-
\begin{bmatrix}
\frac{1}{2} & 0\\
\end{bmatrix}
\right)
&& \text{if }\tilde{c}<0\\
\frac{\partial L}{\partial \hat{p}}
&=
\frac{\partial L}{\partial \hat{t}^R_\text{root}}
\begin{bmatrix}
-\frac{1}{2} & 0\\
\end{bmatrix}
+
\frac{\partial L}{\partial \hat{t}^I_\text{root}}
\frac{1}{\pm2\sqrt{\tilde{c}}}
\frac{\partial \tilde{c}}{\partial \hat{p}}
&& \text{if }\tilde{c}>0
\end{align}
\end{subequations}
where $\frac{\partial \tilde{c}}{\partial \hat{p}} = [-\frac{\hat{b}}{2}~1]$ since $\tilde{c} = -\frac{\hat{b}^2}{4} + \hat{c}$. Equation \ref{eq:grad_in_dldp} illustrates that small values of $\tilde{c}$ cause $\frac{\partial L}{\partial \hat{p}}$ to align with $\frac{\partial \tilde{c}}{\partial \hat{p}}$ as illustrated by the black arrows in Figure \ref{fig:phase_color_1}. Initially, as shown in Figure \ref{fig:example-5a}, the iterates start out strongly attracted to the $\tilde{c}=0$ parabola where the derivatives tend to blow up and the descent direction is fairly orthogonal to the preferred direction (tangent to the $\tilde{c}=0$ parabola) for making progress towards valid solutions on the green ray. Later (after entering the blue region of Figure \ref{fig:phase_color_1}), the iterates are no longer attracted to the $\tilde{c}=0$ parabola allowing them to settle down and converge to a valid solution (on the green ray). See Figure \ref{fig:example-5b}.

\textit{
\RemarkCounter{re:specificform}
When small values of $\tilde{c}$ on the denominator of equation \ref{eq:grad_in_dldp} happen to be cancelled out by equivalently small values in the numerator, the analysis leading to the black arrows in Figure \ref{fig:phase_color_1} needs some modification. For example, when $\hat{t}^I_\text{root,L}=0$, equation \ref{eq:grad_in_dldp}b becomes
\begin{equation}
\label{eq:grad_in_dldp_complex}
\frac{\partial L}{\partial \hat{p}}
=
\left(\frac{-\hat{b}}{2} - \ttargethat^R\right)
\begin{bmatrix}
-\frac{1}{2} & 0\\
\end{bmatrix}
+
(\pm\sqrt{\tilde{c}} - 0)
\frac{1}{\pm2\sqrt{\tilde{c}}}
\begin{bmatrix}
\frac{-\hat{b}}{2} & 1\\
\end{bmatrix}\\
=
\begin{bmatrix}
\frac{\ttargethat^R}{2} & \frac{1}{2}\\
\end{bmatrix}
\end{equation}
after substituting the complex roots, $\hat{t}^R_\text{root} = -\frac{\hat{b}}{2}$ and $\hat{t}^I_\text{root} = \pm \sqrt{\tilde{c}}$, into $\frac{\partial L}{\partial \hat{t}_\text{root}}$ in equation \ref{eq:dldphat}.
In Figure \ref{fig:phase_color_1} where $\ttargethat^R = \frac{1}{2}$, equation \ref{eq:grad_in_dldp_complex} gives $-\frac{\partial L}{\partial \hat{p}} = -[\frac{1}{4}, \frac{1}{2}]$ indicating that the black arrows in the right subfigure of Figure \ref{fig:phase_color_1} should point down and to the left in the entire complex region. 
Note that they still point towards the $\tilde{c}=0$ parabola as long as $\hat{b}<4$.
This boundedness of $\frac{\partial L}{\partial \hat{p}}$ in the complex region is why the upper right subfigure of Figure \ref{fig:example-5a} has many more iterates in the $\tilde{c}>0$ region than in the $\tilde{c}<0$ region where $\frac{\partial L}{\partial \hat{p}}$ does blow up as $\tilde{c}$ vanishes.
These constant values of $\frac{\partial L}{\partial \hat{b}}$ and $\frac{\partial L}{\partial \hat{c}}$ (whenever $\tilde{c}>0$) can also be seen in Figure \ref{fig:example-5b}.
}

\begin{figure}[H]
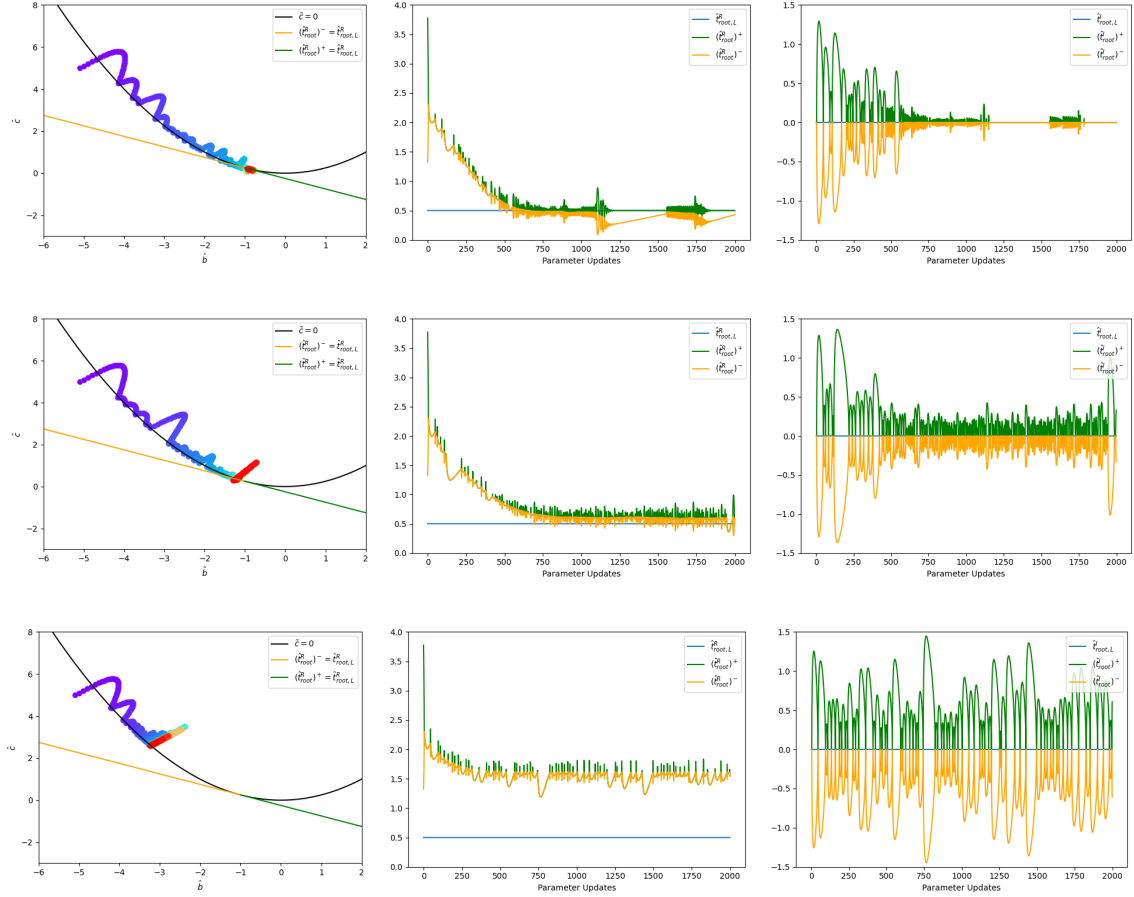

	\centering
    \foreach \etaval in {0.001, 0.01, 0.1}{
	\begin{subfigure}[b]{0.3\textwidth}
		\centering
		\includegraphics[width=\textwidth]{figures/quadratic-section/smoothing-subsection/examples/example14/backprop_\etaval_phase_diagram_1.png}
		\label{fig:}
	\end{subfigure}
	\begin{subfigure}[b]{0.3\textwidth}
		\centering
		\includegraphics[width=\textwidth]{figures/quadratic-section/smoothing-subsection/examples/example14/backprop_\etaval_roots_real.png}
		\label{fig:}
	\end{subfigure}
    \begin{subfigure}[b]{0.3\textwidth}
		\centering
		\includegraphics[width=\textwidth]{figures/quadratic-section/smoothing-subsection/examples/example14/backprop_\etaval_roots_imag.png}
		\label{fig:}
	\end{subfigure}
    }
	\caption{Equation \ref{eq:loss_reg_towards_init} with $\eta = .001$ (top row), $\eta = .01$ (middle row), and $\eta = .1$ (bottom row). The $\eta$-term attracts the iterates towards the initial guess, subjecting them to highly oscillatory behavior near the $\tilde{c}=0$ parabola (preventing convergence).
    } 
	\label{fig:example-7}
\end{figure}

It is not always be possible for the iterates to separate from the $\tilde{c}=0$ parabola. In Figure \ref{fig:example-7}, the initial guess is moved further away from the $\tilde{c}=0$ parabola in order to illustrate how the iterates are still attracted towards it; then, the objective function is modified to prefer solutions close to the initial guess via
\begin{equation}
\label{eq:loss_reg_towards_init} 
L(\hat{p}) =
\frac{1}{2}
\left(
\left|\left|
\hat{t}_\text{root}(\hat{p}) - \hat{t}_\text{root,L}\\  
\right|\right|_2^2
+
\eta
\left|\left|
\hat{p} - \hat{p}_0
\right|\right|_2^2
\right)
\end{equation}
where $\hat{p} = [\hat{b}~\hat{c}]^T$. This relatively common modification of the objective function prevents the iterates from separating far enough away from the $\tilde{c}=0$ parabola to avoid oscillations (and converge).

\begin{figure}[H]
	\centering
	\begin{subfigure}[b]{0.45\textwidth}
		\centering
		\includegraphics[width=\textwidth]{figures/quadratic-section/smoothing-subsection/phase_diagram_zero_loss2.png}
		\label{fig:}
	\end{subfigure}
	\begin{subfigure}[b]{0.45\textwidth}
		\centering
		\includegraphics[width=\textwidth]{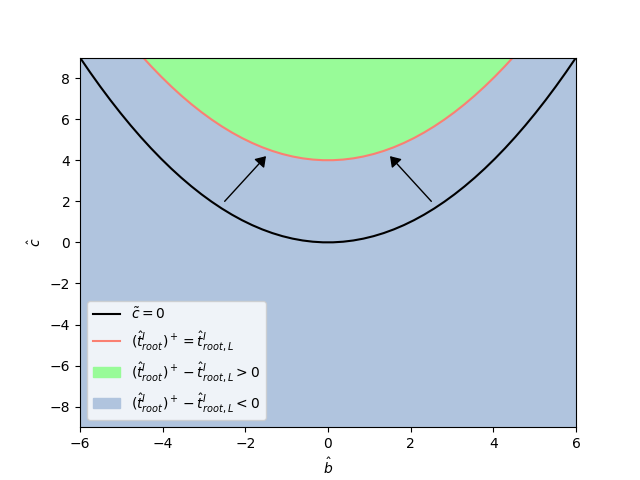}
		\label{fig:}
	\end{subfigure}
	\caption{
    These figures address using $\hat{t}_\text{root}^+$ in the objective function (the figures for using $\hat{t}_\text{root}^-$ illustrate similar behavior).
    Modifying $\ttargethat$ to have a nonzero imaginary part does not affect equation \ref{eq:grad_in_dldp}a, and thus the left subfigure remains identical to that shown in Figure \ref{fig:phase_color_1}; however, as dictated by equation \ref{eq:grad_in_dldp}b, the arrows in the right subfigure change to point away from the $\tilde{c}=0$ parabola and towards $(\hat{t}_\text{root}^I)^+ = \ttargethat^I$ as desired.} 
	\label{fig:phase_color_2}
\end{figure}

Next, consider target roots with nonzero imaginary parts, modifying $\ttargethat = \left[\frac{1}{2}, 0\right]^T$ to $\ttargethat = \left[\frac{1}{2}, 2\right]^T$. The behavior near the $\tilde{c}=0$ parabola, as dictated by equation \ref{eq:grad_in_dldp}, is illustrated in Figure \ref{fig:phase_color_2}. Unfortunately, iterates that start below the $\tilde{c}=0$ parabola in the blue region tend to erroneously converge to the green ray, because the identically zero imaginary component provides no information (below the $\tilde{c}=0$ parabola) to prevent this; in addition, iterates below the green ray will also tend to erroneously converge to it. This can be remedied by including both $\hat{t}^+_\text{root}$ and $\hat{t}^-_\text{root}$ in the objective function, via
\begin{equation}
\label{eq:complexconjugateloss} 
L(\hat{p}) =
\frac{1}{2}
\left(
||\hat{t}^+_\text{root}(\hat{p}) - \hat{t}_\text{root,L}||_2^2
+
||\hat{t}^-_\text{root}(\hat{p}) - \hat{t}^*_\text{root,L}||_2^2
\right)
\end{equation}
where $\ttargethat^* = \left[\frac{1}{2}, -2\right]^T$ is the complex conjugate, in order to create gradients below the $\tilde{c}=0$ parabola that lead towards the intersection point of the green and yellow rays (i.e.~towards the endpoint of the black dashed ray, as desired); in fact, only the real part of $\hat{t}^-_\text{root}$ needs to be included in the objective function. When one desires complex roots but is unconcerned with the precise value of the imaginary components, it is enough to force the real parts of both $\hat{t}^+_\text{root}$ and $\hat{t}^-_\text{root}$ to target the same $\ttargethat^R$ in the objective function; alternatively, when also unconcerned with the precise values of the real components, it is enough to minimize the difference between the real parts of $\hat{t}^+_\text{root}$ and $\hat{t}^-_\text{root}$.

\begin{figure}[H]
	\centering
    \foreach \derivative in {analytic_grad, smooth_grad_0.1}{
	\begin{subfigure}[b]{0.3\textwidth}
		\centering
		\includegraphics[width=\textwidth]{figures/quadratic-section/smoothing-subsection/examples/example18/\derivative_phase_diagram_1.png}
		\label{fig:}
	\end{subfigure}
	\begin{subfigure}[b]{0.3\textwidth}
		\centering
		\includegraphics[width=\textwidth]{figures/quadratic-section/smoothing-subsection/examples/example18/\derivative_roots_real.png}
		\label{fig:}
	\end{subfigure}
    \begin{subfigure}[b]{0.3\textwidth}
		\centering
		\includegraphics[width=\textwidth]{figures/quadratic-section/smoothing-subsection/examples/example18/\derivative_roots_imag.png}
		\label{fig:}
	\end{subfigure}
    }
	\begin{subfigure}[b]{0.3\textwidth}
		\centering
		\includegraphics[width=\textwidth]{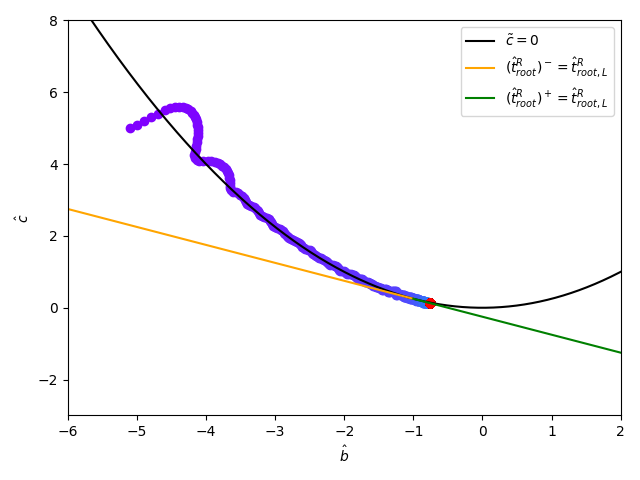}
		\label{fig:}
	\end{subfigure}
	\begin{subfigure}[b]{0.3\textwidth}
		\centering
		\includegraphics[width=\textwidth]{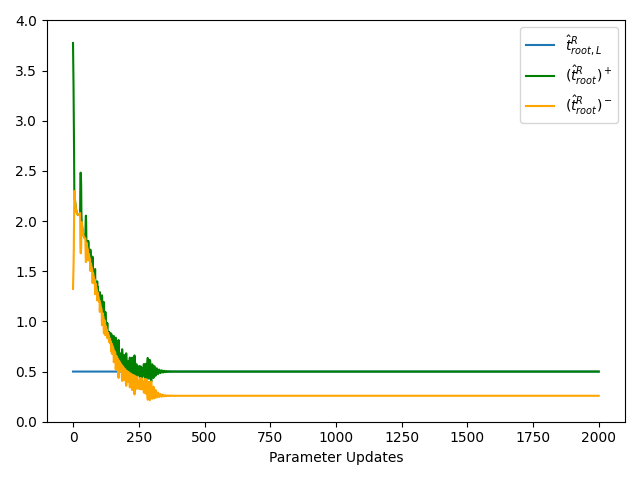}
		\label{fig:}
	\end{subfigure}
	\begin{subfigure}[b]{0.3\textwidth}
		\centering
		\includegraphics[width=\textwidth]{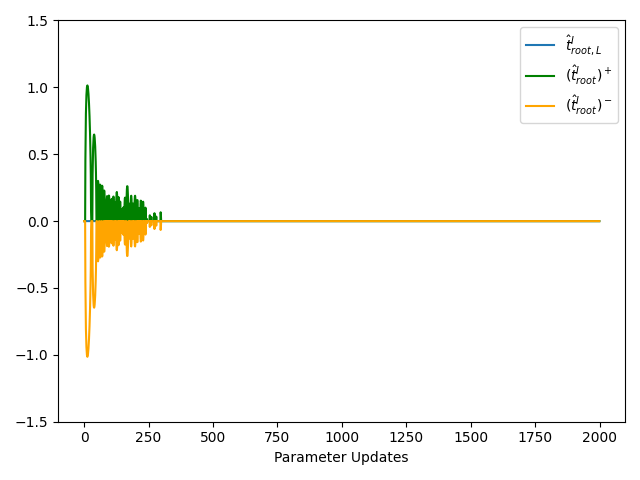}
		\label{fig:}
	\end{subfigure}
	\caption{Replacing backpropagation (via PyTorch) with an implicit layer defined by equations \ref{eq:grad_in_dldp}a and \ref{eq:grad_in_dldp_complex} gives results (top row) similar to Figure \ref{fig:example-3} (as expected). Subsequently limiting the magnitude of $\tilde{c}$ in equation \ref{eq:grad_in_dldp}a (e.g. $|\tilde{c}|\geq0.1$ on the second row) improves convergence. Breaking up the derivatives into components parallel to and orthogonal to $\frac{\partial \tilde{c}}{\partial \hat{p}}$ in equation \ref{eq:orthogonal_clamping_real_side} and subsequently clamping the magnitude of the parallel component (e.g. $\geq1$ on the third row) also improves convergence.
    } 
	\label{fig:example-18}
\end{figure}

To address the oscillations in Figures \ref{fig:example-3}, \ref{fig:example-5a}, and \ref{fig:example-7}, we first replace backpropagation (via PyTorch) with an implicit layer using equation \ref{eq:grad_in_dldp}a and \ref{eq:grad_in_dldp_complex}, obtaining results quite similar to Figures \ref{fig:example-3}, \ref{fig:example-5a}, and \ref{fig:example-7} as expected (see Figures \ref{fig:example-18} and \ref{fig:example-20-bottom-row}, top rows). Then, convergence can be improved in all cases by clamping the magnitude of $\tilde{c}$ in equation \ref{eq:grad_in_dldp}a so that dividing by it no longer causes derivatives to blow up near the $\tilde{c}=0$ parabola (see Figures \ref{fig:example-18} and \ref{fig:example-20-bottom-row}, second rows). 
Alternatively, equation \ref{eq:grad_in_dldp}a can be rewritten as
\begin{equation}
\label{eq:orthogonal_clamping_real_side}
\begin{aligned}
\frac{\partial L}{\partial \hat{p}}
&=
\frac{\partial L}{\partial \hat{t}^R_\text{root}}
\left(
\frac{-1}{\pm2\sqrt{-\tilde{c}}}
+
\frac{\hat{b}}{\hat{b}^2+4}
\right)
\frac{\partial \tilde{c}}{\partial \hat{p}}
+
\frac{\partial L}{\partial \hat{t}^R_\text{root}}
\frac{1}{\hat{b}^2+4}
\begin{bmatrix}
    -2&-\hat{b}\\
\end{bmatrix}
&& \text{if }\tilde{c}<0
\end{aligned}
\end{equation}
by splitting $[-\frac{1}{2}~0]$ into components parallel to and orthogonal to $\frac{\partial \tilde{c}}{\partial \hat{p}}$;
then, (instead of clamping $\tilde{c}$) one can clamp the magnitude of the first term in equation \ref{eq:orthogonal_clamping_real_side} (see Figures \ref{fig:example-18} and \ref{fig:example-20-bottom-row}, bottom rows).

\textit{
\RemarkCounter{re:}
Serendipitously, a reduced number of Newton iterations may lead to erroneously low values of $\tilde{c}$ (as illustrated in Figure \ref{fig:newton_grad}) providing accidental, but beneficial, clamping; unfortunately, increasing the number of Newton iterations would then lead to (perhaps surprising) instability.
}

\clearpage




\foreach \lamb/\row/\rowlabel in {0.1/bottom /bottom-row} {
\begin{figure}[H]
	\centering
	\begin{subfigure}[b]{0.3\textwidth}
		\centering
		\includegraphics[width=\textwidth]{figures/quadratic-section/smoothing-subsection/examples/example20/analytic_\lamb_phase_diagram_1.png}
		\label{fig:}
	\end{subfigure}
	\begin{subfigure}[b]{0.3\textwidth}
		\centering
		\includegraphics[width=\textwidth]{figures/quadratic-section/smoothing-subsection/examples/example20/analytic_\lamb_roots_real.png}
		\label{fig:}
	\end{subfigure}
    \begin{subfigure}[b]{0.3\textwidth}
		\centering
		\includegraphics[width=\textwidth]{figures/quadratic-section/smoothing-subsection/examples/example20/analytic_\lamb_roots_imag.png}
		\label{fig:}
	\end{subfigure}
    \foreach \k in {0.1}{
	\begin{subfigure}[b]{0.3\textwidth}
		\centering
		\includegraphics[width=\textwidth]{figures/quadratic-section/smoothing-subsection/examples/example20/given_\k_\lamb_phase_diagram_1.png}
		\label{fig:}
	\end{subfigure}
	\begin{subfigure}[b]{0.3\textwidth}
		\centering
		\includegraphics[width=\textwidth]{figures/quadratic-section/smoothing-subsection/examples/example20/given_\k_\lamb_roots_real.png}
		\label{fig:}
	\end{subfigure}
    \begin{subfigure}[b]{0.3\textwidth}
		\centering
		\includegraphics[width=\textwidth]{figures/quadratic-section/smoothing-subsection/examples/example20/given_\k_\lamb_roots_imag.png}
		\label{fig:}
	\end{subfigure}
    }
	\begin{subfigure}[b]{0.3\textwidth}
		\centering
		\includegraphics[width=\textwidth]{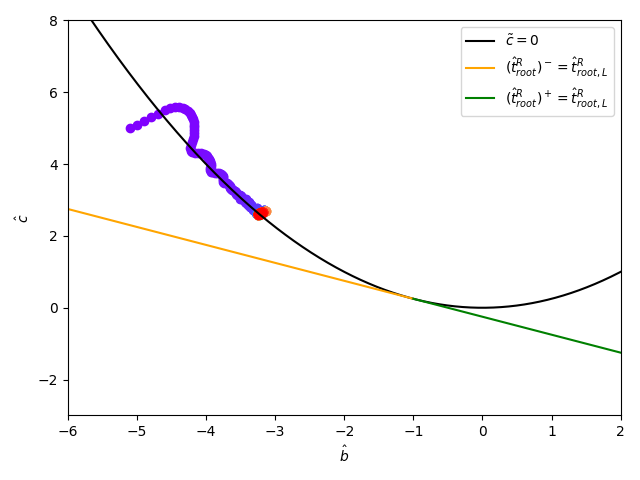}
		\label{fig:}
	\end{subfigure}
	\begin{subfigure}[b]{0.3\textwidth}
		\centering
		\includegraphics[width=\textwidth]{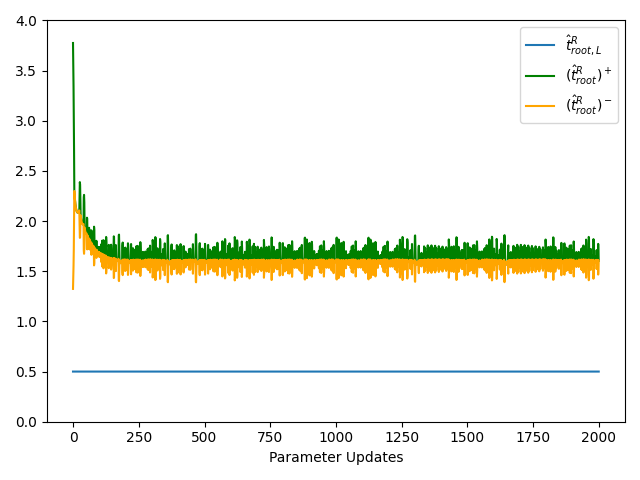}
		\label{fig:}
	\end{subfigure}
	\begin{subfigure}[b]{0.3\textwidth}
		\centering
		\includegraphics[width=\textwidth]{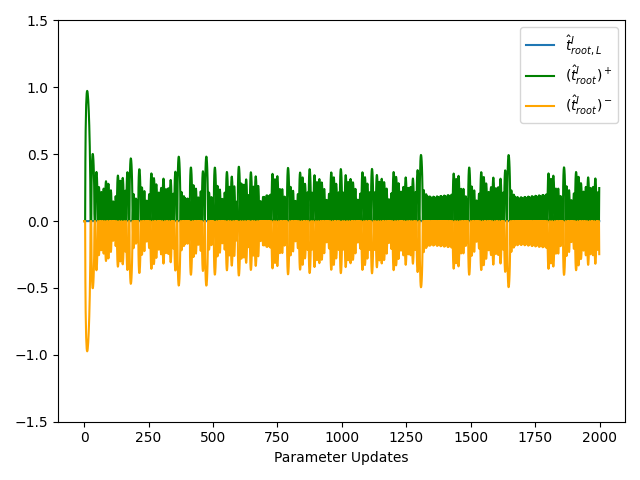}
		\label{fig:}
	\end{subfigure}
	\caption{Replacing backpropagation (via PyTorch) with an implicit layer defined by equations \ref{eq:grad_in_dldp}a and \ref{eq:grad_in_dldp_complex} gives results (top row) similar to the bottom row of Figure \ref{fig:example-7} (as expected). Subsequently limiting the magnitude of $\tilde{c}$ in equation \ref{eq:grad_in_dldp}a (e.g. $|\tilde{c}|\geq0.1$ on the second row) improves convergence. Breaking up the derivatives into components parallel to and orthogonal to $\frac{\partial \tilde{c}}{\partial \hat{p}}$ in equation \ref{eq:orthogonal_clamping_real_side} and subsequently clamping the magnitude of the parallel component (e.g. $\geq1$ on the third row) also improves convergence.
    } 
	\label{fig:example-20-\rowlabel}
\end{figure}
}

\clearpage
\section{Proposed Approach for Quadratic Equations}
\label{section:quadraticproposedapproach}
We begin by leveraging the change of variables robust to $a\to0$ discussed in Section \ref{subsection:computingderivatives-1p-examples},
\begin{subequations}
\label{eq:prop_cov}
\begin{align}
\tilde{t}
&= 
at
+
\begin{bmatrix}
    \frac{b}{2}\\
    0\\
\end{bmatrix}\\
\begin{bmatrix}
    \tilde{t}^R\\
    \tilde{t}^I\\
\end{bmatrix}
&= 
a
\begin{bmatrix}
    \treal\\
    \timag\\    
\end{bmatrix}
+
\begin{bmatrix}
    \frac{b}{2}\\
    0\\
\end{bmatrix}
\end{align}
\end{subequations}
giving $\tilde{t}^2 + \tilde{c} = 0$ with $\tilde{c} = \frac{-b^2}{4}+ac$. In this reduced canonical form, equations \ref{eq:quadratic2d} and \ref{eq:quadraticderivatives} become
\begin{subequations}
    \label{eq:ourapproachf}
    \begin{align}
        \tilde{f}(\tilde{t}_\text{root}; \tilde{p})
        &= 
        \begin{bmatrix}
            (\tilde{t}_\text{root}^R)^2 - (\tilde{t}_\text{root}^I)^2 + \tilde{c}\\
            2\tilde{t}_\text{root}^R\tilde{t}_\text{root}^I\\
        \end{bmatrix}
        =
        a
        f(t_\text{root}; \vec{p})
        = \vec{0}\\
        \tilde{f}_{\theta_1}(\tilde{t}_\text{root}; \tilde{p})
        &=
        \begin{bmatrix}
            2 \tilde{t}_\text{root}^R & - 2 \tilde{t}_\text{root}^I\\
            2 \tilde{t}_\text{root}^I & 2 \tilde{t}_\text{root}^R\\
        \end{bmatrix}
        =
        f_{\theta_1}(t_\text{root}; \vec{p})
        \\
        \tilde{f}_{\theta_1}^{-1}(\tilde{t}_\text{root}; \tilde{p})
        &=
        \frac{1}{s}
        \begin{bmatrix}
            2 \tilde{t}_\text{root}^R & 2 \tilde{t}_\text{root}^I\\
            -2 \tilde{t}_\text{root}^I & 2 \tilde{t}_\text{root}^R\\
        \end{bmatrix}
        \text{ where }
        s
        = 4 ((\tilde{t}_\text{root}^R)^2 + (\tilde{t}_\text{root}^I)^2)\\
        \tilde{f}_{\theta_2}(\tilde{t}_\text{root}; \tilde{p})
        &=
        \begin{bmatrix}
            1 \\ 0 \\        
        \end{bmatrix}
        =
        f_{\theta_2}(t_\text{root}; \vec{p})
        \begin{bmatrix}
        0\\0\\1\\
        \end{bmatrix}
    \end{align}
\end{subequations}
where equation \ref{eq:ourapproachf}a matches equation \ref{eq:quadratic2d} up to a factor of $a$, equations \ref{eq:ourapproachf}b-c match equations \ref{eq:quadraticderivatives}a-b, and equation \ref{eq:ourapproachf}d matches the last column of equation \ref{eq:quadraticderivatives}c.
Note that $\frac{\partial \vec{p}}{\partial \tilde{p}}$ is a 3x3 Jacobian and that $\tilde{f}_{\theta_2}(\tilde{t}_\text{root}; \tilde{p})$ is formally size 2x3; however, $\tilde{f}(\tilde{t}_\text{root}; \tilde{p})$ only requires the last column of $\tilde{f}_{\theta_2}(\tilde{t}_\text{root}; \tilde{p})$, and so only this last column is shown in equation \ref{eq:ourapproachf}d.

The total derivative of equation \ref{eq:ourapproachf}a is $\tilde{f}_{\theta_1}(\tilde{t}_\text{root}, \tilde{p}) d \theta_1 + \tilde{f}_{\theta_2}(\tilde{t}_\text{root}, \tilde{p}) d \theta_2 = 0$, which can be written as $d \theta_1 = - \tilde{f}_{\theta_1}^{-1}(\tilde{t}_\text{root}, \tilde{p}) \tilde{f}_{\theta_2}(\tilde{t}_\text{root}, \tilde{p}) d \theta_2$ when $\tilde{f}_{\theta_1}$ is invertible; in other words, $\frac{\partial \theta_1}{\partial \theta_2} = -\tilde{f}_{\theta_1}^{-1}(\tilde{t}_\text{root}, \tilde{p}) \tilde{f}_{\theta_2}(\tilde{t}_\text{root}, \tilde{p})$. As long as $\theta_1$ and $\theta_2$ are not independent, one can write $d\theta_1 = \frac{\partial \theta_1}{\partial \theta_2} d\theta_2$ leading to $\tilde{f}_{\theta_1}(\tilde{t}_\text{root}, \tilde{p})\frac{\partial \theta_1}{\partial \theta_2} = - \tilde{f}_{\theta_2}(\tilde{t}_\text{root}, \tilde{p})$ or
\begin{equation}
\label{eq:ourapproach2}
\begin{bmatrix}
    2 \tilde{t}_\text{root}^R & - 2 \tilde{t}_\text{root}^I\\
    2 \tilde{t}_\text{root}^I & 2 \tilde{t}_\text{root}^R\\
\end{bmatrix}\\
\begin{bmatrix}
    \frac{\partial \tilde{t}_\text{root}^R}{\partial \tilde{c}}\\
    \frac{\partial \tilde{t}_\text{root}^I}{\partial \tilde{c}}\\
\end{bmatrix}
=
-
\begin{bmatrix}
    1 \\ 0 \\        
\end{bmatrix}
\end{equation}
even when $\tilde{f}_{\theta_1}$ is not invertible.
When the roots are real (and not repeated) with $\tilde{t}_\text{root}^R=\pm\sqrt{-\tilde{c}}$ and $\tilde{t}_\text{root}^I=0$, equation \ref{eq:ourapproach2} gives
$
\frac{\partial \tilde{t}_\text{root}^R}{\partial \tilde{c}}
= \frac{-1}{2\tilde{t}_\text{root}^R} 
= \frac{-1}{\pm2\sqrt{-\tilde{c}}} 
$
and $\frac{\partial \tilde{t}_\text{root}^I}{\partial \tilde{c}} = 0$. When the roots are complex with $\tilde{t}_\text{root}^R = 0$ and $\tilde{t}_\text{root}^I=\pm\sqrt{\tilde{c}}$, equation \ref{eq:ourapproach2} gives $\frac{\partial \tilde{t}_\text{root}^R}{\partial \tilde{c}} = 0$ and
$
\frac{\partial \tilde{t}_\text{root}^I}{\partial \tilde{c}} 
= \frac{1}{2\tilde{t}_\text{root}^I} 
= \frac{1}{\pm2\sqrt{\tilde{c}}} 
$.
In the repeated root case, $\tilde{c}=\tilde{t}_\text{root}^R=\tilde{t}_\text{root}^I=0$ making $\tilde{f}_{\theta_1}$ identically zero so that $\tilde{f}_{\theta_1}^{-1}$ does not exist; then, one can no longer rely on sloppy interpretations of the implicit function theorem in order to write statements such as $d \theta_1 = - \tilde{f}_{\theta_1}^{-1}(\tilde{t}_\text{root}, \tilde{p}) \tilde{f}_{\theta_2}(\tilde{t}_\text{root}, \tilde{p}) d \theta_2$ and $\frac{\partial \theta_1}{\partial \theta_2} = -\tilde{f}_{\theta_1}^{-1}(\tilde{t}_\text{root}, \tilde{p}) \tilde{f}_{\theta_2}(\tilde{t}_\text{root}, \tilde{p})$. However, equation \ref{eq:ourapproach2} is still valid (asymptotically) in spite of the coefficient matrix going to zero.
This can be seen from the real side by plugging in the solution $[\frac{-1}{2\tilde{t}_\text{root}^R}, 0]^T$ to obtain the right hand side via (a trivial) L'Hospital's rule.
Similarly from the complex side, plugging in $[0, \frac{1}{2\tilde{t}_\text{root}^I}]^T$ leads to the right hand side.

In this reduced canonical form, $\tilde{t}_\text{root}$ only depends on $\tilde{c}$ and thus 
\begin{subequations}
\label{eq:ourapproach0}
\begin{align}
\frac{\partial \tilde{t}_\text{root}}{\partial \vec{p}} &= \frac{\partial \tilde{t}_\text{root}}{\partial \tilde{c}}\frac{\partial \tilde{c}}{\partial \vec{p}}\\
\label{eq:ourapproach0.3}
\tilde{f}_{\theta_1}(\tilde{t}_\text{root}, \tilde{p})\frac{\partial \tilde{t}_\text{root}}{\partial \vec{p}} &= \tilde{f}_{\theta_1}(\tilde{t}_\text{root}, \tilde{p})\frac{\partial \tilde{t}_\text{root}}{\partial \tilde{c}}\frac{\partial \tilde{c}}{\partial \vec{p}}
=
-
\begin{bmatrix}
    1 \\ 0\\
\end{bmatrix}
\frac{\partial \tilde{c}}{\partial \vec{p}}
\end{align}
\end{subequations}
since the left hand side of equation \ref{eq:ourapproach2} is $\tilde{f}_{\theta_1}(\tilde{t}_\text{root}, \tilde{p}) \frac{\partial \tilde{t}_\text{root}}{\partial \tilde{c}}$.
Using equation \ref{eq:prop_cov} to expand the left hand side of equation \ref{eq:ourapproach0.3} leads to
\begin{subequations}
\label{eq:ourapproach0.1}
\begin{align}
\label{eq:ourapproach0.1expand}
\tilde{f}_{\theta_1}(\tilde{t}_\text{root}, \tilde{p})
\left(
a
\frac{\partial t_\text{root}}{\partial \vec{p}}
+
\begin{bmatrix}
    \treal_\text{root} & \frac{1}{2} & 0\\
    \timag_\text{root} & 0 & 0 \\
\end{bmatrix}
\right)
&=
-
\begin{bmatrix}
    1 \\ 0\\
\end{bmatrix}
\begin{bmatrix}
    c&
    \frac{-b}{2}&
    a\\
\end{bmatrix}\\
\label{eq:ourapproach0.1ftrick}
a
\tilde{f}_{\theta_1}(\tilde{t}_\text{root}, \tilde{p})
\frac{\partial t_\text{root}}{\partial \vec{p}}
&=
-
a
\begin{bmatrix}
    (\treal_\text{root})^2 - (\timag_\text{root})^2 & \treal_\text{root} & 1\\
    2 \treal_\text{root}\timag_\text{root} & \timag_\text{root} & 0\\
\end{bmatrix}\\
\label{eq:ourapproach0.2withouta}
\tilde{f}_{\theta_1}(\tilde{t}_\text{root}, \tilde{p})
\frac{\partial t_\text{root}}{\partial \vec{p}}
&=
-
\begin{bmatrix}
    (\treal_\text{root})^2 - (\timag_\text{root})^2 & \treal_\text{root} & 1\\
    2 \treal_\text{root}\timag_\text{root} & \timag_\text{root} & 0\\
\end{bmatrix}
\end{align}
\end{subequations}
where equation \ref{eq:quadratic2d} was used to simplify the right hand side of equation \ref{eq:ourapproach0.1ftrick}.
Importantly, equation \ref{eq:ourapproach0.2withouta} was obtained from equation \ref{eq:ourapproach0.1ftrick} by dividing by $a$, which allows for $a\to0$ but not $a=0$.

\textit{
\RemarkCounter{re:}
Note that equation \ref{eq:ourapproach0.2withouta} is identically $f_{\theta_1}(t_\text{root}; \vec{p}) \frac{\partial t_\text{root}}{\partial \vec{p}} = -f_{\theta_2}(t_\text{root}; \vec{p})$. That is, equation \ref{eq:implicitderivative} can be treated more carefully via equation \ref{eq:ourapproach0.2withouta} (similar in spirit to equation \ref{eq:ourapproach2}).
}

From equation \ref{eq:ourapproachf}c, one can obtain
\begin{subequations}
\label{eq:cubicftheta1inverseall}
\begin{align}
\label{eq:cubicftheta1inversectilde}
\tilde{f}^{-1}_{\theta_1}(\tilde{t}_\text{root}; \tilde{p})
&=
\begin{cases}
\frac{1}{\pm 2\sqrt{-\tilde{c}}} I
& \text{if }\tilde{c}<0\\
\frac{1}{\pm 2\sqrt{\tilde{c}}} 
\begin{bmatrix}
    0 & 1\\
    -1 & 0\\
\end{bmatrix}
& \text{if }\tilde{c}>0\\
\end{cases}\\
\label{eq:cubicftheta1inversetilde}
\tilde{f}^{-1}_{\theta_1}(\tilde{t}_\text{root}; \tilde{p})
&=
\begin{cases}
\frac{1}{2 \tilde{t}^R_\text{root}} I
& \text{if } \tilde{t}^I_\text{root}=0\\
\frac{1}{2 \tilde{t}^I_\text{root}} 
\begin{bmatrix}
    0 & 1\\
    -1 & 0\\
\end{bmatrix}
& \text{if } \tilde{t}^I_\text{root}\neq0\\
\end{cases}\\
\label{eq:cubicftheta1inverseoriginal}
\tilde{f}^{-1}_{\theta_1}(\tilde{t}_\text{root}; \tilde{p})
&=
\begin{cases}
\frac{1}{2a t^R_\text{root} +b} I
& \text{if }t^I_\text{root}=0\\
\frac{1}{2a t^I_\text{root}} 
\begin{bmatrix}
    0 & 1\\
    -1 & 0\\
\end{bmatrix}
& \text{if }t^I_\text{root}\neq0\\
\end{cases}
\end{align}
\end{subequations}
which are all identical;
however, equation \ref{eq:cubicftheta1inversectilde} explicitly maintains the $\pm$-sign differentiating between the two coalescing roots as the denominator goes to (and becomes identically equal to) zero.
We proceed using equation \ref{eq:cubicftheta1inverseoriginal}, relying on equation \ref{eq:cubicftheta1inversectilde} only to motivate the algorithm.

\textit{
\RemarkCounter{re:clampingderiv}
As long as the standard definition of $t^\pm_\text{root}$ is used (i.e.~equation \ref{eq:quadraticroots}), both $2a t^R_\text{root} +b$ and $2a t^I_\text{root}$ should be clamped to be positive/negative when using $t^\pm_\text{root}$ in the objective function respectively.
}

\textit{
\RemarkCounter{re:}
This clamping (in Remark \ref{re:clampingderiv}) alleviates issues when: the roots are close to being repeated and $2a t^R_\text{root} +b \approx 0$ for both roots but $2a t^R_\text{root} +b$ is incorrectly the same sign (instead of opposite signs) for both roots due to numerical errors.
Note that similar issues do not arise for complex roots where both $t^I_\text{root}\neq0$ and $a\neq0$, since the $\pm-$sign follows directly (and correctly) from $t^I_\text{root}$ and $a$.
}

\textit{
\RemarkCounter{re:degerateais0}
In the degenerate $a=0$ case (which has two real roots), the sign of $2a t^R_\text{root} +b$ is correctly determined by $b\neq0$ for the finite root, but should be clamped to be opposite the sign of $b$ for the infinite root (due to a nonzero value for $2 a t^R_\text{root}$ from L'Hospital's rule).
This is properly treated by the clamping in Remark \ref{re:clampingderiv}.
The $a=b=0$ case (with two infinite roots) is also correctly treated by the clamping in Remark \ref{re:clampingderiv} (see Section \ref{subsection:quadproposedapproach-rootsolver}).
}

\textit{
\RemarkCounter{re:}
The choice between the $\tilde{c}<0$ and $\tilde{c}>0$ cases is unimportant for truly repeated roots, since (see e.g.~Figures \ref{fig:phase_color_1} and \ref{fig:phase_color_2}) the search directions across the $\tilde{c}=0$ parabola (whether consistent or inconsistent) cannot be significantly improved by any choice of search direction on the $\tilde{c}=0$ parabola; instead, one needs to carefully craft the objective function. 
}

Using equation \ref{eq:cubicftheta1inverseoriginal} in equation \ref{eq:ourapproach0.2withouta} leads to 
\begin{equation}
\label{eq:ourapproachderivctilde}
\frac{\partial t_\text{root}}{\partial \vec{p}}
=
\begin{cases} 
\frac{-1}{2a t^R_\text{root} +b}
\begin{bmatrix}
    (\treal_\text{root})^2 & \treal_\text{root} & 1\\
    0 & 0 & 0\\
\end{bmatrix}
& \text{if }\timag_\text{root}=0\\
\frac{-1}{2a t^I_\text{root}}
\begin{bmatrix}
    2\treal_\text{root}\timag_\text{root} & \timag_\text{root} & 0\\
    -((\treal_\text{root})^2 - (\timag_\text{root})^2) & -\treal_\text{root} & -1\\
\end{bmatrix}
& \text{if }\timag_\text{root}\neq0\\
\end{cases}
\end{equation}
where the rows of the matrices provide directions for $\frac{\partial t^R_\text{root}}{\partial \vec{p}}$ and $\frac{\partial t^I_\text{root}}{\partial \vec{p}}$.
In the $\timag_\text{root}=0$ case, consider the top row of the first matrix in equation \ref{eq:ourapproachderivctilde}. When $(\treal_\text{root})^2>1$, we factor $(\treal_\text{root})^2$ out into the numerator of the scalar multiplier so that each component of the vector is bounded by $1$. One could also factor out the magnitude of the top row, if a unit vector were desired. The scalar multiplier (out front) is then evaluated robustly as follows: Let $N$ be the magnitude of the numerator, $D$ be the magnitude of the denominator, and $M$ be the maximum allowable magnitude of the result. If $N<DM$, then $\frac{N}{D}$ can be robustly computed; otherwise, $\frac{N}{D}$ is set to $M$ without the need for (potentially problematic) division.
In the $\timag_\text{root}\neq0$ case, we consider the top and bottom rows of the second matrix in equation \ref{eq:ourapproachderivctilde} separately.
For the top row, $\timag_\text{root}$ is factored out front (where it cancels) leaving only $2a$ on the denominator.
Note that the $\pm$-sign from equation \ref{eq:cubicftheta1inverseall} is unnecessary for this (non-merging) real part of the complex root, and only the sign of $a$ is required (recall, $a\neq0$ for complex roots).
When $|2\treal_\text{root}|>1$, $2\treal_\text{root}$ is also factored out front; then, the scalar multiplier is robustly evaluated (as discussed above).
For the bottom row, the larger in magnitude between $(\treal_\text{root})^2 - (\timag_\text{root})^2$ and $\treal_\text{root}$ is factored out front when it is larger than $1$; then, the scalar multiplier is robustly evaluated.

\textit{
\RemarkCounter{re:splitupforcancelerror}
It can be problematic to evaluate $(\treal_\text{root})^2 - (\timag_\text{root})^2$ when both $\treal_\text{root}$ and $\timag_\text{root}$ are quite large. This can be alleviated to some degree by instead considering $(\treal_\text{root}+\timag_\text{root})(\treal_\text{root}-\timag_\text{root})$. Since $\treal_\text{root}$ and $\timag_\text{root}$ can only be large when $a$ is small, one would likely prefer to have this term dominate the second row in this case (as opposed to letting perhaps erroneous and commensurate values of $\treal_\text{root}$ and $\timag_\text{root}$ cancel, which is quite likely to happen when they are both clamped to $t_\text{max}$).
}

\subsection{Quadratic Root Solver}
\label{subsection:quadproposedapproach-rootsolver}
When $b^2-4ac <0$, the roots are complex and $a\neq0$. One can (using $\pm t_\text{max}$ as the upper bound) divide $-b$ by $2a$ for the real part and $\sqrt{-(b^2-4ac)}$ by $2a$ for the imaginary part.
Otherwise, the roots are real.
In the $a=0$ case, a pseudo-sign for $a$ is required in order to assign the larger/smaller computed root to $t_\text{root}^\pm$ for Remark \ref{re:clampingderiv}.

When $b=0$, equation \ref{eq:quadform} reduces to $\frac{\pm\sqrt{-ac}}{a}$.
When $a\neq0$, robust division can be used to compute $\pm\frac{\sqrt{|c|}}{\sqrt{|a|}}$.
When $a=0$ and $c\neq0$, $\pm\frac{\sqrt{|c|}}{\sqrt{|a|}} \to \pm t_\text{max}$;
in addition, the pseudo-sign for $a$ is set opposite the sign of $c$ (which is required in order for the roots to be real).
When $a=0$ and $c=0$ making everything a root, we return a repeated root at $0$ (keeping the problem symmetric) and arbitrarily set a pseudo-sign of $a>0$.
This leads to $[0~0~1]$ in the top row of the first matrix in equation \ref{eq:ourapproachderivctilde}, which fixes $c$ to be nonzero creating roots at $\pm t_\text{max}$; then, the top row subsequently becomes $[1~0~0]$, which fixes $a$ to make the roots smaller.

When $b\neq0$, the $-b\pm\sqrt{b^2-4ac}$ terms in equation \ref{eq:quadform} are always nonzero; thus, only division by $2a$ (when $a=0$) is problematic.
Both problematic quotients reduce to $\frac{-b}{a}\to \pm t_\text{max}$ depending on the pseudo-sign of $a$, and we arbitrarily set a pseudo-sign of $a>0$.

\textit{
\RemarkCounter{re:}
Since $a$, $b$, and $c$ do not grow too large in our examples, we use the $\vec{p}_\raw$ values in the root solver (not the normalized coefficients).
}

\textit{
\RemarkCounter{re:reversed_quadratic}
Replacing $t$ with $t^{-1}$ leads to a reversed quadratic $ct^2 + bt + a = 0$, which can be solved to obtain one over the roots (changing the roles of $a$ and $c$ in equation \ref{eq:quadform}). In the case of complex roots, one can thus choose to divide by either $2a$ or $2c$. In the case of real roots and $b=0$, choosing to divide by $c$ instead of $a$ does not help much since the result needs to be flipped anyways. When $b\neq0$, the problematic $\frac{-b}{a}$ case with $a=0$ is robustly replaced by $\frac{a}{-b}$ in equation \ref{eq:quadform} which is identically zero (but unfortunately needs to be flipped). Overall, this strategy of solving for $t^{-1}$ is not necessarily beneficial for the quadratic equation.
}

\subsection{Branch Selection}
\label{subsection:quadproposedapproach-branchselection}
The formulation of the objective function will typically be problem dependent.
As can be seen in equations \ref{eq:loss} and \ref{eq:loss_gradient}, each root used will typically need to be differentiated with respect to its parameters.
The root solver (in Section \ref{subsection:quadproposedapproach-rootsolver}) provides values for both roots for any set of parameters, and those values can be used in equation \ref{eq:ourapproachderivctilde} to robustly compute derivatives.

One particular case that is worth addressing is when $a$ changes sign.
When $a=0$, the quadratic degenerates to a linear function with one root; however, perturbations of $a$ cause the second root to be near $\pm \infty$ depending on the sign of $a$.
As $a$ changes sign, the root of the linear equation switches from being the smaller/larger root to being the larger/smaller root respectively.
As can be seen in equation \ref{eq:quadform}, the linear root only depends on $b$ (it is $t_\text{root}^+$ when $b>0$ and $t_\text{root}^-$ when $b<0$) and not on $a$;
thus, the roles of $t^\pm_\text{root}$ in the objective function do not need to change when $a$ changes sign.
See Figure \ref{fig:normal_fast_example}.

\subsection{Examples}
\label{subsection:quadproposedapproach-examples}
In this section, we show the efficacy for our proposed approach using the root solver from Section \ref{subsection:quadproposedapproach-rootsolver} while computing derivatives according to equation \ref{eq:ourapproachderivctilde} using the robust division discussed in the text (after equation \ref{eq:ourapproachderivctilde}).
The sign of the denominator of each scalar multiplier in equation $\ref{eq:ourapproachderivctilde}$ is chosen according to Remark \ref{re:clampingderiv}.
Similar to equation \ref{eq:lossagain} (and equation \ref{eq:lossagain2d}), we use
\begin{equation}
\label{eq:lossagainwitht} 
L(p) =
\frac{1}{2}
||t_\text{root}(p) - t_\text{root,L}||_2^2
\end{equation}
as the objective function, and choose $t_\text{root}^+$ as the branch under consideration.
Adam was used for the optimization.

\begin{figure}[H]
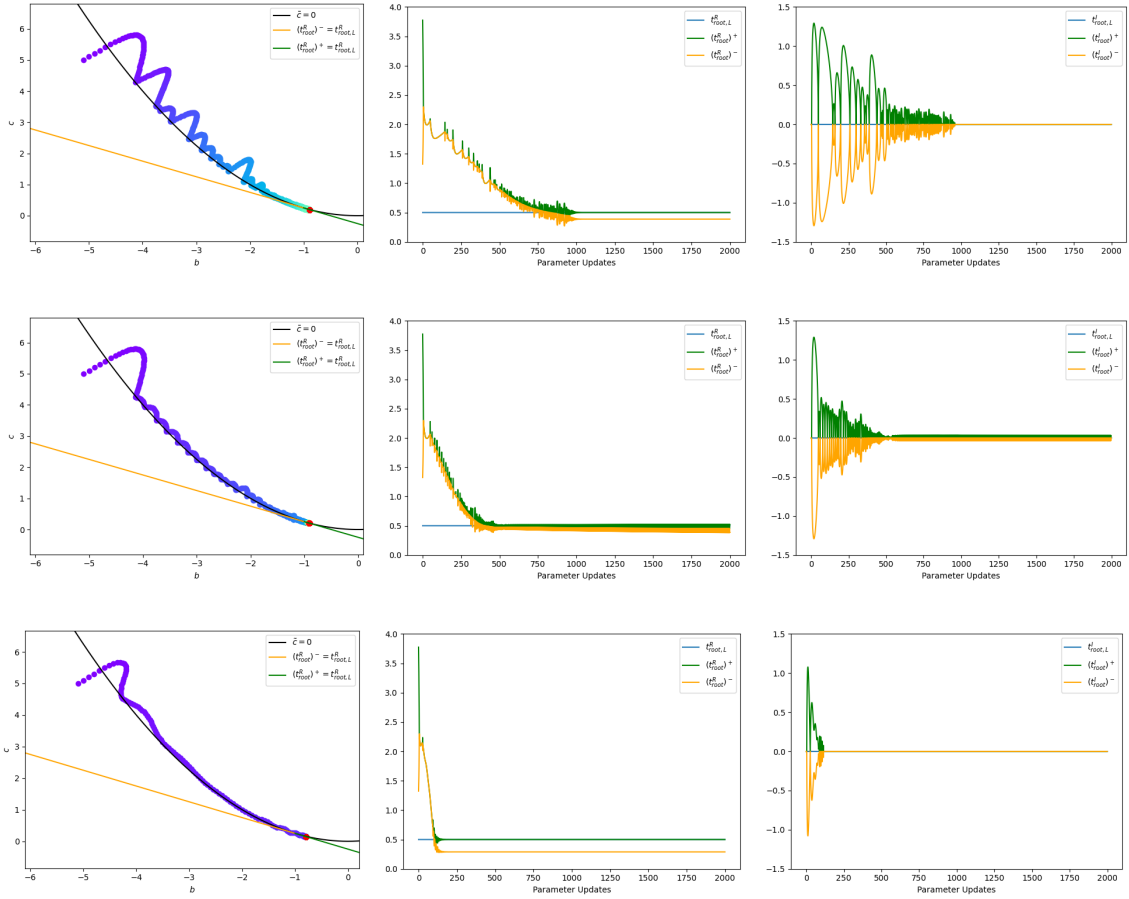

	\centering
    \foreach \deriv in {clamped_given_10000.0, clamped_given_10.0, clamped_given_1.0} {
	\begin{subfigure}[b]{0.3\textwidth}
		\centering
		\includegraphics[width=\textwidth]{figures/quadratic-section/smoothing-subsection/examples_3_parameter/example9/\deriv_normal_phase_diagram_1.png}
		\label{fig:}
	\end{subfigure}
	\begin{subfigure}[b]{0.3\textwidth}
		\centering
		\includegraphics[width=\textwidth]{figures/quadratic-section/smoothing-subsection/examples_3_parameter/example9/\deriv_normal_roots_real.png}
		\label{fig:}
	\end{subfigure}
	\begin{subfigure}[b]{0.3\textwidth}
		\centering
		\includegraphics[width=\textwidth]{figures/quadratic-section/smoothing-subsection/examples_3_parameter/example9/\deriv_normal_roots_imag.png}
		\label{fig:}
	\end{subfigure}
    }
	\caption{
    This example uses the same initial conditions as Figure \ref{fig:example-3}, the root solver from Section \ref{subsection:quadproposedapproach-rootsolver}, and equation \ref{eq:ourapproachderivctilde} with robust division with $M=10000$ (top row), $M=10$ (middle row), and $M=1$ (bottom row). The lack of convergence in the middle row is caused by the iterates settling down towards the left endpoint of the green ray near the $\tilde{c}=0$ parabola (as can be seen by the nonzero imaginary part of the roots). See also Figure \ref{fig:example-22-clamping-large_first_grad_new_t_target}.
    } 
	\label{fig:example-22-clamping-normal-final}
\end{figure}

\begin{figure}[H]
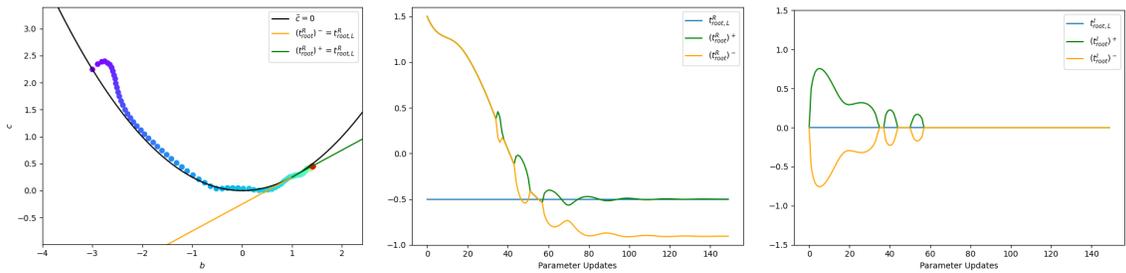

	\centering
    \foreach \subexample/\figcaption/\phasediagramnum in {large_first_grad_new_t_target/Figure \ref{fig:example-5a}/1}{
    \foreach \deriv in {clamped_given_0.01} {
	\begin{subfigure}[b]{0.3\textwidth}
		\centering
		\includegraphics[width=\textwidth]{figures/quadratic-section/smoothing-subsection/examples_3_parameter/example22/\deriv_\subexample_phase_diagram_\phasediagramnum.png}
		\label{fig:}
	\end{subfigure}
	\begin{subfigure}[b]{0.3\textwidth}
		\centering
		\includegraphics[width=\textwidth]{figures/quadratic-section/smoothing-subsection/examples_3_parameter/example22/\deriv_\subexample_roots_real.png}
		\label{fig:}
	\end{subfigure}
	\begin{subfigure}[b]{0.3\textwidth}
		\centering
		\includegraphics[width=\textwidth]{figures/quadratic-section/smoothing-subsection/examples_3_parameter/example22/\deriv_\subexample_roots_imag.png}
		\label{fig:}
	\end{subfigure}
    }
    }
	\caption{
    This example uses the same initial conditions as Figure \ref{fig:example-5a}, the root solver from Section \ref{subsection:quadproposedapproach-rootsolver}, and equation \ref{eq:ourapproachderivctilde} with robust division with $M=.01$. 
    Here, we change $\ttarget^R = \frac{1}{2}$ to $\ttarget^R = -\frac{1}{2}$ so that the direction orthogonal to the $\tilde{c}=0$ parabola tends to point towards the right (instead of towards the left) in the region of interest (near the left endpoint of the green ray).
    Even with very aggressive clamping in the robust division (using $M=.01$) to limit momentum accumulation in Adam, the iterates still safely converge further to the right on the green ray away from the $\tilde{c}=0$ parabola.
    } 
	\label{fig:example-22-clamping-large_first_grad_new_t_target}
\end{figure}

\begin{figure}[H]
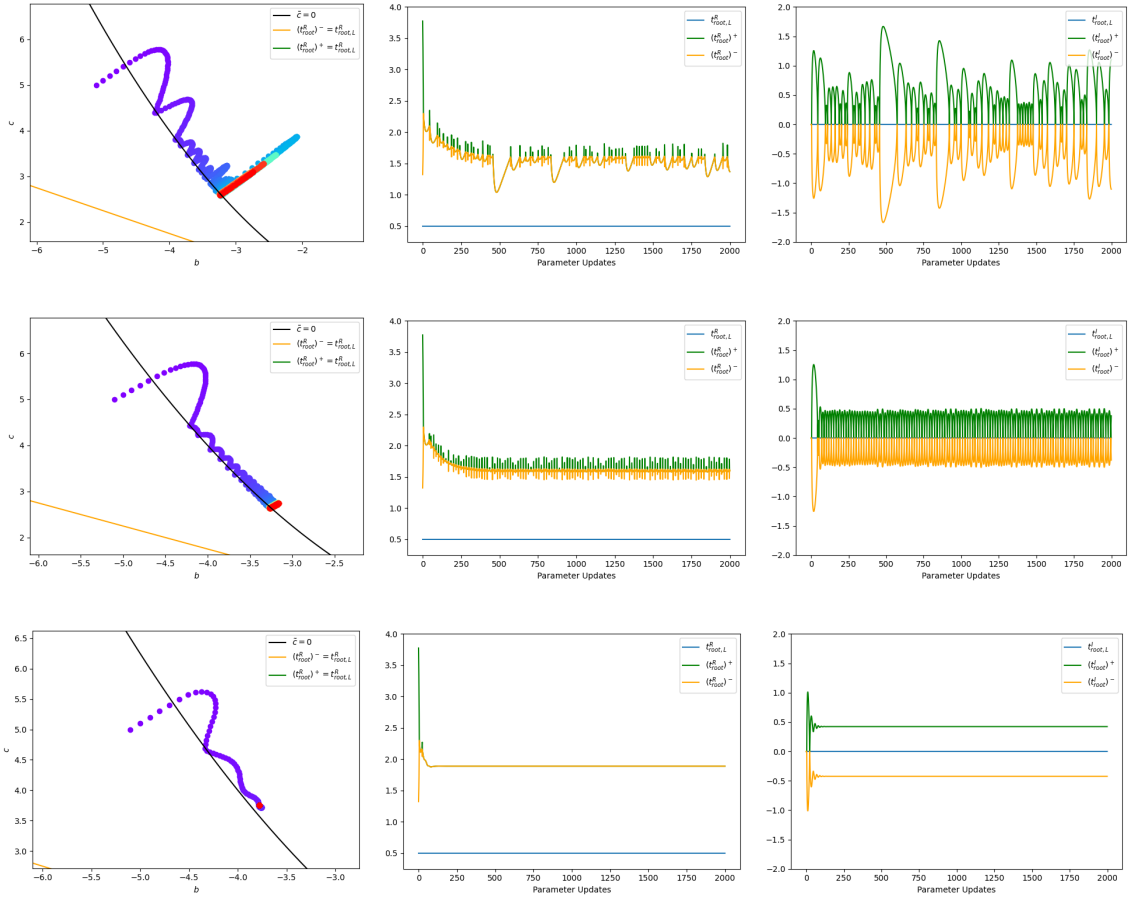

	\centering
    \foreach \deriv in {clamped_given_10000.0, clamped_given_10.0, clamped_given_1.0} {
	\begin{subfigure}[b]{0.3\textwidth}
		\centering
		\includegraphics[width=\textwidth]{figures/quadratic-section/smoothing-subsection/examples_3_parameter/example9/\deriv_reg_0.1_phase_diagram_1.png}
		\label{fig:}
	\end{subfigure}
	\begin{subfigure}[b]{0.3\textwidth}
		\centering
		\includegraphics[width=\textwidth]{figures/quadratic-section/smoothing-subsection/examples_3_parameter/example9/\deriv_reg_0.1_roots_real.png}
		\label{fig:}
	\end{subfigure}
	\begin{subfigure}[b]{0.3\textwidth}
		\centering
		\includegraphics[width=\textwidth]{figures/quadratic-section/smoothing-subsection/examples_3_parameter/example9/\deriv_reg_0.1_roots_imag.png}
		\label{fig:}
	\end{subfigure}
    }
	\caption{
    This example uses the same initial conditions as the bottom row of Figure \ref{fig:example-7}, the root solver from Section \ref{subsection:quadproposedapproach-rootsolver}, and equation \ref{eq:ourapproachderivctilde} with robust division with $M=10000$ (top row), $M=10$ (middle row), and $M=1$ (bottom row).
    } 
	\label{fig:example-22-clamping-reg_0.1-final}
\end{figure}

We first reconsider the examples from Figures \ref{fig:example-3}, \ref{fig:example-5a}, and \ref{fig:example-7} using our proposed approach; however, for the sake of a clean comparison, we remove the first column in equation \ref{eq:ourapproachderivctilde} by setting $da=0$ and holding $a=1$.
Keeping $a=1$ fixed leads to $\hat{t} = at = t$, $\hat{b} = b$, and $\hat{c} = ac = c$ for the change of variables in the beginning of Section \ref{subsection:computingderivatives-1p-examples} (used in Figures \ref{fig:example-3}, \ref{fig:example-5a}, and \ref{fig:example-7}).
Figure \ref{fig:example-22-clamping-normal-final} repeats the example from Figure \ref{fig:example-3} illustrating some of the results one might expect when choosing different values for $M$ in the robust division.
For larger values of $M$, the results most closely match those that would be obtained using analytic derivatives except when the analytic derivatives lead to division by small numbers causing catastrophic overflow.
Smaller values of $M$ can alleviate the oscillations along the $\tilde{c}=0$ parabola allowing for faster convergence.
Figure \ref{fig:example-22-clamping-large_first_grad-final} repeats the example from Figure \ref{fig:example-5a}.
Even with a large value of $M=10000$, the initial derivative is bounded enough to obtain convergence more than twice as fast as in Figure \ref{fig:example-5a}.
The middle row of Figure \ref{fig:example-22-clamping-normal-final} and the bottom row of Figure \ref{fig:example-22-clamping-large_first_grad-final} were chosen to illustrate an occasional lack of convergence caused by the iterates settling down towards the left endpoint of the green ray very close to the problematic $\tilde{c}=0$ parabola.
The derivatives in this region (which tend to point orthogonal to the $\tilde{c}=0$ parabola) can sometimes overcome the momentum from Adam stopping rightward motion (sometimes even driving the iterates to the left).
Although we occasionally observed such behavior, the iterates typically contain enough momentum to continue moving to the right. 
To verify this explanation, Figure \ref{fig:example-22-clamping-large_first_grad_new_t_target} illustrates what one would expect when the direction orthogonal to the $\tilde{c}=0$ parabola tends to point towards the right (instead of towards the left) in the region of interest (near the left endpoint of the green ray).
Finally, Figure \ref{fig:example-22-clamping-reg_0.1-final} repeats the example from Figure \ref{fig:example-7}.

\begin{figure}[H]
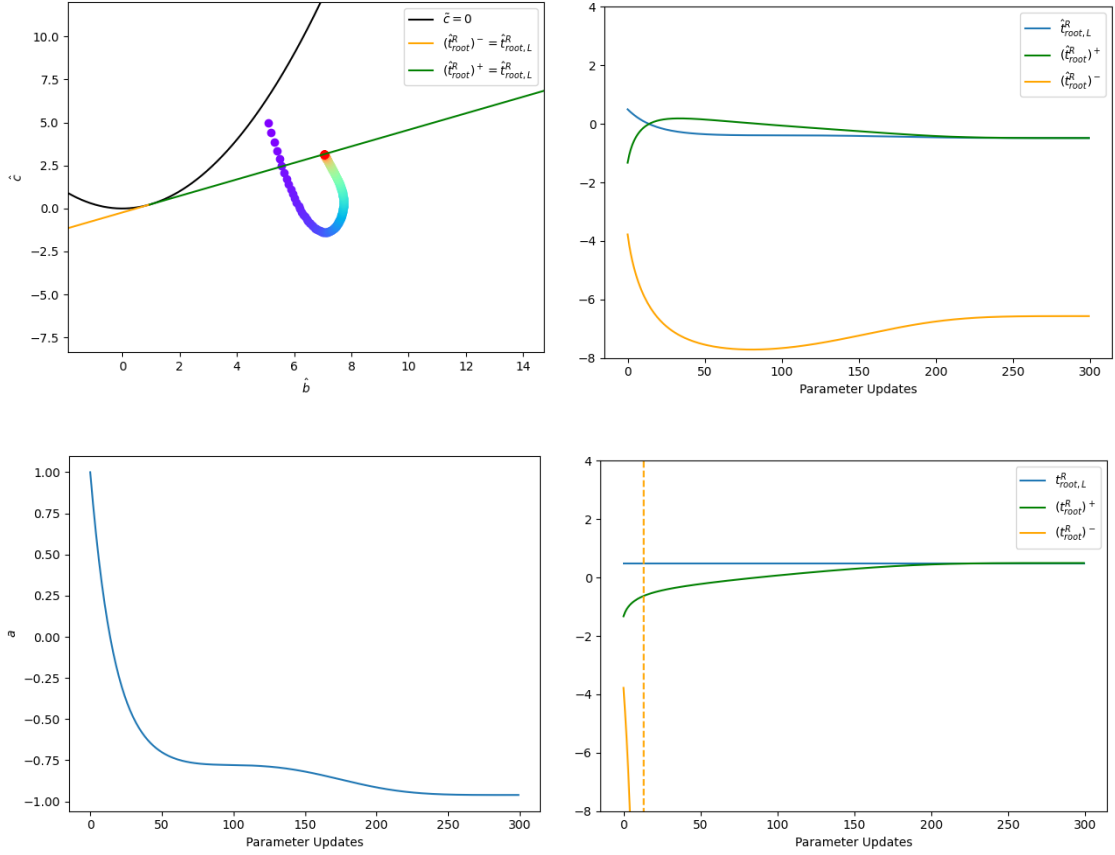

	\centering
    \foreach \deriv in {analytic} {
	\begin{subfigure}[b]{0.45\textwidth}
		\centering
		\includegraphics[width=\textwidth]{figures/quadratic-section/smoothing-subsection/examples_3_parameter/example23/\deriv_normal_fast_phase_diagram_2.png}
		\label{fig:}
	\end{subfigure}
	\begin{subfigure}[b]{0.45\textwidth}
		\centering
		\includegraphics[width=\textwidth]{figures/quadratic-section/smoothing-subsection/examples_3_parameter/example23/hats_\deriv_normal_fast_roots_real.png}
		\label{fig:}
	\end{subfigure}
	\begin{subfigure}[b]{0.45\textwidth}
		\centering
		\includegraphics[width=\textwidth]{figures/quadratic-section/smoothing-subsection/examples_3_parameter/example23/\deriv_normal_fast_a.png}
		\label{fig:}
	\end{subfigure}
	\begin{subfigure}[b]{0.45\textwidth}
		\centering
		\includegraphics[width=\textwidth]{figures/quadratic-section/smoothing-subsection/examples_3_parameter/example23/alt_\deriv_normal_fast_roots_real.png}
		\label{fig:}
	\end{subfigure}
    }
	\caption{
    This example demonstrates what happens when $a\to0$ and subsequently changes sign.
    Even though $(\hat{t}^R_\text{root})^\pm$ are well behaved, $(t^R_\text{root})^- \to -\infty$ before jumping discontinuously towards $+\infty$ when $a$ changes sign;
    meanwhile, $(\hat{t}^R_\text{root})^+\to\ttargethat^R$ and $(t^R_\text{root})^+\to\ttarget^R$ as expected (see Section \ref{subsection:quadproposedapproach-branchselection}).
    Note that the dotted yellow line in the lower right subfigure does not include any actual values of $(t^R_\text{root})^-$, but is shown to indicate the jump from large negative values to large positive values.
    } 
	\label{fig:normal_fast_example}
\end{figure}

In the subsequent examples, $a$ is allowed to vary; thus, the first column in equation \ref{eq:ourapproachderivctilde} is included.
For the sake of visualization only, we still plot some results using the change of variables $\hat{t} = at$, $\hat{b} = b$, and $\hat{c} = ac$.
The line of acceptable solutions
$\hat{c} = - \ttargethat^R\hat{b} -(\ttargethat^R)^2 = - a\ttarget^R\hat{b} -a^2(\ttarget^R)^2$
varies as $a$ varies; thus, to minimize confusion, we only plot it for the last iteration.
In spite of the line moving around, the analysis leading to Figure \ref{fig:t-target} is still valid implying that $(\hat{t}^R_\text{root})^-=\ttargethat^R$ on the left and $(\hat{t}^R_\text{root})^+=\ttargethat^R$ on the right.
Finally, note that $\ttargethat^R = a \ttarget^R$ varies as $a$ varies, as can be seen in Figure \ref{fig:normal_fast_example} (top right).

\begin{figure}[H]
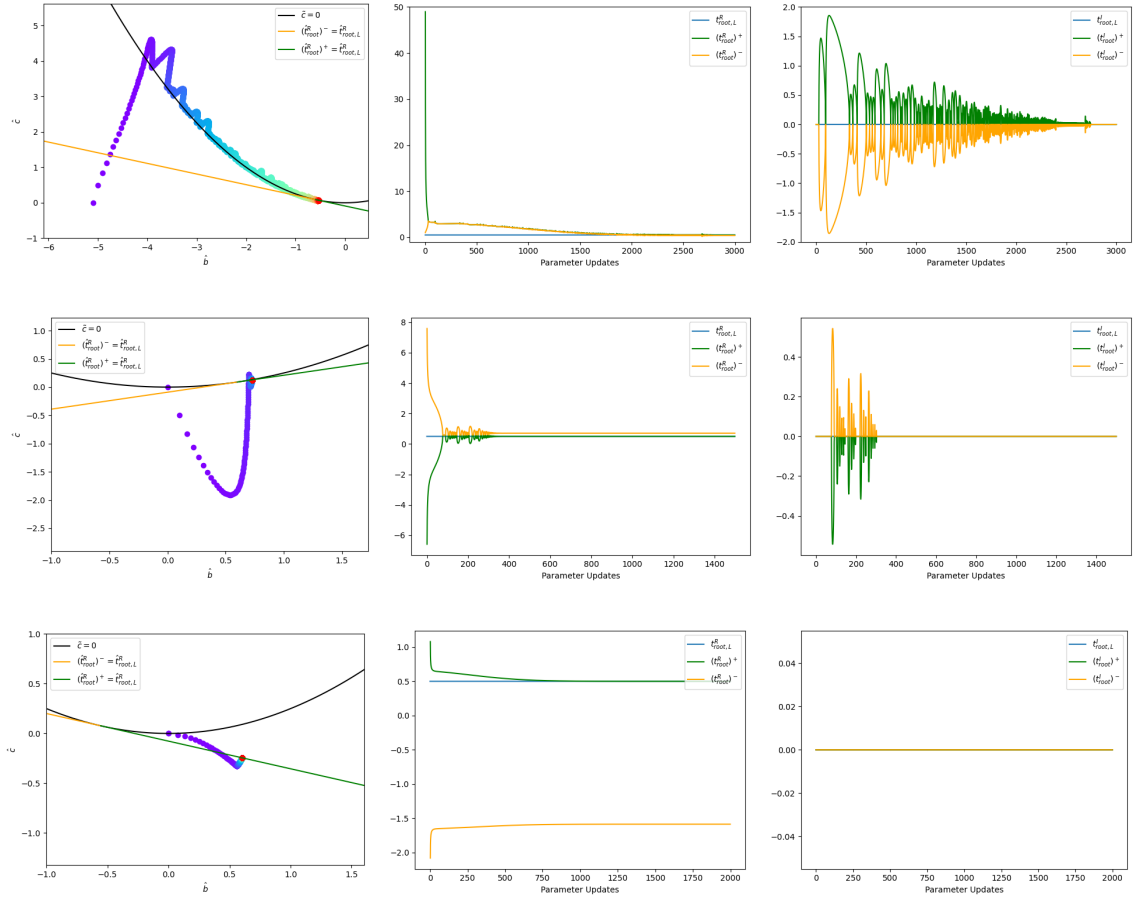

	\centering
    \foreach \subexample/\figcaption/\phasediagramnum/\deriv in {bad_a/$[a~b~c]=[0~-5.1~5]$/1/clamped_given_1000.001, bad_ab/$[a~b~c]=[0~0~5]$/1/clamped_given_1000.0, bad_abc/$[a~b~c]=[0~0~0]$/1/clamped_given_1000.0}{
	\begin{subfigure}[b]{0.3\textwidth}
		\centering
		\includegraphics[width=\textwidth]{figures/quadratic-section/smoothing-subsection/examples_3_parameter/example1/\deriv_\subexample_phase_diagram_\phasediagramnum.png}
		\label{fig:}
	\end{subfigure}
	\begin{subfigure}[b]{0.3\textwidth}
		\centering
		\includegraphics[width=\textwidth]{figures/quadratic-section/smoothing-subsection/examples_3_parameter/example1/\deriv_\subexample_roots_real.png}
		\label{fig:}
	\end{subfigure}
	\begin{subfigure}[b]{0.3\textwidth}
		\centering
		\includegraphics[width=\textwidth]{figures/quadratic-section/smoothing-subsection/examples_3_parameter/example1/\deriv_\subexample_roots_imag.png}
		\label{fig:}
	\end{subfigure}
    }
	\caption{
    For this example, we set the robust division parameter $M=1000$.
    The top row starts with $[a,b,c]^T=[0,-5.1,5]^T$, the middle row starts with $[a,b,c]^T=[0,0,5]^T$, and the bottom row starts with $[a,b,c]^T=[0,0,0]^T$.
    }
	\label{fig:example-22-clamping-robust-all}
\end{figure}

Next, we demonstrate robustness with regard to degeneracies. 
Figure \ref{fig:example-22-clamping-robust-all} (top row) starts with $[a,b,c]^T=[0,-5.1,5]^T$, and Table \ref{tab:bad_a} shows the first few iterations.
In spite of $(t^R_\text{root})^+$ not being the linear root and thus starting out at our maximum allowable value of $10^{150}$, our proposed approach recovers to obtain $(t^R_\text{root})^+=\ttarget$ as desired.
The robust root solver in Section \ref{subsection:quadproposedapproach-rootsolver} used a pseudo-sign of $a>0$ in order to obtain $(t^R_\text{root})^+ = 10^{150}$.
As discussed in Remark \ref{re:clampingderiv}, $2a t^R_\text{root} +b$ is clamped to be positive in equation \ref{eq:ourapproachderivctilde} leading to $\frac{\partial (t^R_\text{root})^+}{\partial a}<0$;
then, $\frac{\partial L}{\partial a} = \frac{\partial L}{\partial (t^R_\text{root})^+} \frac{\partial (t^R_\text{root})^+}{\partial a} = -10^{153}$ since $\frac{\partial L}{\partial (t^R_\text{root})^+} \approx (t^R_\text{root})^+ = 10^{150}$ and $\frac{\partial (t^R_\text{root})^+}{\partial a} = -M = -1000$ via clamping.
Figure \ref{fig:example-22-clamping-robust-all} (middle row) starts with $[a,b,c]^T=[0,0,5]^T$, and Table \ref{tab:bad_ab} shows the first few iterations.
Since $c>0$, the robust root solver in Section \ref{subsection:quadproposedapproach-rootsolver} assumes $a<0$ to obtain $(t^R_\text{root})^\pm = \mp 10^{150}$ respectively;
then, $\frac{\partial L}{\partial a} = 10^{153}$ since $\frac{\partial L}{\partial (t^R_\text{root})^+} \approx -10^{150}$ and $\frac{\partial (t^R_\text{root})^+}{\partial a} = -M = -1000$ via clamping.
Figure \ref{fig:example-22-clamping-robust-all} (bottom row) starts with $[a,b,c]^T=[0,0,0]^T$, and Table \ref{tab:bad_abc} shows the first few iterations.
The robust root solver in Section \ref{subsection:quadproposedapproach-rootsolver} used a pseudo-sign of $a>0$ and sets $(t^R_\text{root})^\pm = 0$;
then, $\frac{\partial L}{\partial c} = 500$ since $\frac{\partial L}{\partial (t^R_\text{root})^+} = (t^R_\text{root})^+ - \ttarget^R = -.5$ and $\frac{\partial (t^R_\text{root})^+}{\partial c} = -M = -1000$ via clamping.
This leads to $c<0$ in the next iteration, and the method recovers in a manner similar to the $a=b=0$ example shown in Table \ref{tab:bad_ab}.
\begin{table}[H]
\footnotesize
\centering
\begin{tabular}{|c|c|c|c|c|c|c|c|c|c|}
\hline
Itr.&$a$&$b$&$c$&$(t_\text{root}^R)^+$&$(t_\text{root}^R)^-$&$\tilde{c}$&$\frac{\partial L}{\partial a}$&$\frac{\partial L}{\partial b}$&$\frac{\partial L}{\partial c}$	 \\ \hline 
0	&	0.00e+00	&	-5.10e+00	&	5.00e+00	&	1.00e+150	&	9.80e-01	&	-6.50e+00	&	-1.00e+153	&	-1.00e+03	&	-1.00e-147	 \\ \hline 
1	&	1.00e-01	&	-5.00e+00	&	5.00e+00	&	4.90e+01	&	1.02e+00	&	-5.75e+00	&	-2.43e+04	&	-4.95e+02	&	-1.01e+01	 \\ \hline 
2	&	1.67e-01	&	-4.91e+00	&	5.07e+00	&	2.83e+01	&	1.07e+00	&	-5.17e+00	&	-4.90e+03	&	-1.73e+02	&	-6.11e+00	 \\ \hline 
\end{tabular}
\caption{Starting with $[a, b, c]^T=[0, -5.1, 5]^T$}
\label{tab:bad_a}
\end{table}

\begin{table}[H]
\footnotesize
\centering
\begin{tabular}{|c|c|c|c|c|c|c|c|c|c|}
\hline
Itr.&$a$&$b$&$c$&$(t_\text{root}^R)^+$&$(t_\text{root}^R)^-$&$\tilde{c}$&$\frac{\partial L}{\partial a}$&$\frac{\partial L}{\partial b}$&$\frac{\partial L}{\partial c}$	 \\ \hline 
0	&	0.00e+00	&	0.00e+00	&	5.00e+00	&	-1.00e+150	&	1.00e+150	&	0.00e+00	&	1.00e+153	&	-1.00e+03	&	1.00e-147	 \\ \hline 
1	&	-1.00e-01	&	1.00e-01	&	5.00e+00	&	-6.59e+00	&	7.59e+00	&	-5.02e-01	&	2.17e+02	&	-3.29e+01	&	5.00e+00	 \\ \hline 
2	&	-1.67e-01	&	1.69e-01	&	4.93e+00	&	-4.95e+00	&	5.96e+00	&	-8.30e-01	&	7.32e+01	&	-1.48e+01	&	2.99e+00	 \\ \hline 
\end{tabular}
\caption{Starting with $[a, b, c]^T=[0, 0, 5]^T$}
\label{tab:bad_ab}
\end{table}

\begin{table}[H]
\footnotesize
\centering
\begin{tabular}{|c|c|c|c|c|c|c|c|c|c|}
\hline
Itr.&$a$&$b$&$c$&$(t_\text{root}^R)^+$&$(t_\text{root}^R)^-$&$\tilde{c}$&$\frac{\partial L}{\partial a}$&$\frac{\partial L}{\partial b}$&$\frac{\partial L}{\partial c}$	 \\ \hline 
0	&	0.00e+00	&	0.00e+00	&	0.00e+00	&	0.00e+00	&	0.00e+00	&	0.00e+00	&	0.00e+00	&	0.00e+00	&	5.00e+02	 \\ \hline 
1	&	0.00e+00	&	0.00e+00	&	-1.00e-01	&	1.00e+150	&	-1.00e+150	&	0.00e+00	&	-1.00e+153	&	-1.00e+03	&	-1.00e-147	 \\ \hline 
2	&	7.44e-02	&	7.44e-02	&	-1.67e-01	&	1.08e+00	&	-2.08e+00	&	-1.38e-02	&	-2.87e+00	&	-2.66e+00	&	-2.46e+00	 \\ \hline 
\end{tabular}
\caption{Starting with $[a, b, c]^T=[0, 0, 0]^T$}
\label{tab:bad_abc}
\end{table}

Finally, we repeat the examples shown in Figures \ref{fig:example-3}, \ref{fig:example-5a}, and \ref{fig:example-7} using the proposed approach with $a$, $b$, and $c$ allowed to vary.
Figure \ref{fig:example-22-clamping-normal} repeats the example shown in Figure \ref{fig:example-3} (and Figure \ref{fig:example-22-clamping-normal-final}).
Figures \ref{fig:example-22-clamping-large_first_grad} and \ref{fig:example-22-clamping-large_first_grad-other} repeat the example shown in Figure \ref{fig:example-5a} (and Figure \ref{fig:example-22-clamping-large_first_grad-final}).
Figure \ref{fig:example-22-clamping-reg_0.1} repeats the example shown in Figure \ref{fig:example-7} (and Figure \ref{fig:example-22-clamping-reg_0.1-final}).

\begin{figure}[H]
	\centering
    \foreach \deriv in {clamped_given_10000.0, clamped_given_10.0, clamped_given_1.0} {
	\begin{subfigure}[b]{0.3\textwidth}
		\centering
		\includegraphics[width=\textwidth]{figures/quadratic-section/smoothing-subsection/examples_3_parameter/example9/\deriv_large_first_grad_phase_diagram_1.png}
		\label{fig:}
	\end{subfigure}
	\begin{subfigure}[b]{0.3\textwidth}
		\centering
		\includegraphics[width=\textwidth]{figures/quadratic-section/smoothing-subsection/examples_3_parameter/example9/\deriv_large_first_grad_roots_real.png}
		\label{fig:}
	\end{subfigure}
	\begin{subfigure}[b]{0.3\textwidth}
		\centering
		\includegraphics[width=\textwidth]{figures/quadratic-section/smoothing-subsection/examples_3_parameter/example9/\deriv_large_first_grad_roots_imag.png}
		\label{fig:}
	\end{subfigure}
    }
	\caption{
    This example uses the same initial conditions as Figure \ref{fig:example-5a}, the root solver from Section \ref{subsection:quadproposedapproach-rootsolver}, and equation \ref{eq:ourapproachderivctilde} with robust division with $M=10000$ (top row), $M=10$ (middle row), and $M=1$ (bottom row). The lack of convergence in the bottom row is caused by the iterates settling down towards the left endpoint of the green ray near the $\tilde{c}=0$ parabola (as can be seen by the nonzero imaginary part of the roots). See also Figure \ref{fig:example-22-clamping-large_first_grad_new_t_target}.
    } 
	\label{fig:example-22-clamping-large_first_grad-final}
\end{figure}


\foreach \subexample/\figcaption/\phasediagramnum in {normal/Figure \ref{fig:example-3}/1}{
\begin{figure}[H]
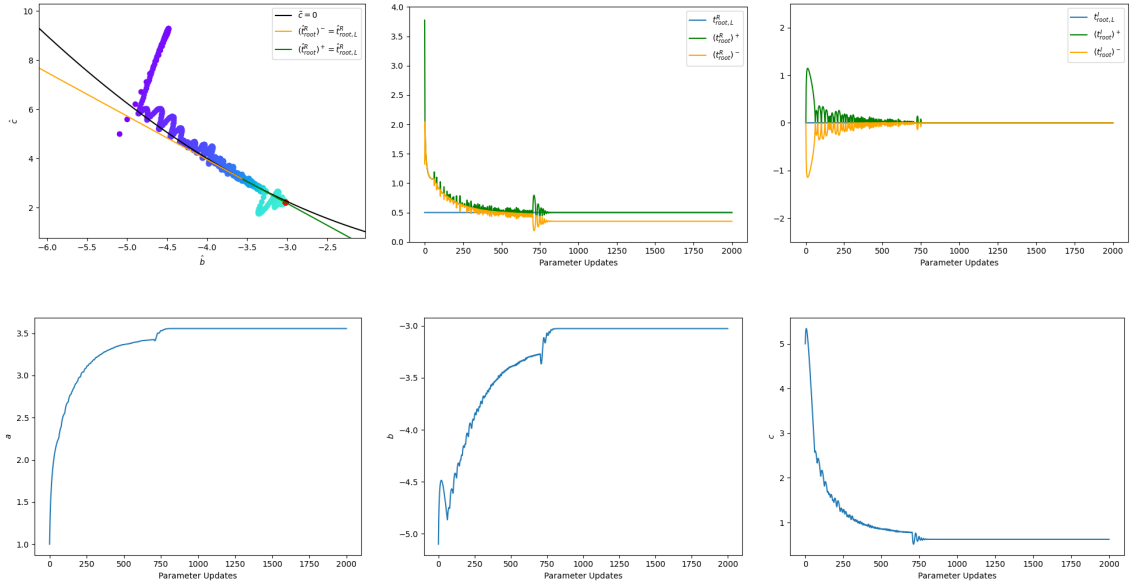

	\centering
    \foreach \deriv in {clamped_given_1000.0} {
	\begin{subfigure}[b]{0.3\textwidth}
		\centering
		\includegraphics[width=\textwidth]{figures/quadratic-section/smoothing-subsection/examples_3_parameter/example1/\deriv_\subexample_phase_diagram_\phasediagramnum.png}
		\label{fig:}
	\end{subfigure}
	\begin{subfigure}[b]{0.3\textwidth}
		\centering
		\includegraphics[width=\textwidth]{figures/quadratic-section/smoothing-subsection/examples_3_parameter/example1/\deriv_\subexample_roots_real.png}
		\label{fig:}
	\end{subfigure}
	\begin{subfigure}[b]{0.3\textwidth}
		\centering
		\includegraphics[width=\textwidth]{figures/quadratic-section/smoothing-subsection/examples_3_parameter/example1/\deriv_\subexample_roots_imag.png}
		\label{fig:}
	\end{subfigure}
	\begin{subfigure}[b]{0.3\textwidth}
		\centering
		\includegraphics[width=\textwidth]{figures/quadratic-section/smoothing-subsection/examples_3_parameter/example1/\deriv_\subexample_a}
		\label{fig:}
	\end{subfigure}
	\begin{subfigure}[b]{0.3\textwidth}
		\centering
		\includegraphics[width=\textwidth]{figures/quadratic-section/smoothing-subsection/examples_3_parameter/example1/\deriv_\subexample_b}
		\label{fig:}
	\end{subfigure}
	\begin{subfigure}[b]{0.3\textwidth}
		\centering
		\includegraphics[width=\textwidth]{figures/quadratic-section/smoothing-subsection/examples_3_parameter/example1/\deriv_\subexample_c}
		\label{fig:}
	\end{subfigure}
    }
	\caption{
    This figure repeats the example shown in Figures \ref{fig:example-3} and \ref{fig:example-22-clamping-normal-final} (using $M=1000$). 
    } 
	\label{fig:example-22-clamping-\subexample}
\end{figure}
}

\foreach \subexample/\figcaption/\phasediagramnum in {large_first_grad/Figure \ref{fig:example-5a}/1}{
\begin{figure}[H]
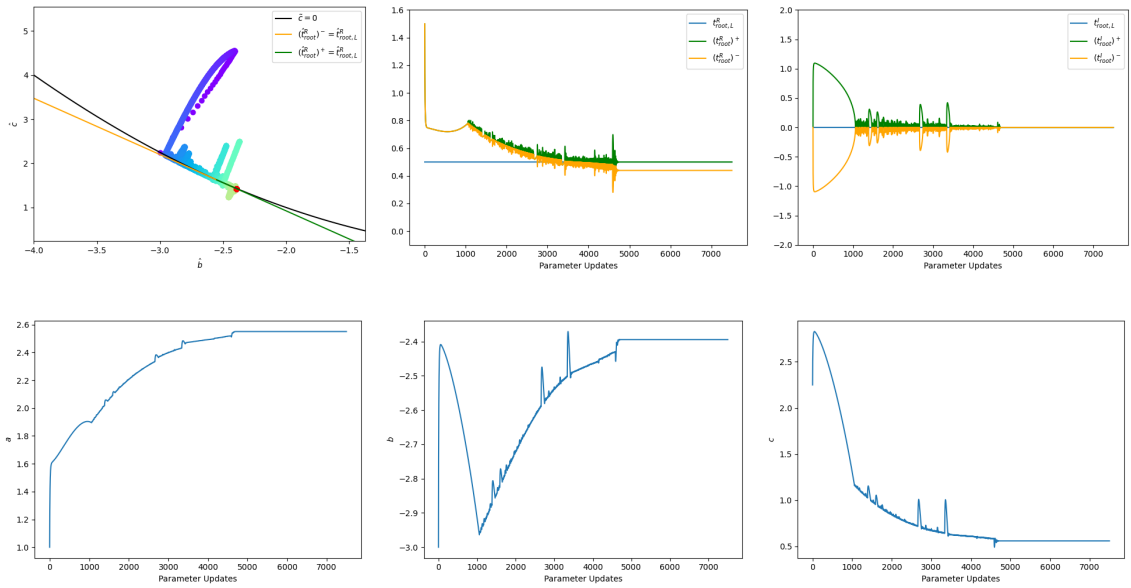

	\centering
    \foreach \deriv in {clamped_given_1000.0} {
	\begin{subfigure}[b]{0.3\textwidth}
		\centering
		\includegraphics[width=\textwidth]{figures/quadratic-section/smoothing-subsection/examples_3_parameter/example1/\deriv_\subexample_phase_diagram_\phasediagramnum.png}
		\label{fig:}
	\end{subfigure}
	\begin{subfigure}[b]{0.3\textwidth}
		\centering
		\includegraphics[width=\textwidth]{figures/quadratic-section/smoothing-subsection/examples_3_parameter/example1/\deriv_\subexample_roots_real.png}
		\label{fig:}
	\end{subfigure}
	\begin{subfigure}[b]{0.3\textwidth}
		\centering
		\includegraphics[width=\textwidth]{figures/quadratic-section/smoothing-subsection/examples_3_parameter/example1/\deriv_\subexample_roots_imag.png}
		\label{fig:}
	\end{subfigure}
	\begin{subfigure}[b]{0.3\textwidth}
		\centering
		\includegraphics[width=\textwidth]{figures/quadratic-section/smoothing-subsection/examples_3_parameter/example1/\deriv_\subexample_a}
		\label{fig:}
	\end{subfigure}
	\begin{subfigure}[b]{0.3\textwidth}
		\centering
		\includegraphics[width=\textwidth]{figures/quadratic-section/smoothing-subsection/examples_3_parameter/example1/\deriv_\subexample_b}
		\label{fig:}
	\end{subfigure}
	\begin{subfigure}[b]{0.3\textwidth}
		\centering
		\includegraphics[width=\textwidth]{figures/quadratic-section/smoothing-subsection/examples_3_parameter/example1/\deriv_\subexample_c}
		\label{fig:}
	\end{subfigure}
    }
	\caption{
    This figure repeats the example shown in Figures \ref{fig:example-5a} and \ref{fig:example-22-clamping-large_first_grad-final} (with $M=1000$). 
    Note that Figures \ref{fig:example-5a} and \ref{fig:example-22-clamping-large_first_grad-final} show 20000 iterations, whereas only 7500 iterations are shown here (for the sake of comparison with Figure \ref{fig:example-22-clamping-large_first_grad-other}).
    } 
	\label{fig:example-22-clamping-\subexample}
\end{figure}
}

\foreach \subexample/\figcaption/\phasediagramnum in {large_first_grad/Figure \ref{fig:example-5a}/1}{
\begin{figure}[H]
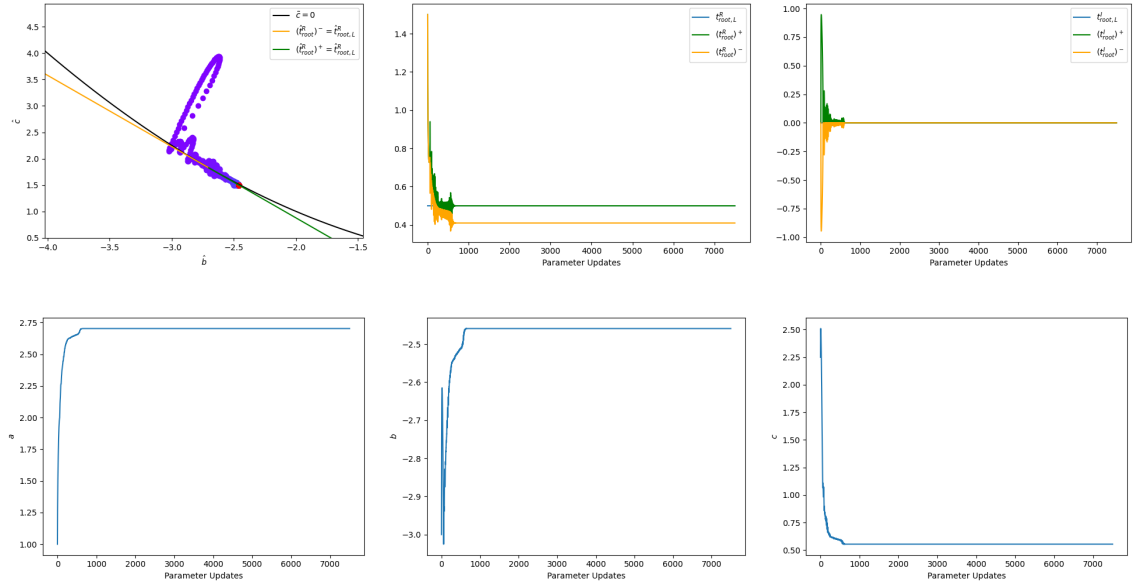

	\centering
    \foreach \deriv in {clamped_given_10.0001} {
	\begin{subfigure}[b]{0.3\textwidth}
		\centering
		\includegraphics[width=\textwidth]{figures/quadratic-section/smoothing-subsection/examples_3_parameter/example1/\deriv_\subexample_phase_diagram_\phasediagramnum.png}
		\label{fig:}
	\end{subfigure}
	\begin{subfigure}[b]{0.3\textwidth}
		\centering
		\includegraphics[width=\textwidth]{figures/quadratic-section/smoothing-subsection/examples_3_parameter/example1/\deriv_\subexample_roots_real.png}
		\label{fig:}
	\end{subfigure}
	\begin{subfigure}[b]{0.3\textwidth}
		\centering
		\includegraphics[width=\textwidth]{figures/quadratic-section/smoothing-subsection/examples_3_parameter/example1/\deriv_\subexample_roots_imag.png}
		\label{fig:}
	\end{subfigure}
	\begin{subfigure}[b]{0.3\textwidth}
		\centering
		\includegraphics[width=\textwidth]{figures/quadratic-section/smoothing-subsection/examples_3_parameter/example1/\deriv_\subexample_a}
		\label{fig:}
	\end{subfigure}
	\begin{subfigure}[b]{0.3\textwidth}
		\centering
		\includegraphics[width=\textwidth]{figures/quadratic-section/smoothing-subsection/examples_3_parameter/example1/\deriv_\subexample_b}
		\label{fig:}
	\end{subfigure}
	\begin{subfigure}[b]{0.3\textwidth}
		\centering
		\includegraphics[width=\textwidth]{figures/quadratic-section/smoothing-subsection/examples_3_parameter/example1/\deriv_\subexample_c}
		\label{fig:}
	\end{subfigure}
    }
	\caption{
    This figure repeats the example shown in Figure \ref{fig:example-22-clamping-large_first_grad} but with $M=10$ in order to demonstrate the accelerated convergence. 
    } 
	\label{fig:example-22-clamping-\subexample-other}
\end{figure}
}

\foreach \subexample/\figcaption/\phasediagramnum in {reg_0.1/the top row of Figure \ref{fig:example-7} (where $\eta=.1$)/2}{
\begin{figure}[H]
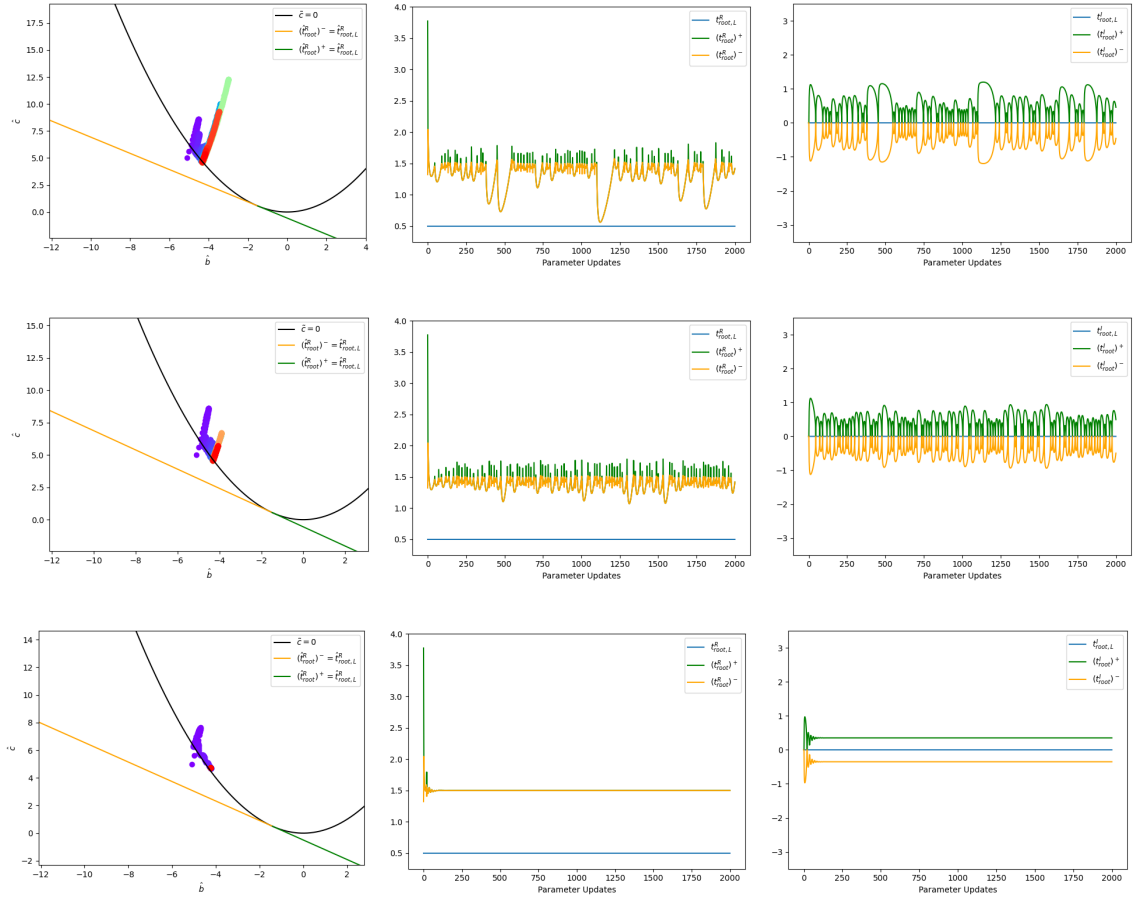

	\centering
    \foreach \deriv in {clamped_given_1000.0, clamped_given_10.0, clamped_given_1.0} {
	\begin{subfigure}[b]{0.3\textwidth}
		\centering
		\includegraphics[width=\textwidth]{figures/quadratic-section/smoothing-subsection/examples_3_parameter/example1/\deriv_\subexample_phase_diagram_\phasediagramnum.png}
		\label{fig:}
	\end{subfigure}
	\begin{subfigure}[b]{0.3\textwidth}
		\centering
		\includegraphics[width=\textwidth]{figures/quadratic-section/smoothing-subsection/examples_3_parameter/example1/\deriv_\subexample_roots_real.png}
		\label{fig:}
	\end{subfigure}
	\begin{subfigure}[b]{0.3\textwidth}
		\centering
		\includegraphics[width=\textwidth]{figures/quadratic-section/smoothing-subsection/examples_3_parameter/example1/\deriv_\subexample_roots_imag.png}
		\label{fig:}
	\end{subfigure}
    }
	\caption{
    This figure repeats the example shown in Figures \ref{fig:example-7} (bottom row) and \ref{fig:example-22-clamping-reg_0.1-final} using $M=1000$ (top row), $M=10$ (middle row), and $M=1$ (bottom row).    
    } 
	\label{fig:example-22-clamping-\subexample}
\end{figure}
}

\clearpage

\section{Cubic Equations}
\label{section:cubic}
In Section \ref{subsection:cubic-rootsolver}, we begin by introducing a novel and robust approach to finding roots (both real and complex) for cubic equations, even when any or all of the coefficients degenerate to be identically zero.
In Section \ref{subsection:cubic-implicitdiff}, we briefly address implicit differentiation.
In Section \ref{subsection:cubic-understandasymptotics}, we analyze the cubic equation in its standard reduced canonical form where the roots are a function of only two parameters.
In Section \ref{subsection:cubic-newcanonicalform}, we propose a new reduced canonical form, which enables a more robust treatment of both the roots and the derivatives of the roots with respect to the parameters.
Finally, Section \ref{section:cubicproposedapproach} details our proposed approach for cubic equations.

\subsection{Cubic Root Solver}
\label{subsection:cubic-rootsolver}
Whereas \cite{bridson2002robust} only considered cubic roots that were in a pre-specified time interval (indicating a potential collision during a time step), we instead would like to find all roots (including complex roots and roots that blow up). If any of the coefficients has magnitude larger than 1, then we divide through by it; in addition, we flip the sign of all of the coefficients whenever $q<0$. The resulting cubic equation can be written as $f(t, \vec{p}) = qt^3 + at^2 + bt + c$ where $0\leq q \leq 1$, $|a| \leq 1$, $|b| \leq 1$, and $|c| \leq 1$.

Rewriting the cubic equation as
\begin{equation}
    \label{eq:cubic_rewrite_q1}
	f(t, \vec{p}) = \frac{1}{3} t^2 \left(qt+3a\right) + \frac{1}{3} t \left(qt^2+3b\right) + \frac{1}{3} \left(qt^3+3c\right) = 0
\end{equation}
facilitates the computation of an interval $[t_\text{left}, t_\text{right}]$ containing the roots. $t_\text{left}$ is determined by $t_\text{left}<0$, $qt_\text{left} < -3a$, $q(t_\text{left})^2 > -3b$, $q(t_\text{left})^3 < -3c$, and $t_\text{right}$ is determined by $t_\text{right}>0$, $qt_\text{right} > -3a$, $q(t_\text{right})^2 > -3b$, $q(t_\text{right})^3 > -3c$. Choosing $t_\text{left} < -3/q$ and $t_\text{right} > 3/q$ is sufficient. When $q$ is large enough, we proceed as follows: To determine monotonic intervals, we examine the roots of the derivative $3qt^2 + 2at + b = 0$. When these critical points are complex or repeated (i.e.~$a^2-3qb\leq0$), the cubic is monotonically increasing with one (possibly repeated) real root (which can be found robustly with a mixture of Newton's method and bisection). Otherwise, the critical points can be computed via
\begin{subequations}
	\label{eq:critical_points}
	\begin{align}
		\leftcrit
		&=
		\begin{cases}
			\frac{-a - \sqrt{a^2-3qb}}{3q}
			& \text{if }a\geq0\\
			\frac{b}{-a + \sqrt{a^2-3qb}}
			& \text{if } a<0\\
		\end{cases}\\
		\rightcrit
		&=
		\begin{cases}
			\frac{-a + \sqrt{a^2-3qb}}{3q}
			& \text{if } a\leq0\\
			\frac{b}{-a - \sqrt{a^2-3qb}}
			& \text{if }a>0\\
		\end{cases}
	\end{align}
\end{subequations}
to avoid cancellation error (see e.g.~\cite{heath2018scientific, harari2023computation, di1750produzioni}). This should result in $t_\text{left} < \leftcrit < \rightcrit < t_\text{right}$; however, critical points can be discarded (leaving less intervals as candidates for sign changes) whenever numerical errors cause these conditions to be violated. The signs of $f(\leftcrit)$ and $f(\rightcrit)$ can be used to determine intervals that contain roots (which can be found robustly with a mixture of Newton's method and bisection). After finding one real root, $r_1$, the cubic can be factored as
\begin{equation}
    \label{eq:cubic_factored}
    (t-r_1)(qt^2 + (qr_1 + a)t + qr_1^2 + ar_1 + b) = 0
\end{equation}
so that the remaining roots are given by
\begin{equation}
    \label{eq:remaining_roots}
    r_{2,3} = \frac{-(qr_1+a) \pm \sqrt{(qr_1+a)^2 - 4q(qr_1^2 + ar_1 + b)}}{2q}
\end{equation}
noting that the most robust way to find $r_2$ and $r_3$ from the quadratic factor in equation \ref{eq:cubic_factored} is to use the robust quadratic root-solver in Section \ref{subsection:quadproposedapproach-rootsolver}. When exactly two roots are found via the iterative solver, we recommend using the root that would give the largest positive number under the square root in equation \ref{eq:remaining_roots} when factoring the cubic in equation \ref{eq:cubic_factored}. A similar strategy can be used even when three roots are found via the iterative solver, especially when one suspects that numerical errors may have led to less accurate than desired roots.

\textit{
\RemarkCounter{re:imag_bound}
Given that the real roots (and thus $r_1$) are bounded by $\pm\frac{3}{q}$, equation \ref{eq:remaining_roots} leads to a bound of $\pm\frac{2}{q}$ on the real part and $\pm\frac{\sqrt{38}}{2q}$ $\left(\approx \pm\frac{3.1}{q}\right)$ on the imaginary parts of the complex roots;
thus, whenever a real root or a real/imaginary part of a complex root grows large, $q$ must be small.
}

When $q$ is too small to robustly use $[-\frac{3}{q}, \frac{3}{q}]$ to bound the roots, one could attempt to use a smaller interval such as $[-t_\text{max},+t_\text{max}]$ and proceed with the approach outlined above. One could also replace $t$ with $t^{-1}$ to obtain a reversed cubic $ct^3 + bt^2 + at + q = 0$, flipping the sign of all of the coefficients if necessary so that $c\geq0$. When $c$ is too small to robustly use $[-\frac{3}{c}, \frac{3}{c}]$, one could (again) attempt to use a smaller interval. Note that either the original or the reversed cubic is guaranteed to have a root in the interval $[-1, 1]$, since the left endpoint evaluates to either $-q+a-b+c$ or $-c+b-a+q$ while the right endpoint evaluates to $q+a+b+c$ for both cubics. Note that we reverse the quadratic factor in equation \ref{eq:cubic_factored} to be 
\begin{equation}
\label{eq:doubleflip}
(c r_1^2 + br_1+a) t^2 + (cr_1+b)t+c=0
\end{equation}
whenever $r_1$ is obtained from the reversed cubic, in order to avoid the problematic case of having to flip an identically zero $r_2$ or $r_3$ (see Remark \ref{re:reversed_quadratic}). 

It is problematic to flip an identically zero $r_1$ from the reversed cubic, which should only occur when $q=0$; however, when $q$ is smaller than the tolerance of the iterative solver, one may also obtain an identically zero $r_1$.
We address this as follows: As $q\to0$, the cubic equation degenerates into a quadratic equation. Assuming that two of the roots are approximately governed by $at^2 + bt + c = 0$, one can uncover the behavior of the remaining root from
\begin{equation}
    \label{eq:cubic_quadratic_division1}
	\frac{qt^3 + at^2 + bt + c}{at^2 + bt + c}
    = \frac{q}{a}\left(t + \frac{a}{q} - \frac{b}{a} + \frac{1}{a} \frac{(b^2-ac)t + bc}{at^2 + bt + c}\right)
    = 0
\end{equation}
indicating that the remaining root $r_1$ blows up like $\frac{-a}{q}$ as $\frac{q}{a} \to 0$ via
\begin{equation}
    \label{eq:root_expansion1}
    r_1
    = -\left(\frac{q}{a}\right)^{-1} + \frac{b}{a}\left(\frac{q}{a}\right)^{0} + \left(\frac{b^2}{a^2} - \frac{c}{a}\right)\left(\frac{q}{a}\right)^{1} + O\left(\left(\frac{q}{a}\right)^2\right)
\end{equation}
where the $O\left(\frac{q}{a}\right)$ term was obtained from the last term in equation \ref{eq:cubic_quadratic_division1} as $t\to\pm\infty$. Plugging this $r_1$ into equation \ref{eq:remaining_roots} gives
\begin{equation}
        r_{2,3}
        \label{eq:quadratic_with_error_terms1}
        = -\frac{b}{2a} + O\left(\frac{q}{a}\right) \pm \sqrt{\frac{b^2-4ac}{(2a)^2} + O\left(\frac{q}{a}\right)}
\end{equation}
as expected.
Note that equation \ref{eq:quadratic_with_error_terms1} is only a valid approximation as $\frac{q}{a} \to 0$ when all of the $\frac{b}{a}$ and $\frac{c}{a}$ terms are robustly bounded even when raised to various powers.
In summary, when $q=0$ identically (or the iterative solver returns $r_1=0$ from the reversed cubic), we use the quadratic root solver (from Section \ref{subsection:quadproposedapproach-rootsolver}) on $at^2 + bt + c = 0$ to find $r_2$ and $r_3$ (i.e.~consistent with equation \ref{eq:doubleflip});
then, we choose $r_1=-\sign{(a)}\infty$ (clamped to a maximum magnitude) consistent with $q>0$ in equation \ref{eq:root_expansion1}.
When $a=0$, the quadratic solver determines a pseudo-sign for $a$ (and thus $\sign{(a)}$) except when $a=b=c=0$ where we choose $r_1=0$ as a triply repeated root.

\textit{
\RemarkCounter{re:}
Notably, one would not want to be compelled into making the code for this cubic solver differentiable.
}

\subsection{Implicit Differentiation}
\label{subsection:cubic-implicitdiff}
Similar to Section \ref{subsection:quadratic-implicit-differentiation}, we write the cubic equation as
\begin{equation}
\label{eq:fcubic2d}
f(t_\text{root}; \vec{p}) = 
\begin{bmatrix}
	q(\treal_\text{root})^3 - 3q\treal_\text{root}(\timag_\text{root})^2 + a (\treal_\text{root})^2 - a (\timag_\text{root})^2 + b \treal_\text{root} + c\\
	-q(\timag_\text{root})^3 + 3q(\treal_\text{root})^2\timag_\text{root} + 2 a \treal_\text{root} \timag_\text{root} + b \timag_\text{root}\\
\end{bmatrix}
= \vec{0}
\end{equation}
letting $\theta_1$ refer to the first variable (i.e.~$t_\text{root}$) and $\theta_2$ refer to the second variable (i.e.~$\vec{p}$)
so that the derivatives 
\begin{subequations}
\label{eq:impldiffcubic} 
\begin{align}
f_{\theta_1}(t_\text{root}; \vec{p}) &= 
\begin{bmatrix}
    3q(\treal_\text{root})^2 - 3q(\timag_\text{root})^2 + 2 a \treal_\text{root} + b
    & - (6q\treal_\text{root}\timag_\text{root} + 2 a \timag_\text{root})\\
    6q\treal_\text{root}\timag_\text{root} + 2 a \timag_\text{root}
    &3q(\treal_\text{root})^2 - 3q(\timag_\text{root})^2 + 2 a \treal_\text{root} + b\\
\end{bmatrix}\\
f_{\theta_1}^{-1}(t_\text{root}; \vec{p}) &= 
\frac{1}{s}
\begin{bmatrix}
    3q(\treal_\text{root})^2 - 3q(\timag_\text{root})^2 + 2 a \treal_\text{root} + b
    & 6q\treal_\text{root}\timag_\text{root} + 2 a \timag_\text{root}\\
    -(6q\treal_\text{root}\timag_\text{root} + 2 a \timag_\text{root})
    &3q(\treal_\text{root})^2 - 3q(\timag_\text{root})^2 + 2 a \treal_\text{root} + b\\
\end{bmatrix}\\
&\text{ where }
s = (3q(\treal_\text{root})^2 - 3q(\timag_\text{root})^2 + 2 a \treal_\text{root} + b)^2 + (6q\treal_\text{root}\timag_\text{root} + 2 a \timag_\text{root})^2 \nonumber\\
f_{\theta_2}(t_\text{root}; \vec{p}) &= 
\begin{bmatrix}
	Re(z_\text{root}^3) & Re(z_\text{root}^2) & Re(z_\text{root}^1) & Re(z_\text{root}^0)\\
	Im(z_\text{root}^3) & Im(z_\text{root}^2) & Im(z_\text{root}^1) & Im(z_\text{root}^0)\\
\end{bmatrix}
\end{align}
\end{subequations}
have compact notation.
The total derivative of equation \ref{eq:fcubic2d} is $f_{\theta_1}(t_\text{root}, \vec{p}) d \theta_1 + f_{\theta_2}(t_\text{root}, \vec{p}) d \theta_2 = 0$, which can be written as $d \theta_1 = - f_{\theta_1}^{-1}(t_\text{root}, \vec{p}) f_{\theta_2}(t_\text{root}, \vec{p}) d \theta_2$ or
\begin{equation}
\label{eq:totdercubic} 
dt_\text{root}
=
- f_{\theta_1}^{-1}(t_\text{root}, \vec{p})
\begin{bmatrix}
	Re(z_\text{root}^3) & Re(z_\text{root}^2) & Re(z_\text{root}^1) & Re(z_\text{root}^0)\\
	Im(z_\text{root}^3) & Im(z_\text{root}^2) & Im(z_\text{root}^1) & Im(z_\text{root}^0)\\
\end{bmatrix}
d\vec{p}
\end{equation}
when $f_{\theta_1}$ is invertible.

\subsection{Understanding the Asymptotics}
\label{subsection:cubic-understandasymptotics}
Similar to completing the square for the quadratic equation, one can make a change of variables $\tilde{t} = t + \frac{a}{3q}$ to obtain
\begin{equation}
\label{eq:covtohatsforcubic}
q \tilde{t}^3 + \left(\frac{3bq-a^2}{3q}\right)\tilde{t} + \left(\frac{2a^3 - 9abq + 27cq^2}{27q^2}\right) = 0
\end{equation}
which can be reduced to a two parameter family by dividing by $q$ and setting $\hat{b} = \frac{3bq-a^2}{3q^2}$ and $\hat{c} = \frac{2a^3 - 9abq + 27cq^2}{27q^3}$ to obtain $\tilde{t}^3 + \hat{b}\tilde{t} + \hat{c} = 0$.

\textit{
\RemarkCounter{re:sum_of_roots}
Since $(\tilde{t}-r_1)(\tilde{t}-r_2)(\tilde{t}-r_3)=\tilde{t}^3 - (r_1 + r_2 + r_3)\tilde{t}^2 + (r_1r_2 + r_1r_3 + r_2r_3)\tilde{t} - r_1 r_2 r_3$, $r_1+r_2+r_3=0$ for this reduced cubic;
furthermore, since any imaginary parts cancel, the sum of the real parts is identically zero.
}

\begin{figure}[H]
	\centering
	\includegraphics[width=.7\textwidth]{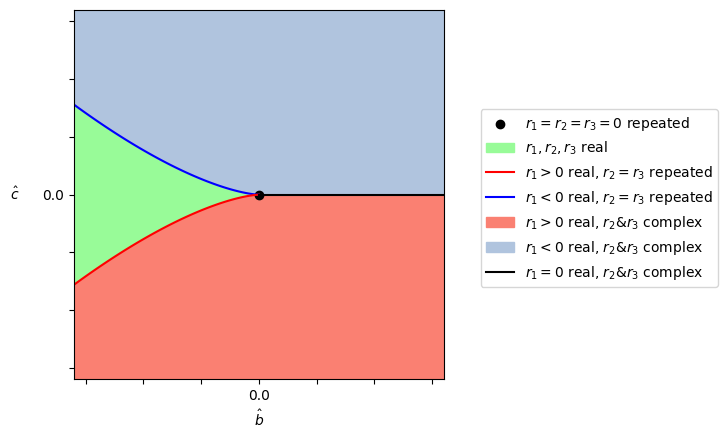}
	\caption{Root classification for the reduced cubic.}
	\label{fig:roots_phase_diagram}
\end{figure}

The critical points can be found from $3\tilde{t}^2 + \hat{b} = 0$ as $\tilde{t} = \pm \sqrt{\frac{-\hat{b}}{3}}$. In order for there to be three real roots, the reduced cubic must be non-negative at $- \sqrt{\frac{-\hat{b}}{3}}$ and non-positive at $\sqrt{\frac{-\hat{b}}{3}}$, i.e.~
$-\left(\sqrt{\frac{-\hat{b}}{3}}\right)^3 - \hat{b} \sqrt{\frac{-\hat{b}}{3}} + \hat{c} \geq 0$
and
$\left(\sqrt{\frac{-\hat{b}}{3}}\right)^3 + \hat{b} \sqrt{\frac{-\hat{b}}{3}} + \hat{c} \leq 0$ which lead to $\hat{c} \geq -2\left(\frac{-\hat{b}}{3}\right)^{3/2}$ and $\hat{c} \leq 2\left(\frac{-\hat{b}}{3}\right)^{3/2}$ respectively.
Graphs of the boundary curves in the valid $\hat{b}\leq0$ region are shown in Figure \ref{fig:roots_phase_diagram}, and the region where both inequalities are strictly valid is shown in green.
Excluding the origin (in Figure \ref{fig:roots_phase_diagram}) where zero is a triply repeated root, the smallest of the three real roots is always strictly negative and the largest is always strictly positive. 
As the cubic becomes negative at $- \sqrt{\frac{-\hat{b}}{3}}$ crossing from the green to the red region, the middle and negative roots merge to become complex while the positive root remains real; similarly, crossing from the green to the blue region merges the two larger roots. 
The boundary between the red and the blue regions occurs when the single real root is zero, which requires $\hat{c}=0$.

For the reduced cubic, equation \ref{eq:impldiffcubic}b has
$s=(3(\trealtil_\text{root})^2 - 3(\timagtil_\text{root})^2 + \hat{b})^2 + (6\trealtil_\text{root}\timagtil_\text{root})^2$ where $s=0$ only when $\timagtil_\text{root}=0$ and $3(\trealtil_\text{root})^2 + \hat{b} = 0$, i.e.~when $\tilde{t}^R_\text{root} = \pm \sqrt{\frac{-\hat{b}}{3}}$ (the boundaries of the green region).
Note that the non-merging real root only takes on the value of a critical point when all three roots are merging (the origin in Figure \ref{fig:roots_phase_diagram}); otherwise, $s\neq0$ for a non-merging real root.

\newcommand{\ttilm}{\tilde{t}^-_\text{root}}
\newcommand{\ttilz}{\tilde{t}^M_\text{root}}
\newcommand{\ttilp}{\tilde{t}^+_\text{root}}

Consider a real root (with $\timagtil=0$) where equation \ref{eq:totdercubic} reduces to
\begin{equation}
\label{eq:cub_red_der}
\begin{bmatrix}
	d\trealtil_\text{root}\\ 
	d\timagtil_\text{root}\\
\end{bmatrix}
=
-
\frac{1}{3(\trealtil_\text{root})^2 + \hat{b}}
\begin{bmatrix}
	\trealtil_\text{root} & 1\\
	0 & 0\\
\end{bmatrix}
\begin{bmatrix}
    d\hat{b}\\
    d\hat{c}\\
\end{bmatrix}
\end{equation}
implying that $d\timagtil_\text{root}$ is identically zero.
Let $\tilde{t}^\pm_\text{root}$ denote the largest/smallest real roots in the green region, noting that $\ttilp$ is also defined in the red region (including the red curve) and $\ttilm$ is also defined in the blue region (including the blue curve).
$\ttilp$ always occurs where the function is increasing with derivative $3(\tilde{t}^+_\text{root})^2 + \hat{b}>0$, except when it is merging where $3(\tilde{t}^+_\text{root})^2 + \hat{b}=0$; similarly, $\ttilm$ also always occurs where the function is increasing, except when it is merging. 
Since the middle root has $-\sqrt{\frac{-\hat{b}}{3}} < \tilde{t}^M_\text{root} < \sqrt{\frac{-\hat{b}}{3}}$ except when it is merging, $3 (\tilde{t}^M_\text{root})^2 + \hat{b}<0$.
Thus, $\frac{\partial\tilde{t}^\pm_\text{root}}{\partial \hat{c}}\to-\infty$ whenever $\tilde{t}^\pm_\text{root}$ is merging on a boundary of the green region, and $\frac{\partial\tilde{t}^M_\text{root}}{\partial \hat{c}}\to+\infty$ on all boundaries of the green region.
On the red curve (excluding the origin), the merging roots are negative implying that $\frac{\partial\tilde{t}^-_\text{root}}{\partial \hat{b}}\to+\infty$ and $\frac{\partial\tilde{t}^M_\text{root}}{\partial \hat{b}}\to-\infty$; similarly, on the blue curve (excluding the origin), the merging roots are positive implying that $\frac{\partial\tilde{t}^+_\text{root}}{\partial \hat{b}}\to-\infty$ and $\frac{\partial\tilde{t}^M_\text{root}}{\partial \hat{b}}\to+\infty$.
At the origin, $\frac{\partial \trealtil_\text{root}}{\partial \hat{b}}$ is a bit more complicated.

\begin{figure}[H]
	\centering
	\begin{subfigure}[b]{0.45\textwidth}
		\centering
	    \includegraphics[width=\textwidth]{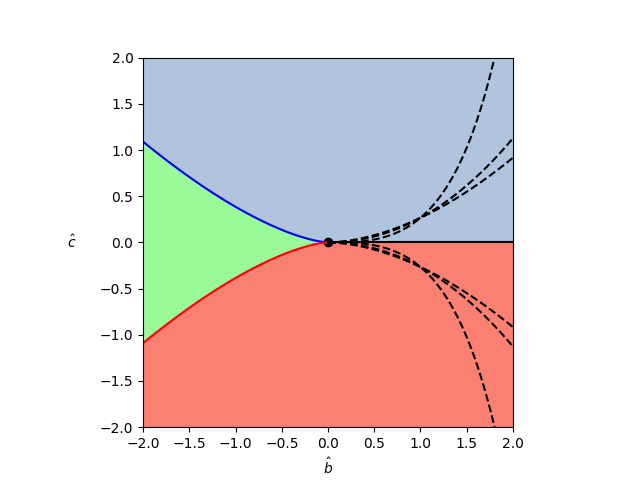}
		\label{fig:}
	\end{subfigure}
	\begin{subfigure}[b]{0.45\textwidth}
		\centering
	    \includegraphics[width=\textwidth]{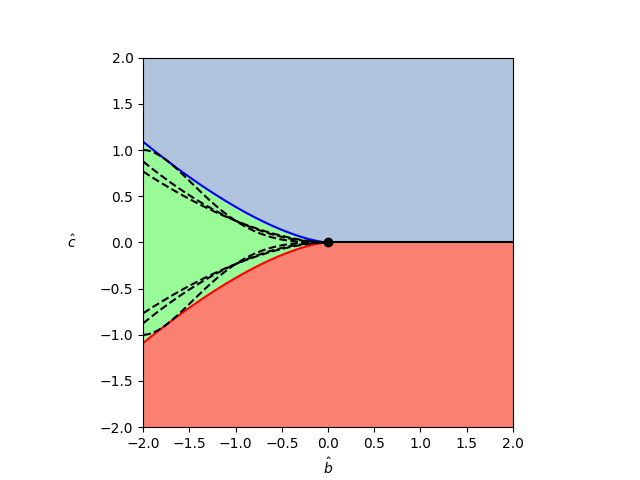}
		\label{fig:}
	\end{subfigure}
	\caption{(Left) Graphs of $\tilde{t}^\pm_\text{root} = \gamma \hat{b}^p$ and thus $\hat{c} = -\gamma^3\hat{b}^{3p} - \gamma \hat{b}^{p+1}$ for $\hat{b}>0$, $\gamma=\pm.25$, and $p=.75, 1, 2$ (black dotted lines). (Right) Graphs of $\tilde{t}^M_\text{root} = \gamma (-\hat{b})^p$ and thus $\hat{c} = -\gamma^3(-\hat{b})^{3p} + \gamma (-\hat{b})^{p+1}$ for $\hat{b}<0$, $\gamma=\pm.25$, and $p=.75, 1, 2$ (black dotted lines).}
	\label{fig:roots_phase_space_all}
\end{figure}

When $\hat{b}\leq0$, one can write $0 \leq 3(\tilde{t}^\pm_\text{root})^2 + \hat{b} \leq 3(\tilde{t}^\pm_\text{root})^2$ leading to
\begin{subequations}
\label{eq:bounds_for_cub_analysis}
\begin{align}
    0
    <
    \frac{1}{3\tilde{t}^+_\text{root}}
    \leq
    \frac{\tilde{t}^+_\text{root}}{3(\tilde{t}^+_\text{root})^2 + \hat{b}}\\
    \frac{\tilde{t}^-_\text{root}}{3(\tilde{t}^-_\text{root})^2 + \hat{b}}
    \leq
    \frac{1}{3\tilde{t}^-_\text{root}}
    <
    0
\end{align}
\end{subequations}
implying that $\frac{\partial\tilde{t}^\pm_\text{root}}{\partial\hat{b}} \to \mp \infty$ respectively approaching the origin from anywhere with $\hat{b}\leq0$.
When $\hat{b}>0$, we choose curves of the form $\tilde{t}^\pm_\text{root} = \gamma \hat{b}^p$ with $p>0$ to illustrate the behavior. See Figure \ref{fig:roots_phase_space_all} left. When $p>1$, $\frac{\partial\tilde{t}^\pm_\text{root}}{\partial \hat{b}}\to0$. When $p=1$, $\frac{\partial\tilde{t}^\pm_\text{root}}{\partial \hat{b}}\to-\gamma$ (where $\gamma>0$ is in the red region, $\gamma<0$ is in the blue region, $\gamma=0$ is the positive $\hat{b}$-axis). When $0<p<1$, {$\frac{\partial\tilde{t}^\pm_\text{root}}{\partial \hat{b}} \to \mp \infty$ respectively}. Thus, both $\frac{\partial\tilde{t}^\pm_\text{root}}{\partial \hat{b}}$ are nonremovable singularities at the origin.
For $\tilde{t}^M_\text{root}$, we choose curves of the form $\tilde{t}^M_\text{root} = \gamma (-\hat{b})^p$.
Choosing $p>\frac{1}{2}$ makes both $3p>\frac{3}{2}$ and $p+1>\frac{3}{2}$, so that the curves are in the green region as they approach the origin.
See Figure \ref{fig:roots_phase_space_all} right.
On these curves, $3 (\tilde{t}^M_\text{root})^2 = 3\gamma^2(-\hat{b})^{2p} < -\hat{b}$ as $\hat{b}\to0$ for $p>\frac{1}{2}$, implying that $3(\tilde{t}^M_\text{root})^2+\hat{b}<0$ and thus that these curves do indeed represent the middle root as $\hat{b}\to0$.
When $p>1$, $\frac{\partial\tilde{t}^M_\text{root}}{\partial \hat{b}}\to0$. When $p=1$, $\frac{\partial\tilde{t}^M_\text{root}}{\partial \hat{b}}\to\gamma$. When $\frac{1}{2}<p<1$, $\frac{\partial\tilde{t}^M_\text{root}}{\partial \hat{b}} \to \sign{(\gamma)} \infty$.
Thus, $\frac{\partial\tilde{t}^M_\text{root}}{\partial \hat{b}}$ is a nonremovable singularity at the origin.

Next, consider the case where the roots are complex with $\timagtil_\text{root}\neq0$.
The second line of equation \ref{eq:fcubic2d} gives $-(\timagtil_\text{root})^2 + 3 (\trealtil_\text{root})^2 + \hat{b} = 0$, implying that $0 < (\timagtil_\text{root})^2 = 3 (\trealtil_\text{root})^2 + \hat{b}$.
Plugging this into equation \ref{eq:impldiffcubic}b leads to $s=4(\timagtil_\text{root})^2\left(12(\trealtil_\text{root})^2+\hat{b}\right)$.
Since $12 (\trealtil_\text{root})^2 + \hat{b} \geq 3 (\trealtil_\text{root})^2 + \hat{b} = (\timagtil_\text{root})^2 > 0$, $s\to0$ implies $\timagtil_\text{root}\to0$ (i.e.~one is approaching the boundaries of the green region).
Equation \ref{eq:totdercubic} can be written as
\begin{equation}
    \label{eq:complexsidederivative}
        \begin{bmatrix}
        	d\trealtil_\text{root}\\ 
        	d\timagtil_\text{root}\\
        \end{bmatrix}
        =
		-
		\frac{1}{2\left(12(\trealtil_\text{root})^2+\hat{b}\right)}
		\begin{bmatrix}
			2\trealtil_\text{root} & -1\\
			\frac{-6(\trealtil_\text{root})^2-\hat{b}}{\pm\sqrt{3(\trealtil_\text{root})^2+\hat{b}}} & \frac{-3\trealtil_\text{root}}{\pm\sqrt{3(\trealtil_\text{root})^2+\hat{b}}}\\
		\end{bmatrix}
        \begin{bmatrix}
            d\hat{b}\\
            d\hat{c}\\
        \end{bmatrix}
\end{equation}
using $(\timagtil_\text{root})^2 = 3(\trealtil_\text{root})^2 + \hat{b}$. $d\tilde{t}^R_\text{root}$ is the same for both complex conjugates, while $d\tilde{t}^I_\text{root}$ differs only in sign. Approaching the red and blue curves (away from the origin), $\frac{\partial\tilde{t}^R_\text{root}}{\partial \hat{b}}$ and $\frac{\partial\tilde{t}^R_\text{root}}{\partial \hat{c}}$ are well-behaved while $\frac{\partial\tilde{t}^I_\text{root}}{\partial \hat{b}} \to \pm \infty$ and $\frac{\partial\tilde{t}^I_\text{root}}{\partial \hat{c}} \to \pm \infty$ depending on which complex conjugate root (and which curve in the $\frac{\partial\tilde{t}^I_\text{root}}{\partial \hat{c}}$ case) is being considered.

Since $12 (\trealtil_\text{root})^2 + \hat{b} > 0$, $\frac{\partial\tilde{t}^R_\text{root}}{\partial \hat{c}} \to +\infty$ approaching the origin. For $\frac{\partial\tilde{t}^I_\text{root}}{\partial \hat{b}}$, note that
$\frac{6(\trealtil_\text{root})^2+\hat{b}}{12(\trealtil_\text{root})^2+\hat{b}}\leq1$. When $\hat{b}\geq0$,
$
\frac{6(\trealtil_\text{root})^2+\hat{b}}{12(\trealtil_\text{root})^2+\hat{b}}
\geq
\frac{6(\trealtil_\text{root})^2+\hat{b}}{12(\trealtil_\text{root})^2+2\hat{b}}
=
\frac{1}{2}
$.
When $\hat{b}\leq0$,
$
\frac{6(\trealtil_\text{root})^2+\hat{b}}{12(\trealtil_\text{root})^2+\hat{b}}
>
\frac{3(\trealtil_\text{root})^2}{12(\trealtil_\text{root})^2}
=
\frac{1}{4}
$
using $3(\trealtil_\text{root})^2+\hat{b}>0$.
Together, all of this implies that
$\frac{1}{4}<\pm2\sqrt{3(\trealtil_\text{root})^2+\hat{b}}\frac{\partial\tilde{t}^I_\text{root}}{\partial \hat{b}}\leq1$;
thus, $\frac{\partial\tilde{t}^I_\text{root}}{\partial \hat{b}} \to \pm \infty$ approaching the origin (and thus on all boundaries of the green region).

Next, consider $\tilde{t}^R_\text{root} = \gamma \hat{b}^p$ with $\hat{b}>0$.
Writing $\hat{c} = -(\trealtil_\text{root})^3 + 3\trealtil_\text{root}(\timagtil_\text{root})^2 - \hat{b} \trealtil_\text{root} = 8(\trealtil_\text{root})^3 + 2\hat{b} \trealtil_\text{root}$ leads to $\hat{c} = 8\gamma^3 \hat{b}^{3p} + 2 \gamma \hat{b}^{p+1}$.
When $p>\frac{1}{2}$,
\begin{subequations}
    \begin{align}
        \frac{\partial\tilde{t}^R_\text{root}}{\partial \hat{b}}
        &\to
        \frac{-\gamma \hat{b}^{p-\frac{1}{2}}}{\sqrt{\hat{b}}}\\
        \frac{\partial\tilde{t}^I_\text{root}}{\partial \hat{c}}
        &\to
        \frac{\pm3\gamma \hat{b}^{p-\frac{1}{2}}}{2\hat{b}}
    \end{align}
\end{subequations}
as $\hat{b}\to0$.
When $0<p<\frac{1}{2}$,
\begin{subequations}
    \label{eq:0p12case}
    \begin{align}
        \frac{\partial\tilde{t}^R_\text{root}}{\partial \hat{b}}
        &\to
        \frac{-1}{12\gamma \hat{b}^{p}}\\
        \frac{\partial\tilde{t}^I_\text{root}}{\partial \hat{c}}
        &\to
        \frac{\pm1}{8\gamma\hat{b}^{2p}\sqrt{3\gamma^2}}
    \end{align}
\end{subequations}
as $\hat{b}\to0$.
When $p>1$, $\frac{\partial\tilde{t}^R_\text{root}}{\partial \hat{b}} \to 0$.
When $p=1$, $\frac{\partial\tilde{t}^R_\text{root}}{\partial \hat{b}}\to -\gamma$.
When $0<p<1$, $\frac{\partial\tilde{t}^R_\text{root}}{\partial \hat{b}}\to \sign{(-\gamma)}\infty$.
Thus, $\frac{\partial\tilde{t}^R_\text{root}}{\partial \hat{b}}$ is a nonremovable singularity.
When $p>\frac{3}{2}$, $\frac{\partial\tilde{t}^I_\text{root}}{\partial \hat{c}} \to 0$.
When $p=\frac{3}{2}$, $\frac{\partial\tilde{t}^I_\text{root}}{\partial \hat{c}} \to \pm\frac{3\gamma}{2}$.
When $0<p<\frac{3}{2}$, $\frac{\partial\tilde{t}^I_\text{root}}{\partial \hat{c}} \to \pm\sign{(\gamma)}\infty$.
Thus, $\frac{\partial\tilde{t}^I_\text{root}}{\partial \hat{c}}$ is a nonremovable singularity.
For completeness, consider $\tilde{t}^R_\text{root} = \gamma (-\hat{b})^p$ with $\hat{b}<0$ where $\hat{c} = 8\gamma^3 (-\hat{b})^{3p} - 2 \gamma (-\hat{b})^{p+1}$.
Choosing $0<p<\frac{1}{2}$ makes $3p<\frac{3}{2}$ and $p+1<\frac{3}{2}$, so that the curves are outside of the green region as they approach the origin.
In this case, equation \ref{eq:0p12case} would have $\hat{b}^p$ and $\hat{b}^{2p}$ replaced with $(-\hat{b})^p$ and $(-\hat{b})^{2p}$ respectively;
thus $\frac{\partial\tilde{t}^R_\text{root}}{\partial \hat{b}} \to \sign{(-\gamma)}\infty$ and $\frac{\partial\tilde{t}^I_\text{root}}{\partial \hat{c}} \to \pm\sign{(\gamma)}\infty$ as $\hat{b}\to0$.

\clearpage

\subsection{A New Canonical Form}
\label{subsection:cubic-newcanonicalform}
We begin with a change of variables robust to $q\to0$,
\begin{subequations}
\label{eq:prop_cubic_cov}
\begin{align}
\tilde{t}
&= 
qt
+
\begin{bmatrix}
    \frac{a}{3}\\
    0\\
\end{bmatrix}\\
\begin{bmatrix}
    \tilde{t}^R\\
    \tilde{t}^I\\
\end{bmatrix}
&= 
q
\begin{bmatrix}
    \treal\\
    \timag\\    
\end{bmatrix}
+
\begin{bmatrix}
    \frac{a}{3}\\
    0\\
\end{bmatrix}
\end{align}
\end{subequations}
giving $\tilde{t}^3 + \hat{b}\tilde{t} + \hat{c} = 0$ with $\hat{b} = -\frac{a^2}{3} + qb$ and $\hat{c} = \frac{2a^3}{27} - \frac{qab}{3} + q^2c$.
In spite of the fact that $\hat{b}$, $\hat{c}$, and $\tilde{t}$ are a different function of $q, a, b, c,$ and $t$ than they were in Section \ref{subsection:cubic-understandasymptotics}, the analysis in Section \ref{subsection:cubic-understandasymptotics} is still valid since it only depended on the cubic equation having the form $\tilde{t}^3 + \hat{b}\tilde{t} + \hat{c} = 0$.
In this section, we consider a further change of variables $\tilde{b} = \hat{c}$ and $\tilde{c} = \frac{\hat{b}^3}{27} + \frac{\hat{c}^2}{4}$ giving $\tilde{t}^3 - 3\left(\frac{\tilde{b}^2}{4}-\tilde{c}\right)^\frac{1}{3}\tilde{t} + \tilde{b} = 0$. 

\begin{figure}[H]
	\centering
	\includegraphics[width=.7\textwidth]{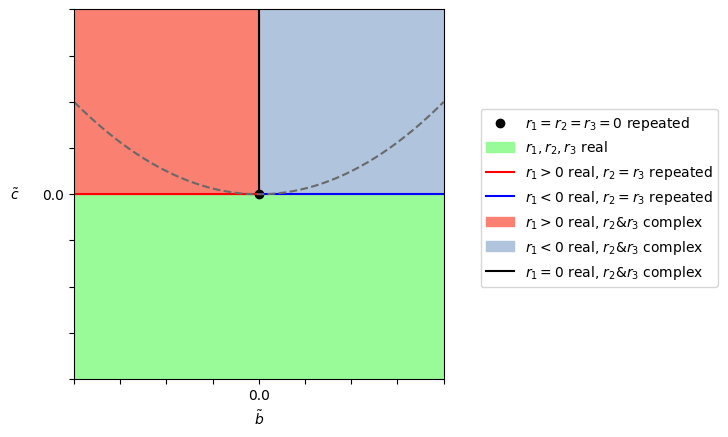}
	\caption{Root classification for the new reduced cubic in terms of $\tilde{b}$ and $\tilde{c}$. Note that the new reduced cubic only has critical points when $\tilde{c}\leq\frac{\tilde{b}^2}{4}$ (on and below the dotted line).}
	\label{fig:roots_tilde_phase_diagram}
\end{figure}

The critical points can be found from $\tilde{t}^2 - \left(\frac{\tilde{b}^2}{4}-\tilde{c}\right)^\frac{1}{3} = 0$ as $\tilde{t} = \pm \left(\frac{\tilde{b}^2}{4}-\tilde{c}\right)^\frac{1}{6}$, implying that $\tilde{c}\leq\frac{\tilde{b}^2}{4}$ is required for the critical points to exist.
In order for there to be three real roots, the reduced cubic must be non-negative at $- \left(\frac{\tilde{b}^2}{4}-\tilde{c}\right)^\frac{1}{6}$ and non-positive at $\left(\frac{\tilde{b}^2}{4}-\tilde{c}\right)^\frac{1}{6}$, i.e.~
$-\left(\frac{\tilde{b}^2}{4}-\tilde{c}\right)^\frac{1}{2} + 3\left(\frac{\tilde{b}^2}{4}-\tilde{c}\right)^\frac{1}{3}\left(\frac{\tilde{b}^2}{4}-\tilde{c}\right)^\frac{1}{6}+\tilde{b}
=
\sqrt{\tilde{b}^2-4\tilde{c}}+\tilde{b}
\geq 0$
and
$\left(\frac{\tilde{b}^2}{4}-\tilde{c}\right)^\frac{1}{2} - 3\left(\frac{\tilde{b}^2}{4}-\tilde{c}\right)^\frac{1}{3}\left(\frac{\tilde{b}^2}{4}-\tilde{c}\right)^\frac{1}{6}+\tilde{b}
=
-\sqrt{\tilde{b}^2-4\tilde{c}}+\tilde{b}
\leq 0$;
equivalently, $\tilde{c} \leq 0$ when $\tilde{b}\leq0$ and $\tilde{c} \leq 0$ when $\tilde{b}\geq0$ respectively.
Graphs of the boundary curves and the region where both inequalities are strictly valid are shown in Figure \ref{fig:roots_tilde_phase_diagram}.
Excluding the origin (in Figure \ref{fig:roots_tilde_phase_diagram}), the smallest of the three real roots is always strictly negative and the largest is always strictly positive. 
Crossing from the green to the red region merges the middle and negative roots to become complex while the positive root remains real; similarly, crossing from the green to the blue region merges the two larger roots. 
The boundary between the red and the blue regions occurs when the single real root is zero, which requires $\tilde{b}=0$.

It is worth briefly discussing the change of variables,
\begin{subequations}
\label{eq:ptildetransformall}
\begin{align}
\tilde{b} &= \hat{c}\\
\label{eq:ptildetransformtildec}
\tilde{c} &= \frac{\hat{b}^3}{27} + \frac{\hat{c}^2}{4}\\
\begin{bmatrix}
d\tilde{b}\\
d\tilde{c}\\
\end{bmatrix}
\label{eq:ptildephatcov}
&=
\begin{bmatrix}
0 & 1\\
\frac{\hat{b}^2}{9} & \frac{\hat{c}}{2}\\
\end{bmatrix}
\begin{bmatrix}
d\hat{b}\\
d\hat{c}\\
\end{bmatrix}
=
\begin{bmatrix}
0 & 1\\
\left(\frac{\tilde{b}^2}{4}-\tilde{c}\right)^\frac{2}{3} & \frac{\tilde{b}}{2}\\
\end{bmatrix}
\begin{bmatrix}
d\hat{b}\\
d\hat{c}\\
\end{bmatrix}\\
\label{eq:ptildetransformhatb}
\hat{b} &= -3\left(\frac{\tilde{b}^2}{4}-\tilde{c}\right)^\frac{1}{3}\\
\hat{c} &= \tilde{b}\\
\begin{bmatrix}
d\hat{b}\\
d\hat{c}\\
\end{bmatrix}
\label{eq:phatptildecov}
&=
\frac{1}{\left(\frac{\tilde{b}^2}{4}-\tilde{c}\right)^\frac{2}{3}}
\begin{bmatrix}
\frac{-\tilde{b}}{2} & 1 \\
\left(\frac{\tilde{b}^2}{4}-\tilde{c}\right)^\frac{2}{3} & 0\\
\end{bmatrix}
\begin{bmatrix}
d\tilde{b}\\
d\tilde{c}\\
\end{bmatrix}
=
\frac{9}{\hat{b}^2}
\begin{bmatrix}
\frac{-\hat{c}}{2} & 1 \\
\frac{\hat{b}^2}{9} & 0\\
\end{bmatrix}
\begin{bmatrix}
d\tilde{b}\\
d\tilde{c}\\
\end{bmatrix}
\end{align}
\end{subequations}
that leads from Figure \ref{fig:roots_phase_diagram} to Figure \ref{fig:roots_tilde_phase_diagram}.
Switching the roles of $b$ and $c$ leads to a rotation and reflection, allowing the sign of $\tilde{c}$ to be used to determine whether the roots are real or complex (similar to the role of $\tilde{c}$ for the reduced quadratic equation).
Notably, the use of $\hat{b}^3$ in equation \ref{eq:ptildetransformtildec} causes compression towards the $\hat{b}=0$ axis, which is subsequently transformed into a parabola by the $\hat{c}^2$ term;
importantly, this also transforms the boundaries of the green region in Figure \ref{fig:roots_phase_diagram} to an independent axis in Figure \ref{fig:roots_tilde_phase_diagram}.
The inverse of a $\hat{b}^3$ mapping contains a cube root (see equation \ref{eq:ptildetransformhatb}), which has non-differentiable cusps at $\hat{b}=0$.
Although this leads to the Jacobian blowing up on the $\tilde{c}=\frac{\tilde{b}^2}{4}$ parabola, all of the interesting behavior we wish to address occurs near the $\tilde{b}$-axis (away from $\tilde{c}=\frac{\tilde{b}^2}{4}$) except for the origin which is highly problematic in any case due to nonremovable singularities.

Once again, equation \ref{eq:impldiffcubic}b has $s=0$ only on the boundaries of the green region where $\tilde{c} = 0$ (and only for merging roots).
On this $\tilde{b}$-axis, the reduced cubic can be factored as
\begin{equation}
\label{eq:factorcubicatmergingroots}
\left(\tilde{t}-\left(\frac{\tilde{b}}{2}\right)^\frac{1}{3}\right)^2\left(\tilde{t}+2\left(\frac{\tilde{b}}{2}\right)^\frac{1}{3}\right) = 0
\end{equation}
where $\ttilm=\ttilz=\left(\frac{\tilde{b}}{2}\right)^\frac{1}{3}$ and $\ttilp = -2\left(\frac{\tilde{b}}{2}\right)^\frac{1}{3}$ when $\tilde{b}\leq0$, while $\ttilp=\ttilz=\left(\frac{\tilde{b}}{2}\right)^\frac{1}{3}$ and $\ttilm = -2\left(\frac{\tilde{b}}{2}\right)^\frac{1}{3}$ when $\tilde{b}\geq0$.

For a real root (with $\timagtil=0$), substituting equation \ref{eq:phatptildecov} into equation \ref{eq:cub_red_der} gives
\begin{equation}
\label{eq:alternativeformderivativeforcubic}
\begin{bmatrix}
	d\trealtil_\text{root}\\ 
	d\timagtil_\text{root}\\
\end{bmatrix}
=
-
\frac{1}{3(\trealtil_\text{root})^2 + \hat{b}}
\frac{9}{\hat{b}^2}
\begin{bmatrix}
\frac{-\hat{c}}{2} \trealtil_\text{root} + \frac{\hat{b}^2}{9} & \trealtil_\text{root}\\
0 & 0 \\
\end{bmatrix}
\begin{bmatrix}
d\tilde{b}\\
d\tilde{c}\\
\end{bmatrix}
\end{equation}
where the problematic $3(\trealtil_\text{root})^2+\hat{b}=0$ corresponds to $\tilde{c}=0$.
Using the reduced cubic $\tilde{t}^3 + \hat{b}\tilde{t} + \hat{c}=0$ to write
\begin{equation}
\frac{-\hat{c}}{2} \trealtil_\text{root} + \frac{\hat{b}^2}{9}
=
\left(
(\trealtil_\text{root})^3 + \hat{b}\trealtil_\text{root}
\right)
\frac{\trealtil_\text{root}}{2}
+\frac{\hat{b}^2}{9}
=
\frac{1}{9}
\left(
3(\trealtil_\text{root})^2 + \hat{b}
\right)
\left(
\frac{3(\trealtil_\text{root})^2}{2} + \hat{b}
\right)
\end{equation}
leads to
\begin{equation}
\label{eq:altalternativeformderivativeforcubicresult}
\frac{\partial \tilde{t}^R_\text{root}}{\partial \tilde{b}}
=
-
\frac{1}{\hat{b}^2}
\left(
\frac{3(\trealtil_\text{root})^2}{2} + \hat{b}
\right)
\end{equation}
removing the need for L'Hospital's rule.
This allows $\frac{\partial \tilde{t}^R_\text{root}}{\partial \tilde{b}}$ to be robustly evaluated on and near the $\tilde{b}$-axis, except near the origin.
For $\frac{\partial \tilde{t}^R_\text{root}}{\partial \tilde{c}}$, 
\begin{equation}
\label{eq:altalternativeformderivativeforcubicresultforc}
\frac{\partial \tilde{t}^R_\text{root}}{\partial \tilde{c}}
=
\frac{9}{\hat{b}^2}
\frac{\partial \tilde{t}^R_\text{root}}{\partial \hat{b}}
\end{equation}
implies that $\frac{\partial \ttilm}{\partial \tilde{c}} \to +\infty$ and $\frac{\partial \ttilz}{\partial \tilde{c}} \to -\infty$ approaching the red curve while $\frac{\partial \ttilp}{\partial \tilde{c}} \to -\infty$ and $\frac{\partial \ttilz}{\partial \tilde{c}} \to +\infty$ approaching the blue curve (similar to $\frac{\partial \tilde{t}^R_\text{root}}{\partial \hat{b}}$).

\textit{
\RemarkCounter{re:btildedernotblowingup}
This new canonical form no longer has the derivatives with respect to both reduced parameters blowing up as roots merge (as was the case for equation \ref{eq:cub_red_der}).
Only derivatives with respect to $\tilde{c}$ blow up, while the derivatives with respect to $\tilde{b}$ remain bounded (away from the origin).
Importantly, this enables one to readily move tangentially to the red and blue curves, changing the value of the roots without changing the closeness to merging.
}

\begin{figure}[H]
	\centering
	\begin{subfigure}[b]{0.45\textwidth}
		\centering
	    \includegraphics[width=\textwidth]{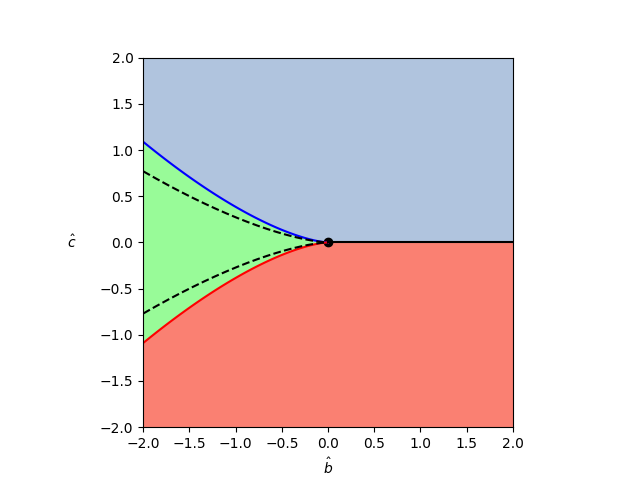}
		\label{fig:}
	\end{subfigure}
	\begin{subfigure}[b]{0.45\textwidth}
		\centering
	    \includegraphics[width=\textwidth]{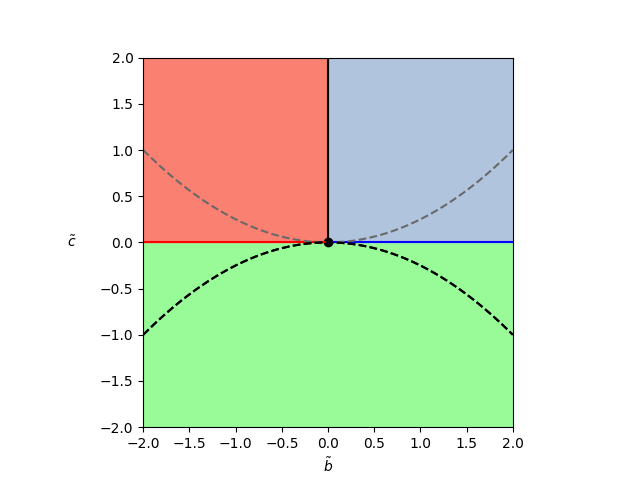}
		\label{fig:}
	\end{subfigure}
	\caption{
    (Left)
    $\ttilp = \sqrt{\frac{2}{3}}(-\hat{b})^\frac{1}{2}$ on $\hat{c}=\sqrt{\frac{2}{27}}(-\hat{b})^\frac{3}{2}$, and $\ttilm = -\sqrt{\frac{2}{3}}(-\hat{b})^\frac{1}{2}$ on $\hat{c}=-\sqrt{\frac{2}{27}}(-\hat{b})^\frac{3}{2}$ (black dotted lines).
    (Right)
    $\ttilp = \sqrt{\frac{2}{3}}(-\hat{b})^\frac{1}{2}$ on $\tilde{c} = -\frac{\tilde{b}^2}{4}$ when $\tilde{b}>0$, and $\ttilm = -\sqrt{\frac{2}{3}}(-\hat{b})^\frac{1}{2}$ on $\tilde{c} = -\frac{(-\tilde{b})^2}{4}$ when $\tilde{b}<0$ (black dotted lines).
    }
	\label{fig:tildephasespacegammacurves}
\end{figure}

Rewriting equation \ref{eq:altalternativeformderivativeforcubicresult} as
\begin{equation}
\label{eq:dtrdbtilderewriteintoterms}
\frac{\partial \tilde{t}^R_\text{root}}{\partial \tilde{b}}
=
-
\frac{1}{2\hat{b}^2}
\left(
\left(3(\trealtil_\text{root})^2 + \hat{b}\right) + \hat{b}
\right)
\end{equation}
shows that $\frac{\partial \ttilz}{\partial \tilde{b}} \to +\infty$ approaching the origin, since $3(\ttilz)^2 + \hat{b} \leq 0$ and $\hat{b}<0$ implies that $\frac{\partial \ttilz}{\partial \tilde{b}} \geq \frac{-1}{2\hat{b}}$.
For $\tilde{t}^\pm_\text{root}$, $3(\tilde{t}^\pm_\text{root})^2 + \hat{b} \geq 0$ and $\hat{b}<0$ cancel when $3(\trealtil_\text{root})^2 + 2\hat{b}=0$ or $\trealtil_\text{root} = \pm\sqrt{\frac{2}{3}}(-\hat{b})^\frac{1}{2}$.
These are $\tilde{t}^R_\text{root} = \gamma (-\hat{b})^p$ curves with $\gamma = \pm\sqrt{\frac{2}{3}}$ and $p=\frac{1}{2}$ implying that 
$
\hat{c}
=
\pm\sqrt{\frac{2}{27}}(-\hat{b})^\frac{3}{2}
$, which can only be satisfied when $\tilde{c} = -\frac{\tilde{b}^2}{4}$ (see Figure \ref{fig:tildephasespacegammacurves}).
$\frac{\partial \ttilp}{\partial \tilde{b}}$ is identically zero on the $\tilde{b}>0$ portion of $\tilde{c} = -\frac{\tilde{b}^2}{4}$, while
$\frac{\partial \ttilm}{\partial \tilde{b}}$ is identically zero on the $\tilde{b}<0$ portion of $\tilde{c} = -\frac{\tilde{b}^2}{4}$.
Examining the roots determined from equation \ref{eq:factorcubicatmergingroots} leads to $\frac{\partial \ttilp}{\partial \tilde{b}}\to-\infty$ and $\frac{\partial \ttilm}{\partial \tilde{b}}\to+\infty$ approaching the origin on the red curve, while $\frac{\partial \ttilp}{\partial \tilde{b}}\to+\infty$ and $\frac{\partial \ttilm}{\partial \tilde{b}}\to-\infty$ approaching the origin on the blue curve.
Thus, both $\frac{\partial \tilde{t}^\pm_\text{root}}{\partial \tilde{b}}$ are nonremovable singularities at the origin.

\textit{
\RemarkCounter{re:}
Choosing a perturbed $\trealtil_\text{root} = \pm\sqrt{\frac{2}{3}(-\hat{b}) (1+\gamma\hat{b})}$ leads to $\frac{\partial \tilde{t}^R_\text{root}}{\partial \tilde{b}} = \gamma$ on the resulting $\hat{c} = \pm\frac{1}{3} (-\hat{b}) \left( 1- 2\gamma\hat{b}\right)\sqrt{\frac{2}{3}(-\hat{b}) (1+\gamma\hat{b})}$ curves. Note that $\gamma=0$ reduces to the curves shown in Figure \ref{fig:tildephasespacegammacurves}.
}

For $\frac{\partial \tilde{t}^R_\text{root}}{\partial \tilde{c}}$, equation \ref{eq:altalternativeformderivativeforcubicresultforc} allows the results obtained from analyzing equation \ref{eq:cub_red_der} to be utilized. Equation \ref{eq:bounds_for_cub_analysis} shows that $\frac{\partial \tilde{t}^\pm_\text{root}}{\partial \tilde{c}} \to \mp \infty$ respectively when approaching the origin from anywhere with $\hat{b}<0$ (i.e.~$\tilde{c}<\frac{\tilde{b}^2}{4}$).
For $\ttilz$, consider $\tilde{t}^M_\text{root} = \gamma (-\hat{b})^p$. When $p>3$, $\frac{\partial\tilde{t}^M_\text{root}}{\partial \tilde{c}}\to0$. When $p=3$, $\frac{\partial\tilde{t}^M_\text{root}}{\partial \tilde{c}}\to9\gamma$. When $\frac{1}{2}<p<3$, $\frac{\partial\tilde{t}^M_\text{root}}{\partial \tilde{c}} \to \sign{(\gamma)} \infty$. Thus, $\frac{\partial\tilde{t}^M_\text{root}}{\partial \tilde{c}}$ is a nonremovable singularity at the origin.

Next, consider the case where the roots are complex with $\timagtil_\text{root}\neq0$.
Equation \ref{eq:complexsidederivative} can be written as
\begin{equation}
\label{eq:complexsidederivativerewrite}
\begin{bmatrix}
	d\trealtil_\text{root}\\ 
	d\timagtil_\text{root}\\
\end{bmatrix}
=
-
\frac{1}{2\left(12(\trealtil_\text{root})^2+\hat{b}\right)}
\frac{9}{\hat{b}^2}
\begin{bmatrix}
-\hat{c}\trealtil_\text{root} - \frac{\hat{b}^2}{9} & 2\trealtil_\text{root} \\
\frac{1}{6}
\frac{3\hat{c}\left(6(\trealtil_\text{root})^2+\hat{b}\right)-2\hat{b}^2\trealtil_\text{root}}{\pm\sqrt{3(\trealtil_\text{root})^2+\hat{b}}}
& \frac{-6(\trealtil_\text{root})^2-\hat{b}}{\pm\sqrt{3(\trealtil_\text{root})^2+\hat{b}}} \\
\end{bmatrix}
\begin{bmatrix}
d\tilde{b}\\
d\tilde{c}\\
\end{bmatrix}
\end{equation}
using equation \ref{eq:phatptildecov}. Approaching the red and blue curves (away from the origin), $\frac{\partial\tilde{t}^R_\text{root}}{\partial \tilde{b}}$ and $\frac{\partial\tilde{t}^R_\text{root}}{\partial \tilde{c}}$ are well-behaved. 
Once again, the reduced cubic can be used to obtain $\hat{c} = 8(\trealtil_\text{root})^3 + 2\hat{b} \trealtil_\text{root}$, which leads to
\begin{equation}
3\hat{c}\left(6(\trealtil_\text{root})^2+\hat{b}\right)-2\hat{b}^2\trealtil_\text{root}
=
6\trealtil_\text{root}
\left(8(\trealtil_\text{root})^2+\frac{2\hat{b}}{3}\right)
\left(3(\trealtil_\text{root})^2+\hat{b}\right)
\end{equation}
and thus
\begin{equation}
\label{eq:dtidbtildeapproachingcurves}
\frac{\partial\tilde{t}^I_\text{root}}{\partial \tilde{b}}
=
-
\frac{1}{2\left(12(\trealtil_\text{root})^2+\hat{b}\right)}
\frac{9}{\hat{b}^2}
\trealtil_\text{root}
\left(8(\trealtil_\text{root})^2+\frac{2\hat{b}}{3}\right)
\left(\pm\sqrt{3(\trealtil_\text{root})^2+\hat{b}}\right)
\end{equation}
removing the need for L'Hospital's rule (except near the origin).
For $\frac{\partial \tilde{t}^I_\text{root}}{\partial \tilde{c}}$,
\begin{equation}
\label{eq:comparisontodtidbhat}
\frac{\partial \tilde{t}^I_\text{root}}{\partial \tilde{c}}
=
\frac{9}{\hat{b}^2}
\frac{\partial \tilde{t}^I_\text{root}}{\partial \hat{b}}
\end{equation}
implies that $\frac{\partial\tilde{t}^I_\text{root}}{\partial \tilde{c}} \to \pm \infty$, depending on which complex conjugate root is being considered, on all boundaries of the green region.

\textit{
\RemarkCounter{re:}
Similar to the $\tilde{t}^I=0$ case (see Remark \ref{re:btildedernotblowingup}), the $\tilde{t}^I\neq0$ case also has derivatives with respect to $\tilde{b}$ remaining bounded (away from the origin).
}

Since $\hat{c}\trealtil_\text{root}\geq0$ (see Figure \ref{fig:roots_phase_diagram} and Remark \ref{re:sum_of_roots}),
\begin{equation}
\frac{\partial \trealtil}{\partial \tilde{b}}
=
\frac{1}{2\left(12(\trealtil_\text{root})^2+\hat{b}\right)}
\left(\frac{9\hat{c}\trealtil_\text{root}}{\hat{b}^2} + 1\right)
\geq
\frac{1}{2\left(12(\trealtil_\text{root})^2+\hat{b}\right)}
\end{equation}
implying that 
$\frac{\partial \trealtil}{\partial \tilde{b}}\to+\infty$ approaching the origin (consistent with $\frac{\partial \ttilz}{\partial \tilde{b}}$ as expected, since $\ttilz$ is continuous with the real part of the complex roots across the $\tilde{b}$-axis).
Considering only the $\hat{b}<0$ (i.e.~$\tilde{c}<\frac{\tilde{b}^2}{4}$) region,
\begin{equation}
\left|\frac{\partial \trealtil}{\partial \tilde{c}}\right|
=\frac{9\left|\trealtil_\text{root}\right|}{\left(12(\trealtil_\text{root})^2+\hat{b}\right)\hat{b}^2}
>\frac{9\left|\trealtil_\text{root}\right|}{12(\trealtil_\text{root})^2\hat{b}^2}
=\frac{9}{12\left|\trealtil_\text{root}\right|\hat{b}^2}
\end{equation}
illustrating that $\frac{\partial \trealtil}{\partial \tilde{c}}\to+\infty$ approaching the origin from the red region (below $\tilde{c}=\frac{\tilde{b}^2}{4}$), while $\frac{\partial \trealtil}{\partial \tilde{c}}\to-\infty$ approaching the origin from the blue region (below $\tilde{c}=\frac{\tilde{b}^2}{4}$).

Consider equation \ref{eq:dtidbtildeapproachingcurves} in the $\hat{b}<0$ (i.e.~$\tilde{c}<\frac{\tilde{b}^2}{4}$) region.
Using $3(\trealtil_\text{root})^2+\hat{b}\geq0$ leads to
$
12 (\trealtil_\text{root})^2 + \hat{b}
\geq
11 (\trealtil_\text{root})^2 + \frac{2\hat{b}}{3}
>
8(\trealtil_\text{root})^2+\frac{2\hat{b}}{3}
>
0
$
and thus
$\frac{8(\trealtil_\text{root})^2+\frac{2\hat{b}}{3}}{12(\trealtil_\text{root})^2+\hat{b}}<1$;
in addition,
\begin{equation}
\frac{8(\trealtil_\text{root})^2+\frac{2\hat{b}}{3}}{12(\trealtil_\text{root})^2+\hat{b}}
\geq
\frac{5(\trealtil_\text{root})^2-\frac{\hat{b}}{3}}{12(\trealtil_\text{root})^2+\hat{b}}
>
\frac{5(\trealtil_\text{root})^2}{12(\trealtil_\text{root})^2}
=
\frac{5}{12}
\end{equation}
again using $3(\trealtil_\text{root})^2+\hat{b}\geq0$.
Equation \ref{eq:dtidbtildeapproachingcurves} can be rewritten as
\begin{equation}
\frac{2\left(12(\trealtil_\text{root})^2+\hat{b}\right)}{9\left(8(\trealtil_\text{root})^2+\frac{2\hat{b}}{3}\right)}
\frac{\partial\tilde{t}^I_\text{root}}{\partial \tilde{b}}
=
\frac{-\trealtil_\text{root}}{\hat{b}^2}\left(\pm\sqrt{3(\trealtil_\text{root})^2+\hat{b}}\right)
\end{equation}
where the coefficient of $\frac{\partial\tilde{t}^I_\text{root}}{\partial \tilde{b}}$ is bounded between $\frac{2}{9}$ and $\frac{8}{15}$.
As discussed at the end of Section \ref{subsection:cubic-understandasymptotics}, choosing $\trealtil_\text{root} =\gamma (-\hat{b})^p$ with $\hat{b}<0$ and $0<p<\frac{1}{2}$ gives curves outside the green region as they approach the origin;
on these curves,
\begin{equation}
\label{eq:remainingtermanalysis}
\frac{-\trealtil_\text{root}}{\hat{b}^2}\left(\pm\sqrt{3(\trealtil_\text{root})^2+\hat{b}}\right)
\to
\sign{(\mp\gamma)}\frac{\sqrt{3}(\trealtil_\text{root})^2}{\hat{b}^2}
\to
\sign{(\mp\gamma)}\infty
\end{equation}
since $\hat{b}\to0$ faster than $(\trealtil_\text{root})^2\to0$, i.e.~$\frac{\partial\tilde{t}^I_\text{root}}{\partial \tilde{b}}\to\sign{(\mp\gamma)}\infty$. 
Choosing a perturbed $\trealtil_\text{root} = \pm\sqrt{\frac{1}{3}(-\hat{b}) (1+3\gamma^2\hat{b}^2)}$ in the blue/red regions respectively leads to 
\begin{equation}
\hat{c}
= \trealtil_\text{root}\left(8 (\trealtil_\text{root})^2 + 2\hat{b}\right) 
=\pm2 \left(\frac{-\hat{b}}{3}\right)^\frac{3}{2} \left(1+12\gamma^2\hat{b}^2\right)\sqrt{1+3\gamma^2\hat{b}^2}
\end{equation}
curves. When $\gamma=0$, these are the boundary curves $\hat{c}=\pm2\left(\frac{-\hat{b}}{3}\right)^\frac{3}{2}$;
when $\gamma\neq0$, the perturbed curves lie outside the green region.
On these curves,
\begin{equation}
\label{eq:thistermgoeslike}
\frac{-\trealtil_\text{root}}{\hat{b}^2}\left(\pm\sqrt{3(\trealtil_\text{root})^2+\hat{b}}\right)
=
\pm|\gamma|\sqrt{1+3\gamma^2\hat{b}^2}
\to
\pm|\gamma|
\end{equation}
approaching the origin in the red region and 
\begin{equation}
\frac{-\trealtil_\text{root}}{\hat{b}^2}\left(\pm\sqrt{3(\trealtil_\text{root})^2+\hat{b}}\right)
=
\mp|\gamma|\sqrt{1+3\gamma^2\hat{b}^2}
\to
\mp|\gamma|
\end{equation}
approaching the origin in the blue region. Thus, $\frac{\partial\tilde{t}^I_\text{root}}{\partial \tilde{b}}$ is a nonremovable singularity at the origin; however, it appears to be close to zero for subsequences very close to the red and blue curves (as expected).


\clearpage

\subsection{Roots in the New Canonical Form}
\label{subsection:cubic-rootsinnewcanonicalform}
\begin{figure}[H]
	\centering
	\begin{subfigure}[b]{0.49\textwidth}
		\centering
	    \includegraphics[width=\linewidth]{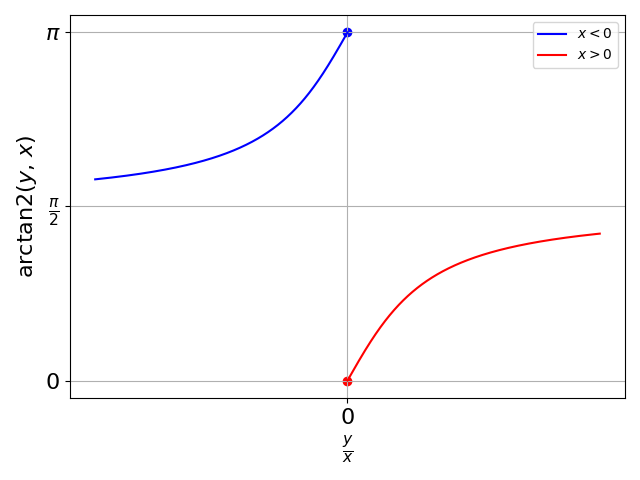}
		\label{}
	\end{subfigure}
	\hfill
	\begin{subfigure}[b]{0.49\textwidth}
		\centering
	    \includegraphics[width=\linewidth]{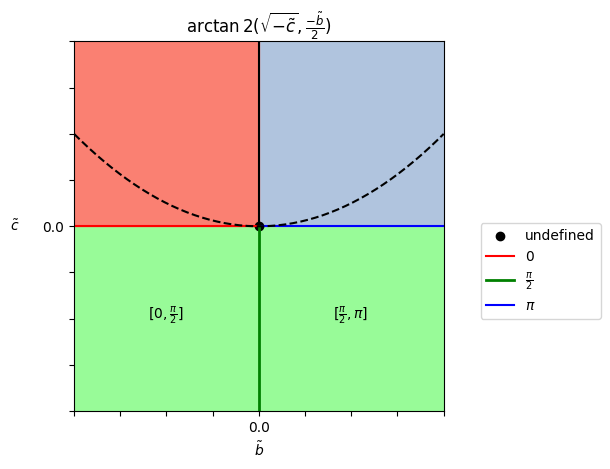}
		\label{}
	\end{subfigure}
    \caption{
    (Left) 
    Since $\sqrt{-\tilde{c}}\geq0$, only the $y\geq0$ case is shown.
    The graph does not reflect the $x=0$ case, which is: $\arctan2(y, x) = \frac{\pi}{2}$ when $y>0$, and $\arctan2(y, x)$ is undefined when $y=0$.
    The undefined case has $\tilde{b}=\tilde{c}=0$ and thus triply repeated roots identically equal to $0$.
    (Right) 
    Values of $\arctan2\left(\sqrt{-\tilde{c}}, \frac{-\tilde{b}}{2}\right)$ in the $\tilde{c}\leq0$ region.
    Note the nonremovable singularity at the origin.
    }
	\label{fig:tilde_phase_space_arctan2}
\end{figure}

In the $\tilde{c}<0$ case, let
\begin{equation}
\label{eq:thetadefinition}
\theta_0 = \arctan2\left(\sqrt{-\tilde{c}},-\frac{\tilde{b}}{2}\right)
\end{equation}
as illustrated in Figure \ref{fig:tilde_phase_space_arctan2} in order to define
\begin{subequations}
    \begin{align}
        \theta^- &= \frac{1}{3}\left(\theta_0 + 2\pi\right) \in \left[\frac{2\pi}{3}, \pi\right]
        &&
        \hspace{-3cm}
        \text{ with }
        \cos(\theta^-) \in \left[-1, -\frac{1}{2}\right] \\       
        \theta^M &= \frac{1}{3}\left(\theta_0 - 2\pi\right) \in \left[-\frac{2\pi}{3}, -\frac{\pi}{3}\right]
        &&
        \hspace{-3cm}
        \text{ with }
        \cos(\theta^M) \in \left[-\frac{1}{2}, \frac{1}{2}\right]\\       
        \theta^+ &= \frac{1}{3}\theta_0 \in \left[0, \frac{\pi}{3}\right]
        &&
        \hspace{-3cm}
        \text{ with }
        \cos(\theta^+) \in \left[\frac{1}{2}, 1\right]  
    \end{align}
\end{subequations}
so that the roots are given by
\begin{equation}
\label{eq:cubic_cardono_real_roots}
\tilde{t}^R_\text{root}
= 2\left(\frac{\tilde{b}^2}{4}-\tilde{c}\right)^\frac{1}{6}\cos\theta
\end{equation}
with $\theta = \theta^\pm$ for $\tilde{t}^\pm_\text{root}$ respectively and $\theta = \theta^M$ for $\tilde{t}^M_\text{root}$.
Equation \ref{eq:cubic_cardono_real_roots} can be verified via a $\cos^3\theta$ identity.
Note that
\begin{equation}
\label{eq:denomrealcase}
3 (\tilde{t}_\text{root}^R)^2 + \hat{b}
=
12\left(\frac{\tilde{b}^2}{4}-\tilde{c}\right)^\frac{1}{3}\left(\cos^2\theta - \frac{1}{4}\right)
\end{equation}
which is nonnegative for $\tilde{t}^\pm_\text{root}$ and nonpositive for $\tilde{t}^M_\text{root}$.

\newcommand{\eps}{\xi_S}
\newcommand{\epsbar}{\xi_D}
\newcommand{\rrootcs}{\tilde{t}_\text{root}^R}
\newcommand{\rrootcsbar}{\bar{\tilde{t}}_\text{root}^R}
In the $\tilde{c}>0$ case, there is one real root and two complex conjugate roots.
Let
$\eps = \left(\frac{-\tilde{b}}{2} + \sqrt{\tilde{c}}\right)^\frac{1}{3}$
and
$\epsbar = \left(\frac{-\tilde{b}}{2} - \sqrt{\tilde{c}}\right)^\frac{1}{3}$,
so that the real root is $\tilde{t}_\text{root}^R=\eps + \epsbar$.
Note that 
$
3 (\tilde{t}_\text{root}^R)^2 + \hat{b}
=
3 \left(\eps^2 + \eps\epsbar + \epsbar^2\right)
>
0
$
unless $\eps = \epsbar = 0$, which is true if and only if $\tilde{b}=\tilde{c}=0$ (i.e.~when $\tilde{t}_\text{root}^R=0$ is a triply repeated real root).
The complex roots are 
\begin{equation}
\label{eq:cubiccomplexrootsanalytic}
\begin{bmatrix}
\tilde{t}^R_\text{root}\\
\tilde{t}^I_\text{root}\\
\end{bmatrix}
=
\begin{bmatrix}
    -\frac{1}{2}(\eps + \epsbar)\\
    \pm\frac{\sqrt{3}}{2}(\eps - \epsbar)\\
\end{bmatrix}   
\end{equation}
noting that $\eps \neq \epsbar$.

\clearpage
\section{Proposed Approach for the Cubic Equation}
\label{section:cubicproposedapproach}
In our newly proposed reduced canonical form (see Section \ref{subsection:cubic-newcanonicalform}), equations \ref{eq:fcubic2d} and \ref{eq:impldiffcubic} become
\begin{subequations}
\begin{align}
\label{eq:fcubictildespace}
\tilde{f}(\tilde{t}_\text{root}; \tilde{p})
&=
\begin{bmatrix}
(\trealtil_\text{root})^3 - 3 \trealtil_\text{root}(\timagtil_\text{root})^2+ \hat{b} \trealtil_\text{root} + \hat{c}\\
-(\timagtil_\text{root})^3 + 3 (\trealtil_\text{root})^2\timagtil_\text{root} + \hat{b} \timagtil_\text{root}\\
\end{bmatrix}
=
q^2 f(t_\text{root}; \vec{p})
=
\vec{0}\\
\tilde{f}_{\theta_1}(\tilde{t}_\text{root}; \tilde{p})
&=
\begin{bmatrix}
    3 (\tilde{t}_\text{root}^R)^2 - 3 (\timagtil_\text{root})^2 + \hat{b}  & - 6 \tilde{t}_\text{root}^R \tilde{t}_\text{root}^I\\
    6 \tilde{t}_\text{root}^R \tilde{t}_\text{root}^I & 3 (\tilde{t}_\text{root}^R)^2 - 3 (\timagtil_\text{root})^2 + \hat{b}\\
\end{bmatrix}
=
q f_{\theta_1}(t_\text{root}; \vec{p})
\\
\label{eq:ftheta1invtildespace}
\tilde{f}_{\theta_1}^{-1}(\tilde{t}_\text{root}; \tilde{p})
&=
\frac{1}{s}
\begin{bmatrix}
    3 (\tilde{t}_\text{root}^R)^2 - 3 (\timagtil_\text{root})^2 + \hat{b} & 6 \tilde{t}_\text{root}^R \tilde{t}_\text{root}^I\\
    -6 \tilde{t}_\text{root}^R \tilde{t}_\text{root}^I & 3 (\tilde{t}_\text{root}^R)^2 - 3 (\timagtil_\text{root})^2 + \hat{b}\\
\end{bmatrix}
\\
&\text{ where }
s
=
\left(3 (\tilde{t}_\text{root}^R)^2 - 3 (\timagtil_\text{root})^2 + \hat{b}\right)^2 + \left(6 \tilde{t}_\text{root}^R \tilde{t}_\text{root}^I\right)^2
\nonumber
\\
\tilde{f}_{\theta_2}(\tilde{t}_\text{root}; \tilde{p})
&=
\label{eq:ftheta2tildespace}
\tilde{f}_{\theta_2}(\tilde{t}_\text{root}; \hat{p})
\frac{\partial \hat{p}}{\partial \tilde{p}}
=
\frac{9}{\hat{b}^2}
\begin{bmatrix}
\tilde{t}_\text{root}^R & 1 \\  
\tilde{t}_\text{root}^I & 0 \\  
\end{bmatrix}
\begin{bmatrix}
\frac{-\hat{c}}{2} & 1 \\
\frac{\hat{b}^2}{9} & 0\\
\end{bmatrix}
\end{align}
\end{subequations}
where equation \ref{eq:ftheta2tildespace} utilizes equation \ref{eq:phatptildecov}.
Formally, $\tilde{f}_{\theta_2}(\tilde{t}_\text{root}; \tilde{p})$ and $\tilde{f}_{\theta_2}(\tilde{t}_\text{root}; \hat{p})$ are size 2x4 and $\frac{\partial \hat{p}}{\partial \tilde{p}}$ is size 4x4;
however, $\tilde{f}(\tilde{t}_\text{root}; \tilde{p})$ only requires the last two columns of $\tilde{f}_{\theta_2}(\tilde{t}_\text{root}; \tilde{p})$, and so only the last two columns are shown in equation \ref{eq:ftheta2tildespace}.

Similar to equation \ref{eq:ourapproach2}, $\tilde{f}_{\theta_1}(\tilde{t}_\text{root}, \tilde{p})\frac{\partial \theta_1}{\partial \theta_2} = - \tilde{f}_{\theta_2}(\tilde{t}_\text{root}, \tilde{p})$
or
$\tilde{f}_{\theta_1}(\tilde{t}_\text{root}, \tilde{p}) \frac{\partial \tilde{t}_\text{root}}{\partial \tilde{p}} = - \tilde{f}_{\theta_2}(\tilde{t}_\text{root}, \tilde{p})$;
then, similar to equation \ref{eq:ourapproach0.3},
\begin{equation}
\label{eq:ourapproach0.3cubic}
\tilde{f}_{\theta_1}(\tilde{t}_\text{root}, \tilde{p})\frac{\partial \tilde{t}_\text{root}}{\partial \vec{p}} = \tilde{f}_{\theta_1}(\tilde{t}_\text{root}, \tilde{p})\frac{\partial \tilde{t}_\text{root}}{\partial \tilde{p}}\frac{\partial \tilde{p}}{\partial \hat{p}}\frac{\partial \hat{p}}{\partial \vec{p}}
=
-\tilde{f}_{\theta_2}(\tilde{t}_\text{root}; \tilde{p})
\frac{\partial \tilde{p}}{\partial \hat{p}}\frac{\partial \hat{p}}{\partial \vec{p}}
= 
-\tilde{f}_{\theta_2}(\tilde{t}_\text{root}; \hat{p})
\frac{\partial \hat{p}}{\partial \vec{p}}
\end{equation}
using equation \ref{eq:ftheta2tildespace}.
Using equation \ref{eq:prop_cubic_cov} to expand the left hand side and inserting $\frac{\partial \hat{p}}{\partial \vec{p}}$ (see the text after equation \ref{eq:prop_cubic_cov}) on the right hand side leads to 
\begin{subequations}
\begin{gather}
\tilde{f}_{\theta_1}(\tilde{t}_\text{root}, \tilde{p})
\left(
q
\frac{\partial t_\text{root}}{\partial \vec{p}}
+
\begin{bmatrix}
    \treal_\text{root} & \frac{1}{3} & 0 & 0\\
    \timag_\text{root} & 0 & 0 & 0\\
\end{bmatrix}
\right)
= 
-
\begin{bmatrix}
\tilde{t}_\text{root}^R & 1 \\  
\tilde{t}_\text{root}^I & 0 \\  
\end{bmatrix}
\begin{bmatrix}
    b & -\frac{2a}{3} & q & 0\\
    -\frac{ab}{3} + 2qc & \frac{2a^2}{9} - \frac{qb}{3} & -\frac{qa}{3} & q^2 \\
\end{bmatrix}\\
\label{eq:ourapproach0.2withacubic}
q \tilde{f}_{\theta_1}(\tilde{t}_\text{root}, \tilde{p})
\frac{\partial t_\text{root}}{\partial \vec{p}}
=
-
q^2
\begin{bmatrix}
(t_\text{root}^R)^3 - 3t_\text{root}^R(t_\text{root}^I)^2 &
(t_\text{root}^R)^2 - (t_\text{root}^I)^2 &     
t_\text{root}^R &     
1 \\    
3(t_\text{root}^R)^2t_\text{root}^I -(t_\text{root}^I)^3 &     
2 t_\text{root}^R t_\text{root}^I &     
t_\text{root}^I &     
0 \\
\end{bmatrix}\\
\label{eq:ourapproach0.2withoutacubic}
\tilde{f}_{\theta_1}(\tilde{t}_\text{root}, \tilde{p})
\frac{\partial t_\text{root}}{\partial \vec{p}}
=
-
q
\begin{bmatrix}
(t_\text{root}^R)^3 - 3t_\text{root}^R(t_\text{root}^I)^2 &
(t_\text{root}^R)^2 - (t_\text{root}^I)^2 &     
t_\text{root}^R &     
1 \\    
3(t_\text{root}^R)^2t_\text{root}^I -(t_\text{root}^I)^3  &     
2 t_\text{root}^R t_\text{root}^I &     
t_\text{root}^I &     
0 \\
\end{bmatrix}
\end{gather}
\end{subequations}
where equation \ref{eq:fcubic2d} was used to obtain the right hand side of equation \ref{eq:ourapproach0.2withacubic}.
Importantly, equation \ref{eq:ourapproach0.2withoutacubic} was obtained from equation \ref{eq:ourapproach0.2withacubic} by dividing by $q$, which allows for $q\to0$ but not $q=0$.

From equation \ref{eq:ftheta1invtildespace}, one can obtain
\begin{subequations}
\label{eq:actualcubicftheta1inverseall}
\begin{align}
\label{eq:actualcubicftheta1inverseftilde}
\tilde{f}^{-1}_{\theta_1}(\tilde{t}_\text{root}; \tilde{p})
&=
\begin{cases}
\frac{1}{3 (\tilde{t}_\text{root}^R)^2 + \hat{b}}I
& \text{if } \tilde{t}_\text{root}^I=0 \\
\frac{1}{2\timagtil_\text{root}}
\frac{1}{(3\trealtil_\text{root})^2 + (\timagtil_\text{root})^2}
\begin{bmatrix}
    3 \tilde{t}_\text{root}^R & \timagtil_\text{root}\\
    -\timagtil_\text{root} & 3 \tilde{t}_\text{root}^R\\
\end{bmatrix}
\begin{bmatrix}
    0 & 1 \\
    -1 & 0 \\
\end{bmatrix}
& \text{if }\tilde{t}_\text{root}^I\neq0\\
\end{cases}\\
\label{eq:actualcubicftheta1inversectilde}
\tilde{f}^{-1}_{\theta_1}(\tilde{t}_\text{root}; \tilde{p})
&=
\begin{cases}
\frac{1}{q}
\frac{1}{3 q (t_\text{root}^R)^2 + 2 a t_\text{root}^R + b}I
& \text{if } t_\text{root}^I=0 \\
\frac{1}{q}
\frac{1}{2 t_\text{root}^I}
\frac{1}{(3 q t_\text{root}^R + a)^2 + (qt_\text{root}^I)^2}
\begin{bmatrix}
    3 q t_\text{root}^R + a & q t_\text{root}^I\\
    -q t_\text{root}^I & 3 q t_\text{root}^R + a\\
\end{bmatrix}
\begin{bmatrix}
    0 & 1 \\
    -1 & 0 \\
\end{bmatrix}
& \text{if } t_\text{root}^I\neq0 \\
\end{cases}
\end{align}
\end{subequations} 
noting that the $\frac{1}{q}$ will cancel with the $q$ on the right hand side of equation \ref{eq:ourapproach0.2withoutacubic}.

In the $\tilde{t}_\text{root}^I=0$ case of equation \ref{eq:actualcubicftheta1inverseftilde}, the denominator is nonnegative for $\tilde{t}^\pm_\text{root}$ and nonpositive for $\tilde{t}^M_\text{root}$.
As can be seen in equation \ref{eq:prop_cubic_cov}, $\tilde{t}^M_\text{root}$ corresponds to $t^M_\text{root}$ while $\tilde{t}^\pm_\text{root}$ correspond to the larger/smaller of $t^\pm_\text{root}$ (respectively) when $q>0$ (and vice versa when $q<0$);
thus, the denominator in the $t_\text{root}^I=0$ case in equation \ref{eq:actualcubicftheta1inversectilde} is nonnegative for $t^\pm_\text{root}$ and nonpositive for $t^M_\text{root}$.

\textit{
\RemarkCounter{re:cubicclampingderivs}
The derivative $3 q (t_\text{root}^R)^2 + 2 a t_\text{root}^R + b$ should be clamped to be positive for $t^\pm_\text{root}$ and negative for $t^M_\text{root}$ when $q>0$ (and vice-versa when $q<0$).
Recall that the cubic root solver determines a pseudo-sign for $q$ when $q=0$.
When there is one real root and two complex conjugate roots, the real root corresponds to either $t^\pm_\text{root}$ and is treated similarly.
}

\textit{
\RemarkCounter{re:quadroleswitch}
In Remark \ref{re:clampingderiv}, the signs for clamping do not change when $a$ changes sign. This is because the standard definition of $t^\pm_\text{root}$ in equation \ref{eq:quadraticroots} switches which of $t^\pm_\text{root}$ is smaller/larger when $a$ changes sign (which seems like a poor convention given Remark \ref{re:cubicclampingderivs}; however, see Remark \ref{re:switchroles}).
}

\textit{
\RemarkCounter{re:othercubicclampingderivs}
The sign of the denominator in the $t_\text{root}^I\neq0$ case of equation \ref{eq:actualcubicftheta1inversectilde} is straightforward based on the signs of $q$ and $t_\text{root}^I$.
}

When $t_\text{root}^I=0$, 
\begin{equation}
\label{eq:ourapproachderivcubicorigreal}
\frac{\partial t_\text{root}}{\partial \vec{p}}
=
\frac{-1}{3 q (t_\text{root}^R)^2 + 2 a t_\text{root}^R + b}
\begin{bmatrix}
(t_\text{root}^R)^3 &
(t_\text{root}^R)^2 &     
t_\text{root}^R &     
1 \\    
0 &     
0 &     
0 &     
0 \\
\end{bmatrix}
\end{equation}
is obtained by substituting equation \ref{eq:actualcubicftheta1inversectilde} into equation \ref{eq:ourapproach0.2withoutacubic}.
When $(\treal_\text{root})^3>1$, we factor $(\treal_\text{root})^3$ out into the numerator of the scalar multiplier so that each component of the vector is bounded by $1$; then, the scalar multiplier is robustly evaluated (as discussed in the text after equation \ref{eq:ourapproachderivctilde}).

When $t_\text{root}^I\neq0$, 
\begin{equation}
\label{eq:ourapproachderivcubicorigcomplex}
\begin{aligned}
&\frac{\partial t_\text{root}}{\partial \vec{p}}
=
\frac{-1}{2 t_\text{root}^I}
\frac{1}{(3 q t_\text{root}^R + a)^2 + (qt_\text{root}^I)^2}
\begin{bmatrix}
    3 q t_\text{root}^R + a & q t_\text{root}^I\\
    -q t_\text{root}^I & 3 q t_\text{root}^R + a \\
\end{bmatrix}
\\ &\hspace{5cm}
\begin{bmatrix}    
3(t_\text{root}^R)^2t_\text{root}^I -(t_\text{root}^I)^3 &     
2 t_\text{root}^R t_\text{root}^I &     
t_\text{root}^I &     
0 \\
-(t_\text{root}^R)^3 + 3t_\text{root}^R(t_\text{root}^I)^2  &
-(t_\text{root}^R)^2 + (t_\text{root}^I)^2 &     
-t_\text{root}^R &     
-1 \\
\end{bmatrix}
\end{aligned}
\end{equation}
is obtained by substituting equation \ref{eq:actualcubicftheta1inversectilde} in equation \ref{eq:ourapproach0.2withoutacubic}.
The 2x2 matrix needs some consideration.
Note that $q$ and $a$ cannot both be zero;
otherwise, the quadratic root solver utilized by the cubic root solver would not find complex roots.

Firstly, consider $q t_\text{root}^I = 0$, which makes $q=0$ since $t_\text{root}^I \neq 0$;
then, $a\neq0$ and
\begin{equation}
\label{eq:part_of_complex_case_der}
\frac{1}{(3 q t_\text{root}^R + a)^2 + (qt_\text{root}^I)^2}
\begin{bmatrix}
    3 q t_\text{root}^R + a & q t_\text{root}^I\\
    -q t_\text{root}^I & 3 q t_\text{root}^R + a \\
\end{bmatrix}
=
\frac{1}{a}I
\end{equation}
so that equation \ref{eq:ourapproachderivcubicorigcomplex} becomes
\begin{equation}
\label{eq:handle_by_factoring}
\frac{\partial t_\text{root}}{\partial \vec{p}}
=
\frac{-1}{2 a t_\text{root}^I}
\begin{bmatrix}    
3(t_\text{root}^R)^2t_\text{root}^I -(t_\text{root}^I)^3 &     
2 t_\text{root}^R t_\text{root}^I &     
t_\text{root}^I &     
0 \\
-(t_\text{root}^R)^3 + 3t_\text{root}^R(t_\text{root}^I)^2 &
-(t_\text{root}^R)^2 + (t_\text{root}^I)^2 &     
-t_\text{root}^R &     
-1 \\
\end{bmatrix}
\end{equation}
which matches equation \ref{eq:ourapproachderivctilde} except for an additional first column.
For the top row, $\timag_\text{root}$ is factored out front (where it cancels) leaving only $2a$ on the denominator.
Note that the $\pm$-sign is unnecessary for this (non-merging) real part of the complex root, and only the sign of $a$ is required.
The larger in magnitude between $2\treal_\text{root}$ and $3(t_\text{root}^R)^2 - (t_\text{root}^I)^2$ is factored out front when it is larger than $1$;
in addition, it is more robust to consider 
$
(\sqrt{3}t_\text{root}^R + t_\text{root}^I)
(\sqrt{3}t_\text{root}^R - t_\text{root}^I)
$
than 
$3(t_\text{root}^R)^2 -(t_\text{root}^I)^2$
in the context of Remark \ref{re:splitupforcancelerror}.
For the bottom row, the larger in magnitude between $t_\text{root}^R\left((t_\text{root}^R)^2 - 3(t_\text{root}^I)^2\right)$, $(\treal_\text{root})^2 - (\timag_\text{root})^2$, and $\treal_\text{root}$ is factored out front when it is larger than $1$;
in addition, it is more robust to consider 
$
(t_\text{root}^R + \sqrt{3}t_\text{root}^I)
(t_\text{root}^R - \sqrt{3}t_\text{root}^I)
$
than 
$(t_\text{root}^R)^2 - 3(t_\text{root}^I)^2$.

Secondly, consider $3 q t_\text{root}^R + a =0$, which makes $q\neq0$ (and $q t_\text{root}^I\neq0$) because $q=0$ would make $a=0$.
From equation \ref{eq:prop_cubic_cov}, $\tilde{t}_\text{root}^R=0$.
From Remark \ref{re:sum_of_roots}, this corresponds to the black rays (not including the origin) in Figures \ref{fig:roots_phase_diagram} and \ref{fig:roots_tilde_phase_diagram}.
Equation \ref{eq:part_of_complex_case_der} becomes
\begin{equation}
\frac{1}{(3 q t_\text{root}^R + a)^2 + (qt_\text{root}^I)^2}
\begin{bmatrix}
    3 q t_\text{root}^R + a & q t_\text{root}^I\\
    -q t_\text{root}^I & 3 q t_\text{root}^R + a \\
\end{bmatrix}
=
\frac{1}{qt_\text{root}^I}
\begin{bmatrix}
    0 & 1\\
    -1 & 0 \\
\end{bmatrix}
\end{equation}
and the $t_\text{root}^I \neq 0$ case of equation \ref{eq:actualcubicftheta1inversectilde} becomes
\begin{equation}
\tilde{f}^{-1}_{\theta_1}(\tilde{t}_\text{root}; \tilde{p})
=
\frac{1}{q}
\frac{1}{2 t_\text{root}^I}
\frac{-1}{qt_\text{root}^I}
I
\end{equation}
so that equation \ref{eq:ourapproachderivcubicorigcomplex} becomes
\begin{equation}
\label{eq:ourapproachderivcubicsecondcase}
\frac{\partial t_\text{root}}{\partial \vec{p}}
=
\frac{1}{2 q (t_\text{root}^I)^2}
\begin{bmatrix}    
(t_\text{root}^R)^3 - 3t_\text{root}^R(t_\text{root}^I)^2  &
(t_\text{root}^R)^2 - (t_\text{root}^I)^2 &     
t_\text{root}^R &     
1 \\
3(t_\text{root}^R)^2t_\text{root}^I -(t_\text{root}^I)^3 &     
2 t_\text{root}^R t_\text{root}^I &     
t_\text{root}^I &     
0 \\
\end{bmatrix}
\end{equation}
where the top and bottom rows are treated similarly to equation \ref{eq:handle_by_factoring}.
Note that the $\pm$ sign is unnecessary for the top row, which only contains $(t_\text{root}^I)^2$.

Finally, consider the case when both $qt_\text{root}^I\neq0$ and $3 q t_\text{root}^R + a \neq 0$.
From equation \ref{eq:ourapproachderivcubicorigcomplex}, one can write
\begin{equation}
\frac{\partial t^R_\text{root}}{\partial \vec{p}}
=
\label{eq:ourapproachderivcubicrealmul}
\frac{-1}{2}
\frac{1}{(3 q t_\text{root}^R + a)^2 + (qt_\text{root}^I)^2}
\begin{bmatrix}
(8q t_\text{root}^R + 3a)(t_\text{root}^R)^2 - a (t_\text{root}^I)^2 \\
5q (t_\text{root}^R)^2 + q (t_\text{root}^I)^2 + 2a t_\text{root}^R \\
2q t_\text{root}^R + a \\
-q\\
\end{bmatrix}^T
\end{equation}
noting that the $\pm$-sign is (again) unnecessary since only $(t_\text{root}^I)^2$ appears.
Since the fourth entry is nonzero, the largest entry can be robustly factored out front.
Unlike equations \ref{eq:ourapproachderivctilde}, \ref{eq:ourapproachderivcubicorigreal}, \ref{eq:handle_by_factoring}, and \ref{eq:ourapproachderivcubicsecondcase}, the parameters contribute to the direction of the vector;
however, we have previously assumed that they can be treated as bounded in the discussion in Section \ref{section:motivation}.
When $8q t_\text{root}^R + 3a$ and $a$ have the same sign, it is more robust to consider
\begin{equation}
\sign{(a)}
\left(\sqrt{|8q t_\text{root}^R + 3a|}t_\text{root}^R + \sqrt{|a|}t_\text{root}^I\right)
\left(\sqrt{|8q t_\text{root}^R + 3a|}t_\text{root}^R - \sqrt{|a|}t_\text{root}^I\right)
\end{equation}
than the first entry of the vector in equation \ref{eq:ourapproachderivcubicrealmul}.
From equation \ref{eq:ourapproachderivcubicorigcomplex}, one can write
\begin{equation}
\begin{aligned}
\frac{\partial t^I_\text{root}}{\partial \vec{p}}
\label{eq:ourapproachderivcubicimagmul}
&=
\frac{1}{2 t_\text{root}^I}
\frac{1}{(3 q t_\text{root}^R + a)^2 + (qt_\text{root}^I)^2}
\\&\hspace{2cm}
\begin{bmatrix}
t_\text{root}^R\left(3q t_\text{root}^R + a\right) (t_\text{root}^R)^2 - \left(q (t_\text{root}^I)^2 + 6 q (t_\text{root}^R)^2 + 3a t_\text{root}^R \right) (t_\text{root}^I)^2 \\
(3q t_\text{root}^R + a)(t_\text{root}^R)^2 - (q t_\text{root}^R + a)(t_\text{root}^I)^2\\
3q (t_\text{root}^R)^2 + q(t_\text{root}^I)^2 + at_\text{root}^R \\
3q t_\text{root}^R + a \\
\end{bmatrix}^T
\end{aligned}
\end{equation}
noting that the fourth entry is nonzero;
thus, the largest entry can be robustly factored out front.
When $t_\text{root}^R\left(3q t_\text{root}^R + a\right)$ and $q (t_\text{root}^I)^2 + 6 q (t_\text{root}^R)^2 + 3a t_\text{root}^R$ have the same sign, it is more robust to consider
\begin{equation}
\begin{aligned}
\sign{\left(t_\text{root}^R\left(3q t_\text{root}^R + a\right)\right)}
&
\left(\sqrt{|t_\text{root}^R\left(3q t_\text{root}^R + a\right)|}t_\text{root}^R + \sqrt{|q (t_\text{root}^I)^2 + 6 q (t_\text{root}^R)^2 + 3a t_\text{root}^R|}t_\text{root}^I \right)
\\&
\left(\sqrt{|t_\text{root}^R\left(3q t_\text{root}^R + a\right)|}t_\text{root}^R - \sqrt{|q (t_\text{root}^I)^2 + 6 q (t_\text{root}^R)^2 + 3a t_\text{root}^R|}t_\text{root}^I \right)
\end{aligned}
\end{equation}
than the first entry of the vector in equation \ref{eq:ourapproachderivcubicimagmul}.
Note that $6 q (t_\text{root}^R)^2$ could be equivalently moved (or partially moved) into the first square root as $-6 q (t_\text{root}^I)^2$; however, that requires carefully considering $3(t_\text{root}^R)^2 - 6(t_\text{root}^I)^2$ to help alleviate cancellation.
When $3q t_\text{root}^R + a$ and $q t_\text{root}^R + a$ have the same sign, it is more robust to consider
\begin{equation}
\sign{(3q t_\text{root}^R + a)}
\left(\sqrt{|3q t_\text{root}^R + a|}t_\text{root}^R + \sqrt{|q t_\text{root}^R + a|}t_\text{root}^I\right)
\left(\sqrt{|3q t_\text{root}^R + a|}t_\text{root}^R - \sqrt{|q t_\text{root}^R + a|}t_\text{root}^I\right)
\end{equation}
than the second entry of the vector in equation \ref{eq:ourapproachderivcubicimagmul}.

\textit{
\RemarkCounter{re:}
Equations \ref{eq:handle_by_factoring} and \ref{eq:ourapproachderivcubicsecondcase} are probably fine alternatives to equations \ref{eq:ourapproachderivcubicrealmul} and \ref{eq:ourapproachderivcubicimagmul} whenever $q t_\text{root}^I$ or $3 q t_\text{root}^R + a$ is small.
}

\subsection{Branch Selection}
\label{subsection:cubicproposedapproach-branchselection}
The root solver (in Section \ref{subsection:cubic-rootsolver}) robustly computes all three roots for any values of the parameters, and those values can be used in equations \ref{eq:ourapproachderivcubicorigreal}, \ref{eq:handle_by_factoring}, \ref{eq:ourapproachderivcubicsecondcase}, \ref{eq:ourapproachderivcubicrealmul}, and \ref{eq:ourapproachderivcubicimagmul} to robustly compute derivatives.
Although the choice of objective function will typically be problem dependent, a few details related to branch selection are discussed here.

For the sake of exposition, consider the change of variables in Section \ref{subsection:cubic-understandasymptotics} highlighted in Figure \ref{fig:roots_phase_diagram} in order to best parallel the discussion in Section \ref{section:branchselection}.
Plugging $\ttargettilde^R$ and $\ttargettilde^I$ into equation \ref{eq:fcubic2d} (top) leads to $\hat{c} = -(\trealtil_\text{root,L})^3 + 3\trealtil_\text{root,L}(\timagtil_\text{root,L})^2 - \hat{b} \trealtil_\text{root,L}$, which can be substituted back into equation \ref{eq:fcubic2d} (top) to obtain
\begin{equation}
\label{eq:branch-selection-cubic-roots-rewrite}
\left( \trealtil_\text{root} - \trealtil_\text{root,L}\right) \left( (\trealtil_\text{root})^2 + \trealtil_\text{root}\trealtil_\text{root,L} + (\trealtil_\text{root,L})^2 + \hat{b} \right) - 3\trealtil_\text{root}(\timagtil_\text{root})^2 + 3\trealtil_\text{root,L}(\timagtil_\text{root,L})^2 = 0
\end{equation}
after refactoring.

\begin{figure}[H]
	\centering
	\begin{subfigure}[b]{0.49\textwidth}
		\centering
		\includegraphics[width=\linewidth]{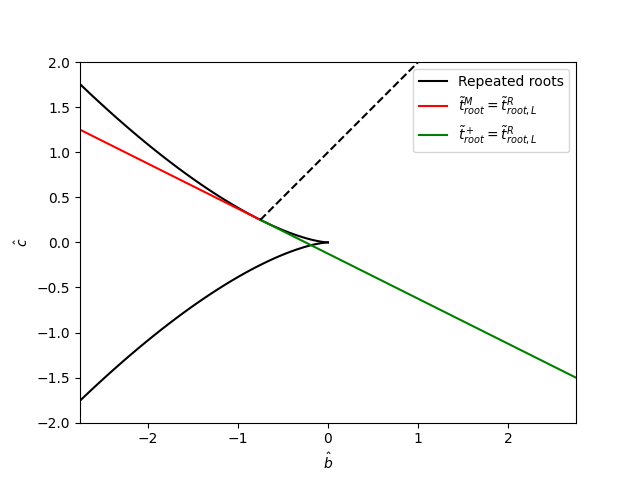}
		\label{}
	\end{subfigure}
	\begin{subfigure}[b]{0.49\textwidth}
		\centering
		\includegraphics[width=\linewidth]{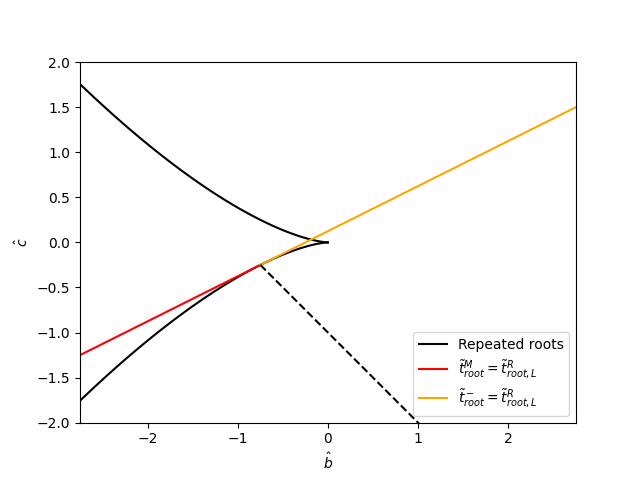}
		\label{}
	\end{subfigure}
	\caption{
    When the target root is real, the line $\hat{c} = -(\trealtil_\text{root,L})^3 - \hat{b} \trealtil_\text{root,L}$ describes the family of solutions.
    $\ttargettilde^R = .5$ is shown to the left (in red/green) and $\ttargettilde^R = -.5$ is shown to the right (in red/yellow).
    The dotted line is $\hat{c} = 8(\trealtil_\text{root,L})^3 + 2 \trealtil_\text{root,L}\hat{b}$, which corresponds to erroneous complex-valued solutions with $\trealtil_\text{root}=\trealtil_\text{root,L}$ obtained when $\timagtil_\text{root,L}$ is not explicitly set to zero in the objective function (similar to the use of equation \ref{eq:badobjectivefunction} in Figure \ref{fig:3.5a}).
    }
	\label{fig:example5cubic}
\end{figure}

\begin{figure}[H]
	\centering
	\begin{subfigure}[b]{0.49\textwidth}
		\centering
	    \includegraphics[width=\linewidth]{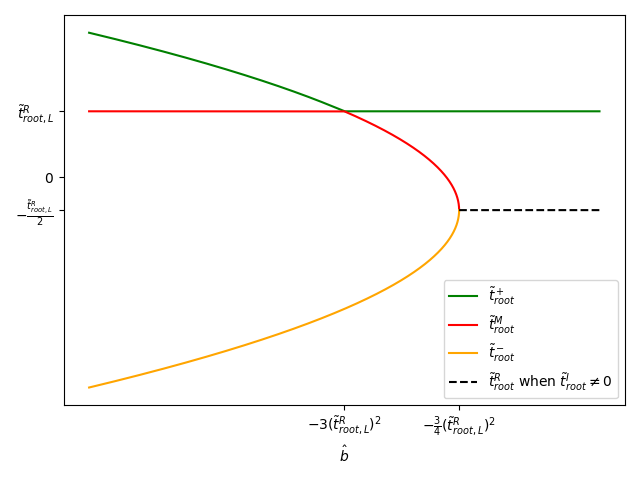}
		\label{}
	\end{subfigure}
	\hfill
	\begin{subfigure}[b]{0.49\textwidth}
		\centering
	    \includegraphics[width=\linewidth]{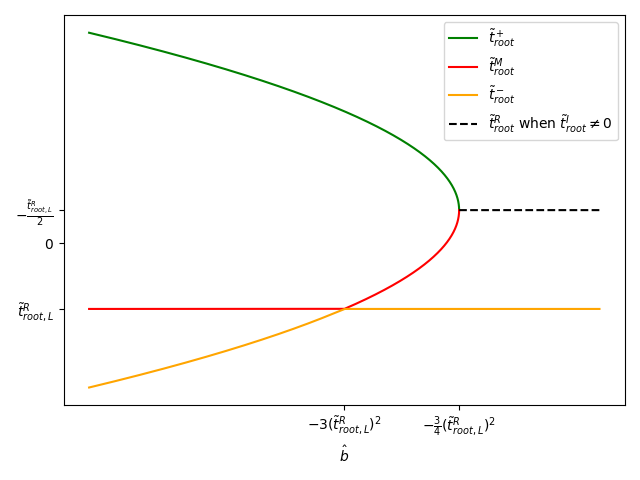}
		\label{}
	\end{subfigure}
	\caption{
    When there are three real roots, one is $\ttargettilde^R$ and the other two are specified by the parabola $\hat{b} = -(\trealtil_\text{root})^2 - \trealtil_\text{root}\trealtil_\text{root,L} - (\trealtil_\text{root,L})^2$.
    When two of the roots are complex, their real part is $\trealtil_\text{root} = -\frac{\trealtil_\text{root,L}}{2}$.
    (Left) $\ttargettilde^R > 0$ and $\ttargettilde^I = 0$. (Right) $\ttargettilde^R < 0$ and $\ttargettilde^I = 0$.
    }
	\label{fig:t-target-cubic}
\end{figure}

When the target root is real (with $\ttargettilde^I = 0$), the line $\hat{c} = -(\trealtil_\text{root,L})^3 - \hat{b} \trealtil_\text{root,L}$ describes the family of solutions (see Figure \ref{fig:example5cubic})
and equation \ref{eq:branch-selection-cubic-roots-rewrite} reduces to 
\begin{equation}
\label{eq:branch-selection-cubic-roots-rewrite-again}
\left( \trealtil_\text{root} - \trealtil_\text{root,L}\right) \left( (\trealtil_\text{root})^2 + \trealtil_\text{root}\trealtil_\text{root,L} + (\trealtil_\text{root,L})^2 + \hat{b} \right) - 3\trealtil_\text{root}(\timagtil_\text{root})^2 = 0
\end{equation}
which has a real root (with $\timagtil_\text{root} = 0$) of $\ttargettilde^R$ as expected.
When there are three real roots, equation \ref{eq:branch-selection-cubic-roots-rewrite-again} dictates that the other two are given by $\hat{b} = -(\trealtil_\text{root})^2 - \trealtil_\text{root}\trealtil_\text{root,L} - (\trealtil_\text{root,L})^2$ as illustrated in Figure \ref{fig:t-target-cubic}.
When two of the roots are complex (with $\timagtil_\text{root} \neq 0$), equation \ref{eq:fcubic2d} (bottom) gives $\hat{b} = (\timagtil_\text{root})^2 - 3 (\trealtil_\text{root})^2$ and equation \ref{eq:branch-selection-cubic-roots-rewrite-again} becomes
\begin{equation}
-2 
\left( \trealtil_\text{root} + \frac{\trealtil_\text{root,L}}{2}\right)
\left( \left(\trealtil_\text{root}-\trealtil_\text{root,L}\right)^2 +(\timagtil_\text{root})^2 \right)
= 0
\end{equation}
where $\trealtil_\text{root} = -\frac{\trealtil_\text{root,L}}{2}$ is the only solution; then, equation \ref{eq:fcubic2d} (bottom) gives $(\timagtil_\text{root})^2 = \frac{3(\ttargettilde^R)^2}{4} + \hat{b}$.
Note that only $\timagtil_\text{root}$ depends on $\hat{b}$, while $\trealtil_\text{root}$ is always $-\frac{\trealtil_\text{root,L}}{2}$ as shown in Figure \ref{fig:t-target-cubic}.

When the target root is complex (with $\ttargettilde^I \neq 0$), equation \ref{eq:fcubic2d} (bottom) leads to a unique $\hat{b} = (\timagtil_\text{root,L})^2 - 3 (\trealtil_\text{root,L})^2$; then, $\hat{c} = 2(\trealtil_\text{root,L})^3 + 2\trealtil_\text{root,L}(\timagtil_\text{root,L})^2$ is also unique.
Eliminating $\hat{b}$ in equation \ref{eq:branch-selection-cubic-roots-rewrite} leads to
\begin{equation}
\label{eq:branch-selection-cubic-roots-rewrite-again-again}
\left( \trealtil_\text{root} + 2\trealtil_\text{root,L}\right)
\left( \left(\trealtil_\text{root}-\trealtil_\text{root,L}\right)^2 +(\timagtil_\text{root,L})^2 \right)
- 3\trealtil_\text{root}(\timagtil_\text{root})^2
= 0
\end{equation}
which has only one real root (with $\timagtil_\text{root} = 0$) of $-2\ttargettilde^R$.
For the complex roots, equation \ref{eq:fcubic2d} (bottom) gives $(\timagtil_\text{root})^2 = 3 (\trealtil_\text{root})^2 + (\timagtil_\text{root,L})^2 - 3 (\trealtil_\text{root,L})^2$. Substituting this into equation \ref{eq:branch-selection-cubic-roots-rewrite-again-again} gives
\begin{equation}
-2\left( \trealtil_\text{root} - \trealtil_\text{root,L}\right)\left( \left(2\trealtil_\text{root}+\trealtil_\text{root,L}\right)^2 + (\timagtil_\text{root,L})^2 \right) = 0
\end{equation}
dictating a real part of $\trealtil_\text{root} = \trealtil_\text{root,L}$; then, equation \ref{eq:fcubic2d} (bottom) gives $(\timagtil_\text{root})^2 = (\ttargettilde^I)^2$.

Next, it is worth briefly commenting on the behavior as $q\to0$ and switches sign.
When a root near $\pm \infty$ changes sign (similar to the bottom right of Figure \ref{fig:normal_fast_example}), $t^M_\text{root}$ becomes one of the $t^\pm_\text{root}$ roots (and vice versa).
As discussed in Remark \ref{re:cubicclampingderivs}, the clamping changes sign when $q$ changes sign keeping the clamping consistent as these two roots change roles;
however, one needs to account for this switching of roles when choosing roots for the objective function.

\textit{
\RemarkCounter{re:switchroles}
Given the need to account for this switching of roles (as $q\to0$ and the root near $\pm\infty$ changes sign) when choosing roots for the objective function, the convention in equation \ref{eq:quadraticroots} now seems prudent (as compared to Remark \ref{re:quadroleswitch}).
}

\clearpage

\subsection{Examples}
\label{subsection:cubicproposedapproach-examples}
In this section, we show the efficacy for our proposed approach using the root solver from Section \ref{subsection:cubic-rootsolver} and equations \ref{eq:ourapproachderivcubicorigreal}, \ref{eq:handle_by_factoring}, \ref{eq:ourapproachderivcubicsecondcase}, \ref{eq:ourapproachderivcubicrealmul}, and \ref{eq:ourapproachderivcubicimagmul} to robustly compute the derivatives.
The sign of the denominator of each scalar multiplier is chosen according to Remark \ref{re:cubicclampingderivs} (and Remark \ref{re:othercubicclampingderivs}).
Equation \ref{eq:lossagainwitht} is used as the objective function, and Adam was used for the optimization.
Each example shows the parameter updates in both reduced spaces (see Figures \ref{fig:roots_phase_diagram} and \ref{fig:roots_tilde_phase_diagram}).
In our newly proposed canonical form, the family of acceptable solutions is cubic instead of linear.
Since the family of acceptable solutions varies as the parameters vary, we (once again) only plot it for the last iteration. 

The roots are shown in the non-reduced space.
When there are three real roots, the largest is labeled $t_\text{root}^+$ and colored green, the smallest is labeled $t_\text{root}^-$ and colored yellow, and the middle root is labeled $t_\text{root}^M$ and colored red.
When two of the roots merge to become complex, the real part of the complex roots is represented by a black dotted line.
The remaining real root changes its label/color (from $t_\text{root}^+$/green to $t_\text{root}^-$/yellow or vice versa) whenever $\hat{c}$ (equal to $\tilde{b})$ changes sign, indicating whether the parameter updates are in the red or blue shaded regions of Figures \ref{fig:roots_phase_diagram} and \ref{fig:roots_tilde_phase_diagram}.
If a parameter update lands exactly on the black ray, we arbitrarily use $t_\text{root}^+$/green for the label/color.
The imaginary parts of the complex roots are colored using the two colors (green/red or red/yellow) not used by the remaining real root.

In each example, we initially choose $t_\text{root}^+$ as the branch under consideration;
however, we switch to $t_\text{root}^M$, or to $t_\text{root}^-$, or to one of the complex roots when appropriate in order to ensure that we are always using the same root.
As discussed at the end of Section \ref{subsection:cubicproposedapproach-branchselection}, such a switch is necessary whenever a root near $\pm \infty$ changes sign.
It is also necessary when $\hat{c}$ (equal to $\tilde{b})$ changes sign while utilizing the single real root (with the other two roots complex) in the objective function.
When the root being utilized merges (switching from being real to complex or vice versa), either of the two new branches may be selected (unless the target root is complex, in which case the sign of the imaginary part matters);
however, we always map the larger/smaller real root to the complex root with positive/negative imaginary part for consistency (noting that merging roots can be considered as an independent quadratic factor along the lines of equations \ref{eq:cubic_factored} and \ref{eq:remaining_roots}).

\begin{figure}[H]
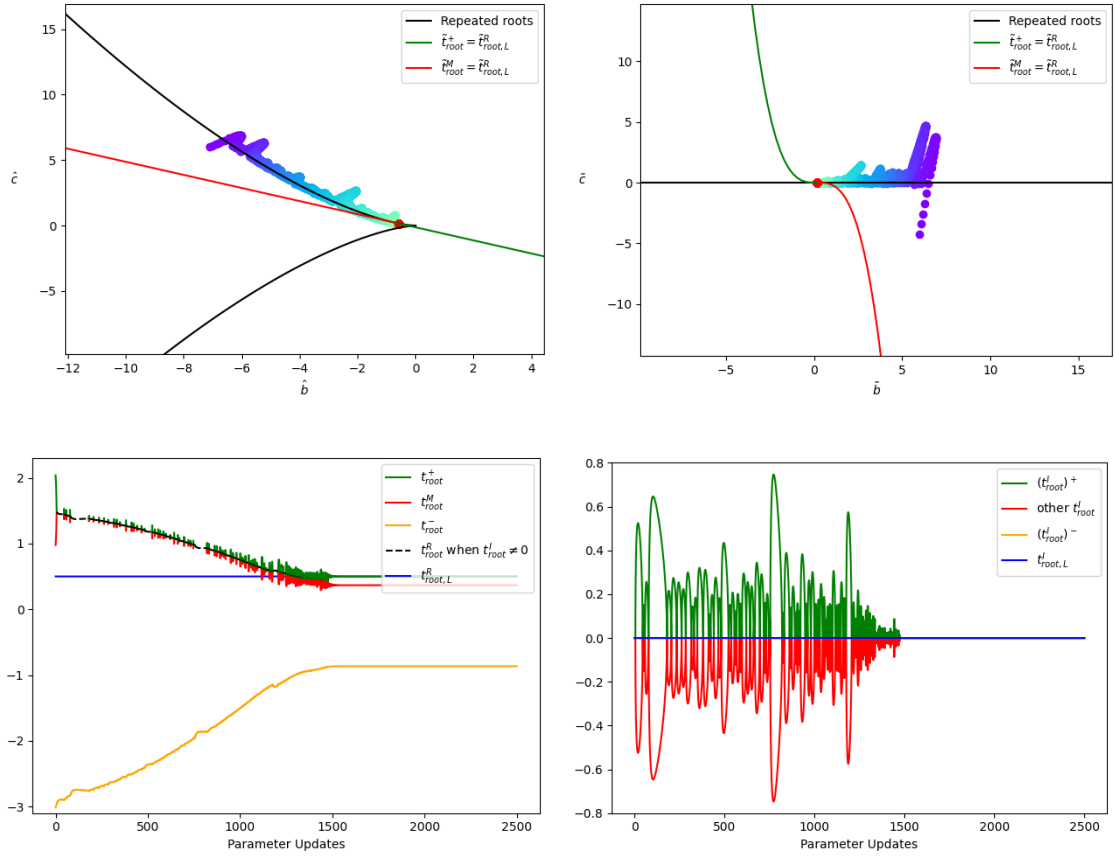

	\centering
    \foreach \deriv in {clamped_given_1000.0} {
	\begin{subfigure}[b]{0.45\textwidth}
		\centering
		\includegraphics[width=\textwidth]{figures/cubic-section/examples_4_parameter/example1/\deriv_normal_phase_diagram_hat_1.png}
		\label{fig:}
	\end{subfigure}
	\begin{subfigure}[b]{0.45\textwidth}
		\centering
		\includegraphics[width=\textwidth]{figures/cubic-section/examples_4_parameter/example1/\deriv_normal_phase_diagram_tilde_1.png}
		\label{fig:}
	\end{subfigure}
	\begin{subfigure}[b]{0.45\textwidth}
		\centering
		\includegraphics[width=\textwidth]{figures/cubic-section/examples_4_parameter/example1/\deriv_normal_new_roots_real.png}
		\label{fig:}
	\end{subfigure}
	\begin{subfigure}[b]{0.45\textwidth}
		\centering
		\includegraphics[width=\textwidth]{figures/cubic-section/examples_4_parameter/example1/\deriv_normal_new_roots_imag.png}
		\label{fig:}
	\end{subfigure}
    }
	\caption{Cubic equation with $q=1$ and $a=0$ held fixed. $M=1000$ is used for the robust division.} 
	\label{fig:cubic-example-1-clamping-normal-1000}
\end{figure}

\begin{figure}[H]
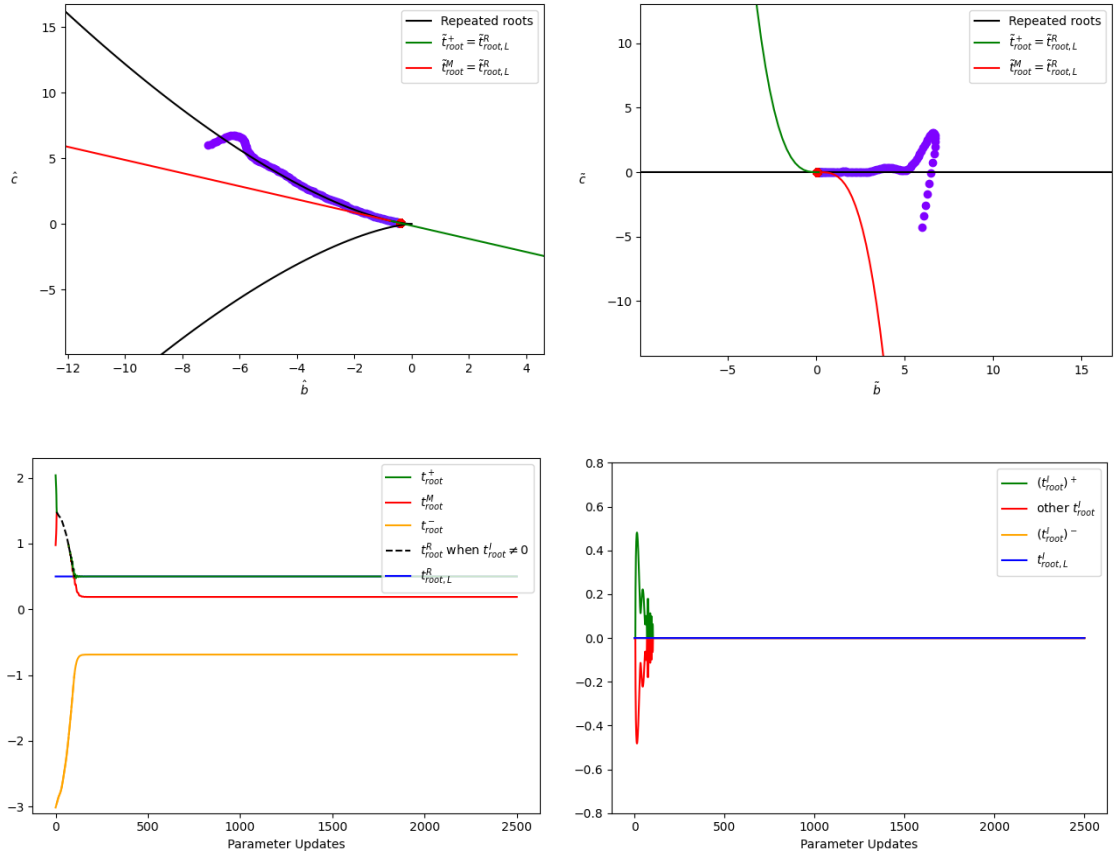

	\centering
    \foreach \deriv in {clamped_given_0.1} {
	\begin{subfigure}[b]{0.45\textwidth}
		\centering
		\includegraphics[width=\textwidth]{figures/cubic-section/examples_4_parameter/example1/\deriv_normal_phase_diagram_hat_1.png}
		\label{fig:}
	\end{subfigure}
	\begin{subfigure}[b]{0.45\textwidth}
		\centering
		\includegraphics[width=\textwidth]{figures/cubic-section/examples_4_parameter/example1/\deriv_normal_phase_diagram_tilde_1.png}
		\label{fig:}
	\end{subfigure}
	\begin{subfigure}[b]{0.45\textwidth}
		\centering
		\includegraphics[width=\textwidth]{figures/cubic-section/examples_4_parameter/example1/\deriv_normal_new_roots_real.png}
		\label{fig:}
	\end{subfigure}
	\begin{subfigure}[b]{0.45\textwidth}
		\centering
		\includegraphics[width=\textwidth]{figures/cubic-section/examples_4_parameter/example1/\deriv_normal_new_roots_imag.png}
		\label{fig:}
	\end{subfigure}
    }
	\caption{Same as Figure \ref{fig:cubic-example-1-clamping-normal-1000}, except $M=.1$ is used in the robust division.} 
	\label{fig:cubic-example-1-clamping-normal-0.1}
\end{figure}

\begin{figure}[H]
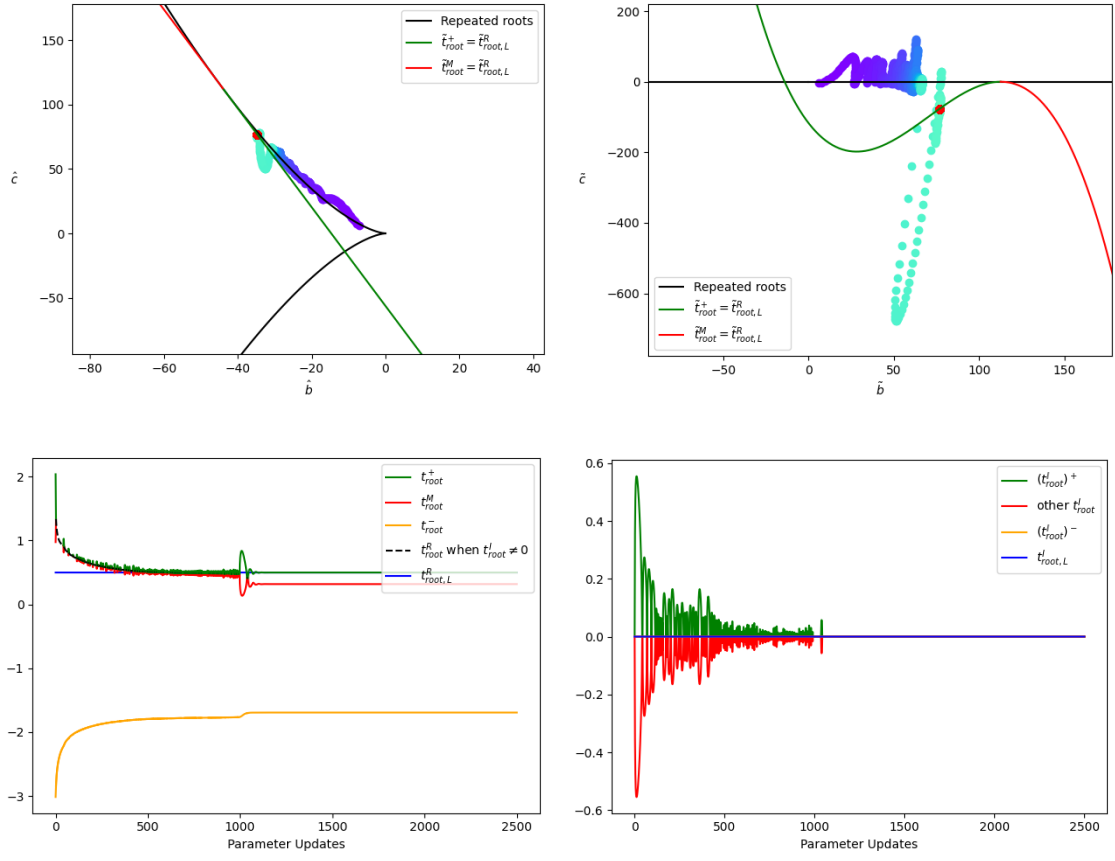

	\centering
    \foreach \deriv in {clamped_given_1000.0} {
	\begin{subfigure}[b]{0.45\textwidth}
		\centering
		\includegraphics[width=\textwidth]{figures/cubic-section/examples_4_parameter/example2/\deriv_normal_phase_diagram_hat_1.png}
		\label{fig:}
	\end{subfigure}
	\begin{subfigure}[b]{0.45\textwidth}
		\centering
		\includegraphics[width=\textwidth]{figures/cubic-section/examples_4_parameter/example2/\deriv_normal_phase_diagram_tilde_1.png}
		\label{fig:}
	\end{subfigure}
	\begin{subfigure}[b]{0.45\textwidth}
		\centering
		\includegraphics[width=\textwidth]{figures/cubic-section/examples_4_parameter/example2/\deriv_normal_new_roots_real.png}
		\label{fig:}
	\end{subfigure}
	\begin{subfigure}[b]{0.45\textwidth}
		\centering
		\includegraphics[width=\textwidth]{figures/cubic-section/examples_4_parameter/example2/\deriv_normal_new_roots_imag.png}
		\label{fig:}
	\end{subfigure}
    }
	\caption{Same as Figure \ref{fig:cubic-example-1-clamping-normal-1000}, except $q$ and $a$ are allowed to vary.} 
	\label{fig:cubic-example-2-clamping-normal-1000}
\end{figure}

We first consider an example similar to that shown in Figure \ref{fig:example-3}.
Setting $dq=da=0$ and holding $q=1$ and $a=0$ removes the first two columns of $\frac{\partial t_\text{root}}{\partial \vec{p}}$.
This leads to $\tilde{t} = qt + \frac{a}{3} = t$, $\hat{b} = b$, and $\hat{c} = c$ for the change of variables in the beginning of Section \ref{subsection:cubic-newcanonicalform}.
Figures \ref{fig:cubic-example-1-clamping-normal-1000} and \ref{fig:cubic-example-1-clamping-normal-0.1} show the results obtained with an initial guess of $[q, a, b, c]^T = [1, 0, -7.1, 6]^T$ choosing $M=1000$ and $M=.1$ for the robust division (respectively).
In Figure \ref{fig:cubic-example-2-clamping-normal-1000}, $q$ and $a$ are allowed to vary.
These examples (as well as many others, omitted for brevity) exhibit the behavior one would expect given the prior discussions in the paper.
Typically, we choose $M=1000$ for the robust division, but smaller values behave as expected (including the occasional lack of convergence illustrated in Figures \ref{fig:example-22-clamping-normal-final} and \ref{fig:example-22-clamping-large_first_grad-final}).
In the subsequent examples, we demonstrate robustness with regard to various degeneracies.

Figure \ref{fig:cubic-example-2-clamping-triple_root} starts out with $t^R_\text{root}=0$ as a triply repeated root, using $[q, a, b, c]^T = [1, 0, 0, 0]^T$.
Tables \ref{tab:triple_root} and \ref{tab:triple_root5} show the first few iterations.
The cubic iterative solver finds one root, and the quadratic solver finds the other two.
Equation \ref{eq:ourapproachderivcubicorigreal} is used to compute the derivatives.
The scalar out front is computed with robust division, and the exactly-zero denominator is clamped to be positive according to Remark \ref{re:cubicclampingderivs}.
This leads to $\frac{\partial t_\text{root}^+}{\partial c}=-1000$ and $\frac{\partial L}{\partial c}=\frac{\partial L}{\partial t_\text{root}^+}\frac{\partial t_\text{root}^+}{\partial c}=(0-\frac{1}{2})(-1000)=500$, while all the other derivatives are identically zero.
As can be seen in Figure \ref{fig:cubic-example-2-clamping-triple_root} and Table \ref{tab:triple_root5}, two of the roots then become complex as the single real root meanders towards a valid solution.

Figures \ref{fig:cubic-example-2-clamping-bad_q} and \ref{fig:cubic-example-2-clamping-bad_q_again} along with Tables \ref{tab:bad_q}, \ref{tab:bad_q5}, \ref{tab:bad_q_again}, and \ref{tab:bad_q_again5} address $q=0$ with $a\neq0$, where two roots are bounded and one root is $\pm\infty$.
Figure \ref{fig:cubic-example-2-clamping-bad_q} starts with $t^2-1=0$, using $[q, a, b, c]^T = [0, 1, 0, -1]^T$.
First, the quadratic root solver finds two roots;
then, the cubic solver finds a root at $-10^{150}$ since $a>0$.
Equation \ref{eq:ourapproachderivcubicorigreal} is used to compute the derivatives.
Since $t_\text{root}^+=1$, the scalar out front is $-\frac{1}{2}$, and all the derivatives are also $-\frac{1}{2}$.
Since
$
\frac{\partial L}{\partial t_\text{root}^+} = 
1-\frac{1}{2} = \frac{1}{2}
$,
all of the derivatives of $L$ are $-\frac{1}{4}$.
Figure \ref{fig:cubic-example-2-clamping-bad_q_again} starts with $-t^2+1=0$, using $[q, a, b, c]^T = [0, -1, 0, 1]^T$.
First, the quadratic root solver finds two roots;
then, the cubic solver finds a root at $10^{150}$ since $a<0$.
Equation \ref{eq:ourapproachderivcubicorigreal} is used to compute the derivatives.
This leads to $\frac{\partial t_\text{root}^+}{\partial q}=-1000$ because of clamping.
Since $\frac{\partial L}{\partial t_\text{root}^+} = 10^{150}-\frac{1}{2} \approx 10^{150}$, $\frac{\partial L}{\partial q}=\frac{\partial L}{\partial t_\text{root}^+}\frac{\partial t_\text{root}^+}{\partial q}=10^{150}(-1000)=-10^{153}$.
Note that $\frac{\partial L}{\partial a}$ is $10^{150}$ times smaller, but still changes $a$ as much as $\frac{\partial L}{\partial q}$ changes $q$ because of the way Adam works.
As can be seen in the figure, $t_\text{root}^+$ is quickly dragged downwards towards $t^R_\text{root,L} = \frac{1}{2}$.
Even after merging with $t_\text{root}^M$ to become a complex root (in iteration 3), the tracked root continues making progress and eventually converges.
Note that in iterations 26-31, the single real root switches from being $t_\text{root}^-$ to $t_\text{root}^+$ as $\hat{c}$ (equal to $\tilde{b}$) changes sign;
however, this has no effect on the complex root being tracked (the tracked complex root switches to being $t_\text{root}^+$ at iteration 41).

Figures \ref{fig:cubic-example-2-clamping-bad_qa_again} and \ref{fig:cubic-example-2-clamping-bad_qa} along with Tables \ref{tab:bad_qa_again}, \ref{tab:bad_qa_again5}, \ref{tab:bad_qa}, and \ref{tab:bad_qa5} address $q=a=0$ with $b\neq0$, where one root is bounded and two roots are $\pm\infty$.
Figure \ref{fig:cubic-example-2-clamping-bad_qa_again} starts with $7.1t + 6=0$, using $[q, a, b, c]^T = [0, 0, 7.1, 6]^T$.
First, the quadratic root solver finds two roots, while choosing a pseudo-sign of $a>0$ to obtain $-10^{150}$ for the unbounded root;
then, the cubic solver also finds a root at $-10^{150}$ using $a>0$.
Equation \ref{eq:ourapproachderivcubicorigreal} is used to compute the derivatives.
Since $t_\text{root}^+$ is not one of the unbounded roots, the derivatives are finite.
Note that $t_\text{root}^+$ immediately merges with $t_\text{root}^M$ to become complex, but later unmerges (at iteration 87) as it converges to the desired solution.
Figure \ref{fig:cubic-example-2-clamping-bad_qa} starts with $-7.1t + 6=0$, using $[q, a, b, c]^T = [0, 0, -7.1, 6]^T$.
First, the quadratic root solver finds two roots, while choosing a pseudo-sign of $a>0$ to obtain $10^{150}$ for the unbounded root;
then, the cubic solver finds a root at $-10^{150}$ using $a>0$.
Equation \ref{eq:ourapproachderivcubicorigreal} is used to compute the derivatives.
This leads to $\frac{\partial t_\text{root}^+}{\partial q}=-1000$ because of clamping.
Since $\frac{\partial L}{\partial t_\text{root}^+} = 10^{150}-\frac{1}{2} \approx 10^{150}$, $\frac{\partial L}{\partial q}=\frac{\partial L}{\partial t_\text{root}^+}\frac{\partial t_\text{root}^+}{\partial q}=10^{150}(-1000)=-10^{153}$.
As can be seen in the figure, $t_\text{root}^+$ is quickly dragged downwards towards $t^R_\text{root,L} = \frac{1}{2}$ and eventually converges.

Figures \ref{fig:cubic-example-2-clamping-bad_qab} and \ref{fig:cubic-example-2-clamping-bad_qab_again} along with Tables \ref{tab:bad_qab}, \ref{tab:bad_qab5}, \ref{tab:bad_qab_again}, and \ref{tab:bad_qab_again5} address $q=a=b=0$ with $c\neq0$, where all three roots are $\pm\infty$.
Figure \ref{fig:cubic-example-2-clamping-bad_qab} starts with $[q, a, b, c]^T = [0, 0, 0, 6]^T$.
First, the quadratic root solver finds roots at $\pm10^{150}$ choosing a pseudo-sign of $a<0$;
then, the cubic solver finds a second root at $10^{150}$ using $a<0$.
Equation \ref{eq:ourapproachderivcubicorigreal} leads to $\frac{\partial t_\text{root}^+}{\partial q}=-1000$ because of clamping,
and $\frac{\partial L}{\partial q}=-10^{153}$.
Figure \ref{fig:cubic-example-2-clamping-bad_qab_again} starts with $[q, a, b, c]^T = [0, 0, 0, -6]^T$.
First, the quadratic root solver finds roots at $\pm10^{150}$ choosing a pseudo-sign of $a>0$;
then, the cubic solver finds a second root at $-10^{150}$ using $a>0$.
Equation \ref{eq:ourapproachderivcubicorigreal} leads to $\frac{\partial t_\text{root}^+}{\partial q}=-1000$ because of clamping,
and $\frac{\partial L}{\partial q}=-10^{153}$.

Figure \ref{fig:cubic-example-2-clamping-bad_qabc} along with Tables \ref{tab:bad_qabc} and \ref{tab:bad_qabc5} address $q=a=b=c=0$, where all three roots are chosen to be identically zero by the quadratic/cubic root solvers.
Equation \ref{eq:ourapproachderivcubicorigreal} leads to $\frac{\partial t_\text{root}^+}{\partial c}=-1000$ as the only nonzero derivative.
Since $\frac{\partial L}{\partial t_\text{root}^+} = 0-\frac{1}{2} = -\frac{1}{2}$, $\frac{\partial L}{\partial c}=-\frac{1}{2}(-1000)=500$;
then, $c$ subsequently becomes negative and the iterations continue similar to Figure \ref{fig:cubic-example-2-clamping-bad_qab_again} (and Tables \ref{tab:bad_qab_again} and \ref{tab:bad_qab_again5}).


\foreach \subexample/\curcap in {
triple_root/{This example starts out with $t^R_\text{root}=0$ as a triply repeated root, using $[q, a, b, c]^T = [1, 0, 0, 0]^T$.},
bad_q/{This example starts with $[q, a, b, c]^T = [0, 1, 0, -1]^T$.},
bad_q_again/{This example starts with $[q, a, b, c]^T = [0, -1, 0, 1]^T$.},
bad_qa_again/{This example starts with $[q, a, b, c]^T = [0, 0, 7.1, 6]^T$.},
bad_qa/{This example starts with $[q, a, b, c]^T = [0, 0, -7.1, 6]^T$.},
bad_qab/{This example starts with $[q, a, b, c]^T = [0, 0, 0, 6]^T$.}, 
bad_qab_again/{This example starts with $[q, a, b, c]^T = [0, 0, 0, -6]^T$.},
bad_qabc/{This example starts out with $t^R_\text{root}=0$ as a triply repeated root, using $[q, a, b, c]^T = [0, 0, 0, 0]^T$.}
}{
\begin{figure}[H]
	\centering
	\begin{subfigure}[b]{0.45\textwidth}
		\centering
		\includegraphics[width=\textwidth]{figures/cubic-section/examples_4_parameter/example2/clamped_given_1000.0_\subexample_phase_diagram_hat_1.png}
		\label{fig:}
	\end{subfigure}
	\begin{subfigure}[b]{0.45\textwidth}
		\centering
		\includegraphics[width=\textwidth]{figures/cubic-section/examples_4_parameter/example2/clamped_given_1000.0_\subexample_phase_diagram_tilde_1.png}
		\label{fig:}
	\end{subfigure}
	\begin{subfigure}[b]{0.45\textwidth}
		\centering
		\includegraphics[width=\textwidth]{figures/cubic-section/examples_4_parameter/example2/clamped_given_1000.0_\subexample_new_roots_real.png}
		\label{fig:}
	\end{subfigure}
	\begin{subfigure}[b]{0.45\textwidth}
		\centering
		\includegraphics[width=\textwidth]{figures/cubic-section/examples_4_parameter/example2/clamped_given_1000.0_\subexample_new_roots_imag.png}
		\label{fig:}
	\end{subfigure}
	\caption{
        \curcap~
        $M=1000$ is used for the robust division.
    }
	\label{fig:cubic-example-2-clamping-\subexample}
\end{figure}
\input{figures/cubic-section/examples_4_parameter/example2/\subexample.tex}
\input{figures/cubic-section/examples_4_parameter/example2/\subexample_5.tex}
}

\clearpage
\section{Conclusions and Future Work}
For general polynomials, one could treat any two real roots or complex conjugate pairs as a quadratic factor.
Subsequently, the polynomial can be written as the quadratic factor times the remaining factor;
moreover, this can be done even when the roots of the quadratic factor have only been found numerically via an iterative solver.
Once a polynomial has been written with all merging, potentially-merging, or close-to-merging pairs of roots in quadratic factors, the product rule can be used to isolate the derivative of each quadratic factor for further consideration along the lines discussed in this paper.
We leave this as future work.
Treating general polynomials would be interesting, since many have used polynomials as approximations to real-world events.
For example, in a pursuer/evader scenario, trajectories can be approximated by polynomials with the pursuer winning when there is a real-valued root to the difference between the polynomials in the allotted time (and the evader winning when the roots stay complex for the entire time).
Importantly, a competitive scenario would be played out near/crossing the numerically-sensitive boundary between real and complex roots.
In the context of differentiable game theory, one could formulate polynomial approximations of strategies with a real-world event occurring when the roots are real and not occurring when the roots are complex.

We briefly summarize some major points here:
Backpropagation through Newton's method is untenable, and there are many cases where it does not work (see Remark \ref{re:shownewtonbackprop});
thus, we utilize implicit differentiation.
In the repeated root case, implicit differentiation yields a coefficient matrix (to the desired derivatives) which is identically equal to zero.
Although numerically untenable, we analytically capture the behavior of the unbounded derivatives and use the results to formulate a robust numerical method for ascertaining search directions;
notably, only the search directions (not the individual derivatives) are required for optimization.
Our approach treats every possible degeneracy (as elucidated somewhat in the examples section), and we provide detailed remarks discussing how to precisely implement various formulas that inherently require L'Hopital's rule in one form or another.

\section{Acknowledgments}
Research supported in part by ONR N00014-19-1-2285 and ONR N00014-21-1-2771.
DJ is supported in part by a Stanford Graduate Fellowship.
RF would like to acknowledge his first PhD advisor, Charles Lange (1942-1993), for introducing him to \cite{bender1978advanced}.
\bibliographystyle{model1-num-names}
\bibliography{refs}

\begin{thebibliography}{85}
\expandafter\ifx\csname natexlab\endcsname\relax\def\natexlab#1{#1}\fi
\providecommand{\url}[1]{\texttt{#1}}
\providecommand{\href}[2]{#2}
\providecommand{\path}[1]{#1}
\providecommand{\DOIprefix}{doi:}
\providecommand{\ArXivprefix}{arXiv:}
\providecommand{\URLprefix}{URL: }
\providecommand{\Pubmedprefix}{pmid:}
\providecommand{\doi}[1]{\href{http://dx.doi.org/#1}{\path{#1}}}
\providecommand{\Pubmed}[1]{\href{pmid:#1}{\path{#1}}}
\providecommand{\bibinfo}[2]{#2}
\ifx\xfnm\relax \def\xfnm[#1]{\unskip,\space#1}\fi
\bibitem[{Lee and Kang(1990)}]{lee1990neural}
\bibinfo{author}{H.~Lee}, \bibinfo{author}{I.~S. Kang},
\newblock \bibinfo{title}{Neural algorithm for solving differential equations},
\newblock \bibinfo{journal}{Journal of Computational Physics}
  \bibinfo{volume}{91} (\bibinfo{year}{1990}) \bibinfo{pages}{110--131}.
\bibitem[{Sirignano and Spiliopoulos(2018)}]{sirignano2018dgm}
\bibinfo{author}{J.~Sirignano}, \bibinfo{author}{K.~Spiliopoulos},
\newblock \bibinfo{title}{Dgm: A deep learning algorithm for solving partial
  differential equations},
\newblock \bibinfo{journal}{Journal of computational physics}
  \bibinfo{volume}{375} (\bibinfo{year}{2018}) \bibinfo{pages}{1339--1364}.
\bibitem[{Raissi and Karniadakis(2018)}]{raissi2018hidden}
\bibinfo{author}{M.~Raissi}, \bibinfo{author}{G.~E. Karniadakis},
\newblock \bibinfo{title}{Hidden physics models: Machine learning of nonlinear
  partial differential equations},
\newblock \bibinfo{journal}{Journal of Computational Physics}
  \bibinfo{volume}{357} (\bibinfo{year}{2018}) \bibinfo{pages}{125--141}.
\bibitem[{Tripathy and Bilionis(2018)}]{tripathy2018deep}
\bibinfo{author}{R.~K. Tripathy}, \bibinfo{author}{I.~Bilionis},
\newblock \bibinfo{title}{Deep uq: Learning deep neural network surrogate
  models for high dimensional uncertainty quantification},
\newblock \bibinfo{journal}{Journal of computational physics}
  \bibinfo{volume}{375} (\bibinfo{year}{2018}) \bibinfo{pages}{565--588}.
\bibitem[{Winovich et~al.(2019)Winovich, Ramani, and Lin}]{winovich2019convpde}
\bibinfo{author}{N.~Winovich}, \bibinfo{author}{K.~Ramani},
  \bibinfo{author}{G.~Lin},
\newblock \bibinfo{title}{Convpde-uq: Convolutional neural networks with
  quantified uncertainty for heterogeneous elliptic partial differential
  equations on varied domains},
\newblock \bibinfo{journal}{Journal of Computational Physics}
  \bibinfo{volume}{394} (\bibinfo{year}{2019}) \bibinfo{pages}{263--279}.
\bibitem[{Raissi et~al.(2019)Raissi, Perdikaris, and
  Karniadakis}]{raissi2019physics}
\bibinfo{author}{M.~Raissi}, \bibinfo{author}{P.~Perdikaris},
  \bibinfo{author}{G.~E. Karniadakis},
\newblock \bibinfo{title}{Physics-informed neural networks: A deep learning
  framework for solving forward and inverse problems involving nonlinear
  partial differential equations},
\newblock \bibinfo{journal}{Journal of Computational Physics}
  \bibinfo{volume}{378} (\bibinfo{year}{2019}) \bibinfo{pages}{686--707}.
\bibitem[{Berg and Nystr{\"o}m(2019)}]{berg2019data}
\bibinfo{author}{J.~Berg}, \bibinfo{author}{K.~Nystr{\"o}m},
\newblock \bibinfo{title}{Data-driven discovery of pdes in complex datasets},
\newblock \bibinfo{journal}{Journal of Computational Physics}
  \bibinfo{volume}{384} (\bibinfo{year}{2019}) \bibinfo{pages}{239--252}.
\bibitem[{Dal~Santo et~al.(2020)Dal~Santo, Deparis, and
  Pegolotti}]{dal2020data}
\bibinfo{author}{N.~Dal~Santo}, \bibinfo{author}{S.~Deparis},
  \bibinfo{author}{L.~Pegolotti},
\newblock \bibinfo{title}{Data driven approximation of parametrized pdes by
  reduced basis and neural networks},
\newblock \bibinfo{journal}{Journal of Computational Physics}
  (\bibinfo{year}{2020}) \bibinfo{pages}{109550}.
\bibitem[{Magiera et~al.(2020)Magiera, Ray, Hesthaven, and
  Rohde}]{magiera2020constraint}
\bibinfo{author}{J.~Magiera}, \bibinfo{author}{D.~Ray}, \bibinfo{author}{J.~S.
  Hesthaven}, \bibinfo{author}{C.~Rohde},
\newblock \bibinfo{title}{Constraint-aware neural networks for riemann
  problems},
\newblock \bibinfo{journal}{Journal of Computational Physics}
  \bibinfo{volume}{409} (\bibinfo{year}{2020}) \bibinfo{pages}{109345}.
\bibitem[{Huang et~al.(2020)Huang, Xu, Farhat, and Darve}]{huang2020learning}
\bibinfo{author}{D.~Z. Huang}, \bibinfo{author}{K.~Xu},
  \bibinfo{author}{C.~Farhat}, \bibinfo{author}{E.~Darve},
\newblock \bibinfo{title}{Learning constitutive relations from indirect
  observations using deep neural networks},
\newblock \bibinfo{journal}{Journal of Computational Physics}
  (\bibinfo{year}{2020}) \bibinfo{pages}{109491}.
\bibitem[{Jagtap et~al.(2020)Jagtap, Kawaguchi, and
  Karniadakis}]{jagtap2020adaptive}
\bibinfo{author}{A.~D. Jagtap}, \bibinfo{author}{K.~Kawaguchi},
  \bibinfo{author}{G.~E. Karniadakis},
\newblock \bibinfo{title}{Adaptive activation functions accelerate convergence
  in deep and physics-informed neural networks},
\newblock \bibinfo{journal}{Journal of Computational Physics}
  \bibinfo{volume}{404} (\bibinfo{year}{2020}) \bibinfo{pages}{109136}.
\bibitem[{Geng et~al.(2020)Geng, Johnson, and Fedkiw}]{geng2020coercing}
\bibinfo{author}{Z.~Geng}, \bibinfo{author}{D.~Johnson},
  \bibinfo{author}{R.~Fedkiw},
\newblock \bibinfo{title}{Coercing machine learning to output physically
  accurate results},
\newblock \bibinfo{journal}{Journal of Computational Physics}
  \bibinfo{volume}{406} (\bibinfo{year}{2020}) \bibinfo{pages}{109099}.
\bibitem[{Alund et~al.(2021)Alund, Iaccarino, and Nordstrom}]{ALUND2021109873}
\bibinfo{author}{O.~Alund}, \bibinfo{author}{G.~Iaccarino},
  \bibinfo{author}{J.~Nordstrom},
\newblock \bibinfo{title}{Learning to differentiate},
\newblock \bibinfo{journal}{Journal of Computational Physics}
  \bibinfo{volume}{424} (\bibinfo{year}{2021}) \bibinfo{pages}{109873}.
\bibitem[{jcp(2020)}]{jcpspecialissue}
\bibinfo{journal}{JCP Special Issue on Machine Learning for Physical Systems,
  Journal of Computational Physics, Guest edited by George E. Karniadakis and
  Jan Hesthaven}  (\bibinfo{year}{2020}).
\bibitem[{Fletcher and Powell(1963)}]{fletcher1963rapidly}
\bibinfo{author}{R.~Fletcher}, \bibinfo{author}{M.~J. Powell},
\newblock \bibinfo{title}{A rapidly convergent descent method for
  minimization},
\newblock \bibinfo{journal}{The computer journal} \bibinfo{volume}{6}
  (\bibinfo{year}{1963}) \bibinfo{pages}{163--168}.
\bibitem[{Broyden(1965)}]{broyden1965class}
\bibinfo{author}{C.~G. Broyden},
\newblock \bibinfo{title}{A class of methods for solving nonlinear simultaneous
  equations},
\newblock \bibinfo{journal}{Mathematics of computation} \bibinfo{volume}{19}
  (\bibinfo{year}{1965}) \bibinfo{pages}{577--593}.
\bibitem[{Broyden(1967)}]{broyden1967quasi}
\bibinfo{author}{C.~G. Broyden},
\newblock \bibinfo{title}{Quasi-newton methods and their application to
  function minimisation},
\newblock \bibinfo{journal}{Mathematics of Computation} \bibinfo{volume}{21}
  (\bibinfo{year}{1967}) \bibinfo{pages}{368--381}.
\bibitem[{Broyden(1969)}]{broyden1969new}
\bibinfo{author}{C.~Broyden},
\newblock \bibinfo{title}{A new double-rank minimisation algorithm. preliminary
  report},
\newblock in: \bibinfo{booktitle}{Notices of the American Mathematical
  Society}, volume~\bibinfo{volume}{16}, \bibinfo{organization}{AMER
  MATHEMATICAL SOC 201 CHARLES ST, PROVIDENCE, RI 02940-2213},
  \bibinfo{year}{1969}, p. \bibinfo{pages}{670}.
\bibitem[{Fletcher(1970)}]{fletcher1970new}
\bibinfo{author}{R.~Fletcher},
\newblock \bibinfo{title}{A new approach to variable metric algorithms},
\newblock \bibinfo{journal}{The computer journal} \bibinfo{volume}{13}
  (\bibinfo{year}{1970}) \bibinfo{pages}{317--322}.
\bibitem[{Goldfarb(1970)}]{goldfarb1970family}
\bibinfo{author}{D.~Goldfarb},
\newblock \bibinfo{title}{A family of variable-metric methods derived by
  variational means},
\newblock \bibinfo{journal}{Mathematics of computation} \bibinfo{volume}{24}
  (\bibinfo{year}{1970}) \bibinfo{pages}{23--26}.
\bibitem[{Shanno(1970)}]{shanno1970conditioning}
\bibinfo{author}{D.~F. Shanno},
\newblock \bibinfo{title}{Conditioning of quasi-newton methods for function
  minimization},
\newblock \bibinfo{journal}{Mathematics of computation} \bibinfo{volume}{24}
  (\bibinfo{year}{1970}) \bibinfo{pages}{647--656}.
\bibitem[{Nocedal(1980)}]{nocedal1980updating}
\bibinfo{author}{J.~Nocedal},
\newblock \bibinfo{title}{Updating quasi-newton matrices with limited storage},
\newblock \bibinfo{journal}{Mathematics of computation} \bibinfo{volume}{35}
  (\bibinfo{year}{1980}) \bibinfo{pages}{773--782}.
\bibitem[{Liu and Nocedal(1989)}]{liu1989limited}
\bibinfo{author}{D.~C. Liu}, \bibinfo{author}{J.~Nocedal},
\newblock \bibinfo{title}{On the limited memory bfgs method for large scale
  optimization},
\newblock \bibinfo{journal}{Mathematical programming} \bibinfo{volume}{45}
  (\bibinfo{year}{1989}) \bibinfo{pages}{503--528}.
\bibitem[{Davidon(1991)}]{davidon1991variable}
\bibinfo{author}{W.~C. Davidon},
\newblock \bibinfo{title}{Variable metric method for minimization},
\newblock \bibinfo{journal}{SIAM Journal on Optimization} \bibinfo{volume}{1}
  (\bibinfo{year}{1991}) \bibinfo{pages}{1--17}.
\bibitem[{Le et~al.(2011)Le, Ngiam, Coates, Lahiri, Prochnow, and
  Ng}]{le2011optimization}
\bibinfo{author}{Q.~V. Le}, \bibinfo{author}{J.~Ngiam},
  \bibinfo{author}{A.~Coates}, \bibinfo{author}{A.~Lahiri},
  \bibinfo{author}{B.~Prochnow}, \bibinfo{author}{A.~Y. Ng},
\newblock \bibinfo{title}{On optimization methods for deep learning},
\newblock in: \bibinfo{booktitle}{Proceedings of the 28th International
  Conference on International Conference on Machine Learning},
  \bibinfo{year}{2011}, pp. \bibinfo{pages}{265--272}.
\bibitem[{Robbins and Monro(1951)}]{robbins1951stochastic}
\bibinfo{author}{H.~Robbins}, \bibinfo{author}{S.~Monro},
\newblock \bibinfo{title}{A stochastic approximation method},
\newblock \bibinfo{journal}{The Annals of Mathematical Statistics}
  (\bibinfo{year}{1951}) \bibinfo{pages}{400--407}.
\bibitem[{Duchi et~al.(2011)Duchi, Hazan, and Singer}]{duchi2011adaptive}
\bibinfo{author}{J.~Duchi}, \bibinfo{author}{E.~Hazan},
  \bibinfo{author}{Y.~Singer},
\newblock \bibinfo{title}{Adaptive subgradient methods for online learning and
  stochastic optimization.},
\newblock \bibinfo{journal}{Journal of machine learning research}
  \bibinfo{volume}{12} (\bibinfo{year}{2011}).
\bibitem[{Tieleman and Hinton(2012)}]{tieleman2012rmsprop}
\bibinfo{author}{T.~Tieleman}, \bibinfo{author}{G.~Hinton},
\newblock \bibinfo{title}{Lecture 6.5: rmsprop: Divide the gradient by a
  running average of its recent magnitude},
\newblock \bibinfo{journal}{Coursera: Neural Networks for machine learning}
  \bibinfo{volume}{4} (\bibinfo{year}{2012}) \bibinfo{pages}{26--31}.
\bibitem[{Zeiler(2012)}]{zeiler2012adadelta}
\bibinfo{author}{M.~D. Zeiler},
\newblock \bibinfo{title}{Adadelta: an adaptive learning rate method},
\newblock \bibinfo{journal}{arXiv preprint arXiv:1212.5701}
  (\bibinfo{year}{2012}).
\bibitem[{Qian(1999)}]{qian1999momentum}
\bibinfo{author}{N.~Qian},
\newblock \bibinfo{title}{On the momentum term in gradient descent learning
  algorithms},
\newblock \bibinfo{journal}{Neural networks} \bibinfo{volume}{12}
  (\bibinfo{year}{1999}) \bibinfo{pages}{145--151}.
\bibitem[{Nesterov(1983)}]{nesterov1983method}
\bibinfo{author}{Y.~E. Nesterov},
\newblock \bibinfo{title}{A method for solving the convex programming problem
  with convergence rate $o(1/k^2)$},
\newblock in: \bibinfo{booktitle}{Doklady Akademii nauk SSSR}, volume
  \bibinfo{volume}{269}, \bibinfo{year}{1983}, pp. \bibinfo{pages}{543--547}.
\bibitem[{Kingma and Ba(2014)}]{kingma2014adam}
\bibinfo{author}{D.~P. Kingma}, \bibinfo{author}{J.~Ba},
\newblock \bibinfo{title}{Adam: A method for stochastic optimization},
\newblock \bibinfo{journal}{arXiv preprint arXiv:1412.6980}
  (\bibinfo{year}{2014}).
\bibitem[{Paszke et~al.(2019)Paszke, Gross, Massa, Lerer, Bradbury, Chanan,
  Killeen, Lin, Gimelshein, Antiga et~al.}]{paszke2019pytorch}
\bibinfo{author}{A.~Paszke}, \bibinfo{author}{S.~Gross},
  \bibinfo{author}{F.~Massa}, \bibinfo{author}{A.~Lerer},
  \bibinfo{author}{J.~Bradbury}, \bibinfo{author}{G.~Chanan},
  \bibinfo{author}{T.~Killeen}, \bibinfo{author}{Z.~Lin},
  \bibinfo{author}{N.~Gimelshein}, \bibinfo{author}{L.~Antiga}, et~al.,
\newblock \bibinfo{title}{Pytorch: An imperative style, high-performance deep
  learning library},
\newblock in: \bibinfo{booktitle}{Advances in neural information processing
  systems}, \bibinfo{year}{2019}, pp. \bibinfo{pages}{8026--8037}.
\bibitem[{Abadi et~al.(2016)Abadi, Agarwal, Barham, Brevdo, Chen, Citro,
  Corrado, Davis, Dean, Devin et~al.}]{abadi2016tensorflow}
\bibinfo{author}{M.~Abadi}, \bibinfo{author}{A.~Agarwal},
  \bibinfo{author}{P.~Barham}, \bibinfo{author}{E.~Brevdo},
  \bibinfo{author}{Z.~Chen}, \bibinfo{author}{C.~Citro}, \bibinfo{author}{G.~S.
  Corrado}, \bibinfo{author}{A.~Davis}, \bibinfo{author}{J.~Dean},
  \bibinfo{author}{M.~Devin}, et~al.,
\newblock \bibinfo{title}{Tensorflow: Large-scale machine learning on
  heterogeneous distributed systems},
\newblock \bibinfo{journal}{arXiv preprint arXiv:1603.04467}
  (\bibinfo{year}{2016}).
\bibitem[{Collobert et~al.(2011)Collobert, Kavukcuoglu, and Farabet}]{torch}
\bibinfo{author}{R.~Collobert}, \bibinfo{author}{K.~Kavukcuoglu},
  \bibinfo{author}{C.~Farabet},
\newblock \bibinfo{title}{Torch7: A matlab-like environment for machine
  learning},
\newblock in: \bibinfo{booktitle}{BigLearn, NIPS Workshop},
  \bibinfo{year}{2011}.
\bibitem[{Jia et~al.(2014)Jia, Shelhamer, Donahue, Karayev, Long, Girshick,
  Guadarrama, and Darrell}]{jia2014caffe}
\bibinfo{author}{Y.~Jia}, \bibinfo{author}{E.~Shelhamer},
  \bibinfo{author}{J.~Donahue}, \bibinfo{author}{S.~Karayev},
  \bibinfo{author}{J.~Long}, \bibinfo{author}{R.~Girshick},
  \bibinfo{author}{S.~Guadarrama}, \bibinfo{author}{T.~Darrell},
\newblock \bibinfo{title}{Caffe: Convolutional architecture for fast feature
  embedding},
\newblock in: \bibinfo{booktitle}{Proceedings of the 22nd ACM international
  conference on Multimedia}, \bibinfo{year}{2014}, pp.
  \bibinfo{pages}{675--678}.
\bibitem[{Al-Rfou et~al.(2016)Al-Rfou, Alain, Almahairi, Angermueller,
  Bahdanau, Ballas, Bastien, Bayer, Belikov, Belopolsky et~al.}]{al2016theano}
\bibinfo{author}{R.~Al-Rfou}, \bibinfo{author}{G.~Alain},
  \bibinfo{author}{A.~Almahairi}, \bibinfo{author}{C.~Angermueller},
  \bibinfo{author}{D.~Bahdanau}, \bibinfo{author}{N.~Ballas},
  \bibinfo{author}{F.~Bastien}, \bibinfo{author}{J.~Bayer},
  \bibinfo{author}{A.~Belikov}, \bibinfo{author}{A.~Belopolsky}, et~al.,
\newblock \bibinfo{title}{Theano: A python framework for fast computation of
  mathematical expressions},
\newblock \bibinfo{journal}{arXiv}  (\bibinfo{year}{2016})
  \bibinfo{pages}{arXiv--1605}.
\bibitem[{Bradbury et~al.(2018)Bradbury, Frostig, Hawkins, Johnson, Leary,
  Maclaurin, Necula, Paszke, Vander{P}las, Wanderman-{M}ilne, and
  Zhang}]{jax2018github}
\bibinfo{author}{J.~Bradbury}, \bibinfo{author}{R.~Frostig},
  \bibinfo{author}{P.~Hawkins}, \bibinfo{author}{M.~J. Johnson},
  \bibinfo{author}{C.~Leary}, \bibinfo{author}{D.~Maclaurin},
  \bibinfo{author}{G.~Necula}, \bibinfo{author}{A.~Paszke},
  \bibinfo{author}{J.~Vander{P}las}, \bibinfo{author}{S.~Wanderman-{M}ilne},
  \bibinfo{author}{Q.~Zhang}, \bibinfo{title}{{JAX}: composable transformations
  of {P}ython+{N}um{P}y programs}, \bibinfo{year}{2018}. \URLprefix
  \url{http://github.com/google/jax}.
\bibitem[{Baydin et~al.(2017)Baydin, Pearlmutter, Radul, and
  Siskind}]{baydin2017automatic}
\bibinfo{author}{A.~G. Baydin}, \bibinfo{author}{B.~A. Pearlmutter},
  \bibinfo{author}{A.~A. Radul}, \bibinfo{author}{J.~M. Siskind},
\newblock \bibinfo{title}{Automatic differentiation in machine learning: a
  survey},
\newblock \bibinfo{journal}{The Journal of Machine Learning Research}
  \bibinfo{volume}{18} (\bibinfo{year}{2017}) \bibinfo{pages}{5595--5637}.
\bibitem[{Schmidhuber(2015)}]{schmidhuber2015deep}
\bibinfo{author}{J.~Schmidhuber},
\newblock \bibinfo{title}{Deep learning in neural networks: An overview},
\newblock \bibinfo{journal}{Neural networks} \bibinfo{volume}{61}
  (\bibinfo{year}{2015}) \bibinfo{pages}{85--117}.
\bibitem[{Heath(2018)}]{heath2018scientific}
\bibinfo{author}{M.~T. Heath}, \bibinfo{title}{Scientific Computing: An
  Introductory Survey, Revised Second Edition}, \bibinfo{publisher}{SIAM},
  \bibinfo{year}{2018}.
\bibitem[{Harari and Albocher(2023)}]{harari2023computation}
\bibinfo{author}{I.~Harari}, \bibinfo{author}{U.~Albocher},
\newblock \bibinfo{title}{Computation of eigenvalues of a real, symmetric
  3$\times$ 3 matrix with particular reference to the pernicious case of two
  nearly equal eigenvalues},
\newblock \bibinfo{journal}{International Journal for Numerical Methods in
  Engineering} \bibinfo{volume}{124} (\bibinfo{year}{2023})
  \bibinfo{pages}{1089--1110}.
\bibitem[{di~Fagnano(1750)}]{di1750produzioni}
\bibinfo{author}{G.~C. di~Fagnano}, \bibinfo{title}{Produzioni matematiche del
  conte Giulio Carlo di Fagnano, marchese de'Toschi, e di Sant'Onorio, nobile
  romano, e patrizio senogagliese...}, volume~\bibinfo{volume}{1},
  \bibinfo{publisher}{Nella stamperia Gavelliana}, \bibinfo{year}{1750}.
\bibitem[{Bridson et~al.(2002)Bridson, Fedkiw, and
  Anderson}]{bridson2002robust}
\bibinfo{author}{R.~Bridson}, \bibinfo{author}{R.~Fedkiw},
  \bibinfo{author}{J.~Anderson},
\newblock \bibinfo{title}{Robust treatment of collisions, contact and friction
  for cloth animation},
\newblock in: \bibinfo{booktitle}{Proceedings of the 29th annual conference on
  Computer graphics and interactive techniques}, \bibinfo{year}{2002}, pp.
  \bibinfo{pages}{594--603}.
\bibitem[{Bailey(2009)}]{bailey2009high}
\bibinfo{author}{D.~H. Bailey},
\newblock \bibinfo{title}{High-precision computation and mathematical physics}
  (\bibinfo{year}{2009}).
\bibitem[{Shewchuk(1997)}]{shewchuk1997adaptive}
\bibinfo{author}{J.~R. Shewchuk},
\newblock \bibinfo{title}{Adaptive precision floating-point arithmetic and fast
  robust geometric predicates},
\newblock \bibinfo{journal}{Discrete \& Computational Geometry}
  \bibinfo{volume}{18} (\bibinfo{year}{1997}) \bibinfo{pages}{305--363}.
\bibitem[{Johnson et~al.(2023)Johnson, Maxfield, Jin, and
  Fedkiw}]{johnson2023softwarebased}
\bibinfo{author}{D.~Johnson}, \bibinfo{author}{T.~Maxfield},
  \bibinfo{author}{Y.~Jin}, \bibinfo{author}{R.~Fedkiw},
  \bibinfo{title}{Software-based automatic differentiation is flawed},
  \bibinfo{year}{2023}. \href{http://arxiv.org/abs/2305.03863}{\tt
  arXiv:2305.03863}.
\bibitem[{LeVeque and Leveque(1992)}]{leveque1992numerical}
\bibinfo{author}{R.~J. LeVeque}, \bibinfo{author}{R.~J. Leveque},
  \bibinfo{title}{Numerical methods for conservation laws},
  volume~\bibinfo{volume}{3}, \bibinfo{publisher}{Springer},
  \bibinfo{year}{1992}.
\bibitem[{Toro(2013)}]{toro2013riemann}
\bibinfo{author}{E.~F. Toro}, \bibinfo{title}{Riemann solvers and numerical
  methods for fluid dynamics: a practical introduction},
  \bibinfo{publisher}{Springer Science \& Business Media},
  \bibinfo{year}{2013}.
\bibitem[{Shu and Osher(1989)}]{shu1989efficient}
\bibinfo{author}{C.-W. Shu}, \bibinfo{author}{S.~Osher},
\newblock \bibinfo{title}{Efficient implementation of essentially
  non-oscillatory shock-capturing schemes, ii},
\newblock in: \bibinfo{booktitle}{Upwind and High-Resolution Schemes},
  \bibinfo{publisher}{Springer}, \bibinfo{year}{1989}, pp.
  \bibinfo{pages}{328--374}.
\bibitem[{Osher et~al.(2004)Osher, Fedkiw, and Piechor}]{osher2004level}
\bibinfo{author}{S.~Osher}, \bibinfo{author}{R.~Fedkiw},
  \bibinfo{author}{K.~Piechor},
\newblock \bibinfo{title}{Level set methods and dynamic implicit surfaces},
\newblock \bibinfo{journal}{Appl. Mech. Rev.} \bibinfo{volume}{57}
  (\bibinfo{year}{2004}) \bibinfo{pages}{B15--B15}.
\bibitem[{Fedkiw et~al.(1999)Fedkiw, Aslam, Merriman, Osher
  et~al.}]{fedkiw1999non}
\bibinfo{author}{R.~P. Fedkiw}, \bibinfo{author}{T.~Aslam},
  \bibinfo{author}{B.~Merriman}, \bibinfo{author}{S.~Osher}, et~al.,
\newblock \bibinfo{title}{A non-oscillatory eulerian approach to interfaces in
  multimaterial flows (the ghost fluid method)},
\newblock \bibinfo{journal}{Journal of computational physics}
  \bibinfo{volume}{152} (\bibinfo{year}{1999}) \bibinfo{pages}{457--492}.
\bibitem[{Liu et~al.(2000)Liu, Fedkiw, and Kang}]{liu2000boundary}
\bibinfo{author}{X.-D. Liu}, \bibinfo{author}{R.~P. Fedkiw},
  \bibinfo{author}{M.~Kang},
\newblock \bibinfo{title}{A boundary condition capturing method for poisson's
  equation on irregular domains},
\newblock \bibinfo{journal}{Journal of computational Physics}
  \bibinfo{volume}{160} (\bibinfo{year}{2000}) \bibinfo{pages}{151--178}.
\bibitem[{Kang et~al.(2000)Kang, Fedkiw, and Liu}]{kang2000boundary}
\bibinfo{author}{M.~Kang}, \bibinfo{author}{R.~P. Fedkiw},
  \bibinfo{author}{X.-D. Liu},
\newblock \bibinfo{title}{A boundary condition capturing method for multiphase
  incompressible flow},
\newblock \bibinfo{journal}{Journal of Scientific Computing}
  \bibinfo{volume}{15} (\bibinfo{year}{2000}) \bibinfo{pages}{323--360}.
\bibitem[{Li and Ito(2006)}]{li2006immersed}
\bibinfo{author}{Z.~Li}, \bibinfo{author}{K.~Ito}, \bibinfo{title}{The immersed
  interface method: numerical solutions of PDEs involving interfaces and
  irregular domains}, \bibinfo{publisher}{SIAM}, \bibinfo{year}{2006}.
\bibitem[{Mo{\"e}s et~al.(1999)Mo{\"e}s, Dolbow, and
  Belytschko}]{moes1999finite}
\bibinfo{author}{N.~Mo{\"e}s}, \bibinfo{author}{J.~Dolbow},
  \bibinfo{author}{T.~Belytschko},
\newblock \bibinfo{title}{A finite element method for crack growth without
  remeshing},
\newblock \bibinfo{journal}{International journal for numerical methods in
  engineering} \bibinfo{volume}{46} (\bibinfo{year}{1999})
  \bibinfo{pages}{131--150}.
\bibitem[{Jameson(1991)}]{jameson1991time}
\bibinfo{author}{A.~Jameson},
\newblock \bibinfo{title}{Time dependent calculations using multigrid, with
  applications to unsteady flows past airfoils and wings},
\newblock in: \bibinfo{booktitle}{10th Computational Fluid Dynamics
  Conference}, \bibinfo{year}{1991}, p. \bibinfo{pages}{1596}.
\bibitem[{Belov et~al.(1995)Belov, Martinelli, and Jameson}]{belov1995new}
\bibinfo{author}{A.~Belov}, \bibinfo{author}{L.~Martinelli},
  \bibinfo{author}{A.~Jameson},
\newblock \bibinfo{title}{A new implicit algorithm with multigrid for unsteady
  incompressible flow calculations},
\newblock in: \bibinfo{booktitle}{33rd Aerospace sciences meeting and exhibit},
  \bibinfo{year}{1995}, p.~\bibinfo{pages}{49}.
\bibitem[{Jiang and Shu(1996)}]{jiang1996efficient}
\bibinfo{author}{G.-S. Jiang}, \bibinfo{author}{C.-W. Shu},
\newblock \bibinfo{title}{Efficient implementation of weighted eno schemes},
\newblock \bibinfo{journal}{Journal of computational physics}
  \bibinfo{volume}{126} (\bibinfo{year}{1996}) \bibinfo{pages}{202--228}.
\bibitem[{Belytschko and Mish(2001)}]{belytschko2001computability}
\bibinfo{author}{T.~Belytschko}, \bibinfo{author}{K.~Mish},
\newblock \bibinfo{title}{Computability in non-linear solid mechanics},
\newblock \bibinfo{journal}{International Journal for Numerical Methods in
  Engineering} \bibinfo{volume}{52} (\bibinfo{year}{2001})
  \bibinfo{pages}{3--21}.
\bibitem[{Kadioglu et~al.(2005)Kadioglu, Sussman, Osher, Wright, and
  Kang}]{kadioglu2005second}
\bibinfo{author}{S.~Y. Kadioglu}, \bibinfo{author}{M.~Sussman},
  \bibinfo{author}{S.~Osher}, \bibinfo{author}{J.~P. Wright},
  \bibinfo{author}{M.~Kang},
\newblock \bibinfo{title}{A second order primitive preconditioner for solving
  all speed multi-phase flows},
\newblock \bibinfo{journal}{Journal of computational physics}
  \bibinfo{volume}{209} (\bibinfo{year}{2005}) \bibinfo{pages}{477--503}.
\bibitem[{Teran et~al.(2005)Teran, Sifakis, Irving, and
  Fedkiw}]{teran2005robust}
\bibinfo{author}{J.~Teran}, \bibinfo{author}{E.~Sifakis},
  \bibinfo{author}{G.~Irving}, \bibinfo{author}{R.~Fedkiw},
\newblock \bibinfo{title}{Robust quasistatic finite elements and flesh
  simulation},
\newblock in: \bibinfo{booktitle}{Proceedings of the 2005 ACM
  SIGGRAPH/Eurographics Symposium on Computer Animation}, SCA '05,
  \bibinfo{publisher}{ACM}, \bibinfo{address}{New York, NY, USA},
  \bibinfo{year}{2005}, pp. \bibinfo{pages}{181--190}.
\bibitem[{Kwatra et~al.(2009)Kwatra, Su, Gr{\'e}tarsson, and
  Fedkiw}]{kwatra2009method}
\bibinfo{author}{N.~Kwatra}, \bibinfo{author}{J.~Su}, \bibinfo{author}{J.~T.
  Gr{\'e}tarsson}, \bibinfo{author}{R.~Fedkiw},
\newblock \bibinfo{title}{A method for avoiding the acoustic time step
  restriction in compressible flow},
\newblock \bibinfo{journal}{Journal of Computational Physics}
  \bibinfo{volume}{228} (\bibinfo{year}{2009}) \bibinfo{pages}{4146--4161}.
\bibitem[{Hughes(2012)}]{hughes2012finite}
\bibinfo{author}{T.~J. Hughes}, \bibinfo{title}{The finite element method:
  linear static and dynamic finite element analysis},
  \bibinfo{publisher}{Courier Corporation}, \bibinfo{year}{2012}.
\bibitem[{De~Borst et~al.(2012)De~Borst, Crisfield, Remmers, and
  Verhoosel}]{de2012nonlinear}
\bibinfo{author}{R.~De~Borst}, \bibinfo{author}{M.~A. Crisfield},
  \bibinfo{author}{J.~J. Remmers}, \bibinfo{author}{C.~V. Verhoosel},
  \bibinfo{title}{Nonlinear finite element analysis of solids and structures},
  \bibinfo{publisher}{John Wiley \& Sons}, \bibinfo{year}{2012}.
\bibitem[{Liu et~al.(1994)Liu, Osher, Chan et~al.}]{liu1994weighted}
\bibinfo{author}{X.-D. Liu}, \bibinfo{author}{S.~Osher},
  \bibinfo{author}{T.~Chan}, et~al.,
\newblock \bibinfo{title}{Weighted essentially non-oscillatory schemes},
\newblock \bibinfo{journal}{Journal of computational physics}
  \bibinfo{volume}{115} (\bibinfo{year}{1994}) \bibinfo{pages}{200--212}.
\bibitem[{Belytschko et~al.(2013)Belytschko, Liu, Moran, and
  Elkhodary}]{belytschko2013nonlinear}
\bibinfo{author}{T.~Belytschko}, \bibinfo{author}{W.~K. Liu},
  \bibinfo{author}{B.~Moran}, \bibinfo{author}{K.~Elkhodary},
  \bibinfo{title}{Nonlinear finite elements for continua and structures},
  \bibinfo{publisher}{John wiley \& sons}, \bibinfo{year}{2013}.
\bibitem[{Shu(2020)}]{shu2020}
\bibinfo{author}{C.-W. Shu},
\newblock \bibinfo{title}{Essentially non-oscillatory and weighted essentially
  non-oscillatory schemes},
\newblock \bibinfo{journal}{Acta Numerica}  (\bibinfo{year}{2020})
  \bibinfo{pages}{1--63}.
\bibitem[{Hu et~al.(2019)Hu, Anderson, Li, Sun, Carr, Ragan-Kelley, and
  Durand}]{hu2019difftaichi}
\bibinfo{author}{Y.~Hu}, \bibinfo{author}{L.~Anderson}, \bibinfo{author}{T.-M.
  Li}, \bibinfo{author}{Q.~Sun}, \bibinfo{author}{N.~Carr},
  \bibinfo{author}{J.~Ragan-Kelley}, \bibinfo{author}{F.~Durand},
\newblock \bibinfo{title}{Difftaichi: Differentiable programming for physical
  simulation},
\newblock \bibinfo{journal}{arXiv preprint arXiv:1910.00935}
  (\bibinfo{year}{2019}).
\bibitem[{Zhuang et~al.(2020)Zhuang, Dvornek, Li, Tatikonda, Papademetris, and
  Duncan}]{zhuang2020adaptive}
\bibinfo{author}{J.~Zhuang}, \bibinfo{author}{N.~Dvornek},
  \bibinfo{author}{X.~Li}, \bibinfo{author}{S.~Tatikonda},
  \bibinfo{author}{X.~Papademetris}, \bibinfo{author}{J.~Duncan},
\newblock \bibinfo{title}{Adaptive checkpoint adjoint method for gradient
  estimation in neural ode},
\newblock \bibinfo{journal}{arXiv preprint arXiv:2006.02493}
  (\bibinfo{year}{2020}).
\bibitem[{Suh et~al.(2022)Suh, Simchowitz, Zhang, and
  Tedrake}]{suh2022differentiable}
\bibinfo{author}{H.~J. Suh}, \bibinfo{author}{M.~Simchowitz},
  \bibinfo{author}{K.~Zhang}, \bibinfo{author}{R.~Tedrake},
\newblock \bibinfo{title}{Do differentiable simulators give better policy
  gradients?},
\newblock in: \bibinfo{booktitle}{International Conference on Machine
  Learning}, \bibinfo{organization}{PMLR}, \bibinfo{year}{2022}, pp.
  \bibinfo{pages}{20668--20696}.
\bibitem[{Metz et~al.(2021)Metz, Freeman, Schoenholz, and
  Kachman}]{metz2021gradients}
\bibinfo{author}{L.~Metz}, \bibinfo{author}{C.~D. Freeman},
  \bibinfo{author}{S.~S. Schoenholz}, \bibinfo{author}{T.~Kachman},
\newblock \bibinfo{title}{Gradients are not all you need},
\newblock \bibinfo{journal}{arXiv preprint arXiv:2111.05803}
  (\bibinfo{year}{2021}).
\bibitem[{Srivastava et~al.(2014)Srivastava, Hinton, Krizhevsky, Sutskever, and
  Salakhutdinov}]{srivastava2014dropout}
\bibinfo{author}{N.~Srivastava}, \bibinfo{author}{G.~Hinton},
  \bibinfo{author}{A.~Krizhevsky}, \bibinfo{author}{I.~Sutskever},
  \bibinfo{author}{R.~Salakhutdinov},
\newblock \bibinfo{title}{Dropout: a simple way to prevent neural networks from
  overfitting},
\newblock \bibinfo{journal}{The journal of machine learning research}
  \bibinfo{volume}{15} (\bibinfo{year}{2014}) \bibinfo{pages}{1929--1958}.
\bibitem[{Stewart(2011)}]{stewart2011dynamics}
\bibinfo{author}{D.~E. Stewart}, \bibinfo{title}{Dynamics with Inequalities:
  impacts and hard constraints}, \bibinfo{publisher}{SIAM},
  \bibinfo{year}{2011}.
\bibitem[{Stewart and Trinkle(1996)}]{stewart1996implicit}
\bibinfo{author}{D.~E. Stewart}, \bibinfo{author}{J.~C. Trinkle},
\newblock \bibinfo{title}{An implicit time-stepping scheme for rigid body
  dynamics with inelastic collisions and coulomb friction},
\newblock \bibinfo{journal}{International Journal for Numerical Methods in
  Engineering} \bibinfo{volume}{39} (\bibinfo{year}{1996})
  \bibinfo{pages}{2673--2691}.
\bibitem[{Ferguson et~al.(2021)Ferguson, Li, Schneider, Gil-Ureta, Langlois,
  Jiang, Zorin, Kaufman, and Panozzo}]{ferguson2021intersection}
\bibinfo{author}{Z.~Ferguson}, \bibinfo{author}{M.~Li},
  \bibinfo{author}{T.~Schneider}, \bibinfo{author}{F.~Gil-Ureta},
  \bibinfo{author}{T.~Langlois}, \bibinfo{author}{C.~Jiang},
  \bibinfo{author}{D.~Zorin}, \bibinfo{author}{D.~M. Kaufman},
  \bibinfo{author}{D.~Panozzo},
\newblock \bibinfo{title}{Intersection-free rigid body dynamics},
\newblock \bibinfo{journal}{ACM Transactions on Graphics} \bibinfo{volume}{40}
  (\bibinfo{year}{2021}) \bibinfo{pages}{183}.
\bibitem[{Agrawal et~al.(2019)Agrawal, Amos, Barratt, Boyd, Diamond, and
  Kolter}]{agrawal2019differentiable}
\bibinfo{author}{A.~Agrawal}, \bibinfo{author}{B.~Amos},
  \bibinfo{author}{S.~Barratt}, \bibinfo{author}{S.~Boyd},
  \bibinfo{author}{S.~Diamond}, \bibinfo{author}{J.~Z. Kolter},
\newblock \bibinfo{title}{Differentiable convex optimization layers},
\newblock in: \bibinfo{booktitle}{Advances in neural information processing
  systems}, \bibinfo{year}{2019}, pp. \bibinfo{pages}{9562--9574}.
\bibitem[{Kolter et~al.(2020)Kolter, Duvenaud, and Johnson}]{kolter2020nips}
\bibinfo{author}{Z.~Kolter}, \bibinfo{author}{D.~Duvenaud},
  \bibinfo{author}{M.~Johnson},
\newblock \bibinfo{title}{Deep implicit layers - neural odes, deep equilibirum
  models, and beyond},
\newblock \bibinfo{journal}{NeurIPS Tutorial}  (\bibinfo{year}{2020}).
\bibitem[{Liang et~al.(2019)Liang, Lin, and Koltun}]{liang2019differentiable}
\bibinfo{author}{J.~Liang}, \bibinfo{author}{M.~Lin},
  \bibinfo{author}{V.~Koltun},
\newblock \bibinfo{title}{Differentiable cloth simulation for inverse
  problems},
\newblock in: \bibinfo{booktitle}{Advances in Neural Information Processing
  Systems}, \bibinfo{year}{2019}, pp. \bibinfo{pages}{772--781}.
\bibitem[{Qiao et~al.(2020)Qiao, Liang, Koltun, and Lin}]{qiao2020scalable}
\bibinfo{author}{Y.-L. Qiao}, \bibinfo{author}{J.~Liang},
  \bibinfo{author}{V.~Koltun}, \bibinfo{author}{M.~C. Lin},
\newblock \bibinfo{title}{Scalable differentiable physics for learning and
  control},
\newblock \bibinfo{journal}{arXiv preprint arXiv:2007.02168}
  (\bibinfo{year}{2020}).
\bibitem[{Bolte et~al.(2021)Bolte, Le, Pauwels, and
  Silveti-Falls}]{bolte2021nips}
\bibinfo{author}{J.~Bolte}, \bibinfo{author}{T.~Le},
  \bibinfo{author}{E.~Pauwels}, \bibinfo{author}{T.~Silveti-Falls},
\newblock \bibinfo{title}{Nonsmooth implicit differentiation for
  machine-learning and optimization},
\newblock in: \bibinfo{editor}{M.~Ranzato}, \bibinfo{editor}{A.~Beygelzimer},
  \bibinfo{editor}{Y.~Dauphin}, \bibinfo{editor}{P.~Liang},
  \bibinfo{editor}{J.~W. Vaughan} (Eds.), \bibinfo{booktitle}{Advances in
  Neural Information Processing Systems}, volume~\bibinfo{volume}{34},
  \bibinfo{publisher}{Curran Associates, Inc.}, \bibinfo{year}{2021}, pp.
  \bibinfo{pages}{13537--13549}. \URLprefix
  \url{https://proceedings.neurips.cc/paper/2021/file/70afbf2259b4449d8ae1429e054df1b1-Paper.pdf}.
\bibitem[{Teran et~al.(2003)Teran, Blemker, Hing, and Fedkiw}]{teran2003finite}
\bibinfo{author}{J.~Teran}, \bibinfo{author}{S.~Blemker},
  \bibinfo{author}{V.~N.~T. Hing}, \bibinfo{author}{R.~Fedkiw},
\newblock \bibinfo{title}{Finite volume methods for the simulation of skeletal
  muscle},
\newblock in: \bibinfo{booktitle}{Proceedings of the 2003 ACM
  SIGGRAPH/Eurographics symposium on Computer animation},
  \bibinfo{organization}{Citeseer}, \bibinfo{year}{2003}, pp.
  \bibinfo{pages}{68--74}.
\bibitem[{Paige and Saunders(1975)}]{paige1975solution}
\bibinfo{author}{C.~C. Paige}, \bibinfo{author}{M.~A. Saunders},
\newblock \bibinfo{title}{Solution of sparse indefinite systems of linear
  equations},
\newblock \bibinfo{journal}{SIAM journal on numerical analysis}
  \bibinfo{volume}{12} (\bibinfo{year}{1975}) \bibinfo{pages}{617--629}.
\bibitem[{Lanczos(1950)}]{lanczos1950iteration}
\bibinfo{author}{C.~Lanczos}, \bibinfo{title}{An iteration method for the
  solution of the eigenvalue problem of linear differential and integral
  operators}, \bibinfo{publisher}{United States Governm. Press Office Los
  Angeles, CA}, \bibinfo{year}{1950}.
\bibitem[{Bender and Orszag(1978)}]{bender1978advanced}
\bibinfo{author}{C.~M. Bender}, \bibinfo{author}{S.~A. Orszag},
  \bibinfo{title}{Advanced mathematical methods for scientists and engineers
  (International Series in Pure and Applied Mathematics)},
  \bibinfo{publisher}{McGraw-Hill}, \bibinfo{year}{1978}.

\end{thebibliography}
\newpage
\section*{Appendix A: Adam Oscillations Near Convergence}
In this paper, we took an extremely thorough approach to the numerical examples (both those illustrated in the paper and those omitted for brevity) in order to provide convincing evidence for some of our (not so obvious) claims. In doing so, we discovered a peculiarity with Adam optimization that does not seem to be addressed in the literature in spite of its extreme popularity for training neural networks. The Adam update contains a division of a so-called first moment by a so-called second moment, where both moments are calculated by averaging new information with older information (as is typical for momentum-style optimization methods). Typically, the default parameters (for Adam) put more weight on the new information when including it into the first moment than when including it into the second moment. This allows a disturbance to increase the numerator faster than the denominator. Under normal circumstances, this works well; however, when both the numerator and denominator are very small (as occurs during convergence), this can lead to the numerator growing faster than the denominator (creating oscillations that disturb the convergence).

\begin{figure}[H]
	\centering
	\begin{subfigure}[b]{0.3\textwidth}
		\centering
		\includegraphics[width=\textwidth]{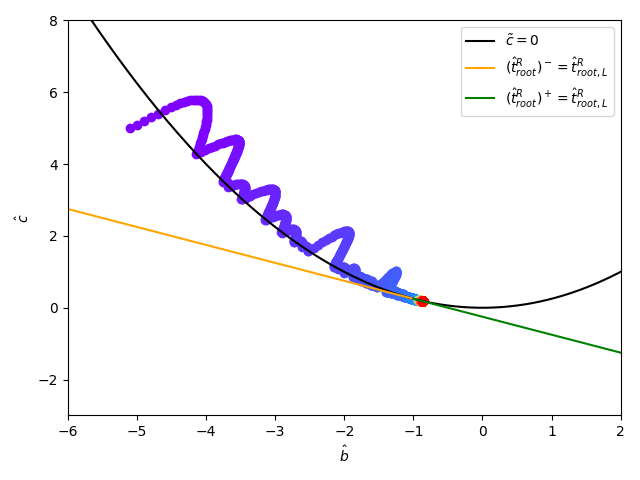}
		\label{fig:}
	\end{subfigure}
	\begin{subfigure}[b]{0.3\textwidth}
		\centering
		\includegraphics[width=\textwidth]{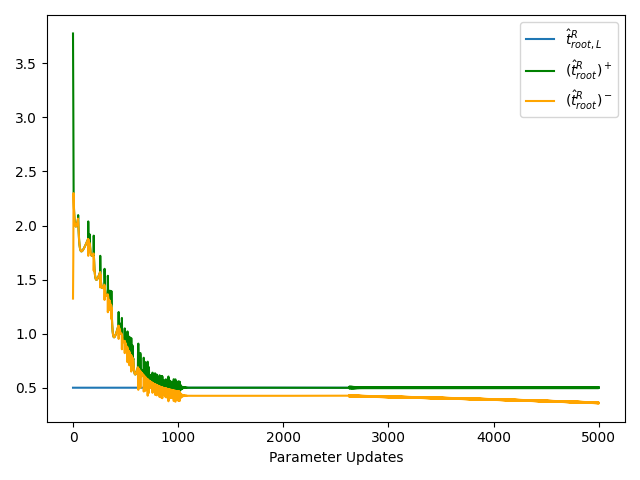}
		\label{fig:}
	\end{subfigure}
	\begin{subfigure}[b]{0.3\textwidth}
		\centering
		\includegraphics[width=\textwidth]{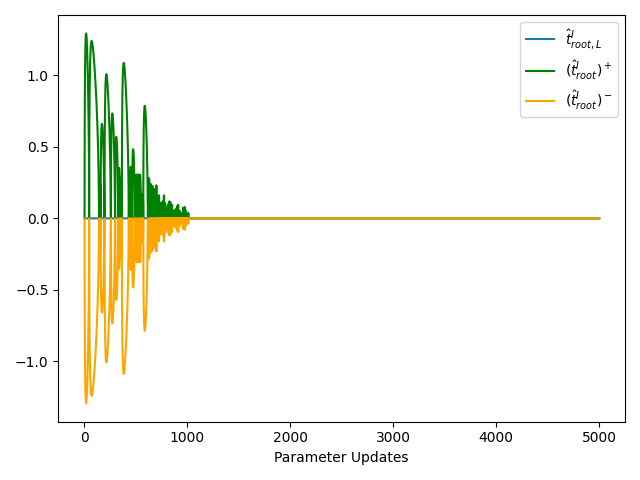}
		\label{fig:}
	\end{subfigure}
	\begin{subfigure}[b]{0.3\textwidth}
		\centering
		\includegraphics[width=\textwidth]{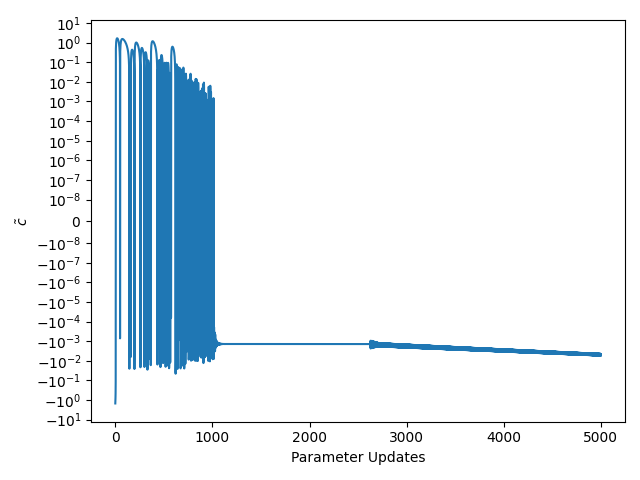}
		\label{fig:}
	\end{subfigure}
	\begin{subfigure}[b]{0.3\textwidth}
		\centering
		\includegraphics[width=\textwidth]{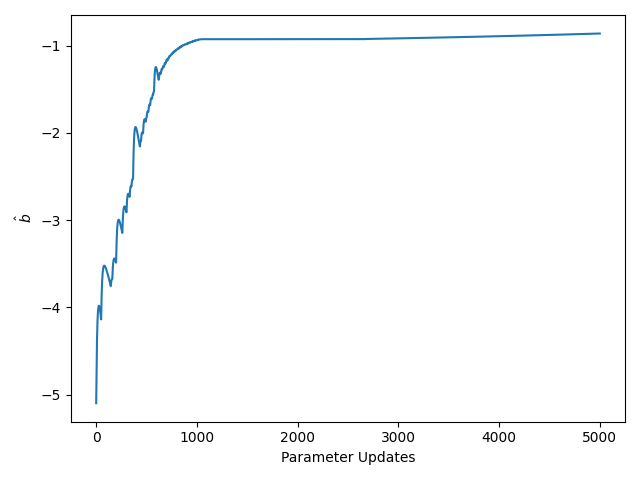}
		\label{fig:}
	\end{subfigure}
	\begin{subfigure}[b]{0.3\textwidth}
		\centering
		\includegraphics[width=\textwidth]{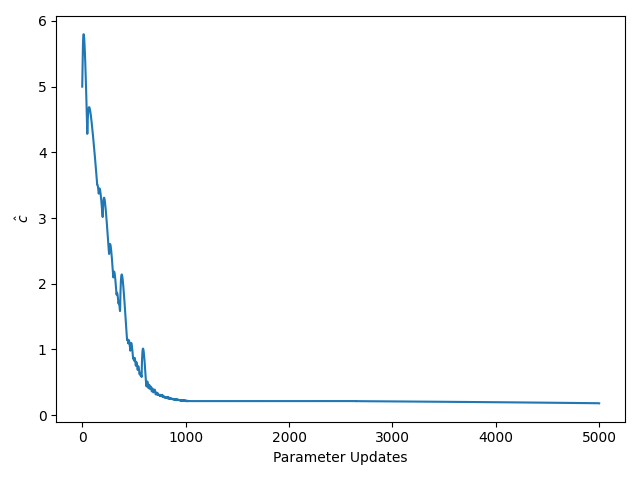}
		\label{fig:}
	\end{subfigure}
	\begin{subfigure}[b]{0.3\textwidth}
		\centering
		\includegraphics[width=\textwidth]{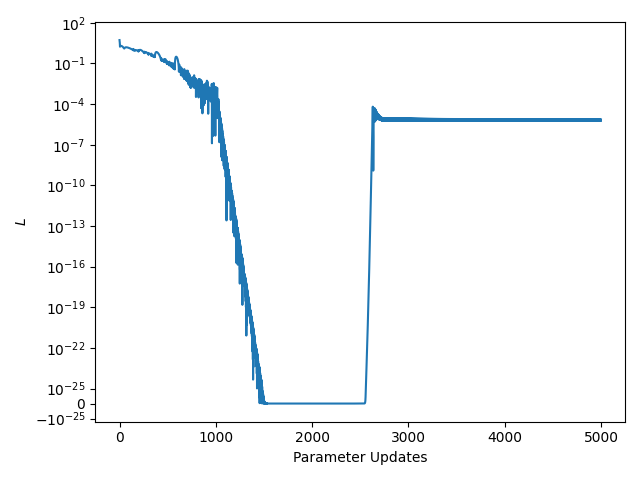}
		\label{fig:}
	\end{subfigure}
	\begin{subfigure}[b]{0.3\textwidth}
		\centering
		\includegraphics[width=\textwidth]{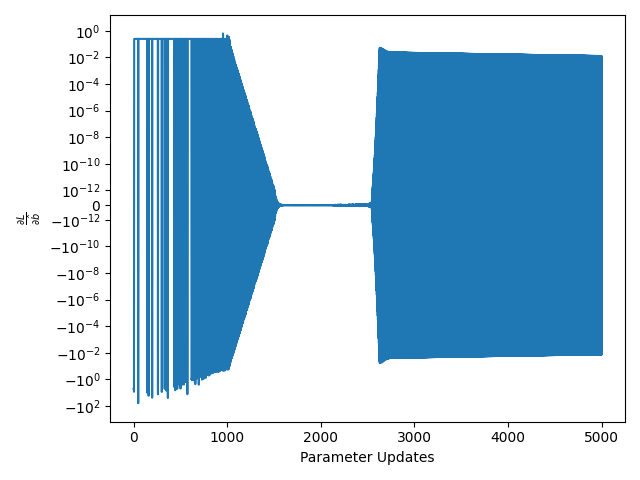}
		\label{fig:}
	\end{subfigure}
	\begin{subfigure}[b]{0.3\textwidth}
		\centering
		\includegraphics[width=\textwidth]{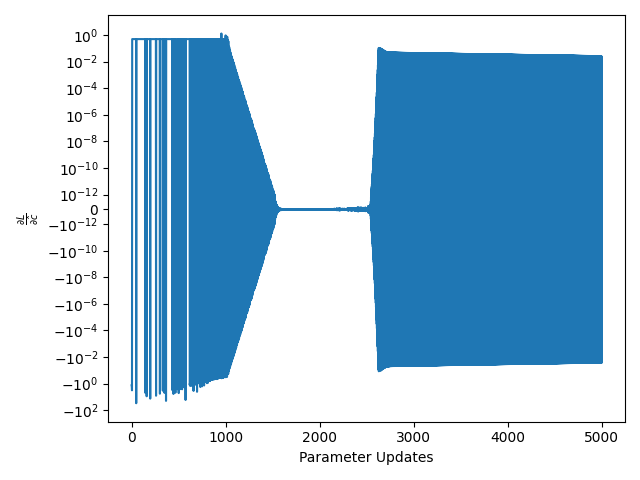}
		\label{fig:}
	\end{subfigure}
	\caption{Adam Oscillations Near Convergence.} 
	\label{fig:appendix_figure}
\end{figure}

We illustrate this in Figure \ref{fig:appendix_figure}, which is the example from Figure \ref{fig:example-3} run for a longer number of parameter updates. The value of $L$ jumps significantly after about 2500 parameter updates, and both $\frac{\partial L}{\partial \hat{b}}$ and $\frac{\partial L}{\partial \hat{c}}$ return to having relatively large magnitudes. Note that $\tilde{c}$ remains bounded away from zero during the disturbance (and thus is not the cause for this behavior). In this particular example, the solution is not too adversely affected, but we have observed other examples that are more dramatic.

\newpage

\section*{Appendix B: Failure of L-BFGS}
\begin{figure}[H]
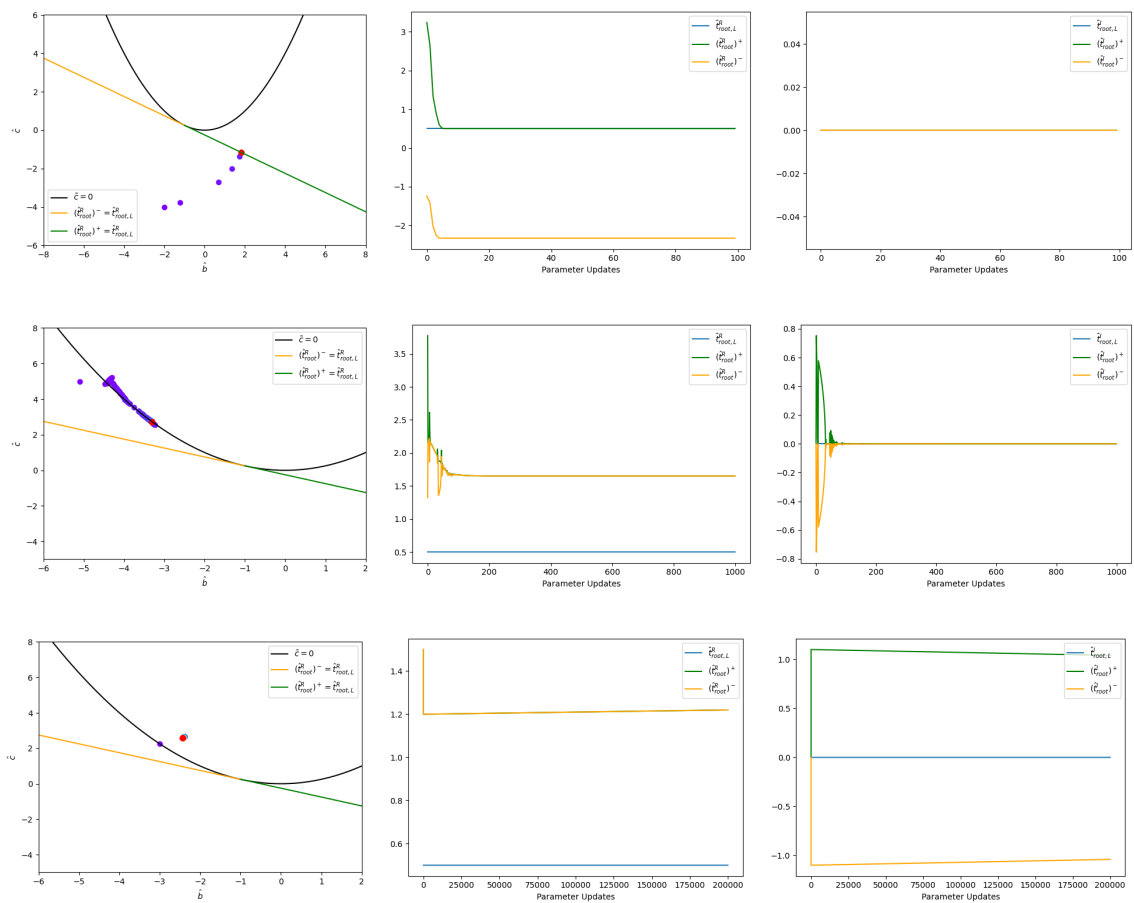

	\centering
    \foreach \subexample in {1, 2, 3}{
	\begin{subfigure}[b]{0.3\textwidth}
		\centering
		\includegraphics[width=\textwidth]{figures/quadratic-section/smoothing-subsection/examples/example27/backprop_lbfgs_\subexample_phase_diagram_1.png}
		\label{fig:}
	\end{subfigure}
	\begin{subfigure}[b]{0.3\textwidth}
		\centering
		\includegraphics[width=\textwidth]{figures/quadratic-section/smoothing-subsection/examples/example27/backprop_lbfgs_\subexample_roots_real.png}
		\label{fig:}
	\end{subfigure}
	\begin{subfigure}[b]{0.3\textwidth}
		\centering
		\includegraphics[width=\textwidth]{figures/quadratic-section/smoothing-subsection/examples/example27/backprop_lbfgs_\subexample_roots_imag.png}
		\label{fig:}
	\end{subfigure}
    }
	\caption{
    The results obtained using the default implementation of L-BFGS in Pytorch on the examples from Figures \ref{fig:example-1} (top row), \ref{fig:example-3} (middle row), and \ref{fig:example-5a} (bottom row).
    } 
	\label{fig:lbfgs}
\end{figure}

Although we focused on the use of Adam, one might also consider so-called second order optimization methods that seek to estimate Hessians;
however, given the various issues addressed in the paper, this seems significantly more difficult.
Figure \ref{fig:lbfgs} shows the results obtained using the default implementation of L-BFGS in Pytorch \cite{paszke2019pytorch} on the examples shown in Figures \ref{fig:example-1}, \ref{fig:example-3}, and \ref{fig:example-5a}.
For the example from Figure \ref{fig:example-1}, L-BFGS performs better than Adam (as expected in this simple case).
For the examples from Figures \ref{fig:example-3} and \ref{fig:example-5a} where large gradients near the $\tilde{c}=0$ parabola are problematic, L-BFGS fails to converge.
For the example from Figure \ref{fig:example-3}, L-BFGS starts out moving in the right direction, but the Hessian approximation deteriorates to the point where the iterates eventually stall out and even start to move in the wrong direction.
For the example from Figure \ref{fig:example-5a}, the initial large gradient causes L-BFGS to jump into the complex region where it gets stuck (even after 200,000 iterations).
Recall from Remark \ref{re:specificform} and equation \ref{eq:grad_in_dldp_complex} that the gradient is constant in this region, meaning that L-BFGS is not getting any new information (which is needed to update its current poor approximation of the Hessian).

\end{document}